\documentclass{article}

\PassOptionsToPackage{numbers, compress}{natbib}

\usepackage[final]{neurips_2021}

\usepackage[utf8]{inputenc}
\usepackage[T1]{fontenc}
\usepackage[colorlinks=true,citecolor=black,urlcolor=black,linkcolor=black]{hyperref}
\usepackage{subcaption}
\usepackage{url}
\usepackage{booktabs}
\usepackage{amsmath,amsfonts,amssymb,mathrsfs}
\usepackage{nicefrac}
\usepackage{microtype}
\usepackage{xcolor}
\usepackage{mfirstuc}
\usepackage{mathtools}
\usepackage{siunitx}

\usepackage{bm}
\newcommand{\mathbold}[1]{\bm{#1}}
\newcommand{\mbf}[1]{\mathbf{#1}}
\newcommand{\vect}[1]{\mbf{#1}}
\newcommand{\vectb}[1]{\mathbold{#1}}

\usepackage{xspace}
\newcommand{\eg}{\textit{e.g.}\xspace}
\newcommand{\ie}{\textit{i.e.}\xspace}

\newcommand{\cf}{\textit{cf.}\xspace}
\newcommand{\etc}{\textit{etc.}\xspace}

\newcommand{\T}{^\mathsf{T}}
\newcommand{\dd}{\,\mathrm{d}}
\newcommand{\E}{\mathbb{E}}
\newcommand{\R}{\mathbb{R}}
\newcommand{\N}{\mathrm{N}}
\newcommand{\Uni}{\mathrm{Uniform}}

\newcommand{\imag}[0]{\mathrm{i}}
\newcommand{\norm}[1]{\|#1\|}
\newcommand{\cbar}{\;|\;}

\newcommand{\vomega}[0]{\mathbold{\omega}}

\renewcommand{\mid}[0]{\,|\,}

\newcommand{\GP}{\mathcal{GP}}

\newcommand{\vf}{\mbf{f}}

\newcommand{\vr}{\mbf{r}}
\newcommand{\vs}{\mbf{s}}

\newcommand{\vw}{\mbf{w}}
\newcommand{\vx}{\mbf{x}}
\newcommand{\vy}{\mbf{y}}

\newcommand{\MK}{\mbf{K}}

\newcommand{\cauchy}{\textrm{Cauchy}}
\newcommand{\tdist}{$t$\textrm{-dist}}
\newcommand{\gaussian}{\textrm{N}}

\usepackage{tabularx}
\usepackage{array}
\usepackage{colortbl}
\newcommand{\PreserveBackslash}[1]{\let\temp=\\#1\let\\=\temp}
\newcolumntype{C}[1]{>{\PreserveBackslash\centering}p{#1}}
\newlength{\tblw}

\usepackage{tikz,pgfplots}
\usetikzlibrary{arrows}
\tikzset{>=stealth'}
\usepgfplotslibrary{groupplots}
\usetikzlibrary{plotmarks,shapes,arrows,patterns}
\pgfplotsset{compat=newest} 

\pgfplotsset{/pgf/number format/.cd, 1000 sep={}}
\pgfkeys{/pgf/number format/.cd,1000 sep={}}

\pgfplotsset{every axis/.append style={
		grid style={line width=0.6pt,dotted,gray}}}

\pgfplotsset{every axis/.append style={
		legend style={inner xsep=1pt, inner ysep=0.5pt, nodes={inner sep=1pt, text depth=0.1em}, draw=none,fill=none}
}}

\usepgfplotslibrary{groupplots}
\usepgfplotslibrary{fillbetween}

\newlength\figureheight
\newlength\figurewidth

\usepackage[capitalize,nameinlink]{cleveref}
\crefname{section}{Sec.}{Sects.}
\crefname{proposition}{Prop.}{Props.}
\crefname{lemma}{Lem.}{Lems.}
\crefname{model}{Mod.}{Mods.}
\crefname{appendix}{App.}{Apps.}

\usepackage[framemethod=tikz]{mdframed}

\definecolor{mycolor0}{rgb}{0.2667,0.4471,0.7098}
\definecolor{mycolor1}{rgb}{0.1647,0.6706,0.3804}
\definecolor{mycolor2}{rgb}{0.8275,0.2627,0.3059}
\definecolor{mycolor3}{rgb}{0.5216,0.4392,0.7176}
\definecolor{mycolor4}{rgb}{0.8118,0.7255,0.4118}
\definecolor{mycolor5}{rgb}{0.2745,0.7176,0.8157}

\definecolor{mylcolor0}{rgb}{0.6902,0.7686,0.8863}
\definecolor{mylcolor1}{rgb}{0.5451,0.8902,0.6941}
\definecolor{mylcolor2}{rgb}{0.9412,0.7490,0.7647}
\definecolor{mylcolor3}{rgb}{0.8627,0.8392,0.9176}
\definecolor{mylcolor4}{rgb}{0.9569,0.9373,0.8667}
\definecolor{mylcolor5}{rgb}{0.7529,0.9020,0.9373}
\definecolor{mylcolor6}{rgb}{0.8750,0.8750,0.8750}

\renewcommand{\paragraph}[1]{\textbf{#1}~~}

\newcommand{\nipstitle}[1]{{%

    \def\toptitlebar{\hrule height4pt \vskip .25in \vskip -\parskip} 
    \def\bottomtitlebar{\vskip .29in \vskip -\parskip \hrule height1pt \vskip .09in} 
    \phantomsection\hsize\textwidth\linewidth\hsize%
    \vskip 0.1in%
    \toptitlebar%
    \begin{minipage}{\textwidth}%
        \centering{\LARGE\bf #1\par}%
    \end{minipage}%
    \bottomtitlebar%
    \addcontentsline{toc}{section}{#1}%
}}

\title{Periodic Activation Functions Induce Stationarity}

\author{
	Lassi Meronen \\
	Aalto University / Saab Finland Oy \\
 	Espoo, Finland \\
 	\texttt{lassi.meronen@aalto.fi} 
 	\And
	Martin Trapp \\
	Aalto University \\
	Espoo, Finland \\
 	\texttt{martin.trapp@aalto.fi} 
	\And
	Arno Solin \\
	Aalto University \\
	Espoo, Finland \\
 	\texttt{arno.solin@aalto.fi}
}

\begin{document}

\maketitle

\begin{abstract}
Neural network models are known to reinforce hidden data biases, making them unreliable and difficult to interpret. We seek to build models that `know what they do not know' by introducing inductive biases in the function space. We show that \emph{periodic} activation functions in Bayesian neural networks establish a connection between the prior on the network weights and translation-invariant, stationary Gaussian process priors. Furthermore, we show that this link goes beyond sinusoidal (Fourier) activations by also covering triangular wave and periodic ReLU activation functions. In a series of experiments, we show that periodic activation functions obtain comparable performance for in-domain data and capture sensitivity to perturbed inputs in deep neural networks for out-of-domain detection.
\end{abstract}

\section{Introduction}
Deep feedforward neural networks \cite{Schmidhuber15,goodfellow2016deep} are an integral part of contemporary artificial intelligence and machine learning systems for visual and auditory perception, medicine, and general data analysis and decision making. However, when these methods have been adopted into real-world use, concerns related to robustness (with respect to data that has not been seen during training), fairness (hidden biases in data being reinforced by the model), and interpretability (why the model acts as it does) have taken a central role. The knowledge gathered by contemporary neural networks has even been characterised as never truly reliable \cite{marcus2020next}. These issues relate to the sensitivity of the trained model to perturbed inputs being fed through it---or the lack thereof.

This motivates {\em Bayesian deep learning}, where the interests are two-fold: encoding prior knowledge into models and performing probabilistic inference under the specified model. We focus on the former. Probabilistic approaches to specifying assumptions about the function space of deep neural networks have gained increasing attention in the machine learning community, comprising, among others, work analysing their posterior distribution \cite{Wenzel+Roth+Veeling:2020,aitchison2021a}, discussing pathologies arising in uncertainty quantification \cite{Foong+Burt+Li:2019}, and calls for better Bayesian priors (\eg, \cite{pearce2019expressive,SunZSG19,Nalisnick0H21,Fortuin2021}).

\begin{figure}[t]
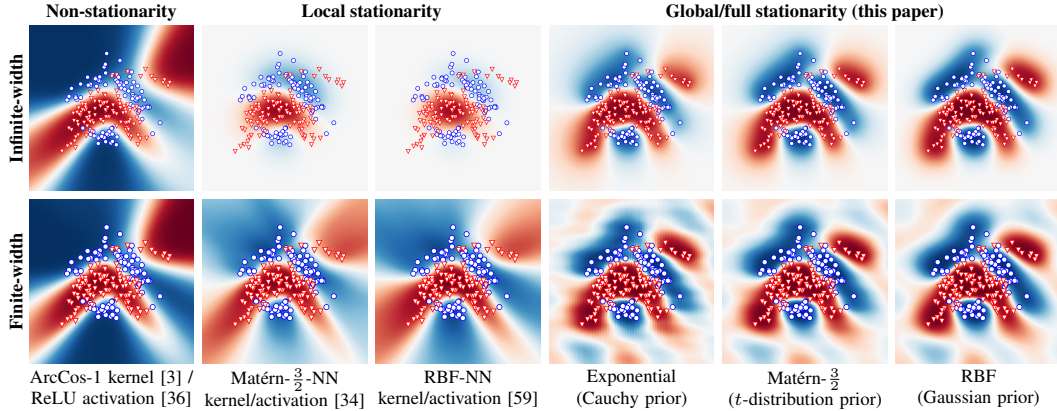

  \centering\scriptsize
  \setlength{\figurewidth}{.165\textwidth}
  \begin{tikzpicture}[inner sep=0]
  
    \node[rotate=90,minimum width=.99\figurewidth] at ({0.45\figurewidth},0) {\bf Infinite-width};
  
    \foreach \file [count=\i] in 
      {GP_ArCos-1_mean,
       GP_Matern32-NN_mean,
       GP_RBF-NN_mean,
       GP_Matern12_mean,
       GP_Matern32_mean,
       GP_RBF_mean} {

       \node at ({\i*\figurewidth},0) {\includegraphics[width=.95\figurewidth]{./img/\file}};  
     }
  
    \node[rotate=90,minimum width=.99\figurewidth] at ({0.45\figurewidth},-\figurewidth) {\bf Finite-width};

    \foreach \file/\cap [count=\i] in 
          {banana_RELU_30_combined_mean/{ArcCos-1 kernel~\cite{cho+saul:2009} / ReLU activation~\cite{NairH10}},
       banana_MaternLS_15_30_combined_mean/{Mat\'ern-$\frac{3}{2}$-NN kernel/activation \cite{meronen2020stationary}},
       banana_RBFLS_30_combined_mean/{RBF-NN \mbox{kernel/activation}~\cite{williams97computing}},
       banana_Matern_SinActivation_5_30_combined_mean/{Exponential\\ (Cauchy prior)},
       banana_Matern_SinActivation_15_30_combined_mean/{Mat\'ern-$\frac{3}{2}$\\ ($t$-distribution prior)},
       banana_RBF_SinActivation_30_combined_mean/{RBF\\ (Gaussian prior)}} {

       \node at ({\i*\figurewidth},-\figurewidth) {\includegraphics[width=.95\figurewidth]{./img/\file}};

       \node[text width=\figurewidth,text centered] at ({\i*\figurewidth},{-1.62*\figurewidth}) {\cap};     
     }

     \tikzstyle{box} = [minimum width=.95\figurewidth,inner sep=1pt,rounded corners=1pt]
     \node[box,fill=white] at (\figurewidth,.55\figurewidth) {\bf Non-stationarity};
     \node[box,fill=white,minimum width=1.95\figurewidth] at (2.5\figurewidth,.55\figurewidth) {\bf Local stationarity};
     \node[box,minimum width=2.95\figurewidth,fill=white] at (5\figurewidth,.55\figurewidth) {\bf Global/full stationarity (this paper)};      
  \end{tikzpicture}
  \caption{Posterior predictive densities of single hidden layer Bayesian neural networks (BNNs) with 30 hidden units and their infinite-width corresponding GPs on the banana classification task. Different activation functions induce different prior assumptions. Estimates obtained through HMC sampling \cite{Ge2018} for 10k iterations.}
  \label{fig:teaser}
  \vspace*{-1em}
\end{figure}

In this paper, we focus on stationary models, which act as a proxy for capturing {\em sensitivity}. Stationarity indicates translation-invariance, meaning that the joint probability distribution does not change when the inputs are shifted. This seemingly naive assumption has strong consequences in the sense that it induces {\em conservative} behaviour across the input domain, both in-distribution and outside the observed data. The resulting model is mean-reverting outside the training data (reversion to the prior), directly leading to higher uncertainty for out-of-distribution (OOD) samples (see \cref{fig:teaser} for examples). These features (together with some direct computational benefits) have made stationary models/priors the standard approach in kernel methods \cite{cortes1995support,hofmann2008kernel}, spatial statistics \cite{Cressie:1991}, and Gaussian process (GP) models \cite{Rasmussen+Williams:2006}, where the kernel is often chosen to induce stationarity.

Neural networks are parametric models, which typically fall into the class of non-stationary models. Non-stationarity increases {\em flexibility} and is often a sought-after property---especially if the interest is solely in optimizing for accuracy on in-domain test data. In fact, all standard neural network activation functions (ReLU, sigmoid, \etc) induce non-stationarity (see \cite{cho+saul:2009,williams97computing}). However, non-stationary models can easily result in over-confidence outside the observed data \cite{hein2019relu}, \cite{pmlr-v161-ulmer21a} or spurious relationships between input variables (as illustrated in \cref{fig:teaser}, where darker shades show higher confidence).

Stationarity in neural networks has been tackled before under the restriction of local stationarity, \ie, translation-invariance is only {\em local}, induced by a Gaussian envelope (as realized by \cite{williams97computing} for the RBF kernel/activation). Recently, \citet{meronen2020stationary} expanded this approach and derived activation functions corresponding to the widely used Mat\'ern class of kernels \cite{Matern:1960,Rasmussen+Williams:2006}. We go one step further and derive activation functions that induce {\em global} stationarity. To do so, we leverage theory from harmonic analysis \cite{wiener1930generalized} of periodic functions, which helps expand the effective support over the entire input domain. We also realize direct links to previous works leveraging harmonic functions in neural networks \cite{GallantW88,CANDES1999197,WongLC02,SitzmannMBLW20,UteuliyevaZTACK20}, and Fourier features in other model families \cite{rahimi2008,SutherlandS15}.

The contributions of this paper are:
{\em (i)}~We show that periodic activation functions establish a direct correspondence between the prior on the network weights and the spectral density of the covariance function of the limiting stationary Gaussian process (GP) of single hidden layer Bayesian neural networks (BNNs).
{\em (ii)}~We leverage this correspondence and show that placing a Student-$t$ prior on the weights of the hidden layer corresponds to a prior on the function space with Mat\'ern covariance.
{\em (iii)}~Finally, we show in a range of experiments that periodic activation functions obtain comparable performance for in-domain data, do not result in overconfident predictions, and enable robust out-of-domain detection.

\subsection{Related Work}
We build upon prior work on exploring the covariance function induced by different activation functions, starting with the seminal work by \citet{williams97computing}, who discussed a sigmoidal (ERF) and a Gaussian (RBF) activation function, resulting in a locally stationary kernel modulated by a Gaussian envelope. 
\citet{cho+saul:2009} introduced non-stationary GP kernels corresponding to the ReLU and step activations, and \cite{tsuchida2018invariance} later extended the approach to the leaky ReLU and analysed alternative weight prior specifications.
More recently, \cite{meronen2020stationary} derived activation functions corresponding to kernels from the Mat\'ern family, which are locally stationary modulated by Gaussian envelope.
In addition to the work connecting neural networks to GPs at initialisation, \cite{jacot2018neural} created a connection between the neural network training and kernel methods by introducing the Neural Tangent Kernel (NTK).

At the same time, uninformative priors have been widely criticised in Bayesian deep learning \cite{Fortuin2021}, and alternative strategies have been proposed to incorporate prior knowledge.
\citet{pearce2019expressive} proposed compositions of BNNs to mimic compositional kernels in their limiting case.
\citet{SunZSG19} proposed a variational objective for BNNs acting on the function space rather than the network parameters.
\citet{Nalisnick0H21} proposed tractable priors on the functional complexity, obtained through the change of variables formula applied to the KL-divergence from a reference model/prior.  
Furthermore, \citet{morales-alvarez2021activationlevel} proposed to encode prior assumptions on the activation level rather than the model parameters.

Recent works have shown the potential of periodic activation functions in neural networks, including approaches to learn implicit neural representations~\cite{SitzmannMBLW20}, methods for recognition of handwritten digits \cite{WongLC02}, and out-of-distribution detection \cite{maennel2019}. 
Earlier work used periodic activation functions to resemble the Fourier series \cite{GallantW88,UteuliyevaZTACK20} and showed that such models are universal function approximators \cite{CANDES1999197}.
In broader terms, \cite{Thorp2021} recently showed that the use of the Fourier transform, replacing self-attention, in Transformers can drastically improve the runtime. 
Inspired by the Fourier duality of covariance and spectral density, \cite{rahimi2008} proposed random Fourier features to approximate a GP with an RBF kernel using Monte Carlo integration.
\cite{SutherlandS15} later analysed the quality of the approximation, and various works have explored (random or deterministic) Fourier features, \eg, in the context of kernel methods (\eg, \cite{lazaro2010sparse,hensman2018variational,solin2020hilbert}).
\cite{NEURIPS2020_55053683} investigated the use of Fourier features in multilayer perceptrons (MLPs) to help represent high-frequency functions.

To the best of our knowledge, none of the previous works has shown that periodic activation functions in BNNs establish a direct correspondence between the prior on the network weights and the density of the spectral density decomposition, thus, inducing global stationarity and enabling principled prior choices for BNNs.

\begin{figure}[t]
  \centering\scriptsize
  \begin{subfigure}[b]{.48\textwidth}
    \centering\tiny

    \pgfplotsset{axis on top,scale only axis, width=\figurewidth,height=\figureheight,yticklabel={\empty},xticklabel={\empty},ticks=none}
    \setlength{\figurewidth}{.3\textwidth}
    \setlength{\figureheight}{\figurewidth}  
    \begin{minipage}[t]{.05\textwidth}
      \centering
      \tikz\node[rotate=90]{Highly non-stationary};
    \end{minipage} 
    \hfill
    \begin{minipage}[t]{.3\textwidth}
      \centering
      % This file was created by tikzplotlib v0.9.8.
\begin{tikzpicture}

\begin{axis}[
height=\figureheight,
tick align=outside,
tick pos=left,
width=\figurewidth,
x grid style={white!69.0196078431373!black},
xmin=-1, xmax=1,
xtick style={color=black},
y dir=reverse,
y grid style={white!69.0196078431373!black},
ymin=-1, ymax=1,
ytick style={color=black}
]
\addplot graphics [includegraphics cmd=\pgfimage,xmin=-2.0364238410596, xmax=1.93708609271523, ymin=1.33112582781457, ymax=-1.31788079470199] {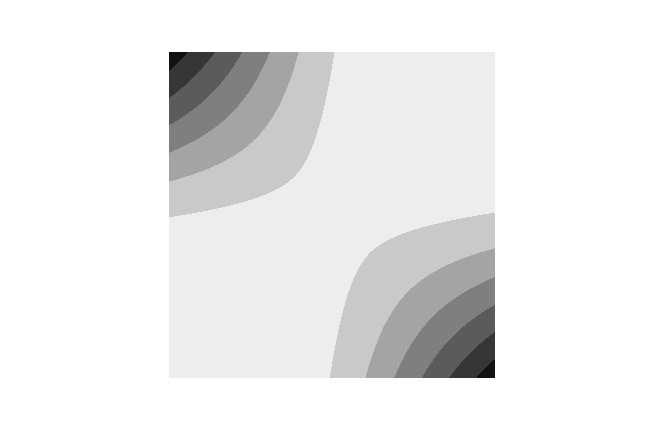};
\end{axis}

\end{tikzpicture}\\
      ArcCos-1 $\sim$ ReLU
    \end{minipage} 
    \hfill 
    \begin{minipage}[t]{.3\textwidth}
      \centering
      % This file was created by tikzplotlib v0.9.8.
\begin{tikzpicture}

\begin{axis}[
height=\figureheight,
tick align=outside,
tick pos=left,
width=\figurewidth,
x grid style={white!69.0196078431373!black},
xmin=-1, xmax=1,
xtick style={color=black},
y dir=reverse,
y grid style={white!69.0196078431373!black},
ymin=-1, ymax=1,
ytick style={color=black}
]
\addplot graphics [includegraphics cmd=\pgfimage,xmin=-2.0364238410596, xmax=1.93708609271523, ymin=1.33112582781457, ymax=-1.31788079470199] {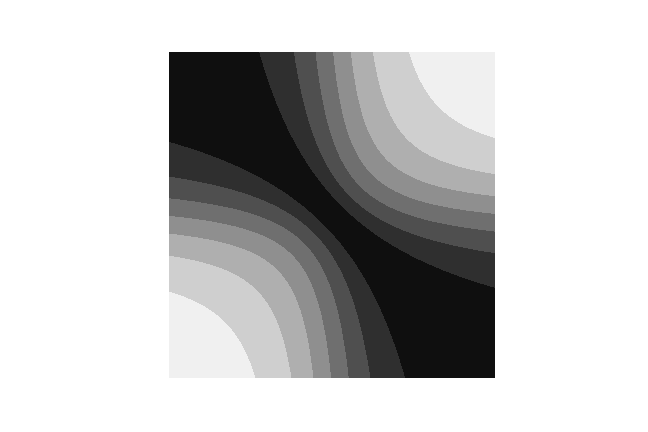};
\end{axis}

\end{tikzpicture}\\
      ArcCos-0 $\sim$ Step
    \end{minipage} 
    \hfill     
    \begin{minipage}[t]{.3\textwidth}
      \centering
      % This file was created by tikzplotlib v0.9.8.
\begin{tikzpicture}

\begin{axis}[
height=\figureheight,
tick align=outside,
tick pos=left,
width=\figurewidth,
x grid style={white!69.0196078431373!black},
xmin=-1, xmax=1,
xtick style={color=black},
y dir=reverse,
y grid style={white!69.0196078431373!black},
ymin=-1, ymax=1,
ytick style={color=black}
]
\addplot graphics [includegraphics cmd=\pgfimage,xmin=-2.0364238410596, xmax=1.93708609271523, ymin=1.33112582781457, ymax=-1.31788079470199] {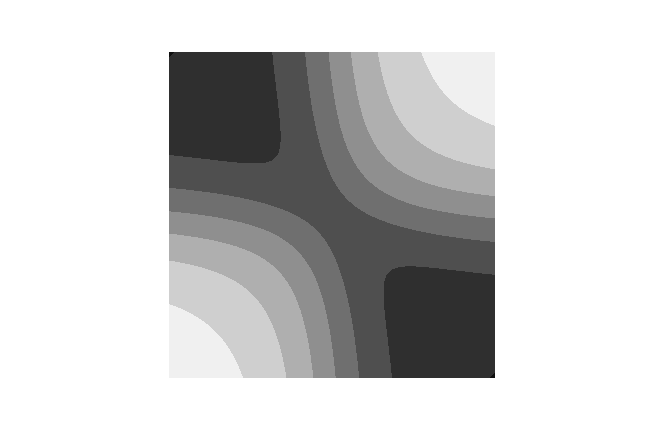};
\end{axis}

\end{tikzpicture}\\
      NN $\sim$ Sigmoid
    \end{minipage}\\[1em]
    \begin{minipage}[t]{.05\textwidth}
      \centering
      \tikz\node[rotate=90]{\hspace*{1em}Locally stationary};
    \end{minipage} 
    \hfill    
    \begin{minipage}[t]{.3\textwidth}
      \centering
      % This file was created by tikzplotlib v0.9.8.
\begin{tikzpicture}

\begin{axis}[
height=\figureheight,
tick align=outside,
tick pos=left,
width=\figurewidth,
x grid style={white!69.0196078431373!black},
xmin=-1, xmax=1,
xtick style={color=black},
y dir=reverse,
y grid style={white!69.0196078431373!black},
ymin=-1, ymax=1,
ytick style={color=black}
]
\addplot graphics [includegraphics cmd=\pgfimage,xmin=-2.0364238410596, xmax=1.93708609271523, ymin=1.33112582781457, ymax=-1.31788079470199] {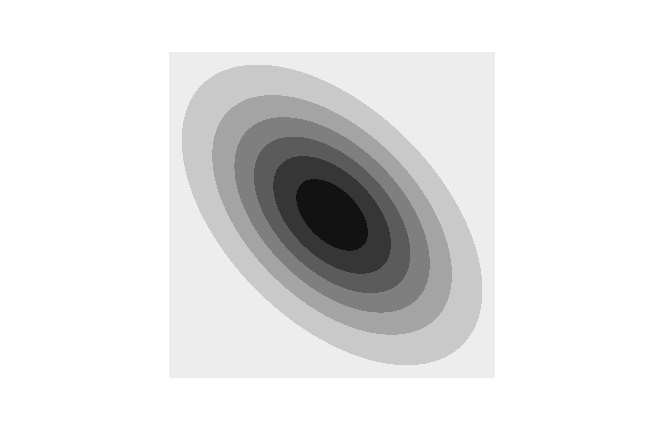};
\end{axis}

\end{tikzpicture}\\
      RBF-NN (local)
    \end{minipage} 
    \hfill 
    \begin{minipage}[t]{.3\textwidth}
      \centering
      % This file was created by tikzplotlib v0.9.8.
\begin{tikzpicture}

\begin{axis}[
height=\figureheight,
tick align=outside,
tick pos=left,
width=\figurewidth,
x grid style={white!69.0196078431373!black},
xmin=-1, xmax=1,
xtick style={color=black},
y dir=reverse,
y grid style={white!69.0196078431373!black},
ymin=-1, ymax=1,
ytick style={color=black}
]
\addplot graphics [includegraphics cmd=\pgfimage,xmin=-2.0364238410596, xmax=1.93708609271523, ymin=1.33112582781457, ymax=-1.31788079470199] {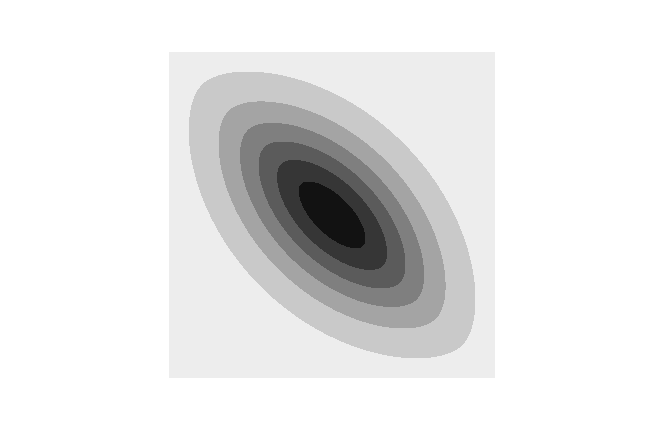};
\end{axis}

\end{tikzpicture}\\
      Mat\'ern-$\nicefrac{3}{2}$-NN (local)
    \end{minipage} 
    \hfill     
    \begin{minipage}[t]{.3\textwidth}
      \centering
      % This file was created by tikzplotlib v0.9.8.
\begin{tikzpicture}

\begin{axis}[
height=\figureheight,
tick align=outside,
tick pos=left,
width=\figurewidth,
x grid style={white!69.0196078431373!black},
xmin=-1, xmax=1,
xtick style={color=black},
y dir=reverse,
y grid style={white!69.0196078431373!black},
ymin=-1, ymax=1,
ytick style={color=black}
]
\addplot graphics [includegraphics cmd=\pgfimage,xmin=-2.0364238410596, xmax=1.93708609271523, ymin=1.33112582781457, ymax=-1.31788079470199] {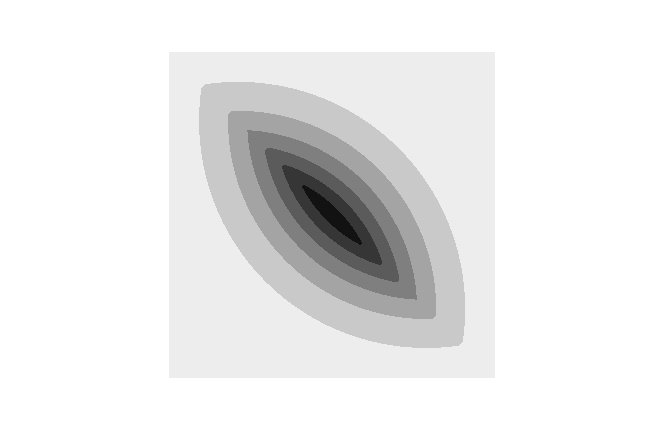};
\end{axis}

\end{tikzpicture}\\
      Exp-NN (local)
    \end{minipage}\\[1em]  
    \begin{minipage}[t]{.05\textwidth}
      \centering
      \tikz\node[rotate=90]{\hspace*{2.5em}Stationary};
    \end{minipage} 
    \hfill      
    \begin{minipage}[t]{.3\textwidth}
      \centering
      % This file was created by tikzplotlib v0.9.8.
\begin{tikzpicture}

\begin{axis}[
height=\figureheight,
tick align=outside,
tick pos=left,
width=\figurewidth,
x grid style={white!69.0196078431373!black},
xmin=-1, xmax=1,
xtick style={color=black},
y dir=reverse,
y grid style={white!69.0196078431373!black},
ymin=-1, ymax=1,
ytick style={color=black}
]
\addplot graphics [includegraphics cmd=\pgfimage,xmin=-2.0364238410596, xmax=1.93708609271523, ymin=1.33112582781457, ymax=-1.31788079470199] {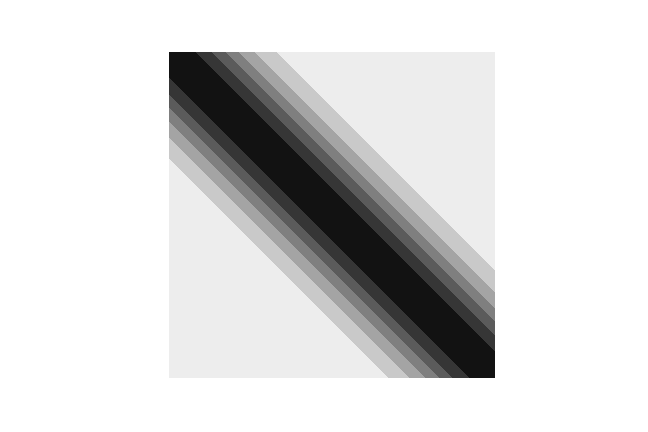};
\end{axis}

\end{tikzpicture}\\
      RBF
    \end{minipage} 
    \hfill 
    \begin{minipage}[t]{.3\textwidth}
      \centering
      % This file was created by tikzplotlib v0.9.8.
\begin{tikzpicture}

\begin{axis}[
height=\figureheight,
tick align=outside,
tick pos=left,
width=\figurewidth,
x grid style={white!69.0196078431373!black},
xmin=-1, xmax=1,
xtick style={color=black},
y dir=reverse,
y grid style={white!69.0196078431373!black},
ymin=-1, ymax=1,
ytick style={color=black}
]
\addplot graphics [includegraphics cmd=\pgfimage,xmin=-2.0364238410596, xmax=1.93708609271523, ymin=1.33112582781457, ymax=-1.31788079470199] {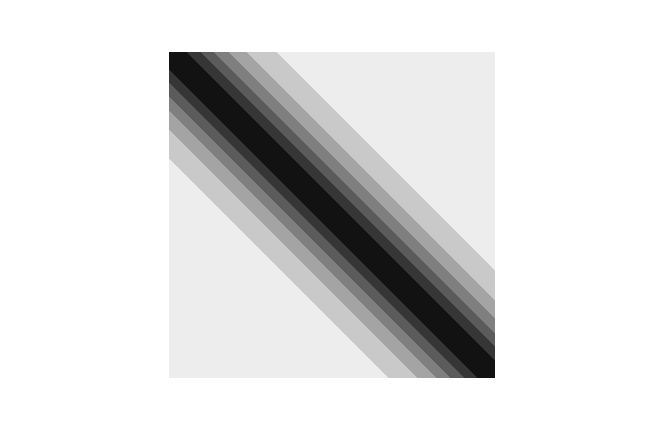};
\end{axis}

\end{tikzpicture}\\
      Mat\'ern-$\nicefrac{3}{2}$
    \end{minipage} 
    \hfill     
    \begin{minipage}[t]{.3\textwidth}
      \centering
      % This file was created by tikzplotlib v0.9.8.
\begin{tikzpicture}

\begin{axis}[
height=\figureheight,
tick align=outside,
tick pos=left,
width=\figurewidth,
x grid style={white!69.0196078431373!black},
xmin=-1, xmax=1,
xtick style={color=black},
y dir=reverse,
y grid style={white!69.0196078431373!black},
ymin=-1, ymax=1,
ytick style={color=black}
]
\addplot graphics [includegraphics cmd=\pgfimage,xmin=-2.0364238410596, xmax=1.93708609271523, ymin=1.33112582781457, ymax=-1.31788079470199] {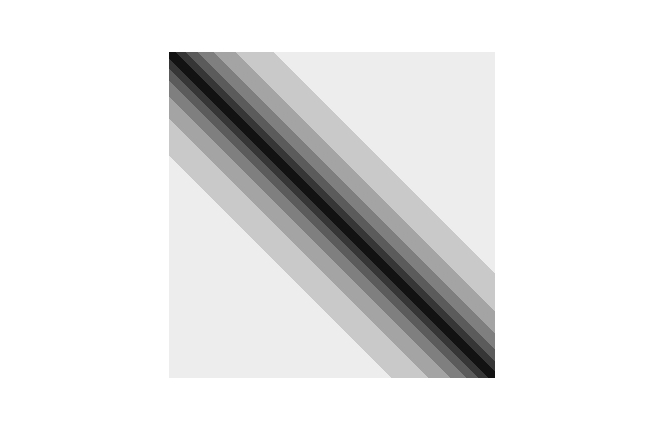};
\end{axis}

\end{tikzpicture}\\
      Exponential
    \end{minipage}
    \caption{Gram matrices}
    \label{fig:gram}
  \end{subfigure}
  \hfill
  \begin{subfigure}[b]{.48\textwidth}
\centering\tiny

    \pgfplotsset{axis on top,scale only axis,width=\figurewidth,height=\figureheight,yticklabel={\empty},xticklabel={\empty},ticks=none}
    \setlength{\figurewidth}{.3\textwidth}
    \setlength{\figureheight}{\figurewidth}  
    \begin{minipage}[t]{.3\textwidth}
      \centering
      % This file was created by tikzplotlib v0.9.8.
\begin{tikzpicture}

\definecolor{color0}{rgb}{0.12156862745098,0.466666666666667,0.705882352941177}

\begin{axis}[
height=\figureheight,
tick align=outside,
tick pos=left,
width=\figurewidth,
x grid style={white!69.0196078431373!black},
xmin=-5.5, xmax=5.5,
xtick style={color=black},
y grid style={white!69.0196078431373!black},
ymin=-2.97558073300145, ymax=2.34945155812376,
ytick style={color=black}
]
\addplot graphics [includegraphics cmd=\pgfimage,xmin=-7.2741935483871, xmax=6.91935483870968, ymin=-3.85720859577052, ymax=3.19581430638208] {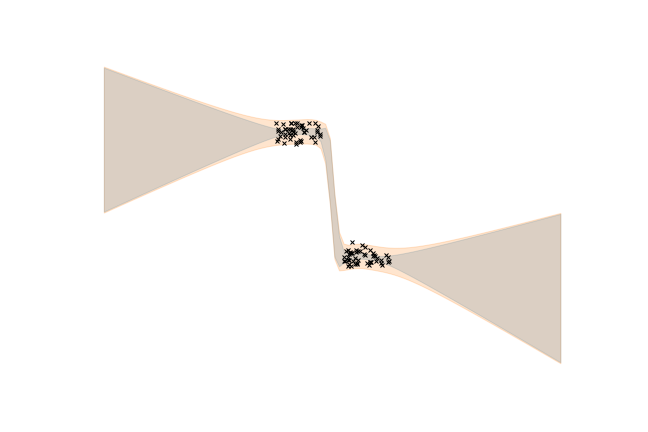};
\addplot [semithick, color0]
table {%
-5 0.91656787311676
-4.8989898989899 0.919638478027137
-4.7979797979798 0.922709865591658
-4.6969696969697 0.925782086234016
-4.5959595959596 0.928855194866428
-4.49494949494949 0.931929251277154
-4.39393939393939 0.935004320798029
-4.29292929292929 0.938080474855207
-4.19191919191919 0.941157791777068
-4.09090909090909 0.944236357621031
-3.98989898989899 0.947316267136725
-3.88888888888889 0.950397624930613
-3.78787878787879 0.953480546714914
-3.68686868686869 0.956565160886354
-3.58585858585859 0.959651610295039
-3.48484848484848 0.962740054323028
-3.38383838383838 0.965830671338755
-3.28282828282828 0.968923661643638
-3.18181818181818 0.972019250928527
-3.08080808080808 0.975117694430804
-2.97979797979798 0.978219281914655
-2.87878787878788 0.981324343744741
-2.77777777777778 0.984433258227277
-2.67676767676768 0.987546460633835
-2.57575757575758 0.990664454345743
-2.47474747474747 0.993787824731652
-2.37373737373737 0.99691725647949
-2.27272727272727 1.0000535556436
-2.17171717171717 1.0031976777313
-2.07070707070707 1.00635076402907
-1.96969696969697 1.00951418899619
-1.86868686868687 1.01268962300006
-1.76767676767677 1.01587911637577
-1.66666666666667 1.01908521389878
-1.56565656565657 1.02231111319185
-1.46464646464646 1.02556088815664
-1.36363636363636 1.02883981073861
-1.26262626262626 1.03215482546562
-1.16161616161616 1.03552642354634
-1.06060606060606 1.03898011733819
-0.959595959595959 1.04241795758721
-0.858585858585859 1.04535267998487
-0.757575757575758 1.04754411239581
-0.656565656565657 1.05016632420802
-0.555555555555555 1.05193857888752
-0.454545454545455 1.04831975695437
-0.353535353535354 1.03574902556776
-0.252525252525253 0.992083778190408
-0.151515151515151 0.852633806323953
-0.0505050505050502 0.404591321560006
0.0505050505050502 -0.495998699442226
0.151515151515151 -0.90436416541564
0.252525252525253 -0.997332160393514
0.353535353535354 -0.999249116677231
0.454545454545454 -0.987231453061566
0.555555555555555 -0.982044081051727
0.656565656565657 -0.983707425209573
0.757575757575758 -0.993322596878338
0.858585858585858 -1.00596665792934
0.959595959595959 -1.01891746112966
1.06060606060606 -1.03183216341142
1.16161616161616 -1.04439807298198
1.26262626262626 -1.05680762925941
1.36363636363636 -1.06917106043899
1.46464646464646 -1.08149815406767
1.56565656565657 -1.09379591569332
1.66666666666667 -1.10606966044288
1.76767676767677 -1.11832349346642
1.86868686868687 -1.13056063544326
1.96969696969697 -1.14278364846906
2.07070707070707 -1.15499459609953
2.17171717171717 -1.16719515891462
2.27272727272727 -1.17938671938351
2.37373737373737 -1.19157042512979
2.47474747474747 -1.20374723674087
2.57575757575758 -1.21591796436772
2.67676767676768 -1.2280832960607
2.77777777777778 -1.24024381991739
2.87878787878788 -1.25240004160134
2.97979797979798 -1.26455239827778
3.08080808080808 -1.27670126984661
3.18181818181818 -1.28884698802299
3.28282828282828 -1.30098984371346
3.38383838383838 -1.31313009311198
3.48484848484848 -1.32526796274952
3.58585858585859 -1.33740365364646
3.68686868686869 -1.34953734479285
3.78787878787879 -1.36166919612216
3.88888888888889 -1.37379935093537
3.98989898989899 -1.38592793804222
4.09090909090909 -1.39805507354223
4.19191919191919 -1.41018086232483
4.29292929292929 -1.42230539944931
4.39393939393939 -1.43442877120844
4.49494949494949 -1.44655105615343
4.5959595959596 -1.45867232590731
4.6969696969697 -1.4707926459826
4.7979797979798 -1.48291207633492
4.8989898989899 -1.49503067199932
5 -1.50714848354222
};
\end{axis}

\end{tikzpicture}\\
      ArcCos-1 $\sim$ ReLU
    \end{minipage} 
    \hfill 
    \begin{minipage}[t]{.3\textwidth}
      \centering
      % This file was created by tikzplotlib v0.9.8.
\begin{tikzpicture}

\definecolor{color0}{rgb}{0.12156862745098,0.466666666666667,0.705882352941177}

\begin{axis}[
height=\figureheight,
tick align=outside,
tick pos=left,
width=\figurewidth,
x grid style={white!69.0196078431373!black},
xmin=-5.5, xmax=5.5,
xtick style={color=black},
y grid style={white!69.0196078431373!black},
ymin=-1.31886536828893, ymax=1.35732508672493,
ytick style={color=black}
]
\addplot graphics [includegraphics cmd=\pgfimage,xmin=-7.2741935483871, xmax=6.91935483870968, ymin=-1.76194325819188, ymax=1.78267986103176] {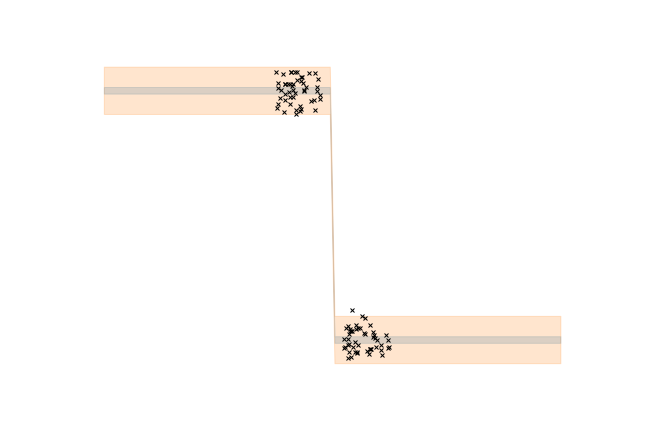};
\addplot [semithick, color0]
table {%
-5 1.04162477558602
-4.8989898989899 1.04162476060543
-4.7979797979798 1.04162473583725
-4.6969696969697 1.0416247799157
-4.5959595959596 1.04162478841176
-4.49494949494949 1.04162474146071
-4.39393939393939 1.0416247725937
-4.29292929292929 1.04162474619058
-4.19191919191919 1.04162479920418
-4.09090909090909 1.04162479456373
-3.98989898989899 1.04162474609088
-3.88888888888889 1.04162476495417
-3.78787878787879 1.04162473645498
-3.68686868686869 1.04162476128735
-3.58585858585859 1.04162475476262
-3.48484848484848 1.04162474639917
-3.38383838383838 1.04162477631997
-3.28282828282828 1.04162469937452
-3.18181818181818 1.04162476286237
-3.08080808080808 1.04162475819689
-2.97979797979798 1.04162478713111
-2.87878787878788 1.04162468572381
-2.77777777777778 1.04162474036946
-2.67676767676768 1.04162477629575
-2.57575757575758 1.04162479361016
-2.47474747474747 1.04162475099183
-2.37373737373737 1.04162481662223
-2.27272727272727 1.0416247786158
-2.17171717171717 1.04162477602566
-2.07070707070707 1.04162479088623
-1.96969696969697 1.04162482439759
-1.86868686868687 1.04162478825144
-1.76767676767677 1.04162475308867
-1.66666666666667 1.04162476449869
-1.56565656565657 1.04162483653308
-1.46464646464646 1.04162482887117
-1.36363636363636 1.04162483410417
-1.26262626262626 1.04162481479612
-1.16161616161616 1.04162486568499
-1.06060606060606 1.04162493052129
-0.959595959595959 1.04162479750576
-0.858585858585859 1.0416248560354
-0.757575757575758 1.04162485534219
-0.656565656565657 1.04162478314765
-0.555555555555555 1.04162475752518
-0.454545454545455 1.04162470347028
-0.353535353535354 1.0416246431619
-0.252525252525253 1.04162448519042
-0.151515151515151 1.04162438276781
-0.0505050505050502 1.04162399698575
0.0505050505050502 -1.00316439510478
0.151515151515151 -1.00316478126754
0.252525252525253 -1.00316487486047
0.353535353535354 -1.00316480416952
0.454545454545454 -1.0031648366803
0.555555555555555 -1.00316486518951
0.656565656565657 -1.00316497774063
0.757575757575758 -1.00316508482625
0.858585858585858 -1.00316516987094
0.959595959595959 -1.00316515652615
1.06060606060606 -1.00316528119551
1.16161616161616 -1.00316529373028
1.26262626262626 -1.00316527267502
1.36363636363636 -1.00316532123578
1.46464646464646 -1.00316535638785
1.56565656565657 -1.00316537216883
1.66666666666667 -1.00316540429531
1.76767676767677 -1.00316539009126
1.86868686868687 -1.00316539952292
1.96969696969697 -1.00316542292039
2.07070707070707 -1.00316539537013
2.17171717171717 -1.00316539393186
2.27272727272727 -1.00316539407189
2.37373737373737 -1.00316539892317
2.47474747474747 -1.00316542005412
2.57575757575758 -1.00316542875453
2.67676767676768 -1.00316542908852
2.77777777777778 -1.00316543496102
2.87878787878788 -1.00316541652814
2.97979797979798 -1.00316541899321
3.08080808080808 -1.0031654827109
3.18181818181818 -1.0031654089232
3.28282828282828 -1.00316547095869
3.38383838383838 -1.00316547452393
3.48484848484848 -1.00316545358983
3.58585858585859 -1.00316540673474
3.68686868686869 -1.00316547804473
3.78787878787879 -1.00316538237133
3.88888888888889 -1.00316545432548
3.98989898989899 -1.00316543813891
4.09090909090909 -1.00316545569798
4.19191919191919 -1.00316546844407
4.29292929292929 -1.00316546011662
4.39393939393939 -1.00316543472616
4.49494949494949 -1.00316539653075
4.5959595959596 -1.00316542232125
4.6969696969697 -1.00316545481204
4.7979797979798 -1.00316543810116
4.8989898989899 -1.0031654268118
5 -1.00316543136163
};
\end{axis}

\end{tikzpicture}\\
      ArcCos-0 $\sim$ Step
    \end{minipage} 
    \hfill     
    \begin{minipage}[t]{.3\textwidth}
      \centering
      % This file was created by tikzplotlib v0.9.8.
\begin{tikzpicture}

\definecolor{color0}{rgb}{0.12156862745098,0.466666666666667,0.705882352941177}

\begin{axis}[
height=\figureheight,
tick align=outside,
tick pos=left,
width=\figurewidth,
x grid style={white!69.0196078431373!black},
xmin=-5.5, xmax=5.5,
xtick style={color=black},
y grid style={white!69.0196078431373!black},
ymin=-1.35800317661076, ymax=1.39204840732447,
ytick style={color=black}
]
\addplot graphics [includegraphics cmd=\pgfimage,xmin=-7.2741935483871, xmax=6.91935483870968, ymin=-1.81330973024242, ymax=1.82914269881086] {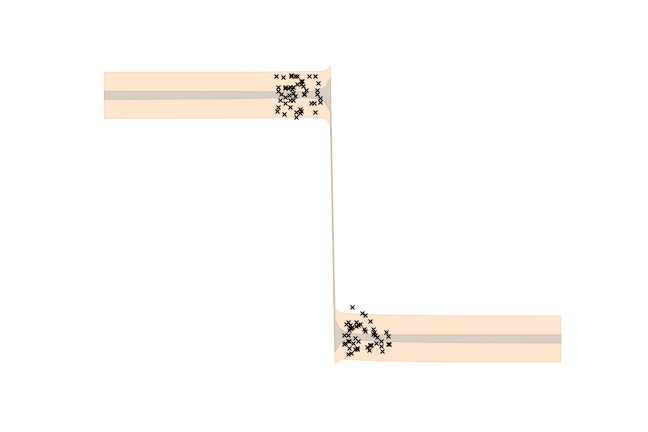};
\addplot [semithick, color0]
table {%
-5 1.02923549844823
-4.8989898989899 1.02927555489855
-4.7979797979798 1.02931719440788
-4.6969696969697 1.02936051179036
-4.5959595959596 1.02940560950981
-4.49494949494949 1.02945259837452
-4.39393939393939 1.02950159848372
-4.29292929292929 1.02955274015199
-4.19191919191919 1.02960616503746
-4.09090909090909 1.029662027316
-3.98989898989899 1.02972049515848
-3.88888888888889 1.02978175214771
-3.78787878787879 1.02984599925887
-3.68686868686869 1.02991345671772
-3.58585858585859 1.02998436633409
-3.48484848484848 1.03005899414768
-3.38383838383838 1.03013763343484
-3.28282828282828 1.0302206079783
-3.18181818181818 1.03030827610541
-3.08080808080808 1.03040103508172
-2.97979797979798 1.03049932606416
-2.87878787878788 1.03060364013622
-2.77777777777778 1.03071452470093
-2.67676767676768 1.03083259100728
-2.57575757575758 1.03095852273137
-2.47474747474747 1.03109308531714
-2.37373737373737 1.03123713637098
-2.27272727272727 1.03139163706382
-2.17171717171717 1.03155766365821
-2.07070707070707 1.03173641902627
-1.96969696969697 1.03192924226179
-1.86868686868687 1.03213761445611
-1.76767676767677 1.03236315588492
-1.66666666666667 1.03260760832513
-1.56565656565657 1.0328727900608
-1.46464646464646 1.03316050461003
-1.36363636363636 1.03347237144394
-1.26262626262626 1.03380952713504
-1.16161616161616 1.03417211454923
-1.06060606060606 1.03455842987348
-0.959595959595959 1.03496352206993
-0.858585858585859 1.03537692226737
-0.757575757575758 1.03577899286416
-0.656565656565657 1.03613512055496
-0.555555555555555 1.03638680783292
-0.454545454545455 1.0364393657221
-0.353535353535354 1.03614809444674
-0.252525252525253 1.03530468546436
-0.151515151515151 1.03355729351253
-0.0505050505050502 1.02808418403503
0.0505050505050502 -0.995550167724143
0.151515151515151 -1.00074899745272
0.252525252525253 -1.0025620017138
0.353535353535354 -1.00432646577051
0.454545454545454 -1.00628310429119
0.555555555555555 -1.00828351826216
0.656565656565657 -1.01018124372678
0.757575757575758 -1.01189796768769
0.858585858585858 -1.01340859701242
0.959595959595959 -1.01471846599841
1.06060606060606 -1.01584702097508
1.16161616161616 -1.01681828263676
1.26262626262626 -1.0176559851861
1.36363636363636 -1.01838146893703
1.46464646464646 -1.01901302940213
1.56565656565657 -1.0195659607959
1.66666666666667 -1.02005288420197
1.76767676767677 -1.02048415638454
1.86868686868687 -1.02086826642267
1.96969696969697 -1.02121218430047
2.07070707070707 -1.0215216525056
2.17171717171717 -1.02180142349248
2.27272727272727 -1.02205545069646
2.37373737373737 -1.02228704067568
2.47474747474747 -1.02249897438479
2.57575757575758 -1.02269360364908
2.67676767676768 -1.02287292821494
2.77777777777778 -1.02303865737048
2.87878787878788 -1.02319225954547
2.97979797979798 -1.02333500246967
3.08080808080808 -1.02346798557056
3.18181818181818 -1.02359216662007
3.28282828282828 -1.02370838346993
3.38383838383838 -1.0238173719838
3.48484848484848 -1.02391978096738
3.58585858585859 -1.02401618444641
3.68686868686869 -1.024107092134
3.78787878787879 -1.02419295804501
3.88888888888889 -1.02427418784964
3.98989898989899 -1.02435114513616
4.09090909090909 -1.02442415659492
4.19191919191919 -1.024493516689
4.29292929292929 -1.02455949137216
4.39393939393939 -1.02462232159452
4.49494949494949 -1.0246822259556
4.5959595959596 -1.02473940342256
4.6969696969697 -1.02479403533855
4.7979797979798 -1.02484628744005
4.8989898989899 -1.02489631133831
5 -1.02494424616675
};
\end{axis}

\end{tikzpicture}\\
      NN $\sim$ Sigmoid
    \end{minipage}\\[1em]
    \begin{minipage}[t]{.3\textwidth}
      \centering
      % This file was created by tikzplotlib v0.9.8.
\begin{tikzpicture}

\definecolor{color0}{rgb}{0.12156862745098,0.466666666666667,0.705882352941177}

\begin{axis}[
height=\figureheight,
tick align=outside,
tick pos=left,
width=\figurewidth,
x grid style={white!69.0196078431373!black},
xmin=-5.5, xmax=5.5,
xtick style={color=black},
y grid style={white!69.0196078431373!black},
ymin=-1.41337325563841, ymax=1.44368623510855,
ytick style={color=black}
]
\addplot graphics [includegraphics cmd=\pgfimage,xmin=-7.2741935483871, xmax=6.91935483870968, ymin=-1.88639635013294, ymax=1.89778840582329] {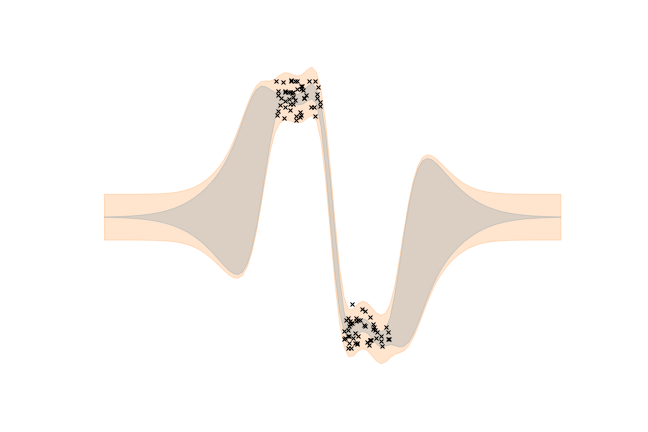};
\addplot [semithick, color0]
table {%
-5 4.57673000985602e-20
-4.8989898989899 4.83595512180778e-19
-4.7979797979798 4.79797380289063e-18
-4.6969696969697 4.46950329669543e-17
-4.5959595959596 3.90895716008855e-16
-4.49494949494949 3.20947749930582e-15
-4.39393939393939 2.47370409095209e-14
-4.29292929292929 1.78963999757227e-13
-4.19191919191919 1.21520223391156e-12
-4.09090909090909 7.74376688376863e-12
-3.98989898989899 4.63049727009492e-11
-3.88888888888889 2.59788463839024e-10
-3.78787878787879 1.36731312154363e-09
-3.68686868686869 6.75000648131544e-09
-3.58585858585859 3.12502286559685e-08
-3.48484848484848 1.35653674256573e-07
-3.38383838383838 5.52010072478776e-07
-3.28282828282828 2.10522161011385e-06
-3.18181818181818 7.52267682390285e-06
-3.08080808080808 2.51796107013088e-05
-2.97979797979798 7.89216085738926e-05
-2.87878787878788 0.000231564686400064
-2.77777777777778 0.000635814140584138
-2.67676767676768 0.00163311885433181
-2.57575757575758 0.00392270690192546
-2.47474747474747 0.00880840472870228
-2.37373737373737 0.0184859741103779
-2.27272727272727 0.0362545118340763
-2.17171717171717 0.066448021956416
-2.07070707070707 0.113857297461217
-1.96969696969697 0.182538685205864
-1.86868686868687 0.274220070410416
-1.76767676767677 0.386896387524759
-1.66666666666667 0.514370897659234
-1.56565656565657 0.647168799063429
-1.46464646464646 0.774480567104576
-1.36363636363636 0.886134230687958
-1.26262626262626 0.973764459753778
-1.16161616161616 1.03144750848879
-1.06060606060606 1.05708600256085
-0.959595959595959 1.05523061307242
-0.858585858585859 1.03947542273048
-0.757575757575758 1.03013478861497
-0.656565656565657 1.04398904523731
-0.555555555555555 1.0787851906594
-0.454545454545455 1.10239292090031
-0.353535353535354 1.05820845766129
-0.252525252525253 0.890246498754803
-0.151515151515151 0.576993720101715
-0.0505050505050502 0.15316881738262
0.0505050505050502 -0.29743726531704
0.151515151515151 -0.677441646558914
0.252525252525253 -0.920076071317091
0.353535353535354 -1.01623242689895
0.454545454545454 -1.00984936678892
0.555555555555555 -0.967639572713259
0.656565656565657 -0.943743392006094
0.757575757575758 -0.959230243656604
0.858585858585858 -1.00335502089426
0.959595959595959 -1.04929308903066
1.06060606060606 -1.07133696934163
1.16161616161616 -1.05454271598784
1.26262626262626 -0.995738565419508
1.36363636363636 -0.900027699110664
1.46464646464646 -0.777100333972598
1.56565656565657 -0.638966393661308
1.66666666666667 -0.498452641553999
1.76767676767677 -0.367496865560118
1.86868686868687 -0.255217530026026
1.96969696969697 -0.166515948742389
2.07070707070707 -0.101878561831332
2.17171717171717 -0.0583812570397199
2.27272727272727 -0.0313137757390657
2.37373737373737 -0.0157159251980286
2.47474747474747 -0.00738022451295201
2.57575757575758 -0.00324323524596753
2.67676767676768 -0.00133402349457303
2.77777777777778 -0.000513741059956149
2.87878787878788 -0.000185289061526522
2.97979797979798 -6.26050764719731e-05
3.08080808080808 -1.9822041679222e-05
3.18181818181818 -5.88278803030514e-06
3.28282828282828 -1.63689719975549e-06
3.38383838383838 -4.27130203006663e-07
3.48484848484848 -1.04541081717446e-07
3.58585858585859 -2.40037431500915e-08
3.68686868686869 -5.17135774037701e-09
3.78787878787879 -1.0454995250947e-09
3.88888888888889 -1.9837604007221e-10
3.98989898989899 -3.53302275028884e-11
4.09090909090909 -5.90656822721316e-12
4.19191919191919 -9.27022019462593e-13
4.29292929292929 -1.36596872471888e-13
4.39393939393939 -1.88978837084259e-14
4.49494949494949 -2.4548764991634e-15
4.5959595959596 -2.99439804265141e-16
4.6969696969697 -3.42980619631957e-17
4.7979797979798 -3.68913330372037e-18
4.8989898989899 -3.72637441860046e-19
5 -3.53481791437967e-20
};
\end{axis}

\end{tikzpicture}\\
      RBF-NN (local)
    \end{minipage} 
    \hfill 
    \begin{minipage}[t]{.3\textwidth}
      \centering
      % This file was created by tikzplotlib v0.9.8.
\begin{tikzpicture}

\definecolor{color0}{rgb}{0.12156862745098,0.466666666666667,0.705882352941177}

\begin{axis}[
height=\figureheight,
tick align=outside,
tick pos=left,
width=\figurewidth,
x grid style={white!69.0196078431373!black},
xmin=-5.5, xmax=5.5,
xtick style={color=black},
y grid style={white!69.0196078431373!black},
ymin=-1.39508141346796, ymax=1.41460107287846,
ytick style={color=black}
]
\addplot graphics [includegraphics cmd=\pgfimage,xmin=-7.2741935483871, xmax=6.91935483870968, ymin=-1.86026063306174, ymax=1.86117312368849] {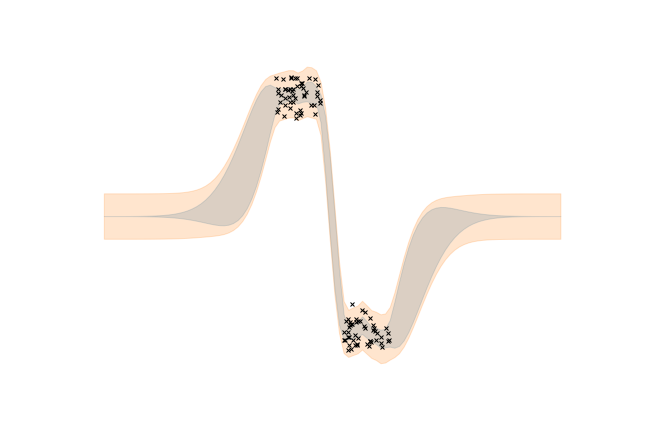};
\addplot [semithick, color0]
table {%
-5 3.25076826194551e-06
-4.8989898989899 5.56016721527952e-06
-4.7979797979798 9.41801920657758e-06
-4.6969696969697 1.57975169861882e-05
-4.5959595959596 2.62399106031476e-05
-4.49494949494949 4.31584473780042e-05
-4.39393939393939 7.0288414024983e-05
-4.29292929292929 0.000113344224528076
-4.19191919191919 0.000180965002400259
-4.09090909090909 0.000286055019383411
-3.98989898989899 0.000447654330756494
-3.88888888888889 0.000693506762026969
-3.78787878787879 0.00106352449878598
-3.68686868686869 0.00161437650160035
-3.58585858585859 0.0024254449621007
-3.48484848484848 0.00360639017180853
-3.38383838383838 0.00530652615506039
-3.28282828282828 0.00772612021914537
-3.18181818181818 0.011129568803211
-3.08080808080808 0.0158601468483866
-2.97979797979798 0.022355655056233
-2.87878787878788 0.0311637781500537
-2.77777777777778 0.0429553039375367
-2.67676767676768 0.0585325367122303
-2.57575757575758 0.0788292880700128
-2.47474747474747 0.104897789316003
-2.37373737373737 0.137876822110976
-2.27272727272727 0.178934427105917
-2.17171717171717 0.229177884839778
-2.07070707070707 0.289523468471403
-1.96969696969697 0.3605189719377
-1.86868686868687 0.442113458132756
-1.76767676767677 0.533371270850273
-1.66666666666667 0.63213127963128
-1.56565656565657 0.734617650904774
-1.46464646464646 0.835015093974021
-1.36363636363636 0.925029266302618
-1.26262626262626 0.993461375624015
-1.16161616161616 1.02905603151883
-1.06060606060606 1.0408768223878
-0.959595959595959 1.05265280349373
-0.858585858585859 1.05307543006977
-0.757575757575758 1.03097426999942
-0.656565656565657 1.04755508306576
-0.555555555555555 1.0736515864942
-0.454545454545455 1.06585601244749
-0.353535353535354 1.04674908243507
-0.252525252525253 0.926087355854931
-0.151515151515151 0.61215274653815
-0.0505050505050502 0.175255256372026
0.0505050505050502 -0.293575221044083
0.151515151515151 -0.70080408849841
0.252525252525253 -0.956277268251663
0.353535353535354 -1.01055285333248
0.454545454545454 -0.992450400105085
0.555555555555555 -0.977276175822871
0.656565656565657 -0.939653787141166
0.757575757575758 -0.977510946257457
0.858585858585858 -1.01620152627823
0.959595959595959 -1.03208080957935
1.06060606060606 -1.05694367128267
1.16161616161616 -1.05051075218363
1.26262626262626 -1.0100098815187
1.36363636363636 -0.93717419170416
1.46464646464646 -0.843787137230334
1.56565656565657 -0.740845269852567
1.66666666666667 -0.636469697245873
1.76767676767677 -0.5363302944201
1.86868686868687 -0.444083189388775
1.96969696969697 -0.361792830309165
2.07070707070707 -0.290318241176601
2.17171717171717 -0.229650749609412
2.27272727272727 -0.179197065875806
2.37373737373737 -0.138006861454415
2.47474747474747 -0.104947873126719
2.57575757575758 -0.0788341189943769
2.67676767676768 -0.058514224357337
2.77777777777778 -0.0429273387247925
2.87878787878788 -0.0311339175203255
2.97979797979798 -0.0223279698433835
3.08080808080808 -0.0158364348209858
3.18181818181818 -0.0111103030180581
3.28282828282828 -0.00771106409361625
3.38383838383838 -0.00529511620580563
3.48484848484848 -0.00359796158588379
3.58585858585859 -0.00241935443527138
3.68686868686869 -0.00161006063771314
3.78787878787879 -0.00106051987915506
3.88888888888889 -0.000691448871358342
3.98989898989899 -0.000446266192852299
4.09090909090909 -0.000285132045992613
4.19191919191919 -0.000180359677522237
4.29292929292929 -0.000112952417106483
4.39393939393939 -7.00380097796819e-05
4.49494949494949 -4.3000372464896e-05
4.5959595959596 -2.61413103993711e-05
4.6969696969697 -1.57367305381633e-05
4.7979797979798 -9.38097217965858e-06
4.8989898989899 -5.53784140506902e-06
5 -3.23746229390501e-06
};
\end{axis}

\end{tikzpicture}\\
      Mat\'ern-$\nicefrac{3}{2}$-NN (local)
    \end{minipage} 
    \hfill     
    \begin{minipage}[t]{.3\textwidth}
      \centering
      % This file was created by tikzplotlib v0.9.8.
\begin{tikzpicture}

\definecolor{color0}{rgb}{0.12156862745098,0.466666666666667,0.705882352941177}

\begin{axis}[
height=\figureheight,
tick align=outside,
tick pos=left,
width=\figurewidth,
x grid style={white!69.0196078431373!black},
xmin=-5.5, xmax=5.5,
xtick style={color=black},
y grid style={white!69.0196078431373!black},
ymin=-1.41584218396233, ymax=1.45191228742009,
ytick style={color=black}
]
\addplot graphics [includegraphics cmd=\pgfimage,xmin=-7.2741935483871, xmax=6.91935483870968, ymin=-1.89063597061504, ymax=1.9077143226067] {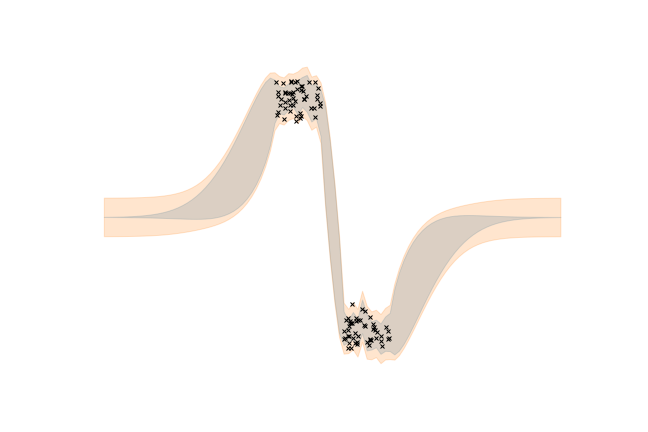};
\addplot [semithick, color0]
table {%
-5 0.000413832765163344
-4.8989898989899 0.000574320554716505
-4.7979797979798 0.000791895039900092
-4.6969696969697 0.00108483743135152
-4.5959595959596 0.00147654088436079
-4.49494949494949 0.00199668735641326
-4.39393939393939 0.00268261553353992
-4.29292929292929 0.00358088668481343
-4.19191919191919 0.0047490473256365
-4.09090909090909 0.00625757686223828
-3.98989898989899 0.0081919947280316
-3.88888888888889 0.0106550848268712
-3.78787878787879 0.0137691754870643
-3.68686868686869 0.0176783909756747
-3.58585858585859 0.0225507666028902
-3.48484848484848 0.0285800945840197
-3.38383838383838 0.0359873435124918
-3.28282828282828 0.0450214722854438
-3.18181818181818 0.0559594416957999
-3.08080808080808 0.0691052159954174
-2.97979797979798 0.0847875450112184
-2.87878787878788 0.103356327302904
-2.77777777777778 0.125177378603808
-2.67676767676768 0.150625469146072
-2.57575757575758 0.180075549497423
-2.47474747474747 0.213892157340793
-2.37373737373737 0.252417086159207
-2.27272727272727 0.295955498661886
-2.17171717171717 0.344760779201207
-2.07070707070707 0.399018535166916
-1.96969696969697 0.458830270919673
-1.86868686868687 0.52419736173663
-1.76767676767677 0.595006041374189
-1.66666666666667 0.671014176992669
-1.56565656565657 0.751840631628869
-1.46464646464646 0.836958000607883
-1.36363636363636 0.925689449554178
-1.26262626262626 1.01721027572322
-1.16161616161616 1.03248258741894
-1.06060606060606 1.01964332146784
-0.959595959595959 1.05880780214105
-0.858585858585859 1.06521860260565
-0.757575757575758 1.0525130493041
-0.656565656565657 1.09892508952139
-0.555555555555555 1.08726589763725
-0.454545454545455 1.00048467914433
-0.353535353535354 1.0206107977575
-0.252525252525253 0.916534025518617
-0.151515151515151 0.544663169549071
-0.0505050505050502 0.161929097314886
0.0505050505050502 -0.224332241236241
0.151515151515151 -0.606669139512494
0.252525252525253 -0.977753213878791
0.353535353535354 -1.00266961813287
0.454545454545454 -0.956641072497011
0.555555555555555 -1.01778047171202
0.656565656565657 -0.873688578027896
0.757575757575758 -1.01359191507293
0.858585858585858 -1.04085640101046
0.959595959595959 -1.02400727703714
1.06060606060606 -1.06976387152799
1.16161616161616 -1.02597242055852
1.26262626262626 -1.01073862287129
1.36363636363636 -0.919800066690835
1.46464646464646 -0.831633141273615
1.56565656565657 -0.747057302474617
1.66666666666667 -0.666745078541889
1.76767676767677 -0.591220518718265
1.86868686868687 -0.520862335113305
1.96969696969697 -0.455911120079185
2.07070707070707 -0.396479916060626
2.17171717171717 -0.342567356530209
2.27272727272727 -0.294072582914094
2.37373737373737 -0.250811168686172
2.47474747474747 -0.212531341565433
2.57575757575758 -0.178929880336109
2.67676767676768 -0.149667166059437
2.77777777777778 -0.124380980298972
2.87878787878788 -0.102698757981872
2.97979797979798 -0.0842481132235374
3.08080808080808 -0.0686655576682694
3.18181818181818 -0.055603418866412
3.28282828282828 -0.0447350385495012
3.38383838383838 -0.0357583863343812
3.48484848484848 -0.0283982634965465
3.58585858585859 -0.0224072950547917
3.68686868686869 -0.0175659182528693
3.78787878787879 -0.0136815738122324
3.88888888888889 -0.0105872954899435
3.98989898989899 -0.0081398759603488
4.09090909090909 -0.00621776516733782
4.19191919191919 -0.00471883313452739
4.29292929292929 -0.0035581045166824
4.39393939393939 -0.00266554830871665
4.49494949494949 -0.00198398411527136
4.5959595959596 -0.00146714689744055
4.6969696969697 -0.00107793552382654
4.7979797979798 -0.000786856878257707
4.8989898989899 -0.00057066663640228
5 -0.000411199895579928
};
\end{axis}

\end{tikzpicture}\\
      Exp-NN (local)
    \end{minipage}\\[1em] 
    \begin{minipage}[t]{.3\textwidth}
      \centering
      % This file was created by tikzplotlib v0.9.8.
\begin{tikzpicture}

\definecolor{color0}{rgb}{0.12156862745098,0.466666666666667,0.705882352941177}

\begin{axis}[
height=\figureheight,
tick align=outside,
tick pos=left,
width=\figurewidth,
x grid style={white!69.0196078431373!black},
xmin=-5.5, xmax=5.5,
xtick style={color=black},
y grid style={white!69.0196078431373!black},
ymin=-1.95870343222169, ymax=2.03489920516907,
ytick style={color=black}
]
\addplot graphics [includegraphics cmd=\pgfimage,xmin=-7.2741935483871, xmax=6.91935483870968, ymin=-2.61989592185592, ymax=2.66964399521793] {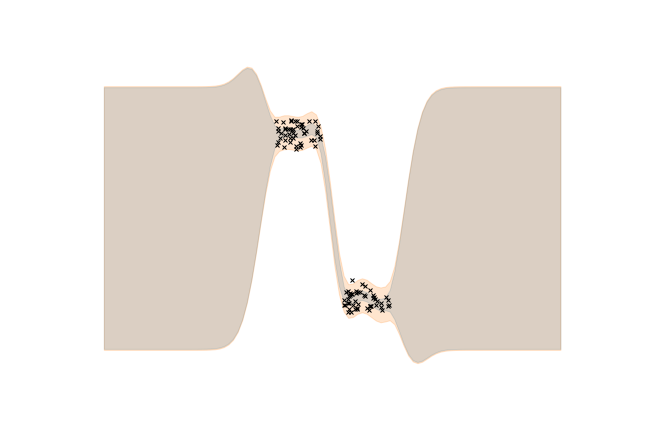};
\addplot [semithick, color0]
table {%
-5 1.99588044368504e-20
-4.8989898989899 2.3654041780787e-19
-4.7979797979798 2.61925192284587e-18
-4.6969696969697 2.7097708806312e-17
-4.5959595959596 2.61909452793985e-16
-4.49494949494949 2.36488615687676e-15
-4.39393939393939 1.99472844114931e-14
-4.29292929292929 1.57160063683328e-13
-4.19191919191919 1.15651628963747e-12
-4.09090909090909 7.9482952360521e-12
-3.98989898989899 5.10111401986473e-11
-3.88888888888889 3.05685486860333e-10
-3.78787878787879 1.71019740901712e-09
-3.68686868686869 8.93128898415312e-09
-3.58585858585859 4.35315557847215e-08
-3.48484848484848 1.97984708913163e-07
-3.38383838383838 8.40040044341315e-07
-3.28282828282828 3.32429343556784e-06
-3.18181818181818 1.22660552816944e-05
-3.08080808080808 4.21866637138229e-05
-2.97979797979798 0.000135191763938053
-2.87878787878788 0.000403506195372652
-2.77777777777778 0.00112117686783359
-2.67676767676768 0.00289867609917022
-2.57575757575758 0.00696922826733137
-2.47474747474747 0.0155729726460646
-2.37373737373737 0.0323219033318853
-2.27272727272727 0.0622745712772124
-2.17171717171717 0.111329349912031
-2.07070707070707 0.184627157320068
-1.96969696969697 0.284100978678391
-1.86868686868687 0.406071969525816
-1.76767676767677 0.540429084359499
-1.66666666666667 0.672719646059594
-1.56565656565657 0.788993701122153
-1.46464646464646 0.881079250119085
-1.36363636363636 0.948819057145257
-1.26262626262626 0.997306664016713
-1.16161616161616 1.03102328780669
-1.06060606060606 1.05005777048154
-0.959595959595959 1.05265923297516
-0.858585858585859 1.04260361631266
-0.757575757575758 1.03358446698812
-0.656565656565657 1.04259785554546
-0.555555555555555 1.07263781276428
-0.454545454545455 1.09651346632328
-0.353535353535354 1.05788569012184
-0.252525252525253 0.896333632080561
-0.151515151515151 0.585153986682203
-0.0505050505050502 0.157499330113467
0.0505050505050502 -0.299614356712421
0.151515151515151 -0.683642557206097
0.252525252525253 -0.925066415517398
0.353535353535354 -1.01659776635948
0.454545454545454 -1.00675904980102
0.555555555555555 -0.965430395282508
0.656565656565657 -0.945591276288284
0.757575757575758 -0.963798330842789
0.858585858585858 -1.00563187337114
0.959595959595959 -1.04535851748059
1.06060606060606 -1.06341009497082
1.16161616161616 -1.05219961995386
1.26262626262626 -1.01198028218814
1.36363636363636 -0.944611775920716
1.46464646464646 -0.851329093943726
1.56565656565657 -0.734866400665611
1.66666666666667 -0.602295572095357
1.76767676767677 -0.465202035144413
1.86868686868687 -0.336741275075129
1.96969696969697 -0.227614304029144
2.07070707070707 -0.143362592331665
2.17171717171717 -0.0840526137117101
2.27272727272727 -0.0458552138196213
2.37373737373737 -0.0232792702928025
2.47474747474747 -0.0110004950182356
2.57575757575758 -0.00484042529274022
2.67676767676768 -0.00198410848405825
2.77777777777778 -0.000757945316408704
2.87878787878788 -0.000269940928815875
2.97979797979798 -8.96621039980932e-05
3.08080808080808 -2.77837439544991e-05
3.18181818181818 -8.03393239467698e-06
3.28282828282828 -2.16829222930297e-06
3.38383838383838 -5.46312111075042e-07
3.48484848484848 -1.28518225537171e-07
3.58585858585859 -2.82322351590702e-08
3.68686868686869 -5.79199664704444e-09
3.78787878787879 -1.10981706979661e-09
3.88888888888889 -1.98630205535813e-10
3.98989898989899 -3.32074206109427e-11
4.09090909090909 -5.18610884841413e-12
4.19191919191919 -7.56626994407917e-13
4.29292929292929 -1.03126434173049e-13
4.39393939393939 -1.31316154735748e-14
4.49494949494949 -1.56219984525921e-15
4.5959595959596 -1.73633797025439e-16
4.6969696969697 -1.80309711590933e-17
4.7979797979798 -1.74943711584798e-18
4.8989898989899 -1.58590991694962e-19
5 -1.34327775518323e-20
};
\end{axis}

\end{tikzpicture}\\
      RBF
    \end{minipage} 
    \hfill 
    \begin{minipage}[t]{.3\textwidth}
      \centering
      % This file was created by tikzplotlib v0.9.8.
\begin{tikzpicture}

\definecolor{color0}{rgb}{0.12156862745098,0.466666666666667,0.705882352941177}

\begin{axis}[
height=\figureheight,
tick align=outside,
tick pos=left,
width=\figurewidth,
x grid style={white!69.0196078431373!black},
xmin=-5.5, xmax=5.5,
xtick style={color=black},
y grid style={white!69.0196078431373!black},
ymin=-2.38218559206964, ymax=2.41741191958385,
ytick style={color=black}
]
\addplot graphics [includegraphics cmd=\pgfimage,xmin=-7.2741935483871, xmax=6.91935483870968, ymin=-3.17682094168115, ymax=3.18026185521089] {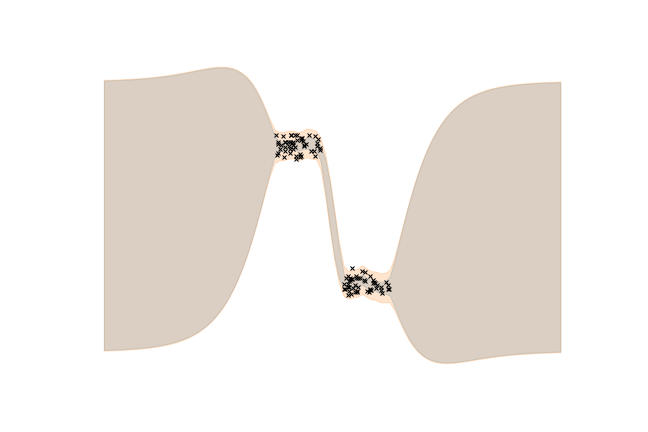};
\addplot [semithick, color0]
table {%
-5 0.0127480648412799
-4.8989898989899 0.0147831160010664
-4.7979797979798 0.017133352785627
-4.6969696969697 0.0198454653623716
-4.5959595959596 0.0229725795094615
-4.49494949494949 0.0265750299294342
-4.39393939393939 0.0307211975673649
-4.29292929292929 0.0354884077978445
-4.19191919191919 0.0409638829281617
-4.09090909090909 0.0472457378789553
-3.98989898989899 0.0544440018299983
-3.88888888888889 0.0626816406616355
-3.78787878787879 0.0720955446893135
-3.68686868686869 0.0828374328690267
-3.58585858585859 0.0950746075891253
-3.48484848484848 0.108990472424455
-3.38383838383838 0.124784697659003
-3.28282828282828 0.142672883562202
-3.18181818181818 0.162885527582622
-3.08080808080808 0.18566604665254
-2.97979797979798 0.211267537037904
-2.87878787878788 0.239947868386143
-2.77777777777778 0.27196260185351
-2.67676767676768 0.307555089583688
-2.57575757575758 0.346942948427228
-2.47474747474747 0.390299897387567
-2.37373737373737 0.437731697007934
-2.27272727272727 0.489244618976645
-2.17171717171717 0.5447044924717
-2.07070707070707 0.603783904157977
-1.96969696969697 0.665894551780976
-1.86868686868687 0.73010104322973
-1.76767676767677 0.795011564902781
-1.66666666666667 0.858639780127943
-1.56565656565657 0.918231017654535
-1.46464646464646 0.970044220131039
-1.36363636363636 1.00907918025102
-1.26262626262626 1.02873622177512
-1.16161616161616 1.02919048550961
-1.06060606060606 1.03282849394148
-0.959595959595959 1.05150103415843
-0.858585858585859 1.05626266365714
-0.757575757575758 1.02670929078379
-0.656565656565657 1.04785607563405
-0.555555555555555 1.07625218490204
-0.454545454545455 1.06493880897133
-0.353535353535354 1.04583828947681
-0.252525252525253 0.922660887536299
-0.151515151515151 0.605921863635242
-0.0505050505050502 0.174307007232811
0.0505050505050502 -0.286070709051655
0.151515151515151 -0.690948135309065
0.252525252525253 -0.952536533119483
0.353535353535354 -1.01026894320209
0.454545454545454 -0.992275157493244
0.555555555555555 -0.978667908318681
0.656565656565657 -0.932861155252253
0.757575757575758 -0.97920242976693
0.858585858585858 -1.0184683437073
0.959595959595959 -1.02448827419932
1.06060606060606 -1.05625636938126
1.16161616161616 -1.05020363473106
1.26262626262626 -1.02950148938507
1.36363636363636 -0.992969022661474
1.46464646464646 -0.942361387316077
1.56565656565657 -0.883025566788195
1.66666666666667 -0.818972281182199
1.76767676767677 -0.753164447631715
1.86868686868687 -0.68774789556096
1.96969696969697 -0.624235375311979
2.07070707070707 -0.563652853237182
2.17171717171717 -0.506655411564423
2.27272727272727 -0.453618701116272
2.37373737373737 -0.40471077489788
2.47474747474747 -0.359948215837643
2.57575757575758 -0.319239725664459
2.67676767676768 -0.282419733661849
2.77777777777778 -0.249274088870499
2.87878787878788 -0.219559496680114
2.97979797979798 -0.193018033787619
3.08080808080808 -0.169387810350185
3.18181818181818 -0.14841063346314
3.28282828282828 -0.129837352527565
3.38383838383838 -0.113431427006113
3.48484848484848 -0.0989711442497723
3.58585858585859 -0.0862508244013761
3.68686868686869 -0.0750812766673554
3.78787878787879 -0.0652897130868153
3.88888888888889 -0.0567192795356753
3.98989898989899 -0.0492283268172509
4.09090909090909 -0.0426895154651058
4.19191919191919 -0.0369888248205692
4.29292929292929 -0.0320245188293164
4.39393939393939 -0.0277061068427057
4.49494949494949 -0.0239533267119749
4.5959595959596 -0.0206951689821102
4.6969696969697 -0.0178689545071036
4.7979797979798 -0.0154194729007355
4.8989898989899 -0.0132981855707384
5 -0.0114624943897615
};
\end{axis}

\end{tikzpicture}\\
      Mat\'ern-$\nicefrac{3}{2}$
    \end{minipage} 
    \hfill     
    \begin{minipage}[t]{.3\textwidth}
      \centering
      % This file was created by tikzplotlib v0.9.8.
\begin{tikzpicture}

\definecolor{color0}{rgb}{0.12156862745098,0.466666666666667,0.705882352941177}

\begin{axis}[
height=\figureheight,
tick align=outside,
tick pos=left,
width=\figurewidth,
x grid style={white!69.0196078431373!black},
xmin=-5.5, xmax=5.5,
xtick style={color=black},
y grid style={white!69.0196078431373!black},
ymin=-2.51775734057808, ymax=2.53466651786341,
ytick style={color=black}
]
\addplot graphics [includegraphics cmd=\pgfimage,xmin=-7.2741935483871, xmax=6.91935483870968, ymin=-3.35425135687635, ymax=3.33770077350974] {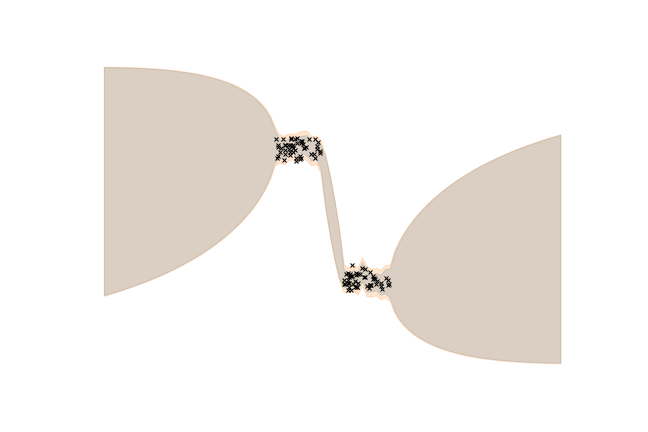};
\addplot [semithick, color0]
table {%
-5 0.530443476111254
-4.8989898989899 0.54069428933464
-4.7979797979798 0.551143199389198
-4.6969696969697 0.561794034493604
-4.5959595959596 0.572650696846818
-4.49494949494949 0.583717164057701
-4.39393939393939 0.594997490602386
-4.29292929292929 0.606495809309701
-4.19191919191919 0.618216332875343
-4.09090909090909 0.630163355405264
-3.98989898989899 0.642341253988981
-3.88888888888889 0.654754490303204
-3.78787878787879 0.667407612246505
-3.68686868686869 0.680305255605489
-3.58585858585859 0.693452145753381
-3.48484848484848 0.706853099381088
-3.38383838383838 0.720513026261958
-3.28282828282828 0.73443693105078
-3.18181818181818 0.748629915117145
-3.08080808080808 0.763097178414544
-2.97979797979798 0.777844021385555
-2.87878787878788 0.792875846903712
-2.77777777777778 0.808198162253038
-2.67676767676768 0.823816581145587
-2.57575757575758 0.839736825778506
-2.47474747474747 0.855964728930313
-2.37373737373737 0.872506236097898
-2.27272727272727 0.889367407674581
-2.17171717171717 0.906554421171046
-2.07070707070707 0.924073573478073
-1.96969696969697 0.941931283173777
-1.86868686868687 0.960134092875328
-1.76767676767677 0.978688671635757
-1.66666666666667 0.997601817387524
-1.56565656565657 1.01688045943307
-1.46464646464646 1.0365316609835
-1.36363636363636 1.05656262174623
-1.26262626262626 1.07698068056324
-1.16161616161616 1.03476244165408
-1.06060606060606 1.01345971950166
-0.959595959595959 1.06100968370913
-0.858585858585859 1.06470119708916
-0.757575757575758 1.05549152304255
-0.656565656565657 1.10061880468207
-0.555555555555555 1.08635393332927
-0.454545454545455 1.00082860836945
-0.353535353535354 1.02123882889441
-0.252525252525253 0.910853345665605
-0.151515151515151 0.534068341946097
-0.0505050505050502 0.157479007476815
0.0505050505050502 -0.219052630632505
0.151515151515151 -0.595664524133535
0.252525252525253 -0.972494654181563
0.353535353535354 -1.00310044371931
0.454545454545454 -0.956813415014135
0.555555555555555 -1.01769457052927
0.656565656565657 -0.871292315813966
0.757575757575758 -1.01450759874721
0.858585858585858 -1.04383300834725
0.959595959595959 -1.02494855900728
1.06060606060606 -1.07037589177861
1.16161616161616 -1.01737064180444
1.26262626262626 -1.04497538011281
1.36363636363636 -1.0251640973679
1.46464646464646 -1.00572840904508
1.56565656565657 -0.986661194395475
1.66666666666667 -0.967955467669667
1.76767676767677 -0.949604375558356
1.86868686868687 -0.931601194681536
1.96969696969697 -0.913939329125117
2.07070707070707 -0.896612308024349
2.17171717171717 -0.879613783193137
2.27272727272727 -0.862937526798189
2.37373737373737 -0.846577429077313
2.47474747474747 -0.830527496100795
2.57575757575758 -0.814781847575683
2.67676767676768 -0.799334714691088
2.77777777777778 -0.78418043800479
2.87878787878788 -0.769313465369863
2.97979797979798 -0.754728349900157
3.08080808080808 -0.740419747975188
3.18181818181818 -0.726382417281827
3.28282828282828 -0.712611214894127
3.38383838383838 -0.699101095388751
3.48484848484848 -0.685847108996751
3.58585858585859 -0.67284439978974
3.68686868686869 -0.660088203901058
3.78787878787879 -0.647573847780386
3.88888888888889 -0.635296746481401
3.98989898989899 -0.623252401982
4.09090909090909 -0.611436401536348
4.19191919191919 -0.59984441605813
4.29292929292929 -0.588472198534502
4.39393939393939 -0.577315582470108
4.49494949494949 -0.566370480360492
4.5959595959596 -0.55563288219471
4.6969696969697 -0.545098853985971
4.7979797979798 -0.534764536330503
4.8989898989899 -0.524626142993373
5 -0.514679959521482
};
\end{axis}

\end{tikzpicture}\\
      Exponential
    \end{minipage}
    \caption{Regression results with corresponding prior}
    \label{fig:regression}
  \end{subfigure}
  \definecolor{pred}{HTML}{FFE5CF}
  \definecolor{post}{HTML}{DBD0C3}
  \caption{{\bf Left:}~Gram/covariance matrices (for $\kappa(x,x')$) reflecting different prior beliefs on how infinite-width NN model functions ought to behave outside the observed data. {\bf Right:}~1D regression results corresponding to the model induced by the priors, showing the posterior \protect\tikz[baseline]\protect\node[fill=post,minimum width=14pt,minimum height=6pt,anchor=base,yshift=3pt,rounded corners=1pt,draw=post!70!black!30!]{}; and predictive \protect\tikz[baseline]\protect\node[fill=pred,minimum width=14pt,minimum height=6pt,anchor=base,yshift=3pt,rounded corners=1pt,draw=pred!70!black!30!]{}; 95\% intervals. See \cref{fig:gram-regression-finite} in appendix for the corresponding finite-width results.}
  \label{fig:gram-regression}
  \vspace*{-1em}
\end{figure}

\section{Encoding Inductive Biases into Infinitely-wide Networks}
\label{sec:background}
We are concerned with the notions of {\em similarity} and {\em bias} in the input space when performing supervised learning. For this, we consider modelling assumptions present in random (untrained) models that induce the {\it a~priori} (before observing data) characteristics in the model. We argue that these assumptions, when combined with data, either reinforce or suppress data biases depending on the prior assumptions.
For analysis, we consider the mean function, $\mu(\cdot)$, and covariance (kernel) function, $\kappa(\cdot,\cdot)$, induced by model functions $f(\vx)$, representing the input--output mapping of a single-layer: $\mu(\vx) \coloneqq \mathrm{E}[f(\vx)]$ and $\kappa(\vx,\vx') \coloneqq \mathrm{E}[(f(\vx)-\mu(\vx))(f(\vx')-\mu(\vx'))^*]$, where $\vx$, $\vx'\in \mathbb{R}^d$ are two inputs and the expectations are taken over model functions.
In probabilistic machine learning, rather than inferring the covariance structure from the expectations over model functions, one typically directly encodes assumptions by choosing a suitable parametric kernel family. However, both approaches, building a model by choosing a form for $f$ or choosing a kernel family, are equivalent.

Specification of prior knowledge through the covariance function and doing inference under this model, is at the core of Gaussian process (GP) models \cite{Rasmussen+Williams:2006}. These models admit the form of a prior $f(\vx) \sim \GP(\mu(\vx), \kappa(\vx,\vx'))$ and a likelihood (observation) model $\vy \mid \vf\sim\prod_{i=1}^{n} p(y_{i} \mid f(\vx_{i}))$, where the data $\mathcal{D} = \{(\vx_i, y_i)\}_{i=1}^n$ are input--output pairs and $\vx_{i} \in \mathbb{R}^d$. This non-parametric machine learning paradigm covers regression and classification tasks, under which GPs are known to offer a convenient machinery for learning and inference, while offering meaningful uncertainty estimates.

\citet{Neal:1995} showed that an untrained single-layer network converges to a GP in the limit of infinite width. This link is well-understood and generalizes to deep models \cite{matthews2018gaussian}, \cite{lee2018deep}. By placing an i.i.d. Gaussian prior on the weights and biases of a network, the distribution on the output of an untrained network converges to a Gaussian distribution in the limit of infinite width. By application of the multivariate Central Limit Theorem, the joint distribution of outputs for any collection of inputs is multivariate Gaussian, as in a GP, and completely characterized by some kernel function $\kappa(\cdot,\cdot)$. Let $\sigma(\cdot)$ be some non-linear (activation) function, such as the ReLU or sigmoid, and $\vw$ and $b$ be the network weights and biases. Then, the associated kernel for the infinite-width network can be formulated in terms of \cite{Neal:1995,Gal+Ghahramani:2016}:
\begin{equation}\label{eq:covfun-nn}
  \kappa(\vx,\vx') = \int p(\vw)\,p(b)\,\sigma(\vw\T\vx+b)\,\sigma(\vw\T\vx'+b)\,\dd\vw \dd b ,
\end{equation}
where priors $p(\vw)$ and $p(b)$ are chosen suitably. From the formulation in \cref{eq:covfun-nn} one can read that the covariance function corresponding to an untrained (random) single-layer neural network is {\em fully characterized} by the choice of activation function $\sigma(\cdot)$ and the priors on the network weights and biases.

Many of the covariance functions corresponding to commonly used activation functions under Gaussian priors on the weights have closed-form representations. The ArcCos kernel  \cite{cho+saul:2009} covers the ReLU (ArcCos-1) and the step (ArcCos-0) activations, the so-called `neural network kernel' \cite{williams97computing,Rasmussen+Williams:2006} links to sigmoidal activations. Gram matrices $\MK_{ij} = \kappa(x_i,x'_j)$, evaluated for input pairs $x$ and $x'$ over $x,x' \in [-3,3]$, are visualized in \cref{fig:gram} to illustrate these covariance structures. The models induced by the ReLU, step, and sigmoid activations are by nature local, with the concentration of the non-linearity around the origin. As can be seen in the 1D regression results in \cref{fig:regression}, the strong inductive bias in the prior (model) is reflected in the posterior. The ReLU extrapolates by matching the local bias of the data, while the step and sigmoidal saturate outside the data, and priors are clearly sensitive to translations and {\em highly non-stationary}.

\paragraph{Stationarity}
For a {\em stationary} (homogeneous) covariance function the covariance structure of the model functions $f(\vx)$ is the same regardless of the absolute position in the input space, and thus the parametrization can be relaxed to $\kappa(\vx,\vx') \triangleq \kappa(\vx-\vx') = \kappa(\vr)$. 
Stationarity implies translation-invariance, \ie\ the notion of similarity between inputs $\vx$, and $\vx'$ is only a function of their distance $\vx-\vx'$. This implies {\em weak stationarity} under stochastic process theory \cite{papoulis1991probability}, and results in {\em reversion to the prior} outside observed data (see \cref{fig:regression}). For stationary covariance functions, the best-known family is the Mat\'ern \cite{Matern:1960, Stein:1999} family of kernels, which features models with sample functions of varying degrees of differentiability (smoothness):
\begin{equation}\label{eq:matern}
  \kappa_\mathrm{Mat.}(\vx,\vx') = \frac{2^{1-\nu}}{\Gamma(\nu)}\left(\sqrt{2\nu}\,\frac{\|\vx-\vx'\|}{\ell}\right)^\nu \mathrm{K}_\nu\!\left(\sqrt{2\nu}\,\frac{\|\vx-\vx'\|}{\ell}\right) , 
\end{equation}
where $\nu$ is a smoothness parameter, $\ell$ a characteristic length-scale parameter, $\mathrm{K}_\nu(\cdot)$ the modified Bessel function, and $\Gamma(\cdot)$ the gamma function. For the Mat\'ern class, the processes 
are $\lceil \nu \rceil -1$ times mean-square differentiable, and the family includes the RBF (squared exponential) and the exponential (Ornstein--Uhlenbeck) kernels as limiting cases as $\nu\to\infty$ and $\nu=\nicefrac{1}{2}$.

\paragraph{Local stationarity} Neural networks do not naturally exhibit stationarity. Formally, this stems from the problem of representing non-linearities over an infinite input domain with a finite set of local non-linear mappings. However, to bridge the gap between neural networks and widely-used stationary kernels, the typical approach is to restrict the invariance to be {\em local} (see \cref{fig:gram}). This is the approach in the RBF-NN~\cite{williams1998computation,Rasmussen+Williams:2006} and general Mat\'ern-NN~\cite{meronen2020stationary} activation functions, where the prior is a composite covariance function with a Gaussian decay envelope (see discussion in \cite{genton2001classes}), \ie,
\begin{equation}\label{eq:Mat-NN}
  \kappa_\textrm{Mat-NN}(\vx,\vx') \propto \exp(-\vx{}\T\vx/2\sigma_\mathrm{m}^2)\,\kappa_\textrm{Mat.}(\vx,\vx')\exp(-{\vx'}{}\T{\vx'}/2\sigma_\mathrm{m}^2) ,
\end{equation}
where $\sigma_\mathrm{m}^2 = 2\sigma_\mathrm{b}^2 + \ell^2$ (see \cref{fig:gram} for the decay effect of the envelope). One motivation for this approach is to assume a Gaussian input density on the training/testing inputs, thus restricting our interest only to inputs in the vicinity of the training data. The local behaviour is highlighted in \cref{fig:regression}, where the locally stationary models revert to the mean (as expected), but the marginal uncertainty of the model functions drops to zero when far enough from training samples. This motivates us to seek activation functions that induce (globally) stationary behaviour (bottom-rows in \cref{fig:gram-regression}).

\section{Global Stationarity-inducing Activation Functions}
\label{sec:global-stationarity}
We derive a direct connection between (fully) stationary covariance functions and periodic activation functions. Our main result leverages the spectral duality of the spectral density of stationary covariance functions to establish a direct connection between the weight prior and the induced prior on the function space. Furthermore, we show that this connection is not restricted to the (sinusoidal) Fourier basis and establishes an equivalence relationship between a Student-$t$ distribution on the weights and a prior on the function space with Mat\'ern covariance structure.

\subsection{Spectral Duality Under Harmonic Functions}
For stationarity-inducing priors, the covariance function can be written equivalently in terms of its spectral density function. This stems from \emph{Bochner's theorem} (see \cite{DaPrato:1992} and 36A in \cite{loomis1953introduction} for a more general formulation) which states a bounded continuous positive definite function $\kappa(\vr)$ can be represented through the following norm-preserving isomorphism
\begin{equation}\label{eq:bochner}
  \kappa(\vr) = \int 
       \omega(\vr) \, \mu(\!\dd \vomega),
\end{equation}
where $\mu$ is a positive measure and $\vr \in \mathbb{R}^d$. If the measure $\mu(\vomega)$ has a density, it is called the \emph{spectral density} $S(\vectb{\omega})$ corresponding to the covariance function $\kappa(\vect{r})$. With $\omega(\vr)=(2\pi)^{-d}\exp(\imag \, \vomega\T \vect{r})$, this gives rise to the Fourier duality of covariance and spectral density, which is known as the \emph{Wiener--Khinchin theorem} (\eg, \cite{Rasmussen+Williams:2006}):
\begin{equation} \label{eq:duality}
  \kappa(\vr) = \frac{1}{(2\pi)^d} \int S(\vomega) \,
       e^{ \imag \, \vomega\T \vect{r}} \dd \vomega
  \quad \text{and} \quad
  S(\vomega) = \int \kappa(\vr) \,
       e^{ -\imag \, \vomega\T \vr} \dd \vr.
\end{equation}
For $d>1$, if the covariance function is \emph{isotropic} (it only depends on the Euclidean norm $\norm{\vr}$ such that $\kappa(r) \triangleq \kappa(\norm{\vr})$), then it is invariant to all rigid motions of the input space. Moreover, the spectral density will also only depend on the norm of the dual input variable $\vomega$. In the neural network case, we can assume the previous layer will take care of scaling the inputs, and thus we effectively are interested in the isotropic case, which brings us to analysing 1D projections.

An important question is whether we are restricted to the (sinusoidal) Fourier basis. Let $\psi_j(x)$ be any sequence of bounded, square-integrable (over its period) periodic functions, such that these functions define an inner product space (generalising a basis of a vector space). In signal processing terms, they define a {\em frame} that provides a redundant yet stable way of representing a signal. From a generalised harmonic perspective \cite{wiener1930generalized}, we could resort to any convenient basis, such as Gabor wavelets \cite{gabor1946theory}. However, in this work we seek to choose $\psi(x)$ suitably, such that we have uniform convergence to $\omega(r)$ (\cf, \cref{eq:bochner}), ensuring that we retain the spectral density $S(\omega)$.

\begin{figure}
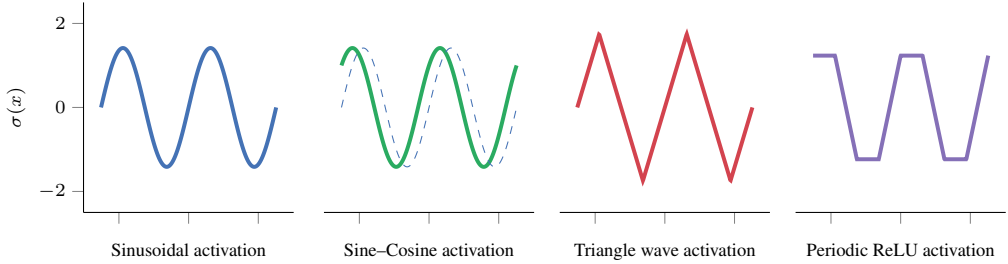

  \centering\scriptsize
  \begin{subfigure}{\textwidth}

  \pgfplotsset{axis on top,scale only axis,width=\figurewidth,height=\figureheight,
    xticklabel={\empty}, 
    xtick align=outside,
    ytick align=outside,  
    axis x line*= bottom, axis y line*= left}
  \setlength{\figurewidth}{.2\textwidth}
  \setlength{\figureheight}{\figurewidth}  
  \begin{minipage}[t]{.3\textwidth}
    \raggedleft
    \pgfplotsset{xlabel={Sinusoidal activation\vphantom{g}}}
    \input{img/sin_activation.tex}
  \end{minipage}
  \pgfplotsset{axis y line=none}
  \begin{minipage}[t]{.22\textwidth}
    \raggedleft
    \pgfplotsset{xlabel={Sine--Cosine activation\vphantom{g}}}
    \input{img/sincosactivation.tex}
  \end{minipage}
  \begin{minipage}[t]{.22\textwidth}
    \raggedleft
    \pgfplotsset{xlabel={Triangle wave activation}}
    \input{img/triangle_activation.tex}
  \end{minipage}
  \begin{minipage}[t]{.22\textwidth}
    \raggedleft
    \pgfplotsset{xlabel={Periodic ReLU activation\vphantom{g}}}    
    \input{img/relu_activation.tex}
  \end{minipage}
  \end{subfigure}
  \caption{Sketches of types of periodic activations functions $\sigma(\cdot)$ visualized over two repeating periods each. The two leftmost are infinitely differentiable and the two rightmost are piecewise linear. The dashed line overlays the Sinusoidal activation for comparison.}
  \label{fig:activations}
  \vspace*{-1em}
\end{figure}

\subsection{Types of Periodic Activation Functions}
We show how various periodic activation functions (see \cref{fig:activations}) under the view of \cref{eq:covfun-nn} directly link back to the spectral duality. These periodic activation functions are continuous, bounded, and centered at zero, while not necessarily differentiable and never monotonic over $\R$ (but monotonic within half-period).

\paragraph{Sinusoidal activation}
By assuming the activation function to be a sinusoid, $\sigma_{\sin}(x) = \sqrt{2}\sin(x)$ (see \cref{fig:activations}), placing a uniform prior on the bias term, $p(b) = \Uni(-\pi,\pi)$,
and substituting this into \cref{eq:covfun-nn}, we can obtain an expression for the covariance function $\kappa(x,x')$:
\begin{equation}
  \kappa(x,x') = \int p(w) \int_{-\pi}^{\pi} \frac{1}{\pi}\sin(w x+b)\sin(w x'+b) \dd b \dd w .
\end{equation}
By solving the inner integral and applying Euler's formula ($\cos(z) = \frac{1}{2} (e^{\imag z} + e^{-\imag z})$), we obtain:
\begin{equation}
  \kappa(x,x') = \frac{1}{2} \int p(w)  e^{\imag w (x- x')} \dd w + \frac{1}{2} \int p(w)e^{-\imag w (x - x')} \dd w ,
\end{equation}
which simplifies under the assumption that the prior on the weight is symmetric and has support on the real line, \ie,
\begin{equation}
  \kappa(x,x') = \int p(w)  e^{\imag w (x- x')} \dd w .
\end{equation}
By letting $r = x-x'$, we find that we recover the spectral density decomposition of a stationary process given by the Wiener--Khinchin theorem in \cref{eq:duality}, where $p(w) = \frac{1}{2\pi}S(w)$. Therefore, we obtain a direct connection between the prior on the weights and the spectral density. A detailed derivation can be found in \cref{app:sin_activation}.

\paragraph{Sine--Cosine activation}
An alternative way of using sinusoidal activation functions is the use of a composite sine--cosine activation, \ie,
  $\sigma_{\sin\cos}(x) = \sin(x) + \cos(x)$ (\cref{fig:activations} compares the sin--cos to the sinusoidal activation).
Using such construction has the benefit of removing the need to integrate over the bias, which can result in reduced variance \cite{SutherlandS15} of the estimates, giving us a covariance function in the form of:
\begin{align}
  \kappa(x,x') &= \int p(w)\, \left[\sin(wx) + \cos(wx) \right]\left[\sin(wx') + \cos(wx') \right] \dd w \nonumber \\
  &= \int p(w)\, \sin(w (x+x')) \dd w + \int p(w)\, \cos(w (x-x')) \dd w .
\end{align}
By application of Euler's formula and under the usual assumption that $p(w)$ is symmetric and has support on the entire real line, the above reduces to:
\begin{align}
  \kappa(x,x') &= \frac{1}{2} \left[ \int p(w)\, e^{\imag w(x-x')}\dd w + \int p(w)\, e^{-\imag w(x-x')}\dd w \right] 
  = \int p(w)\, e^{\imag w(x-x')}\dd w  .
\end{align}
Again, by letting $r = x-x'$, we find that we recover the spectral density decomposition of a stationary process given by the Wiener--Khinchin theorem. A detailed derivation can be found in \cref{app:sin_cos_activation}.

\paragraph{Triangle wave activation}
Inspired by the success of piecewise linear activation functions (\eg, ReLU~\cite{NairH10}, leaky ReLU~\cite{tsuchida2018invariance}, PReLU~\cite{HeZRS15}), we show that the triangle wave, a periodic piecewise linear function, can be used instead of  sinusoidal activations. The triangle wave is given by:
\begin{equation}
  \psi(x) = \frac{4}{p} \bigg( x - \frac{p}{2} \bigg\lfloor \frac{2x}{p} + \frac{1}{2} \bigg\rfloor \bigg) (-1)^{\lfloor \frac{2x}{p} + \frac{1}{2} \rfloor} ,
\end{equation}
where $p$ is the period  and $\lfloor\bullet\rfloor$ denotes the floor function.
By considering a period of $p = 2\pi$, assuming a uniform prior on the biases, $b \sim \Uni(-\pi,\pi)$, and appropriate rescaling, we get the activation
\begin{equation}
	\sigma_{\text{triangle}}(x) = \frac{\pi}{2\sqrt{2}} \bigg( x - \pi \bigg\lfloor \frac{x}{\pi} + \frac{1}{2} \bigg\rfloor \bigg) (-1)^{\lfloor \frac{x}{\pi} +\frac{1}{2} \rfloor},
\end{equation}
for which we obtain a direct correspondence of the network weight priors to the spectral density. As shown in the derivations in \cref{app:triangle_wave}, we find that we again recover the spectral density decomposition of a stationary process given by the Wiener--Khinchin theorem. 

\paragraph{Periodic ReLU activation}
To go even further, we define a piecewise linear periodic activation function with a repeating rectified linear pattern, which we call the periodic ReLU function. It can be defined in terms of the sum of two triangle waves, with the second one being shifted by \nicefrac{1}{4} of a period. The resulting periodic function is piecewise linear and defined as:
\begin{equation}
  \psi(x) = \frac{8}{\pi^2} \bigg( \bigg((x+\frac{\pi}{2}) - \pi \bigg\lfloor \frac{(x+\frac{\pi}{2})}{\pi} + \frac{1}{2} \bigg\rfloor\bigg) (-1)^{\lfloor \frac{(x+\frac{\pi}{2})}{\pi} + \frac{1}{2} \rfloor} +  \bigg(x - \pi \bigg\lfloor \frac{x}{\pi} + \frac{1}{2} \bigg\rfloor\bigg) (-1)^{\lfloor \frac{x}{\pi} + \frac{1}{2} \rfloor} \bigg) 
	,
\end{equation}
assuming a period of $p = 2 \pi$.
It is again possible to obtain a correspondence between the network weight prior and the spectral density by considering a weighted version of the equation above.
In particular, we show in \cref{app:periodic_relu} that from the following weighted periodic ReLU activation function, 
\begin{equation}
  \sigma_{\text{pReLU}}(x) = \frac{\pi}{4} \bigg( \bigg((x+\frac{\pi}{2}) - \pi \bigg\lfloor \frac{(x+\frac{\pi}{2})}{\pi} + \frac{1}{2} \bigg\rfloor\bigg) (-1)^{\lfloor \frac{(x+\frac{\pi}{2})}{\pi} + \frac{1}{2} \rfloor} +  \bigg(x - \pi \bigg\lfloor \frac{x}{\pi} + \frac{1}{2} \bigg\rfloor\bigg) (-1)^{\lfloor \frac{x}{\pi} + \frac{1}{2} \rfloor} \bigg)  ,
\end{equation}
we again recover the spectral density decomposition of a stationary process given by the Wiener--Khinchin theorem, providing a direct connection between the prior on the weights and the spectral density.
Note that, choosing a piecewise linear periodic activation function has potential computational benefits compared to sinusoidal activation functions and can help prevent vanishing gradients.

\begin{figure}[t]
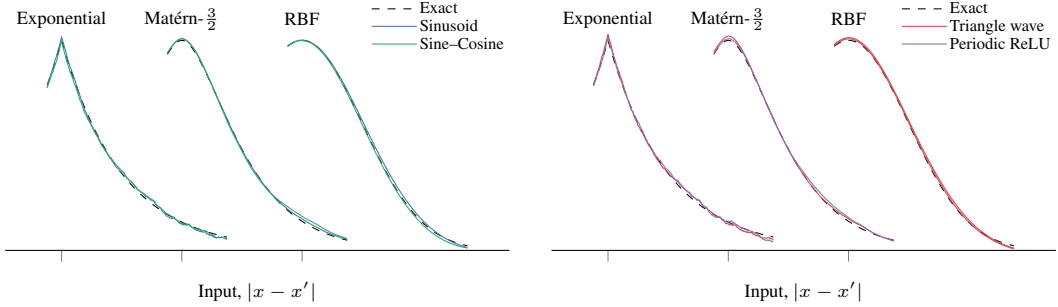

  \centering\scriptsize
    \pgfplotsset{axis on top,scale only axis,width=\figurewidth,height=\figureheight,
      xticklabel={\empty}, 
      xtick align=outside,
      ytick align=outside,  
      xtick={25,225,425},
      axis x line*= bottom, axis y line=none}

    \pgfplotsset{legend style={legend cell align=left,align=left, inner xsep=1pt, inner ysep=1pt, row sep=0pt},legend style={at={(1.00,1.00)},anchor=north east},legend style={rounded corners=1pt,draw=black},y tick label style={rotate=90}, legend style={nodes={scale=0.9, transform shape}}}  
    \setlength{\figurewidth}{.48\textwidth}
    \setlength{\figureheight}{.5\figurewidth}  
    \begin{minipage}{.48\textwidth}
      \input{img/kernels_p1.tex}
    \end{minipage}
    \hfill
    \begin{minipage}{.48\textwidth}
      \input{img/kernels_p2.tex}
    \end{minipage}\\[-2em]
  \caption{Covariance functions calculated by MC integration with 5000 samples and compared to their exact dashed counterparts (peaks shifted for clarity). Regardless of the type of periodic activation function used, we can recover the behaviour of the stationary kernels (see also \cref{fig:gram-finite}).}
  \label{fig:activations}
  \vspace*{-1em}  
\end{figure}

\subsection{Kernel Functions}
\label{sec:kernel-func}
We have established that is it possible to obtain a direct correspondence of the prior on the weights and the spectral density of a stationary covariance function by using periodic activation functions in random neural networks.
In \cref{app:Matern} we show that by placing a Student-$t$ distribution on the weights with degrees of freedom of $u = 2\nu$ we recover the spectral density of the Mat\'ern family, \ie,
\begin{equation}
p(w) =\frac{\Gamma(\frac{u+1}{2})}{\sqrt{u\pi}\Gamma(\frac{u}{2})} \left(1+\frac{w^2}{u}\right)^{-\frac{u+1}{2}} \!\!\!= \frac{1}{2\pi}2\sqrt{\pi}\frac{\Gamma(\nu+\frac{1}{2})}{\Gamma(\nu)}(2\nu)^{\nu}\left( 2\nu+w^2\right)^{-(\nu+\frac{1}{2})} = \frac{1}{2\pi}S_{\text{Mat.}}(w)  ,
\end{equation}
where $p(w)$ denotes the probability density function of a Student-$t$ distribution and $S_{\text{Mat.}}(w)$ denotes the spectral density of the Mat\'ern family. 
This means that a Student-$t$ prior on the weights in combination with an appropriately scaled periodic activation function corresponds directly to a prior in the function space with Mat\'ern covariance structure.
\cref{fig:activations} verifies this result and shows that we recover the exact covariance function (dashed) for various examples from the Mat\'ern family by Monte Carlo (MC) integration (5000 samples) with all of the discussed periodic activation functions.

Recall that the Mat\'ern family contains the exponential kernel ($\nu = \nicefrac{1}{2}$) and the RBF Kernel ($\nu = \infty$) as special cases. Similarly, the Student-$t$ distribution has the Cauchy ($u = 1$) and the Normal distribution ($u = \infty$) as limiting cases, resulting in correspondence to the respective special cases in the Mat\'ern family. \cref{tab:priors} in \cref{app:Matern} summarises the set of priors on the weights that correspond to the spectral density of kernel functions in the Mat\'ern family. This correspondence is exact for the sinusoidal and sine--cosine activations. For the triangle wave and periodic ReLU activations, the obtained correspondence is approximate, and the introduced approximation error is analyzed in detail in \cref{app:derivations}.

\begin{table}[t!]
  \caption{Examples of UCI regression tasks, showing the globally stationary NN model directly gives competitive RMSE and for most data sets also improved mean negative log predictive density (NLPD) compared to non-stationary and locally stationary NN models. KFAC Laplace was used as the inference method. High std is due to variability in 10-fold-cv splits of the small data sets, not method performance.}  
  \label{tbl:benchmarks}
  \scriptsize
  \tiny
  \setlength{\tabcolsep}{0pt}
  \setlength{\tblw}{0.09\textwidth}

  \begin{subfigure}{\textwidth}

  \label{tbl:benchmarks_reg}

  \setlength{\tabcolsep}{0pt}
  \setlength{\tblw}{0.09\textwidth}  
  \begin{tabularx}{\columnwidth}{l @{\extracolsep{\fill}} C{\tblw}  C{\tblw} C{\tblw} C{\tblw} C{\tblw}  C{\tblw} C{\tblw} C{\tblw}}
  \toprule

%NLPD and RMSE (this table has been automatically generated, do not alter manually)
                          &        \multicolumn{2}{c}{\sc boston}         &       \multicolumn{2}{c}{\sc concrete}        &        \multicolumn{2}{c}{\sc airfoil}        &       \multicolumn{2}{c}{\sc elevators}       \\
($n$, $d$)                &         \multicolumn{2}{c}{(506, 12)}         &         \multicolumn{2}{c}{(1030, 5)}         &         \multicolumn{2}{c}{(1503, 5)}         &        \multicolumn{2}{c}{(16599, 18)}        \\
($c$, $n_\textrm{batch}$) &          \multicolumn{2}{c}{(1, 50)}          &          \multicolumn{2}{c}{(1, 50)}          &          \multicolumn{2}{c}{(1, 50)}          &         \multicolumn{2}{c}{(1, 500)}          \\
\midrule
                          &         NLPD          &         RMSE          &         NLPD          &         RMSE          &         NLPD          &         RMSE          &         NLPD          &         RMSE          \\
\midrule
ReLU                      &    $0.51{\pm}0.32$    &    $0.37{\pm}0.07$    &    $0.78{\pm}0.16$    &    $0.48{\pm}0.04$    &    $0.51{\pm}0.53$    &    $0.41{\pm}0.21$    &    $0.38{\pm}0.03$    &    $0.35{\pm}0.01$    \\
loc RBF                   &    $0.52{\pm}0.30$    &    $0.37{\pm}0.08$    &    $0.78{\pm}0.22$    &    $0.44{\pm}0.05$    &    $0.10{\pm}0.15$    &    $0.26{\pm}0.03$    &    $0.41{\pm}0.04$    &  $\bf 0.35{\pm}0.01$  \\
glob RBF (sin)            &    $0.42{\pm}0.34$    &    $0.36{\pm}0.07$    &    $0.74{\pm}0.15$    &    $0.49{\pm}0.05$    &    $0.14{\pm}0.17$    &    $0.29{\pm}0.05$    &    $0.38{\pm}0.03$    &    $0.35{\pm}0.01$    \\
glob RBF (prelu)          &    $0.39{\pm}0.30$    &  $\bf 0.36{\pm}0.07$  &    $0.74{\pm}0.14$    &    $0.49{\pm}0.04$    &  $\bf 0.05{\pm}0.12$  &  $\bf 0.26{\pm}0.03$  &    $0.74{\pm}0.73$    &    $0.46{\pm}0.21$    \\
loc Mat-3/2               &    $0.71{\pm}0.38$    &    $0.40{\pm}0.08$    &    $0.84{\pm}0.28$    &  $\bf 0.42{\pm}0.04$  &    $0.11{\pm}0.18$    &    $0.26{\pm}0.03$    &    $0.43{\pm}0.04$    &    $0.35{\pm}0.01$    \\
glob Mat-3/2 (sin)        &    $0.43{\pm}0.27$    &    $0.39{\pm}0.08$    &    $0.73{\pm}0.16$    &    $0.49{\pm}0.05$    &    $0.07{\pm}0.15$    &    $0.27{\pm}0.03$    &  $\bf 0.37{\pm}0.03$  &    $0.35{\pm}0.01$    \\
glob Mat-3/2 (prelu)      &  $\bf 0.38{\pm}0.22$  &    $0.38{\pm}0.08$    &  $\bf 0.72{\pm}0.18$  &    $0.48{\pm}0.05$    &    $0.08{\pm}0.12$    &    $0.27{\pm}0.03$    &    $0.39{\pm}0.03$    &    $0.36{\pm}0.01$    \\

  \bottomrule
  \end{tabularx}
  \end{subfigure}
\end{table}

\section{Experiments}\label{sec:experiments}

To assess the performance of stationarity-inducing activation functions, we compared the in-domain and out-of-domain predictive uncertainty to non-stationary and locally stationary models on various benchmark data sets.
In all experiments, the compared models differ only by the activation function and the respective weight priors of the last hidden layer of the network. Replacing the prior effects the initialization of the NN weights and the training objective, as we maximize the log joint density. A detailed description of the experimental setup is available in \cref{app:details}. The illustrative toy BNN examples are implemented using {Turing.jl}~\cite{Ge2018}, GP regression results use GPflow~\cite{GPflow:2017}, and all other experiments are implemented using {PyTorch}~\cite{PyTorch}.

\begin{figure}[b]
  \centering\scriptsize
  \setlength{\figurewidth}{.26\textwidth}
  \setlength{\figureheight}{.75\figurewidth}
  \pgfplotsset{scale only axis,y tick label style={rotate=90},
    xtick={0,90,180,270,360},
    xticklabels={$0^\circ$,$90^\circ$,$180^\circ$,$270^\circ$,$360^\circ$}}
  \begin{subfigure}[t]{.33\textwidth}
    \centering
    \pgfplotsset{ylabel={Accuracy}}
    % This file was created by tikzplotlib v0.9.8.
\begin{tikzpicture}

\definecolor{color0}{rgb}{0,0.75,0.75}

\begin{axis}[
height=\figureheight,
legend cell align={left},
legend style={
  fill opacity=0.8,
  draw opacity=1,
  text opacity=1,
  at={(0.5,0.91)},
  anchor=north,
  draw=white!80!black
},
tick align=outside,
tick pos=left,
width=\figurewidth,
x grid style={white!69.0196078431373!black},
xlabel={Rotation angle},
xmin=-18, xmax=378,
xtick style={color=black},
y grid style={white!69.0196078431373!black},
ymin=0.068925, ymax=1.034175,
ytick style={color=black}
]
\addplot [semithick, mycolor0]
table {%
0 0.9892
10 0.9767
20 0.9247
30 0.7596
40 0.5214
50 0.3456
60 0.2285
70 0.1657
80 0.1555
90 0.1679
100 0.1804
110 0.2023
120 0.2274
130 0.2688
140 0.3121
150 0.3622
160 0.4064
170 0.432
180 0.4567
190 0.447
200 0.421
210 0.3736
220 0.3067
230 0.2357
240 0.1924
250 0.1691
260 0.1691
270 0.1707
280 0.1675
290 0.1747
300 0.2212
310 0.3326
320 0.5076
330 0.747
340 0.9135
350 0.9769
360 0.9892
};
%\addlegendentry{ReLU}
\label{plt:relu}
\addplot [semithick, mycolor1]
table {%
0 0.9869
10 0.9705
20 0.9073
30 0.7529
40 0.5175
50 0.3309
60 0.2111
70 0.1471
80 0.1398
90 0.1413
100 0.1498
110 0.1669
120 0.1978
130 0.2595
140 0.3299
150 0.4027
160 0.4504
170 0.4749
180 0.4909
190 0.4704
200 0.4359
210 0.3698
220 0.2758
230 0.2168
240 0.1787
250 0.1584
260 0.1494
270 0.1433
280 0.1304
290 0.1449
300 0.2069
310 0.3524
320 0.5615
330 0.7975
340 0.9276
350 0.976
360 0.9869
};
%\addlegendentry{local RBF}
\label{plt:locrbf}
\addplot [semithick, mycolor2]
table {%
0 0.9903
10 0.9755
20 0.9245
30 0.7868
40 0.5515
50 0.355
60 0.2207
70 0.1539
80 0.1378
90 0.1332
100 0.1418
110 0.1718
120 0.2158
130 0.2711
140 0.3202
150 0.3695
160 0.402
170 0.4224
180 0.4515
190 0.4232
200 0.3916
210 0.3404
220 0.276
230 0.2298
240 0.1945
250 0.1627
260 0.14
270 0.124
280 0.1191
290 0.1359
300 0.195
310 0.3275
320 0.5314
330 0.7676
340 0.9229
350 0.9767
360 0.9903
};
%\addlegendentry{global RBF (sin)}
\label{plt:sin}
\addplot [semithick, mycolor3]
table {%
0 0.9897
10 0.9763
20 0.9164
30 0.7767
40 0.5314
50 0.3448
60 0.218
70 0.1574
80 0.1388
90 0.1383
100 0.1499
110 0.1797
120 0.2228
130 0.271
140 0.3191
150 0.3705
160 0.4173
170 0.445
180 0.4718
190 0.4575
200 0.4201
210 0.358
220 0.2749
230 0.2168
240 0.1731
250 0.1391
260 0.1235
270 0.1188
280 0.1128
290 0.1311
300 0.1964
310 0.3225
320 0.5135
330 0.7563
340 0.9176
350 0.9764
360 0.9897
};
%\addlegendentry{global RBF (sincos)}
\label{plt:sincos}

\end{axis}

\end{tikzpicture}
  \end{subfigure}
  \hfill
  \begin{subfigure}[t]{.33\textwidth}
    \centering
    \pgfplotsset{ylabel={Mean confidence}}
    % This file was created by tikzplotlib v0.9.8.
\begin{tikzpicture}

\definecolor{color0}{rgb}{0,0.75,0.75}

\begin{axis}[
height=\figureheight,
legend cell align={left},
legend style={
  fill opacity=0.8,
  draw opacity=1,
  text opacity=1,
  at={(0.5,0.91)},
  anchor=north,
  draw=white!80!black
},
tick align=outside,
tick pos=left,
width=\figurewidth,
x grid style={white!69.0196078431373!black},
xlabel={Rotation angle},
xmin=-18, xmax=378,
xtick style={color=black},
y grid style={white!69.0196078431373!black},
ymin=0.512388862028329, ymax=1.00652925617658,
ytick style={color=black}
]
\addplot [semithick, mycolor0]
table {%
0 0.984068329169837
10 0.966521468506187
20 0.91783526157386
30 0.828161127217575
40 0.750341363780955
50 0.718219261976443
60 0.706227720386081
70 0.711236307722443
80 0.721514782318743
90 0.754984715961048
100 0.743562840642788
110 0.744764666573623
120 0.741749172343267
130 0.74272298530097
140 0.757907341829815
150 0.787403896601699
160 0.820955436722585
170 0.849403350794013
180 0.869150217185363
190 0.841203803075071
200 0.809517415721511
210 0.781863681865864
220 0.746798541865626
230 0.733589482792771
240 0.727950931215205
250 0.727665711957863
260 0.724756847989657
270 0.741240734015841
280 0.713701678881356
290 0.711491846584462
300 0.71078372036255
310 0.720406602363406
320 0.761759861602808
330 0.836164341028648
340 0.915834770859647
350 0.966097811882515
360 0.98395844158303
};
%\addlegendentry{ReLU}
\addplot [semithick, mycolor1]
table {%
0 0.958259915705385
10 0.917976521172311
20 0.849106507169721
30 0.743887432017347
40 0.648927010273931
50 0.613432653628408
60 0.600794689179102
70 0.58152295936688
80 0.56033808234537
90 0.567269015953904
100 0.546321197089067
110 0.554620928823511
120 0.576439352674568
130 0.601070604406425
140 0.635314624532101
150 0.685998004134381
160 0.739057817201999
170 0.786523710022735
180 0.815711104998899
190 0.77552136848401
200 0.733506855095494
210 0.689623261747073
220 0.640524120025762
230 0.617504932206942
240 0.599113527733617
250 0.582765491719736
260 0.566676234765494
270 0.56715183604012
280 0.534849789035067
290 0.53651337653024
300 0.556903131681167
310 0.590750159749451
320 0.647930553979521
330 0.755912434726358
340 0.859508921477645
350 0.930154487436159
360 0.957820053192658
};
%\addlegendentry{local RBF}
\addplot [semithick, mycolor2]
table {%
0 0.973383692028658
10 0.945022857596868
20 0.890994889299596
30 0.794705293800362
40 0.694878054518462
50 0.649009576543292
60 0.630236547252261
70 0.617915193784712
80 0.603844115253948
90 0.6046234597482
100 0.606850666209523
110 0.617453586467781
120 0.635720877210883
130 0.657725232092645
140 0.681561614948893
150 0.713140756411046
160 0.749779914241938
170 0.785300642438991
180 0.811221573935527
190 0.79421019414012
200 0.763818978328421
210 0.735397130745452
220 0.696602999914795
230 0.671834105906334
240 0.652180165262424
250 0.642605713615368
260 0.629458357295745
270 0.617328443803396
280 0.587710048868399
290 0.576086276085613
300 0.578464377715147
310 0.61075142851317
320 0.674523341997117
330 0.779489288765921
340 0.88148299557544
350 0.947015585829072
360 0.973411464927044
};
%\addlegendentry{global RBF (sin)}
\addplot [semithick, mycolor3]
table {%
0 0.972183020710178
10 0.9416555846991
20 0.884953411105809
30 0.785724010062016
40 0.692544744391847
50 0.653932900114844
60 0.636762352048403
70 0.609833313667315
80 0.581242099310882
90 0.574802136328446
100 0.573009610463592
110 0.580031819964382
120 0.613669566079524
130 0.646722352378602
140 0.678210227458184
150 0.719897068651972
160 0.765496381694334
170 0.798406807657693
180 0.829990317673866
190 0.80140678133945
200 0.766745819143833
210 0.729739661839142
220 0.692481587099175
230 0.665876531434043
240 0.639691893173463
250 0.613822446669342
260 0.587640989021099
270 0.579782876134606
280 0.566606796467056
290 0.567538316273252
300 0.584844004962637
310 0.61838945913941
320 0.672568548142038
330 0.766613567712507
340 0.871872972824346
350 0.944254319094574
360 0.97225519011973
};
%\addlegendentry{global RBF (sincos)}
\end{axis}

\end{tikzpicture}  
  \end{subfigure}  
  \hfill
  \begin{subfigure}[t]{.33\textwidth}
    \centering
    \pgfplotsset{ylabel={NLPD}}
    % This file was created by tikzplotlib v0.9.8.
\begin{tikzpicture}

\definecolor{color0}{rgb}{0,0.75,0.75}

\begin{axis}[
height=\figureheight,
legend cell align={left},
legend style={
  fill opacity=0.8,
  draw opacity=1,
  text opacity=1,
  at={(0.5,0.09)},
  anchor=south,
  draw=white!80!black
},
tick align=outside,
tick pos=left,
width=\figurewidth,
x grid style={white!69.0196078431373!black},
xlabel={Rotation angle},
xmin=-18, xmax=378,
xtick style={color=black},
y grid style={white!69.0196078431373!black},
ymin=-0.308801740942441, ymax=7.30017719033611,
ytick style={color=black}
]
\addplot [semithick, mycolor0]
table {%
0 0.0370609377520388
10 0.077336778075992
20 0.24846223731164
30 0.829031744320276
40 1.99616366244491
50 3.37950969060473
60 4.66984557212764
70 5.78052871021982
80 6.47419683178307
90 6.95431451164163
100 6.86651423246632
110 6.68328399745368
120 6.2043103969384
130 5.74391359296388
140 5.24349968093845
150 4.89296722770501
160 4.74815707661248
170 4.84349524858205
180 4.93982868262083
190 4.71212340075796
200 4.44298270496928
210 4.39590207070205
220 4.5100267700196
230 4.91693023437199
240 5.35801897473976
250 5.75825375530119
260 6.02282891993057
270 6.18042728585904
280 6.05208867109949
290 5.6895789092791
300 4.81217636632488
310 3.56498657207274
320 2.15185529237223
330 0.915080067077576
340 0.288547268319715
350 0.0775604957936011
360 0.0371567733232657
};
%\addlegendentry{ReLU}
\addplot [semithick, mycolor1]
table {%
0 0.0638031638322996
10 0.133829412648976
20 0.321658362852487
30 0.761007806223396
40 1.54837895165075
50 2.29244706224378
60 2.89948183459255
70 3.20006007956548
80 3.21167558068066
90 3.22730960345172
100 3.18703653472016
110 3.16063626593804
120 3.12534376037424
130 2.96944091948812
140 2.76461094514823
150 2.63126056098193
160 2.59090077813178
170 2.60327764456328
180 2.64202677719928
190 2.60538685797967
200 2.60568185278255
210 2.67713405444793
220 2.81945144624391
230 2.94771153283649
240 3.06729424632273
250 3.11046265707284
260 3.0991549913964
270 3.08211019176078
280 3.04600639463576
290 2.91422997732226
300 2.61809374693107
310 2.0390317984838
320 1.29696018424257
330 0.625281120750819
340 0.271830452138431
350 0.111618981220991
360 0.0643450837394918
};
%\addlegendentry{local RBF}
\addplot [semithick, mycolor2]
table {%
0 0.0434822047965909
10 0.0961054996317059
20 0.247230615938501
30 0.643875948823449
40 1.43644730649875
50 2.3042488861293
60 3.00368623562015
70 3.39974087387132
80 3.4737095546295
90 3.49731927818416
100 3.51613926280293
110 3.43546246626798
120 3.3509717269355
130 3.22145350963649
140 3.08491841976214
150 2.97528749325912
160 2.91871535131634
170 2.96443375167719
180 3.07059194960285
190 3.11755024718269
200 3.09933505066528
210 3.14393909336099
220 3.20724211885029
230 3.2750713432129
240 3.336214839031
250 3.38921902589571
260 3.38999632202462
270 3.39940990809313
280 3.39015492689056
290 3.2467686102812
300 2.8548785653196
310 2.21404102066735
320 1.41094360492257
330 0.665866802807915
340 0.256757646947588
350 0.0930735408648269
360 0.0434354069716858
};
%\addlegendentry{global RBF (sin)}
\addplot [semithick, mycolor3]
table {%
0 0.0441401128310133
10 0.098156828567223
20 0.262861422082474
30 0.675578199307758
40 1.49269520828426
50 2.35622403251929
60 3.06792930196772
70 3.46015023551872
80 3.52001392344442
90 3.4929668193202
100 3.41066712109728
110 3.30676952049626
120 3.25461567292126
130 3.19895850609489
140 3.14028281141145
150 3.12685675645441
160 3.12789888428603
170 3.15951285844132
180 3.26459440328963
190 3.23452843554918
200 3.20247929573847
210 3.20879620119175
220 3.25918540899313
230 3.3076839681552
240 3.33304520495258
250 3.36462062166211
260 3.35774770899389
270 3.38865701340212
280 3.42107453331796
290 3.32636104450345
300 2.94912548925957
310 2.3133109746084
320 1.48580762574665
330 0.71620951338062
340 0.276728078247363
350 0.0959071689075344
360 0.0440367720405573
};
%\addlegendentry{global RBF (sincos)}
\end{axis}

\end{tikzpicture}
  \end{subfigure}\\[-1.2em]
  \caption{Rotated MNIST: The models have been trained on unrotated digits. The test-set digits are rotated at test time to show the sensitivity of the trained model to perturbations. All models perform equally in terms of accuracy, while ReLU (\ref{plt:relu}) shows overconfidence in terms of mean confidence and NLPD. The stationary RBF models (\ref{plt:locrbf}~local, \ref{plt:sin}~sin, \ref{plt:sincos}~sin--cos) capture uncertainty.}
  \label{fig:mnist}
\end{figure}
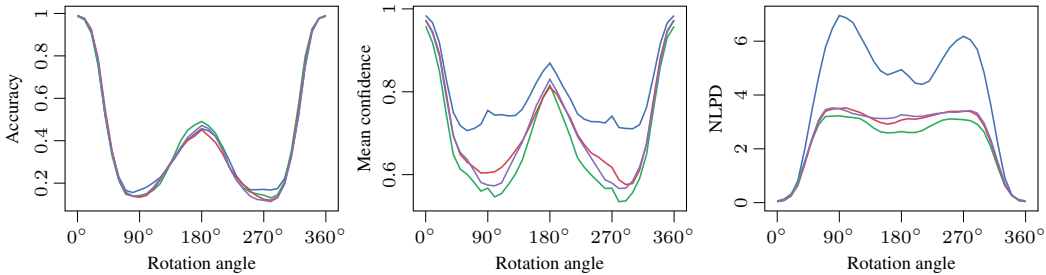

\paragraph{Illustrative toy examples}
\cref{fig:teaser} shows predictive densities for non-stationary, locally stationary, and globally stationary activation functions on the banana classification task. The top row illustrates the predictive densities of infinite-width BNNs (GP), the bottom row shows corresponding results for a single hidden layer BNN using 30 hidden units. We observe that global stationarity-inducing activation functions revert to the prior outside the data, leading to conservative behaviour (high uncertainty) for out-of-domain samples. Moreover, we see that the finite-width BNNs result in similar behaviour to their infinite-width counterpart, while locally stationary activation functions in finite-width BNNs exhibit a slower reversion to the mean than their infinite-width corresponding GPs. Additionally, we include a 1D toy regression study that highlights the differences between different prior assumptions encoded by choice of the activation function. \cref{fig:gram-regression} shows the corresponding prior covariance as well as posterior predictions for the infinite-width (GP) model. We replicated the experiment with a finite-width network and observed that the behaviour translates to finite width (see \cref{fig:gram-regression-finite} in \cref{app:toy}).

\paragraph{UCI benchmarks}
\cref{tbl:benchmarks} shows results on UCI \cite{UCI} regression data sets comparing deep neural networks with ReLU, locally stationary RBF \cite{williams1998computation}, and locally stationary Mat\'ern-$\nicefrac{3}{2}$ \cite{meronen2020stationary} against global stationary models. An extended results table can be found in \cref{app:uci_reg} and additional results on UCI classification tasks in \cref{app:uci_class}. \cref{tbl:benchmarks} lists root mean square error (RMSE) and negative log predictive density (NLPD), which captures the predictive uncertainty, while the RMSE only accounts for the mean. Global stationary models provide better estimates of the target distribution in all cases while obtaining comparable RMSEs. It is important to note that the large standard deviations in \cref{tbl:benchmarks} are due to the small number of data points and the fact that some splits in the 10-fold CV end up being harder than others. The posterior inference is performed using KFAC Laplace \cite{ritter2018a_kfac_laplace}.

\begin{figure}[t]
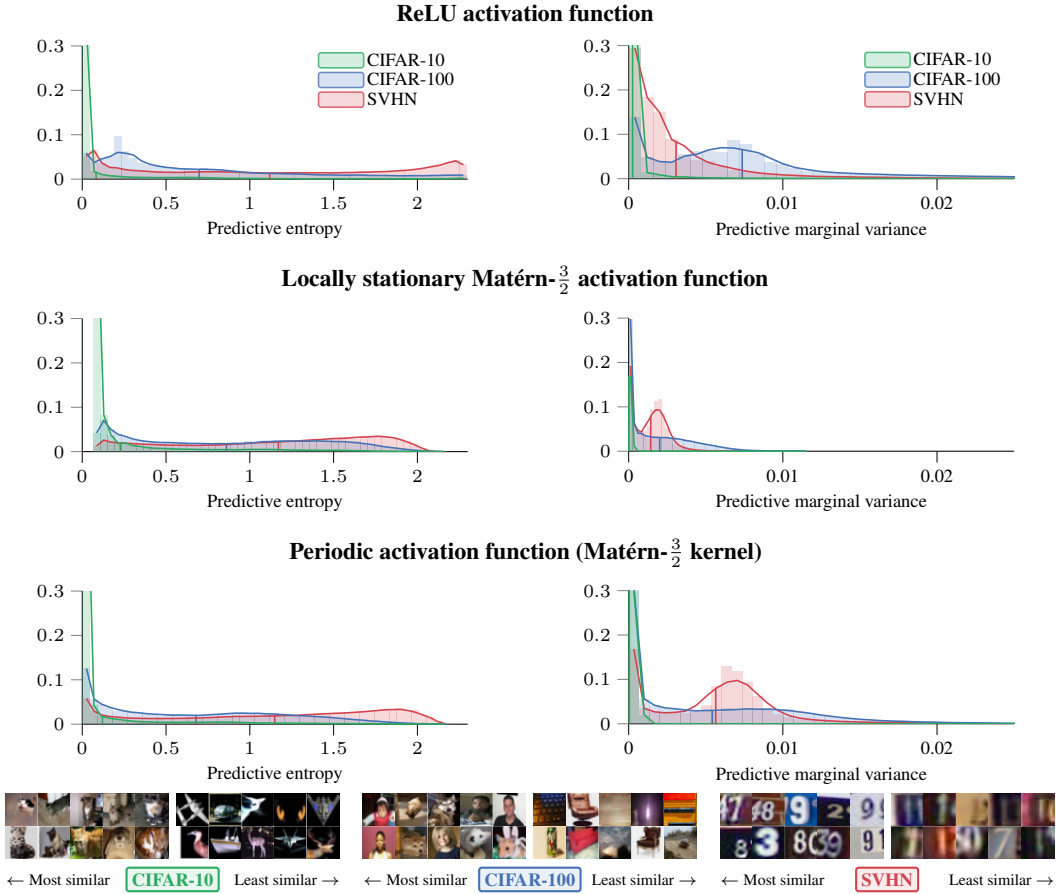


  \newcommand{\plotimages}[5]{%
    \foreach \x [count=\i] in {0,1,2,3,4,5,6,7,8,9} {

        \node[draw=white,fill=black!20,minimum size=\figurewidth,inner sep=0pt]
          (\i) at ({\figurewidth*mod(\i-1,5)},{\figureheight*int((\i-1)/5)})
          {\includegraphics[width=\figurewidth]{./fig/cifarimgs_#3/#2/#1_image_low\x.png}};
        \node[draw=white,fill=black!20,minimum size=\figurewidth,inner sep=0pt]
          (\i) at ({\figurewidth*mod(\i-1,5)+5.25*\figurewidth},{\figureheight*int((\i-1)/5)})
          {\includegraphics[width=\figurewidth]{./fig/cifarimgs_#3/#2/#1_image_high\x.png}};  
    }%
    \node[draw=#4,thick,fill=#4!10,rounded corners=1pt,inner sep=2pt] at ({4.625*\figurewidth},{-1.15*\figurewidth}) {\bf\color{#4} #5};  
    \node at ({1.1*\figurewidth},{-1.15*\figurewidth}) {\tiny $\leftarrow$~Most similar};
    \node at ({8.15*\figurewidth},{-1.15*\figurewidth}) {\tiny Least similar~$\rightarrow$};                
  }

  \centering\scriptsize

  \textbf{\small ReLU activation function}\\
  \vspace*{0.5em}
  \pgfplotsset{every x tick scale label/.style={at={(rel axis cs:0.9,-0.45)},anchor=south west,inner sep=1pt},scaled ticks=false}
  \begin{subfigure}{.48\textwidth}
    \centering
    \setlength{\figurewidth}{\linewidth}
    \setlength{\figureheight}{.5\textwidth}
    \pgfplotsset{xlabel={Predictive entropy}, axis x line*=bottom, axis y line*=left, ymode=log}
    \pgfplotsset{legend image code/.code={\draw[#1,fill=#1,fill opacity=0.2] (0cm,-0.1cm) rectangle (0.6cm,0.1cm);}, legend reversed=true, legend style={inner xsep=1pt, inner ysep=1pt, row sep=0pt, rounded corners=1pt, draw opacity=0.8, draw=white!80!black},}
     \input{./fig/cifarimgs_swag/relu_entropy.tex}
  \end{subfigure}
  \hfill
  \begin{subfigure}{.48\textwidth}
    \centering
    \setlength{\figurewidth}{\linewidth}
    \setlength{\figureheight}{.5\textwidth}
    \pgfplotsset{xlabel={Predictive entropy}, axis x line*=bottom, axis y line*=left, ymode=log}
    \pgfplotsset{legend image code/.code={\draw[#1,fill=#1,fill opacity=0.2] (0cm,-0.1cm) rectangle (0.6cm,0.1cm);}, legend reversed=true, legend style={inner xsep=1pt, inner ysep=1pt, row sep=0pt, rounded corners=1pt, draw opacity=0.8, draw=white!80!black},}
     \pgfplotsset{xlabel={Predictive marginal variance}}
    \input{./fig/cifarimgs_swag/relu_variance.tex}
  \end{subfigure}\\

   \vspace*{1em}
  \textbf{\small Locally stationary Mat\'ern-$\frac32$ activation function}\\
  \vspace*{0.5em}
  \pgfplotsset{every x tick scale label/.style={at={(rel axis cs:0.9,-0.45)},anchor=south west,inner sep=1pt}}
  \begin{subfigure}{.48\textwidth}
    \centering
    \setlength{\figurewidth}{\linewidth}
    \setlength{\figureheight}{.5\textwidth}

    \pgfplotsset{xlabel={Predictive entropy}, axis x line*=bottom, axis y line*=left, ymode=log}
     \input{./fig/cifarimgs_swag/matern32_local_entropy_nolegend.tex}
  \end{subfigure}
  \hfill
  \begin{subfigure}{.48\textwidth}
    \centering
    \setlength{\figurewidth}{\linewidth}
    \setlength{\figureheight}{.5\textwidth}
    \pgfplotsset{xlabel={Predictive entropy}, axis x line*=bottom, axis y line*=left, ymode=log}
     \pgfplotsset{xlabel={Predictive marginal variance}}

    \input{./fig/cifarimgs_swag/matern32_local_variance_nolegend.tex}
  \end{subfigure}\\

 \vspace*{1em}
  \textbf{\small Periodic activation function (Mat\'ern-$\frac32$ kernel)} \\
  \vspace*{0.5em}
  \pgfplotsset{every x tick scale label/.style={at={(rel axis cs:0.9,-0.45)},anchor=south west,inner sep=1pt}}
  \begin{subfigure}{.48\textwidth}
    \centering
    \setlength{\figurewidth}{\linewidth}
    \setlength{\figureheight}{.5\textwidth}
    \pgfplotsset{xlabel={Predictive entropy}, axis x line*=bottom, axis y line*=left, ymode=log}

     \input{./fig/cifarimgs_swag/matern32_sin_entropy_nolegend.tex}
  \end{subfigure}
  \hfill
  \begin{subfigure}{.48\textwidth}
    \centering
    \setlength{\figurewidth}{\linewidth}
    \setlength{\figureheight}{.5\textwidth}
    \pgfplotsset{xlabel={Predictive entropy}, axis x line*=bottom, axis y line*=left, ymode=log}
    \pgfplotsset{xlabel={Predictive marginal variance}}

    \input{./fig/cifarimgs_swag/matern32_sin_variance_nolegend.tex}
  \end{subfigure}\\

  \setlength{\figurewidth}{0.031\textwidth}
  \setlength{\figureheight}{\figurewidth}  
  \begin{subfigure}{.32\textwidth}
    \centering
    \begin{tikzpicture}[outer sep=0]
      \plotimages{cifar10}{matern32_sin}{swag}{mycolor1}{CIFAR-10}
    \end{tikzpicture}
  \end{subfigure}
  \hfill
  \begin{subfigure}{.32\textwidth}
    \centering
    \begin{tikzpicture}[outer sep=0]
      \plotimages{cifar100}{matern32_sin}{swag}{mycolor0}{CIFAR-100}
    \end{tikzpicture}
  \end{subfigure}
  \hfill
  \begin{subfigure}{.32\textwidth}
    \centering
    \begin{tikzpicture}[outer sep=0]as inference method
      \plotimages{svhn}{matern32_sin}{swag}{mycolor2}{SVHN}
    \end{tikzpicture}
  \end{subfigure}
  
  \caption{OOD detection experiment results for models trained on CIFAR-10 and tested on CIFAR-10, CIFAR-100, and SVHN. Top rows show predictive entropy and predictive marginal variance histograms of test images for in-domain and OOD data sets. Top to bottom: ReLU activation function, local stationary Mat\'ern-$\nicefrac32$ activation function \cite{meronen2020stationary}, and sinusoidal activation function that induces a globally stationary Mat\'ern-$\nicefrac32$ model. The last row shows sample images from each test set: left-side images with the lowest entropy/highest confidence and right-side images with the highest entropy/lowest confidence. See \cref{app:cifar} for histograms for more models.}
  \label{fig:histograms}
\end{figure}

\paragraph{Detection of distribution shift with rotated MNIST}
\cref{fig:mnist} demonstrates the predictive behaviour of different activation functions under covariate shift. We evaluate the predictive accuracy, mean confidence of the predicted class, and NLPD on MNIST \cite{MNIST} for different rotations of the input. Models are trained on unrotated digits using KFAC Laplace \cite{ritter2018a_kfac_laplace}. The results indicate that all models obtain similar accuracy results, while only local and global stationary models do not result in over-confident uncertainty estimates. For an ideally calibrated model, the mean confidence would decrease as low as the accuracy curve when the digits are rotated, keeping the NLPD values as low as possible.

\paragraph{Out-of-distribution detection using CIFAR-10, CIFAR-100, and SVHN}
\cref{fig:histograms} shows model performance on OOD detection in image classification for ReLU, locally stationary Mat\'ern-$\nicefrac{3}{2}$, and globally stationary Mat\'ern-$\nicefrac{3}{2}$ models. The models have been trained using SWAG \cite{maddox19_SWAG} on CIFAR-10 \cite{CIFAR}, and tested on CIFAR-10, CIFAR-100 \cite{CIFAR}, and SVHN \cite{SVHN} test set images. Both CIFAR-100 (more similar) and SVHN (more dissimilar to CIFAR-10) images are OOD data, and the models should show high uncertainties for the respective test images. We use two metrics to measure model uncertainty: predictive entropy and predictive marginal variance. Predictive entropy captures uncertainty present in the predicted mean class weights, while the predictive marginal variance captures uncertainty related to the variability in the predicted class weights. From these two metrics, the predictive marginal variance could be more suitable as an OOD detection metric, as predictive entropy can justifiably give high values for some in-distribution samples, for example, those that the model confidently predicts being in between two known classes. 

The histograms of predictive entropies for different test sets show that all models can separate between in-distribution and OOD data to some extent, and all models consider SVHN more OOD than CIFAR-100, which is intuitive as CIFAR-100 resembles CIFAR-10 more. Also, the predictive marginal variance histograms show that the models have higher variance on OOD data compared to in-distribution data. However, the ReLU model considers CIFAR-100 more OOD than SVHN and the locally stationary Mat\'ern-$\nicefrac{3}{2}$ model has overall relatively small predictive marginal variance values for all data sets. We consider predictive marginal variance a better metric for OOD detection than predictive entropy, as in-distribution samples that are hard to classify are expected to have high predictive entropy, not necessarily allowing the detection of OOD samples on this metric. See \cref{fig:histograms_app} in \cref{app:cifar} for results on different models, and \cref{tbl:cifar_auroc} in \cref{app:cifar} for numerical results showing the area under the receiver operating characteristic curve and the area under the precision-recall curve for the OOD detection experiment.

\section{Discussion and Conclusions}
\label{sec:discussion}
We have shown that \emph{periodic} activation functions in Bayesian neural networks (BNNs) establish a direct connection between the prior on the network weights and the spectral density of the limiting stationary process. This link goes beyond the Fourier (sinusoidal) basis, which we have illustrated by deriving the correspondence also for the triangle wave and a new periodic variant of the ReLU activation function.  
Moreover, we have shown that for BNNs with a periodic activation function, placing a Student-$t$ distribution on the network weights corresponds to a prior in the function space with Mat\'ern covariance structure. This correspondence is exact for the sinusoidal and sin--cos activations and approximate for the triangle and periodic ReLU activations.

Our work can help build neural network models that are more robust and less vulnerable to OOD data---occurring either naturally or maliciously. As demonstrated in the experiments, the modelling assumptions have practical importance in reducing sensitivity to input shift. Perturbation of inputs can be captured in the last layer of a deep model, and the principled choice of {\it a~priori} stationarity has shown to be effective based on the experiments. Using periodic activation functions in BNNs induces \emph{global} stationarity, \ie, the model is not only locally translation-invariant, which induces conservative behaviour across the input domain. The resulting model reverts to the prior for out-of-distribution (OOD) samples, resulting in well-specified uncertainties.

In contrast to models with non-stationary or locally stationary activations, optimising and performing approximate inference in models with periodic activation functions has proven to be more challenging. In particular, we found that the uncertainty over the bias terms is often not accurately estimated when using KFAC Laplace or SWAG, which required us to use a relatively large number of hidden units. However, the experiments using dynamic HMC indicate that the number of hidden units can be drastically reduced using a more accurate approximate inference scheme, as seen in \cref{fig:teaser} and \cref{fig:comparisons_app} in \cref{app:toy}.

The codes and data to replicate the results are available under MIT license at \url{https://github.com/AaltoML/PeriodicBNN}.

\begin{ack}
We acknowledge the computational resources provided by the Aalto Science-IT project. This research was supported by the Academy of Finland grants 324345 and 339730, and Saab Finland~Oy. We thank William J.\ Wilkinson and the anonymous reviewers for feedback on the manuscript.
\end{ack}

\phantomsection%
\addcontentsline{toc}{section}{References}
\begingroup
\small
\bibliographystyle{abbrvnat}
\bibliography{bibliography}
\endgroup

\clearpage
\appendix

\nipstitle{
    {\Large Supplementary Material:} \\
    Periodic Activation Functions Induce Stationarity}
\pagestyle{empty}

This supplementary document is organized as follows. 
\cref{app:derivations} includes further details and derivations for the periodic activation functions in the Methods section of the main paper. 
\cref{app:Matern} provides a full derivation of the correspondence between the Student-$t$ prior and Mat\'ern family under periodic activation functions. 
\cref{app:insights} presents and discusses further links to existing work.
\cref{app:details} includes details on the experiments, baseline methods, data sets, and additional tables and result plots.

\section{Derivations for Activation Functions}\label{app:derivations}

In this section we will introduce the detailed derivations for the activation functions presented in the main text.
In each derivation, we start from the definition of the covariance function of a random single-layer neural network given as:
\begin{equation}\label{eq:app-nn-cov}
  \kappa(x,x') = \int \int p(w)\, p(b)\, \sigma(w x+b) \sigma(w x'+b) \dd b \dd w ,
\end{equation}
where $\sigma(\cdot)$ denotes the activation function and $p(w)$ and $p(b)$ the priors over weights and biases respectively.

\subsection{Sinusoidal Activation Function}\label{app:sin_activation}
Let the activation function $\sigma(\cdot)$ and prior $p(b)$ take the following form:
\begin{align}
  \sigma(x) &= \sqrt{2}\sin(x) , \\
  p(b) &= \Uni(-\pi,\pi) .
\end{align}
Then the covariance function $\kappa(x,x')$ from \cref{eq:app-nn-cov} can now be written as:
\begin{equation} \label{app:eq:sin_covariance}
  \kappa(x,x') = \int p(w) \int_{-\pi}^{\pi} \frac{1}{\pi}\sin(w x+b)\sin(w x'+b) \dd b \dd w .
\end{equation}
The inner integral (over $b$) can now be solved by considering that $\sin(x) \sin(y) = \frac{1}{2}[\cos(x-y)-\cos(x+y)]$, \ie,
\begin{align}
  &\int_{-\pi}^{\pi}\sin(w x+b)\sin(w x'+b) \dd b \nonumber \\
  &\qquad\qquad= \int_{-\pi}^{\pi}\frac{\cos(w x+b-w x'-b) - \cos(w x+b+w x'+b) }{2} \dd b \nonumber \\
  &\qquad\qquad= \frac{1}{2}\int_{-\pi}^{\pi}\cos(w (x- x')) - \cos(w (x+ x')+2b) \dd b \nonumber \\
  &\qquad\qquad= \vphantom{\int} \pi\cos(w (x- x')) .
\end{align}

Plugging the above back into \cref{app:eq:sin_covariance} gives us:
\begin{equation}
  \kappa(x,x') =  \int p(w) \cos(w (x- x')) \dd w ,
\end{equation}
which, by application of Euler's formula $\cos(z) = \frac{1}{2} (e^{\imag z} + e^{-\imag z})$, can be written as:
\begin{equation}
  \kappa(x,x') =\frac{1}{2} \int p(w)  e^{\imag w (x- x')} \dd w + \frac{1}{2} \int p(w)e^{-\imag w (x- x')} \dd w .
\end{equation}
Since the integration is over the entire real line we can do a change of variables $w = -w$ in the second integral and by assuming a symmetric prior on $w$, we obtain
\begin{align}
  \kappa(x,x') &= \frac{1}{2} \int p(w)  e^{\imag w (x - x')} \dd w +\frac{1}{2} \int p(-w)e^{\imag w (x- x')} \dd w \nonumber \\
&=\int p(w)  e^{\imag w (x - x')} \dd w .
\end{align}
By letting $r = x-x'$, we find that we recover the spectral density decomposition of a stationary process given by the Wiener--Khinchin theorem, \ie,
\begin{equation}
  \kappa(r) = \int p(w) \,  e^{\imag w r} \dd w ,
\end{equation}
where $p(w) = \frac{1}{2\pi} S(w)$, which was to be shown.

\subsection{Sine--Cosine Activation Function} \label{app:sin_cos_activation}
Let us assume a bias-free random single-layer neural network. Following \cref{eq:app-nn-cov}, in the infinite limit, the corresponding covariance function would take the form:
\begin{equation}
  \kappa(x,x') = \int p(w)\, \sigma(w x) \sigma(w x') \dd w .
\end{equation}
Let the activation function be given as:
\begin{equation}
  \sigma(x) = \sin(x) + \cos(x) .
\end{equation}
This gives the covariance function in the form
\begin{align}
	\kappa(x,x') &= \int p(w)\, \left[\sin(wx) + \cos(wx) \right]\left[\sin(wx') + \cos(wx') \right] \dd w \nonumber \\
	&= \int p(w)\, \sin(w (x+x')) \dd w + \int p(w)\, \cos(w (x-x')) \dd w .
\end{align}
By application of Euler's formula we get
\begin{multline}
	\kappa(x,x') = \frac{1}{2} \Big[ \int p(w)\, e^{\imag w(x-x')} \dd w + \int p(w)\, e^{-\imag w(x-x')}\dd w \\
	+ \int p(w)\, \imag e^{\imag w(x+x')}\dd w - \int p(w)\, \imag e^{-\imag w(x+x')}\dd w \Big] ,
\end{multline}
where under assumption that $p(w)$ is symmetric and has support on the entire real line the above reduces to:
\begin{align}
	\kappa(x,x') &= \frac{1}{2} \left[ \int p(w)\, e^{\imag w(x-x')}\dd w + \int p(w)\, e^{-\imag w(x-x')}\dd w \right] \nonumber \\
	&= \int p(w)\, e^{\imag w(x-x')}\dd w ,
\end{align}
where the last step follows from application of the change of variables and the symmetricity of $p(w)$. Again, by letting $r = x-x'$, we find that we recover the spectral density decomposition of a stationary process given by the Wiener--Khinchin theorem, \ie,
\begin{equation}
  \kappa(r) = \int p(w) \,  e^{\imag w r} \dd w 
\end{equation}
confirming the statement in the main paper.

\subsection{Triangle Wave Activation} \label{app:triangle_wave}
The triangle wave is a periodic, piecewise linear function and it can be written in a parametric form as 
\begin{equation}
  \psi(x) = \frac{4}{p} \bigg( x - \frac{p}{2} \bigg\lfloor \frac{2x}{p} + \frac{1}{2} \bigg\rfloor \bigg) (-1)^{\lfloor \frac{2x}{p} + \frac{1}{2} \rfloor} ,
\end{equation}
where $p$ is the period of the triangle wave. In the following derivation we will assume that $p = 2\pi$. For analysis, we note that the Fourier series approximation to $\psi(x)$ can be given in the following form (in the limit of $n$):
\begin{equation}
\psi(x) = \lim_{n\to\infty}\frac{8}{\pi^2} \sum^{n-1}_{k=0} (-1)^k (2k + 1)^{-2} \sin( (2k +1) x) .
\end{equation}

Let $\lambda_k \coloneqq 2k + 1$, and let us assume the triangle wave activation function and a uniform prior on $b$ (as in the preceding derivation):
\begin{align}
\sigma(z) &= \sqrt{2} \sum^{n-1}_{k=0} (-1)^k \lambda_k^{-2} \sin(\lambda_k z), \\
p(b) &= \mathrm{Uniform}(-\pi, \pi) ,
\end{align}
where $n$ is the number of harmonics to include in the approximation and $k$ is the harmonic label.

The covariance function $\kappa(x,x') $ is, therefore, given as:
\begin{multline}\label{app:eq:triangle_wave}
\kappa(x,x')  = \int p(w)\, \int^{\pi}_{-\pi} 2 p(b) \left[ \sum^{n-1}_{k=0} (-1)^k \lambda_k^{-2} \sin(\lambda_k (wx + b))\right] \\
\left[ \sum^{n-1}_{j=0} (-1)^j \lambda_j^{-2} \sin(\lambda_j (wx' + b))\right] \dd w \dd b .
\end{multline}

Now, let us solve the inner integrals by ignoring the constant terms and assuming that $k \neq j$:
\begin{multline}
\int^{\pi}_{-\pi} \frac{1}{\pi} \sin(\lambda_k (wx + b)) \sin(\lambda_j (wx' + b)) \dd b  \\
=\int^{\pi}_{-\pi} \frac{1}{2\pi} \cos(\lambda_k (wx + b) - \lambda_j (wx' + b)) - \cos(\lambda_k (wx + b) + \lambda_j (wx' + b)) \dd b  .
\end{multline}

By solving the definite integral above, we obtain:
\begin{multline}
	= \frac{\sin(w (\lambda_k x - \lambda_j x')) - \sin(w (\lambda_k x - \lambda_j x')) + 2 \pi (\lambda_k - \lambda_j))}{\lambda_j - \lambda_k}\\
	+ \frac{\sin(w (\lambda_k x + \lambda_j x')) - \sin(w (\lambda_k x + \lambda_j x') + 2 \pi (\lambda_k + \lambda_j)) }{\lambda_j + \lambda_k} ,
\end{multline}
where we recognise that $2 \pi (\lambda_k - \lambda_j)$ will always be an even multiple of $2\pi$, as both $\lambda_k$ and $\lambda_j$ are odd. Thus, the above cancels out and \cref{app:eq:triangle_wave} reduces to containing only indexes $k = j$:
\begin{equation}
\kappa(x,x')  = \int p(w)\, \int^{\pi}_{-\pi} 2 p(b) \sum^{n-1}_{k=0} \frac{(-1)^{2k}}{\lambda_k^{4}} \sin(\lambda_k (wx + b)) \sin(\lambda_k (wx' + b)) \dd w \dd b  .
\end{equation}

The solution to the first integral for every summand is given as:
\begin{align}
	&\int^{\pi}_{-\pi} 2p(b) \sin(\lambda_k (wx + b)) \sin(\lambda_k (wx' + b)) \dd b \nonumber \\
	&\qquad\qquad=\int^{\pi}_{-\pi} \frac{1}{2\pi} \cos(w \lambda_k (x - x')) - \cos(w \lambda_k (x+y) + 2 b \lambda_k) \dd b \nonumber \\
	&\qquad\qquad= \cos(w \lambda_k (x - x')) - \underbrace{\frac{\cos(w \lambda_k (x+x') + 2 \lambda_k \pi) \sin(2 \lambda_k \pi))}{\lambda_k}}_{=0} .
\end{align}

Inserting this into the equation of the covariance function results in:
\begin{equation}
	\kappa(x,x')  = \int p(w)\,\sum^{n-1}_{k=0}\frac{(-1)^{2k}}{\lambda_k^{4}}\cos(w \lambda_k (x - x')) \dd w .
\end{equation}

We now take into account that the exact solution is given when $n \to \infty$ and that $(-1)^{2k} = 1$:
\begin{equation}
	\kappa(x,x')  = \int \lim_{n\to\infty}p(w)\,\sum^{n-1}_{k=0}\frac{1}{\lambda_k^{4}}\cos(w \lambda_k (x - x')) \dd w .
\end{equation}

Here we can use the dominated convergence theorem to take the limit outside of the integral. For this we have $f_n(w) = p(w)\,\sum^{n-1}_{k=0}\frac{1}{\lambda_k^{4}}\cos(w \lambda_k (x - x'))$, and we need a function $g(w)$ for which $|f_n(w)| \leq g(w)$ for all $n$ and $\int g(w) \dd w < \infty$:
\begin{align}
	|f_n(w)| &= p(w)\,\sum^{n-1}_{k=0}\frac{1}{(2k+1)^{4}}|\cos(w \lambda_k (x - x'))| \\
	&\leq p(w)\,\sum^{n-1}_{k=0}\frac{1}{(2k+1)^{4}} \\
	&\leq p(w)\,\sum^{\infty}_{k=1}\frac{1}{k^{4}} = \frac{\pi^4}{90}p(w) 
\end{align}

Hence, by choosing $g(w) = \frac{\pi^4}{90}p(w) $ the requirements to use the dominated convergence theorem are satisfied since $\int \frac{\pi^4}{90}p(w) \dd w = \frac{\pi^4}{90} < \infty$.

We obtain: 
\begin{equation}\label{eq:triangle_kernel}
	\kappa(x,x')  = \lim_{n\to\infty} \int p(w)\,\sum^{n-1}_{k=0}\frac{1}{\lambda_k^{4}}\cos(w \lambda_k (x - x')) \dd w .
\end{equation}

It is also possible to write the density on the weights in \cref{eq:triangle_kernel} in the form of a mixture density (assuming the density $p$ is in the location-scale family, which is the case for the prior distributions discussed in this paper):
\begin{equation}
\begin{aligned}
\kappa(x,x') &= \lim_{n\to\infty} \int \sum^{n-1}_{k=0} \frac{1}{\lambda^4_k} p(w) e^{\mathrm{i} \lambda_k w (x - x')}
\mathrm{d}w, \\
\kappa(x,x') &= \lim_{n\to\infty} \int \sum^{n-1}_{k=0} \pi_k p(w) e^{\mathrm{i} \lambda_k w (x - x')} \mathrm{d}w, \\
\kappa(x,x') &= \lim_{n\to\infty} \int \sum^{n-1}_{k=0} \pi_k p(w \mid \lambda_k) e^{\mathrm{i}  w (x - x')} \mathrm{d}w ,\\
\end{aligned}
\end{equation}
where $p(w \mid \lambda_k)$ denotes the density function of $p(w)$ with scale parameter $\lambda_k$. Let us now denote $\lim_{n\to\infty}\sum^{n-1}_{k=0} \pi_k p(w \mid \lambda_k)$ as $\hat{p}(w) $, \ie,

\begin{equation}
\kappa(x,x') = \int \hat{p}(w) e^{\mathrm{i}  w (x - x')} \mathrm{d}w .
\end{equation}

By letting $r = x-x'$, we find that we again recover the spectral density decomposition of a stationary process given by the Wiener--Khinchin theorem, \ie,
\begin{equation}\label{eq:triangle_final}
\kappa(r) = \int \hat{p}(w) \,  e^{\imag w r} \dd w ,
\end{equation}
which recovers the statement in the main paper.

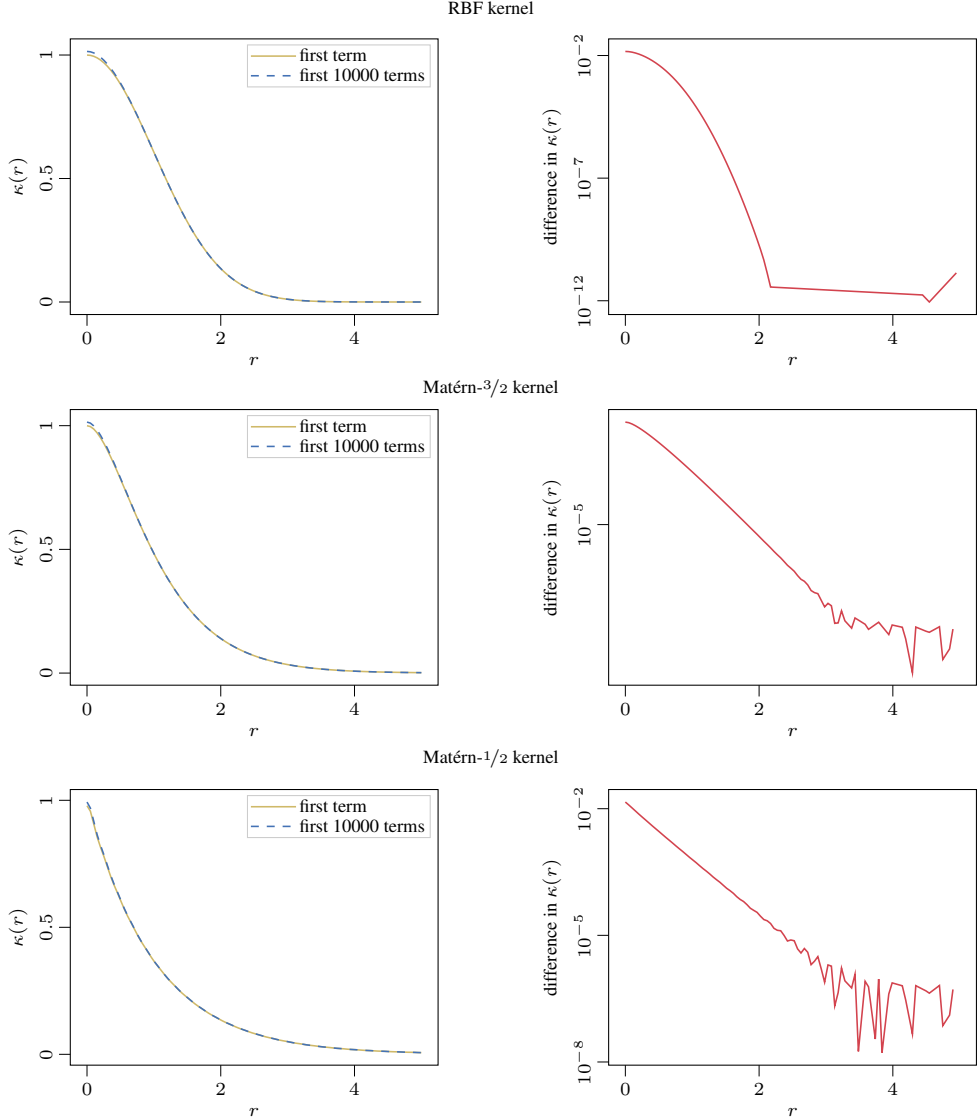
\begin{figure}[t]
  \centering\scriptsize
  \setlength{\figurewidth}{.35\textwidth}
  \setlength{\figureheight}{.75\figurewidth}
  \pgfplotsset{scale only axis,y tick label style={rotate=90}}

    \centering
    RBF kernel
    \\[0.5em]
  \begin{subfigure}[t]{.49\textwidth}
    \centering
    % This file was created by tikzplotlib v0.9.6.
\begin{tikzpicture}

\begin{axis}[
height=\figureheight,
legend cell align={left},
legend style={fill opacity=0.8, draw opacity=1, text opacity=1, draw=white!80!black},
tick align=outside,
tick pos=left,
width=\figurewidth,
x grid style={white!69.0196078431373!black},
xlabel={$r$},
xmin=-0.25, xmax=5.25,
xtick style={color=black},
y grid style={white!69.0196078431373!black},
ylabel={$\kappa(r)$},
ymin=-0.0507249156334001, ymax=1.06530520567872,
ytick style={color=black}
]
\addplot [semithick, mycolor4]
table {%
0 0.9999
0.0505050505050505 0.998625560344961
0.101010101010101 0.994811979254287
0.151515151515152 0.988488346446275
0.202020202020202 0.979702733973313
0.252525252525253 0.968521587672942
0.303030303030303 0.955028890355668
0.353535353535354 0.939325111470696
0.404040404040404 0.92152596162243
0.454545454545455 0.901760973529418
0.505050505050505 0.880171933755585
0.555555555555556 0.856911191746135
0.606060606060606 0.83213987432709
0.656565656565657 0.806026034852721
0.707070707070707 0.778742766599739
0.757575757575758 0.750466309816826
0.808080808080808 0.721374181064118
0.858585858585859 0.691643352154402
0.909090909090909 0.66144850418381
0.95959595959596 0.63096037987313
1.01010101010101 0.600344254798983
1.06060606060606 0.569758545151138
1.11111111111111 0.539353566486903
1.16161616161616 0.509270454646312
1.21212121212121 0.479640256623524
1.26262626262626 0.450583195838309
1.31313131313131 0.422208112990651
1.36363636363636 0.394612080578854
1.41414141414141 0.367880186276956
1.46464646464646 0.342085477751888
1.51515151515152 0.317289059195888
1.56565656565657 0.293540327886417
1.61616161616162 0.270877337484784
1.66666666666667 0.249327273556418
1.71717171717172 0.228907025940632
1.76767676767677 0.209623842107299
1.81818181818182 0.191476045495162
1.86868686868687 0.17445380300743
1.91919191919192 0.158539926314908
1.96969696969697 0.143710692350308
2.02020202020202 0.129936669331632
2.07070707070707 0.117183535787241
2.12121212121212 0.10541288132961
2.17171717171717 0.0945829792980507
2.22222222222222 0.0846495228236677
2.27272727272727 0.0755663173255895
2.32323232323232 0.0672859238921409
2.37373737373737 0.0597602494040063
2.42424242424242 0.0529410805923683
2.47474747474747 0.04678056047179
2.52525252525253 0.0412316067279534
2.57575757575758 0.0362482726614065
2.62626262626263 0.0317860521816963
2.67676767676768 0.0278021311071121
2.72727272727273 0.0242555876529369
2.77777777777778 0.0211075454880257
2.82828282828283 0.0183212831109747
2.87878787878788 0.0158623035507098
2.92929292929293 0.0136983685414742
2.97979797979798 0.0117995013697478
3.03030303030303 0.0101379625523225
3.08080808080808 0.00868820239282764
3.13131313131313 0.00742679429075069
3.18181818181818 0.00633235245455581
3.23232323232323 0.00538543741043198
3.28282828282828 0.00456845241135571
3.33333333333333 0.00386553354745879
3.38383838383838 0.00326243604702686
3.43434343434343 0.00274641894558807
3.48484848484848 0.00230612999509657
3.53535353535354 0.00193149239162249
3.58585858585859 0.00161359462258372
3.63636363636364 0.00134458447667597
3.68686868686869 0.00111756802359097
3.73737373737374 0.000926514157799928
3.78787878787879 0.000766165111774877
3.83838383838384 0.000631953179024618
3.88888888888889 0.00051992374568147
3.93939393939394 0.000426664610074893
3.98989898989899 0.000349241471428986
4.04040404040404 0.000285139389909776
4.09090909090909 0.000232209958960105
4.14141414141414 0.000188623885328199
4.19191919191919 0.000152828640527207
4.24242424242424 0.000123510827786433
4.29292929292929 9.95628990719761e-05
4.34343434343434 8.00538557537732e-05
4.39393939393939 6.42035724095704e-05
4.44444444444444 5.13603946365974e-05
4.49494949494949 4.09816773079524e-05
4.54545454545454 3.26169483255896e-05
4.5959595959596 2.58934036027969e-05
4.64646464646465 2.05034609256959e-05
4.6969696969697 1.61941227926503e-05
4.74747474747475 1.2757920749916e-05
4.7979797979798 1.00252356790496e-05
4.84848484848485 7.85780959375591e-06
4.8989898989899 6.14328452318945e-06
4.94949494949495 4.79062280754166e-06
5 3.72628050671091e-06
};
\addlegendentry{first term}
\addplot [semithick, dashed, mycolor0]
table {%
0 1.01457656380089
0.0505050505050505 1.0130301913321
0.101010101010101 1.00849870301748
0.151515151515152 1.0011435424954
0.202020202020202 0.991126153404086
0.252525252525253 0.978607979381789
0.303030303030303 0.963750464066739
0.353535353535354 0.946715051097174
0.404040404040404 0.927663184111329
0.454545454545455 0.90675630674744
0.505050505050505 0.884155862643406
0.555555555555556 0.86002329542024
0.606060606060606 0.834520048357389
0.656565656565657 0.807807561243112
0.707070707070707 0.780047252579385
0.757575757575758 0.751400453504212
0.808080808080808 0.722028230921124
0.858585858585859 0.692091020999415
0.909090909090909 0.661748008070277
0.95959595959596 0.631156226820727
1.01010101010101 0.600469419455111
1.06060606060606 0.569836723592245
1.11111111111111 0.539401289439646
1.16161616161616 0.509298925461086
1.21212121212121 0.479656856447136
1.26262626262626 0.450592654668298
1.31313131313131 0.422213380453366
1.36363636363636 0.394614947366421
1.41414141414141 0.367881711099338
1.46464646464646 0.342086270384814
1.51515151515152 0.317289461869953
1.56565656565657 0.293540527809228
1.61616161616162 0.270877434491798
1.66666666666667 0.249327319556446
1.71717171717172 0.22890704725747
1.76767676767677 0.209623851759675
1.81818181818182 0.191476049766323
1.86868686868687 0.174453804851941
1.91919191919192 0.158539927091041
1.96969696969697 0.143710692667437
2.02020202020202 0.129936669456951
2.07070707070707 0.1171835358331
2.12121212121212 0.105412881343022
2.17171717171717 0.0945829793016724
2.22222222222222 0.0846495228083332
2.27272727272727 0.0755663173093251
2.32323232323232 0.0672859238743488
2.37373737373737 0.0597602493788201
2.42424242424242 0.0529410805893202
2.47474747474747 0.0467805604418295
2.52525252525253 0.0412316067068001
2.57575757575758 0.0362482726466236
2.62626262626263 0.0317860521542113
2.67676767676768 0.0278021310694899
2.72727272727273 0.024255587650621
2.77777777777778 0.0211075454384795
2.82828282828283 0.0183212831068273
2.87878787878788 0.0158623035452743
2.92929292929293 0.0136983685197664
2.97979797979798 0.0117995013403692
3.03030303030303 0.010137962537699
3.08080808080808 0.00868820239089352
3.13131313131313 0.00742679428083962
3.18181818181818 0.00633235237181573
3.23232323232323 0.00538543740937541
3.28282828282828 0.00456845231442653
3.33333333333333 0.00386553354608883
3.38383838383838 0.00326243601678386
3.43434343434343 0.00274641883776019
3.48484848484848 0.00230612997994207
3.53535353535354 0.00193149239009173
3.58585858585859 0.00161359462216267
3.63636363636364 0.00134458447623778
3.68686868686869 0.00111756802326504
3.73737373737374 0.000926514156784085
3.78787878787879 0.000766165103174164
3.83838383838384 0.000631953048193121
3.88888888888889 0.000519923630263658
3.93939393939394 0.000426664609063365
3.98989898989899 0.000349241465125678
4.04040404040404 0.000285139207763287
4.09090909090909 0.00023220995883335
4.14141414141414 0.000188623692824691
4.19191919191919 0.000152828639813002
4.24242424242424 0.000123510631541308
4.29292929292929 9.95628989523214e-05
4.34343434343434 8.00535722974614e-05
4.39393939393939 6.42035711270255e-05
4.44444444444444 5.13603963583402e-05
4.49494949494949 4.0981343344619e-05
4.54545454545454 3.26169492150218e-05
4.5959595959596 2.5893403433816e-05
4.64646464646465 2.05033683734386e-05
4.6969696969697 1.61937369254392e-05
4.74747474747475 1.27578967737778e-05
4.7979797979798 1.00252354436757e-05
4.84848484848485 7.85780958456808e-06
4.8989898989899 6.1432839330064e-06
4.94949494949495 4.79063648045864e-06
5 3.72624442336949e-06
};
\addlegendentry{first 10000 terms}
\end{axis}

\end{tikzpicture}
  \end{subfigure}
  \hfill
  \begin{subfigure}[t]{.49\textwidth}
    \centering
    % This file was created by tikzplotlib v0.9.6.
\begin{tikzpicture}

\begin{axis}[
height=\figureheight,
log basis y={10},
tick align=outside,
tick pos=left,
width=\figurewidth,
x grid style={white!69.0196078431373!black},
xlabel={$r$},
xmin=-0.25, xmax=5.25,
xtick style={color=black},
y grid style={white!69.0196078431373!black},
ylabel={difference in $\kappa(r)$},
ymin=2.74307250753277e-13, ymax=0.0475882750455984,
ymode=log,
ytick style={color=black}
]
\addplot [semithick, mycolor2]
table {%
0 0.0146765638008928
0.0505050505050505 0.0144046309871433
0.101010101010101 0.0136867237631948
0.151515151515152 0.0126551960491207
0.202020202020202 0.0114234194307729
0.252525252525253 0.0100863917088464
0.303030303030303 0.00872157371107118
0.353535353535354 0.00738993962647772
0.404040404040404 0.00613722248889925
0.454545454545455 0.00499533321802126
0.505050505050505 0.0039839288878214
0.555555555555556 0.00311210367410442
0.606060606060606 0.00238017403029911
0.656565656565657 0.00178152639039086
0.707070707070707 0.00130448597964572
0.757575757575758 0.000934143687386424
0.808080808080808 0.000654049857005679
0.858585858585859 0.000447668845013016
0.909090909090909 0.000299503886467489
0.95959595959596 0.000195846947597
1.01010101010101 0.000125164656127885
1.06060606060606 7.8178441106358e-05
1.11111111111111 4.77229527435785e-05
1.16161616161616 2.84708147746393e-05
1.21212121212121 1.65998236125731e-05
1.26262626262626 9.45882998920622e-06
1.31313131313131 5.2674627147864e-06
1.36363636363636 2.86678756705738e-06
1.41414141414141 1.52482238158225e-06
1.46464646464646 7.92632926460701e-07
1.51515151515152 4.02674064925446e-07
1.56565656565657 1.99922811472408e-07
1.61616161616162 9.7007014143724e-08
1.66666666666667 4.60000277313632e-08
1.71717171717172 2.13168377116713e-08
1.76767676767677 9.65237614836134e-09
1.81818181818182 4.27116178536302e-09
1.86868686868687 1.84451101614691e-09
1.91919191919192 7.76133074475993e-10
1.96969696969697 3.1712954484675e-10
2.02020202020202 1.2531981008479e-10
2.07070707070707 4.58591914442863e-11
2.12121212121212 1.34116051597744e-11
2.17171717171717 3.62164465084192e-12
2.22222222222222 -1.53344698050617e-11
2.27272727272727 -1.62644064882755e-11
2.32323232323232 -1.77921150035232e-11
2.37373737373737 -2.51862211642084e-11
2.42424242424242 -3.04809649742666e-12
2.47474747474747 -2.99605618092613e-11
2.52525252525253 -2.11532250049551e-11
2.57575757575758 -1.47829179453218e-11
2.62626262626263 -2.74850142645278e-11
2.67676767676768 -3.76222282494343e-11
2.72727272727273 -2.31589400434551e-12
2.77777777777778 -4.95461900673622e-11
2.82828282828283 -4.14739423360011e-12
2.87878787878788 -5.43541253672508e-12
2.92929292929293 -2.17078438558005e-11
2.97979797979798 -2.93785915733702e-11
3.03030303030303 -1.46235107356674e-11
3.08080808080808 -1.93412126592296e-12
3.13131313131313 -9.91106675896081e-12
3.18181818181818 -8.27400804120915e-11
3.23232323232323 -1.05656455806624e-12
3.28282828282828 -9.69291816949047e-11
3.33333333333333 -1.36996664013012e-12
3.38383838383838 -3.02430021130451e-11
3.43434343434343 -1.07827887677742e-10
3.48484848484848 -1.51545013517274e-11
3.53535353535354 -1.5307637969697e-12
3.58585858585859 -4.21049913684746e-13
3.63636363636364 -4.38189198467831e-13
3.68686868686869 -3.2592309927304e-13
3.73737373737374 -1.01584268340921e-12
3.78787878787879 -8.60071237319959e-12
3.83838383838384 -1.30831497111711e-10
3.88888888888889 -1.15417812351741e-10
3.93939393939394 -1.01152804665369e-12
3.98989898989899 -6.30330732271159e-12
4.04040404040404 -1.82146489406156e-10
4.09090909090909 -1.26754184405498e-13
4.14141414141414 -1.92503508278244e-10
4.19191919191919 -7.14204113601638e-13
4.24242424242424 -1.96245124453121e-10
4.29292929292929 -1.19654693054791e-13
4.34343434343434 -2.83456311821564e-10
4.39393939393939 -1.28254497950772e-12
4.44444444444444 1.72174282480914e-12
4.49494949494949 -3.33963333373831e-10
4.54545454545454 8.89432231749969e-13
4.5959595959596 -1.68980962171128e-13
4.64646464646465 -9.25522572670827e-11
4.6969696969697 -3.85867211098089e-10
4.74747474747475 -2.39761381779071e-11
4.7979797979798 -2.35373915182576e-13
4.84848484848485 -9.18783752963497e-15
4.8989898989899 -5.90183059724408e-13
4.94949494949495 1.36729169839363e-11
5 -3.60833414183589e-11
};
\end{axis}

\end{tikzpicture}  
  \end{subfigure}  
  \\
      \centering
    Mat\'ern-$\nicefrac{3}{2}$ kernel
    \\[0.5em]
  \begin{subfigure}[t]{.49\textwidth}
    \centering
    % This file was created by tikzplotlib v0.9.6.
\begin{tikzpicture}

\begin{axis}[
height=\figureheight,
legend cell align={left},
legend style={fill opacity=0.8, draw opacity=1, text opacity=1, draw=white!80!black},
tick align=outside,
tick pos=left,
width=\figurewidth,
x grid style={white!69.0196078431373!black},
xlabel={$r$},
xmin=-0.25, xmax=5.25,
xtick style={color=black},
y grid style={white!69.0196078431373!black},
ylabel={$\kappa(r)$},
ymin=-0.0489678768674302, ymax=1.06513509397651,
ytick style={color=black}
]
\addplot [semithick, mycolor4]
table {%
0 0.999818679668737
0.0505050505050505 0.996319035405353
0.101010101010101 0.986309189715655
0.151515151515152 0.970910909869111
0.202020202020202 0.951233923551267
0.252525252525253 0.928079820870525
0.303030303030303 0.902089771259414
0.353535353535354 0.873915215401228
0.404040404040404 0.844160935203224
0.454545454545455 0.813282134287701
0.505050505050505 0.781631906711647
0.555555555555556 0.749558450760766
0.606060606060606 0.717390719214043
0.656565656565657 0.68537065321854
0.707070707070707 0.65366896792973
0.757575757575758 0.622453263579594
0.808080808080808 0.591890307288861
0.858585858585859 0.562095434267354
0.909090909090909 0.533132270143701
0.95959595959596 0.50506332349965
1.01010101010101 0.477961040330274
1.06060606060606 0.451868773450409
1.11111111111111 0.426791106120421
1.16161616161616 0.402731884200585
1.21212121212121 0.379710413087872
1.26262626262626 0.357731536973034
1.31313131313131 0.336770289131068
1.36363636363636 0.316800050785253
1.41414141414141 0.297811731036234
1.46464646464646 0.279791567819131
1.51515151515152 0.262702527278854
1.56565656565657 0.246504302447252
1.61616161616162 0.231173966735314
1.66666666666667 0.216690519781771
1.71717171717172 0.203014761079443
1.76767676767677 0.190102376091458
1.81818181818182 0.177924906470122
1.86868686868687 0.166460224904516
1.91919191919192 0.15567217795021
1.96969696969697 0.145517773899787
2.02020202020202 0.135967545321996
2.07070707070707 0.127001194442043
2.12121212121212 0.118588000980437
2.17171717171717 0.110689007086255
2.22222222222222 0.103276028598826
2.27272727272727 0.0963317276898095
2.32323232323232 0.0898315515320301
2.37373737373737 0.0837417473134957
2.42424242424242 0.0780364361297114
2.47474747474747 0.0727013710847407
2.52525252525253 0.0677179815230849
2.57575757575758 0.0630580146712054
2.62626262626263 0.0586982319138027
2.67676767676768 0.0546271126672452
2.72727272727273 0.0508314115998503
2.77777777777778 0.0472881952319428
2.82828282828283 0.0439768586652596
2.87878787878788 0.0408880458780055
2.92929292929293 0.0380129765729285
2.97979797979798 0.0353336159100381
3.03030303030303 0.0328318411062895
3.08080808080808 0.0304998712611165
3.13131313131313 0.028332483229286
3.18181818181818 0.0263160148025752
3.23232323232323 0.0244346338222291
3.28282828282828 0.0226816069120229
3.33333333333333 0.021054399456347
3.38383838383838 0.0195431641023521
3.43434343434343 0.0181341782112339
3.48484848484848 0.0168213311465782
3.53535353535354 0.0156039926792894
3.58585858585859 0.0144755787342824
3.63636363636364 0.0134243239999328
3.68686868686869 0.0124444308051403
3.73737373737374 0.011536494268761
3.78787878787879 0.0106966662517935
3.83838383838384 0.00991501669638875
3.88888888888889 0.00918586192344786
3.93939393939394 0.00851045491651195
3.98989898989899 0.00788718796106786
4.04040404040404 0.00730786705429746
4.09090909090909 0.00676682135198753
4.14141414141414 0.00626550427915354
4.19191919191919 0.00580408756177252
4.24242424242424 0.00537602654606277
4.29292929292929 0.00497564276788447
4.34343434343434 0.00460423690865655
4.39393939393939 0.00426333455298627
4.44444444444444 0.00394795573351847
4.49494949494949 0.00365245616135553
4.54545454545454 0.00337772670957151
4.5959595959596 0.00312625724313917
4.64646464646465 0.00289454100591306
4.6969696969697 0.00267705378875626
4.74747474747475 0.00247410639172525
4.7979797979798 0.00228880034966918
4.84848484848485 0.00211899735247053
4.8989898989899 0.00195940641453613
4.94949494949495 0.00180966484118127
5 0.00167317163210415
};
\addlegendentry{first term}
\addplot [semithick, dashed, mycolor0]
table {%
0 1.01449404984724
0.0505050505050505 1.01038311407448
0.101010101010101 0.999101301949927
0.151515151515152 0.982206954943217
0.202020202020202 0.961016587221668
0.252525252525253 0.936434978632757
0.303030303030303 0.909150937662217
0.353535353535354 0.879833059359677
0.404040404040404 0.849086945056304
0.454545454545455 0.81735949922706
0.505050505050505 0.784990564026826
0.555555555555556 0.752313406359344
0.606060606060606 0.719642227538132
0.656565656565657 0.687204826861619
0.707070707070707 0.655158768984932
0.757575757575758 0.623660109404048
0.808080808080808 0.592865669789257
0.858585858585859 0.562882035393305
0.909090909090909 0.533765306616764
0.95959595959596 0.505571824026656
1.01010101010101 0.478368876616609
1.06060606060606 0.452195351041325
1.11111111111111 0.427052170827208
1.16161616161616 0.402940313793661
1.21212121212121 0.379876655264464
1.26262626262626 0.357863938665511
1.31313131313131 0.336875586624265
1.36363636363636 0.316883743694829
1.41414141414141 0.297878202173277
1.46464646464646 0.279844269058986
1.51515151515152 0.26274427421835
1.56565656565657 0.246537384193414
1.61616161616162 0.231200146480576
1.66666666666667 0.216711187699232
1.71717171717172 0.203031090630085
1.76767676767677 0.190115294700987
1.81818181818182 0.177935089168085
1.86868686868687 0.166468230693773
1.91919191919192 0.155678501115968
1.96969696969697 0.145522770006772
2.02020202020202 0.135971459692334
2.07070707070707 0.127004265079878
2.12121212121212 0.11859043512118
2.17171717171717 0.110690921340948
2.22222222222222 0.103277513163382
2.27272727272727 0.0963328994054339
2.32323232323232 0.0898324884452404
2.37373737373737 0.0837424706501152
2.42424242424242 0.0780369897428174
2.47474747474747 0.0727018209008182
2.52525252525253 0.0677183431714072
2.57575757575758 0.0630582802481161
2.62626262626263 0.0586984361675496
2.67676767676768 0.0546272901712827
2.72727272727273 0.0508315486340417
2.77777777777778 0.0472882869535053
2.82828282828283 0.0439769377428471
2.87878787878788 0.0408881194706925
2.92929292929293 0.0380130224024877
2.97979797979798 0.0353336446091528
3.03030303030303 0.0328318780233366
3.08080808080808 0.030499902361042
3.13131313131313 0.0283324921313771
3.18181818181818 0.0263160239639572
3.23232323232323 0.0244346554591024
3.28282828282828 0.0226816175773326
3.33333333333333 0.0210543959558111
3.38383838383838 0.0195431704171694
3.43434343434343 0.0181341912521458
3.48484848484848 0.0168213304242844
3.53535353535354 0.0156039882903571
3.58585858585859 0.0144755870547932
3.63636363636364 0.0134243298055749
3.68686868686869 0.0124444238995039
3.73737373737374 0.0115364937291625
3.78787878787879 0.010696675902998
3.83838383838384 0.00991501588163256
3.88888888888889 0.00918585347157431
3.93939393939394 0.00851045892507906
3.98989898989899 0.00788719586859785
4.04040404040404 0.00730786155409931
4.09090909090909 0.00676681591117986
4.14141414141414 0.00626551101886671
4.19191919191919 0.00580409051077955
4.24242424242424 0.00537601918083223
4.29292929292929 0.00497564303434296
4.34343434343434 0.00460424374618944
4.39393939393939 0.00426333147466474
4.44444444444444 0.00394794958983278
4.49494949494949 0.00365246124942496
4.54545454545454 0.00337773141833329
4.5959595959596 0.00312625043078665
4.64646464646465 0.00289453811201407
4.6969696969697 0.00267706077536738
4.74747474747475 0.00247410707600121
4.7979797979798 0.00228879335467553
4.84848484848485 0.00211899879521146
4.8989898989899 0.00195941233969295
4.94949494949495 0.00180966141675087
5 0.0016731672618396
};
\addlegendentry{first 10000 terms}
\end{axis}

\end{tikzpicture}
  \end{subfigure}
  \hfill
  \begin{subfigure}[t]{.49\textwidth}
    \centering
    % This file was created by tikzplotlib v0.9.6.
\begin{tikzpicture}

\begin{axis}[
height=\figureheight,
log basis y={10},
tick align=outside,
tick pos=left,
width=\figurewidth,
x grid style={white!69.0196078431373!black},
xlabel={$r$},
xmin=-0.25, xmax=5.25,
xtick style={color=black},
y grid style={white!69.0196078431373!black},
ylabel={difference in $\kappa(r)$},
ymin=1.09290295916067e-10, ymax=0.0357797283815805,
ymode=log,
ytick style={color=black}
]
\addplot [semithick, mycolor2]
table {%
0 0.0146753701784991
0.0505050505050505 0.0140640786691248
0.101010101010101 0.0127921122342727
0.151515151515152 0.0112960450741064
0.202020202020202 0.00978266367040082
0.252525252525253 0.00835515776223206
0.303030303030303 0.00706116640280274
0.353535353535354 0.0059178439584493
0.404040404040404 0.00492600985307956
0.454545454545455 0.00407736493935895
0.505050505050505 0.00335865731517926
0.555555555555556 0.0027549555985783
0.606060606060606 0.00225150832408871
0.656565656565657 0.00183417364307981
0.707070707070707 0.00148980105520191
0.757575757575758 0.00120684582445418
0.808080808080808 0.000975362500395538
0.858585858585859 0.000786601125950415
0.909090909090909 0.00063303647306312
0.95959595959596 0.000508500527006128
1.01010101010101 0.000407836286334839
1.06060606060606 0.000326577590915478
1.11111111111111 0.000261064706787106
1.16161616161616 0.000208429593076043
1.21212121212121 0.000166242176592279
1.26262626262626 0.000132401692477802
1.31313131313131 0.000105297493196821
1.36363636363636 8.36929095762162e-05
1.41414141414141 6.6471137043278e-05
1.46464646464646 5.27012398555682e-05
1.51515151515152 4.17469394953796e-05
1.56565656565657 3.30817461624278e-05
1.61616161616162 2.61797452611734e-05
1.66666666666667 2.06679174615809e-05
1.71717171717172 1.63295506425132e-05
1.76767676767677 1.29186095291223e-05
1.81818181818182 1.0182697963379e-05
1.86868686868687 8.00578925694295e-06
1.91919191919192 6.32316575757685e-06
1.96969696969697 4.99610698451169e-06
2.02020202020202 3.91437033755992e-06
2.07070707070707 3.07063783500094e-06
2.12121212121212 2.43414074294068e-06
2.17171717171717 1.91425469345252e-06
2.22222222222222 1.48456455555113e-06
2.27272727272727 1.17171562438734e-06
2.32323232323232 9.36913210305979e-07
2.37373737373737 7.23336619573578e-07
2.42424242424242 5.5361310603852e-07
2.47474747474747 4.49816077469078e-07
2.52525252525253 3.61648322316777e-07
2.57575757575758 2.65576910740695e-07
2.62626262626263 2.04253746846628e-07
2.67676767676768 1.7750403748934e-07
2.72727272727273 1.37034191410867e-07
2.77777777777778 9.17215624979262e-08
2.82828282828283 7.90775875059158e-08
2.87878787878788 7.35926869513404e-08
2.92929292929293 4.5829559203181e-08
2.97979797979798 2.86991147444016e-08
3.03030303030303 3.69170471276048e-08
3.08080808080808 3.10999254925226e-08
3.13131313131313 8.90209113021956e-09
3.18181818181818 9.16138197942784e-09
3.23232323232323 2.16368732479377e-08
3.28282828282828 1.06653096909604e-08
3.33333333333333 -3.50053591322008e-09
3.38383838383838 6.31481733223693e-09
3.43434343434343 1.30409119544761e-08
3.48484848484848 -7.22293870547563e-10
3.53535353535354 -4.38893234255833e-09
3.58585858585859 8.32051074046991e-09
3.63636363636364 5.80564207819378e-09
3.68686868686869 -6.90563643074293e-09
3.73737373737374 -5.39598534257668e-10
3.78787878787879 9.65120445887979e-09
3.83838383838384 -8.14756182659915e-10
3.88888888888889 -8.45187354998889e-09
3.93939393939394 4.00856710845721e-09
3.98989898989899 7.90752998323929e-09
4.04040404040404 -5.5001981545677e-09
4.09090909090909 -5.44080767592159e-09
4.14141414141414 6.73971317210337e-09
4.19191919191919 2.94900703249473e-09
4.24242424242424 -7.36523054524485e-09
4.29292929292929 2.66458498494881e-10
4.34343434343434 6.83753289664607e-09
4.39393939393939 -3.07832152910553e-09
4.44444444444444 -6.14368569386292e-09
4.49494949494949 5.08806942542059e-09
4.54545454545454 4.70876177632862e-09
4.5959595959596 -6.8123525141274e-09
4.64646464646465 -2.89389899283538e-09
4.6969696969697 6.98661111473223e-09
4.74747474747475 6.84275961414238e-10
4.7979797979798 -6.99499365071701e-09
4.84848484848485 1.44274092108918e-09
4.8989898989899 5.92515682134795e-09
4.94949494949495 -3.42443040135358e-09
5 -4.37026454695039e-09
};
\end{axis}

\end{tikzpicture}  
  \end{subfigure}  
   \\
       \centering
    Mat\'ern-$\nicefrac{1}{2}$ kernel
    \\[0.5em]
  \begin{subfigure}[t]{.49\textwidth}
    \centering
    % This file was created by tikzplotlib v0.9.6.
\begin{tikzpicture}

\begin{axis}[
height=\figureheight,
legend cell align={left},
legend style={fill opacity=0.8, draw opacity=1, text opacity=1, draw=white!80!black},
tick align=outside,
tick pos=left,
width=\figurewidth,
x grid style={white!69.0196078431373!black},
xlabel={$r$},
xmin=-0.25, xmax=5.25,
xtick style={color=black},
y grid style={white!69.0196078431373!black},
ylabel={$\kappa(r)$},
ymin=-0.0426849430382151, ymax=1.04237773048233,
ytick style={color=black}
]
\addplot [semithick, mycolor4]
table {%
0 0.978691436038913
0.0505050505050505 0.957061187828367
0.101010101010101 0.906967618351842
0.151515151515152 0.855760426017867
0.202020202020202 0.815418388028983
0.252525252525253 0.779057971942736
0.303030303030303 0.739649221724383
0.353535353535354 0.700497171981491
0.404040404040404 0.666584527176005
0.454545454545455 0.635894231480491
0.505050505050505 0.60428329740698
0.555555555555556 0.572759178585609
0.606060606060606 0.54463598994761
0.656565656565657 0.519303520714132
0.707070707070707 0.493794014470279
0.757575757575758 0.468193348552385
0.808080808080808 0.444952103749011
0.858585858585859 0.424146515610083
0.909090909090909 0.403528584214963
0.95959595959596 0.382692210909287
1.01010101010101 0.363503208384421
1.06060606060606 0.346434454727824
1.11111111111111 0.329767026455276
1.16161616161616 0.312806064269156
1.21212121212121 0.296964479795588
1.26262626262626 0.282954634575349
1.31313131313131 0.269484534448605
1.36363636363636 0.255691259128637
1.41414141414141 0.242611781795201
1.46464646464646 0.231096298761082
1.51515151515152 0.220213802058353
1.56565656565657 0.209015769663229
1.61616161616162 0.19821672065113
1.66666666666667 0.188731538517681
1.71717171717172 0.179940534938032
1.76767676767677 0.170870807258425
1.81818181818182 0.161957331117188
1.86868686868687 0.154123817164657
1.91919191919192 0.147019790250038
1.96969696969697 0.139695390372789
2.02020202020202 0.132344223722883
2.07070707070707 0.125855079803232
2.12121212121212 0.120108088957494
2.17171717171717 0.11421359920498
2.22222222222222 0.108159980360172
2.27272727272727 0.102766889609979
2.32323232323232 0.0981080956720768
2.37373737373737 0.0933828613241893
2.42424242424242 0.088409548169719
2.47474747474747 0.0839126282287908
2.52525252525253 0.0801234700283943
2.57575757575758 0.0763514938051761
2.62626262626263 0.0722796756649523
2.67676767676768 0.0685185831509647
2.72727272727273 0.0654219965246082
2.77777777777778 0.0624239074760684
2.82828282828283 0.0591057560072263
2.87878787878788 0.0559522313282459
2.92929292929293 0.0534054676677251
2.97979797979798 0.0510321155487685
3.03030303030303 0.0483447284171805
3.08080808080808 0.0456963725451769
3.13131313131313 0.0435850822577318
3.18181818181818 0.0417124145946875
3.23232323232323 0.0395529304663114
3.28282828282828 0.0373280272571482
3.33333333333333 0.0355613502829606
3.38383838383838 0.0340863030245863
3.43434343434343 0.0323679945580032
3.48484848484848 0.0305012187336072
3.53535353535354 0.0290076815909525
3.58585858585859 0.0278448687900161
3.63636363636364 0.0264940470623813
3.68686868686869 0.0249329233389736
3.73737373737374 0.0236569863593937
3.78787878787879 0.0227360165472286
3.83838383838384 0.021689604115143
3.88888888888889 0.0203916051325627
3.93939393939394 0.0192907382274526
3.98989898989899 0.0185540188266385
4.04040404040404 0.0177576691554123
4.09090909090909 0.0166878583514848
4.14141414141414 0.0157300510809445
4.19191919191919 0.0151309696515082
4.24242424242424 0.0145376282024066
4.29292929292929 0.0136667687488809
4.34343434343434 0.0128284021680237
4.39393939393939 0.012329795981455
4.44444444444444 0.0118986132635703
4.49494949494949 0.0112016760599826
4.54545454545454 0.0104657008902512
4.5959595959596 0.0100385452717702
4.64646464646465 0.00973406510267424
4.6969696969697 0.00918907811581937
4.74747474747475 0.00854345705544303
4.7979797979798 0.00816571882434071
4.84848484848485 0.00795727631824153
4.8989898989899 0.0075444647618705
4.94949494949495 0.00698084686324711
5 0.00663646253097671
};
\addlegendentry{first term}
\addplot [semithick, dashed, mycolor0]
table {%
0 0.993056699867762
0.0505050505050505 0.969341125696071
0.101010101010101 0.917374208008142
0.151515151515152 0.864564950015574
0.202020202020202 0.822861194795716
0.252525252525253 0.785383985077725
0.303030303030303 0.745046812718638
0.353535353535354 0.705090727834846
0.404040404040404 0.67049222082616
0.454545454545455 0.639235837355082
0.505050505050505 0.60714418949821
0.555555555555556 0.575198769703398
0.606060606060606 0.546719622531769
0.656565656565657 0.521092548865897
0.707070707070707 0.495326459216972
0.757575757575758 0.469500751292658
0.808080808080808 0.446073554779987
0.858585858585859 0.425112038345252
0.909090909090909 0.404353804113468
0.95959595959596 0.383396610035775
1.01010101010101 0.36411079060153
1.06060606060606 0.346957838477374
1.11111111111111 0.330212408973
1.16161616161616 0.313187326463818
1.21212121212121 0.297295317786637
1.26262626262626 0.283238474251382
1.31313131313131 0.269725109821223
1.36363636363636 0.255898980116393
1.41414141414141 0.24279250328424
1.46464646464646 0.231249544974775
1.51515151515152 0.220343953664815
1.56565656565657 0.209129993974387
1.61616161616162 0.198315391544737
1.66666666666667 0.18881369821247
1.71717171717172 0.180011356403955
1.76767676767677 0.170934088003433
1.81818181818182 0.162010794792955
1.86868686868687 0.154167717348189
1.91919191919192 0.14705888623963
1.96969696969697 0.139730423626615
2.02020202020202 0.132372650115711
2.07070707070707 0.125878745762628
2.12121212121212 0.120130211982695
2.17171717171717 0.114232667358821
2.22222222222222 0.10817463019502
2.27272727272727 0.102780059927506
2.32323232323232 0.0981209441371806
2.37373737373737 0.0933927983942178
2.42424242424242 0.0884168853757194
2.47474747474747 0.0839203943370911
2.52525252525253 0.0801309190065067
2.57575757575758 0.0763562666196035
2.62626262626263 0.0722834902925959
2.67676767676768 0.0685234609021649
2.72727272727273 0.0654259668578015
2.77777777777778 0.0624259285139701
2.82828282828283 0.0591081914969619
2.87878787878788 0.0559553642578494
2.92929292929293 0.05340707545952
2.97979797979798 0.0510328904889865
3.03030303030303 0.0483467180444686
3.08080808080808 0.0456982410757348
3.13131313131313 0.043585296079199
3.18181818181818 0.0417128523936262
3.23232323232323 0.0395546009100308
3.28282828282828 0.0373288638959202
3.33333333333333 0.0355610140866984
3.38383838383838 0.0340868696621411
3.43434343434343 0.0323691948776813
3.48484848484848 0.0305012365405576
3.53535353535354 0.0290073830519134
3.58585858585859 0.0278456823655358
3.63636363636364 0.0264946499582123
3.68686868686869 0.0249324145066144
3.73737373737374 0.0236570212935867
3.78787878787879 0.0227369381006471
3.83838383838384 0.0216896207633982
3.88888888888889 0.0203909435416038
3.93939393939394 0.0192911454893418
3.98989898989899 0.0185547658628249
4.04040404040404 0.0177572553480824
4.09090909090909 0.0166874252472196
4.14141414141414 0.0157306892428268
4.19191919191919 0.0151312647683614
4.24242424242424 0.0145370310974829
4.29292929292929 0.0136668148997176
4.34343434343434 0.0128290443255536
4.39393939393939 0.0123295519969102
4.44444444444444 0.0118980949584964
4.49494949494949 0.0112021597550793
4.54545454545454 0.0104661273859196
4.5959595959596 0.0100379653021632
4.64646464646465 0.00973382732423678
4.6969696969697 0.00918972597725231
4.74747474747475 0.00854352967206266
4.7979797979798 0.00816511594890959
4.84848484848485 0.00795740517278717
4.8989898989899 0.00754498747288185
4.94949494949495 0.00698055518704713
5 0.00663608757635527
};
\addlegendentry{first 10000 terms}
\end{axis}

\end{tikzpicture}
  \end{subfigure}
  \hfill
  \begin{subfigure}[t]{.49\textwidth}
    \centering
    % This file was created by tikzplotlib v0.9.6.
\begin{tikzpicture}

\begin{axis}[
height=\figureheight,
log basis y={10},
tick align=outside,
tick pos=left,
width=\figurewidth,
x grid style={white!69.0196078431373!black},
xlabel={$r$},
xmin=-0.25, xmax=5.25,
xtick style={color=black},
y grid style={white!69.0196078431373!black},
ylabel={difference in $\kappa(r)$},
ymin=8.40565347922707e-09, ymax=0.0284518722842284,
ymode=log,
ytick style={color=black}
]
\addplot [semithick, mycolor2]
table {%
0 0.0143652638288493
0.0505050505050505 0.0122799378677046
0.101010101010101 0.0104065896562997
0.151515151515152 0.00880452399770681
0.202020202020202 0.00744280676673281
0.252525252525253 0.00632601313498871
0.303030303030303 0.00539759099425541
0.353535353535354 0.00459355585335497
0.404040404040404 0.0039076936501552
0.454545454545455 0.00334160587459065
0.505050505050505 0.00286089209123008
0.555555555555556 0.00243959111778846
0.606060606060606 0.00208363258415833
0.656565656565657 0.00178902815176496
0.707070707070707 0.00153244474669278
0.757575757575758 0.00130740274027263
0.808080808080808 0.00112145103097666
0.858585858585859 0.000965522735169311
0.909090909090909 0.000825219898504981
0.95959595959596 0.000704399126487842
1.01010101010101 0.000607582217109937
1.06060606060606 0.000523383749549577
1.11111111111111 0.000445382517723414
1.16161616161616 0.000381262194661502
1.21212121212121 0.000330837991048438
1.26262626262626 0.000283839676032904
1.31313131313131 0.00024057537261829
1.36363636363636 0.000207720987756066
1.41414141414141 0.000180721489038804
1.46464646464646 0.000153246213693392
1.51515151515152 0.000130151606461432
1.56565656565657 0.000114224311158401
1.61616161616162 9.86708936077108e-05
1.66666666666667 8.21596947892245e-05
1.71717171717172 7.08214659223827e-05
1.76767676767677 6.32807450083039e-05
1.81818181818182 5.34636757672513e-05
1.86868686868687 4.39001835321473e-05
1.91919191919192 3.90959895925358e-05
1.96969696969697 3.50332538260889e-05
2.02020202020202 2.84263928283512e-05
2.07070707070707 2.36659593963462e-05
2.12121212121212 2.21230252014648e-05
2.17171717171717 1.90681538410759e-05
2.22222222222222 1.46498348477658e-05
2.27272727272727 1.31703175272752e-05
2.32323232323232 1.28484651038124e-05
2.37373737373737 9.93707002847066e-06
2.42424242424242 7.33720600046284e-06
2.47474747474747 7.76610830036106e-06
2.52525252525253 7.44897811238987e-06
2.57575757575758 4.77281442740063e-06
2.62626262626263 3.814627643553e-06
2.67676767676768 4.87775120018574e-06
2.72727272727273 3.97033319332607e-06
2.77777777777778 2.02103790165886e-06
2.82828282828283 2.43548973553193e-06
2.87878787878788 3.13292960352207e-06
2.92929292929293 1.60779179488035e-06
2.97979797979798 7.74940217905529e-07
3.03030303030303 1.9896272880765e-06
3.08080808080808 1.86853055787056e-06
3.13131313131313 2.13821467198727e-07
3.18181818181818 4.37798938775213e-07
3.23232323232323 1.67044371935515e-06
3.28282828282828 8.36638771949116e-07
3.33333333333333 -3.36196262260002e-07
3.38383838383838 5.66637554738625e-07
3.43434343434343 1.20031967811712e-06
3.48484848484848 1.78069503378731e-08
3.53535353535354 -2.98539039048779e-07
3.58585858585859 8.13575519766074e-07
3.63636363636364 6.02895831022798e-07
3.68686868686869 -5.08832359264605e-07
3.73737373737374 3.4934193010594e-08
3.78787878787879 9.21553418554988e-07
3.83838383838384 1.66482552708958e-08
3.88888888888889 -6.6159095886073e-07
3.93939393939394 4.07261889118909e-07
3.98989898989899 7.47036186427713e-07
4.04040404040404 -4.13807329929794e-07
4.09090909090909 -4.33104265250284e-07
4.14141414141414 6.38161882216687e-07
4.19191919191919 2.95116853251884e-07
4.24242424242424 -5.97104923717034e-07
4.29292929292929 4.61508366626784e-08
4.34343434343434 6.4215752985583e-07
4.39393939393939 -2.4398454482287e-07
4.44444444444444 -5.18305073879052e-07
4.49494949494949 4.8369509672401e-07
4.54545454545454 4.26495668373952e-07
4.5959595959596 -5.79969606980327e-07
4.64646464646465 -2.37778437469152e-07
4.6969696969697 6.47861432945263e-07
4.74747474747475 7.26166196358263e-08
4.7979797979798 -6.02875431117075e-07
4.84848484848485 1.28854545639392e-07
4.8989898989899 5.22711011347217e-07
4.94949494949495 -2.91676199977442e-07
5 -3.74954621439223e-07
};
\end{axis}

\end{tikzpicture}  
  \end{subfigure}  
  \\[-0.5em]
 \caption{Numerical integration results on the error introduced by approximating the mixture in \cref{eq:triangle_final} using only its first component. The top row figures correspond to the RBF kernel, with $p(w)$ being standard normal distribution. The middle and bottom row figures correspond to the Mat\'ern-$\nicefrac{3}{2}$ and Mat\'ern-$\nicefrac{1}{2}$ kernels, with $p(w)$ being Student-t distribution with degrees of freedom 3 and 1 respectively. The figures on the left compare the kernel values $\kappa(r)$ from equation \cref{eq:triangle_kernel} with $r = x-x'$, while the sum has either only the first term or the 100 first terms included. The figures on the right show the difference between the curves on the left side figures on a logarithmic scale. The figures show that the error introduced by approximating the mixture in \cref{eq:triangle_final} using only its first component is very small when $p(w)$ is one of the distributions used in the experiments.}
  \label{fig:triangle_approx}
\end{figure}

We again obtain a connection between the prior (in this case, a mixture) and the spectral density through the Wiener–Khinchin theorem. In this case, the spectral density also has to admit a mixture density. However, working with a mixture density as prior could be potentially challenging for inference. For this reason, in practise the mixture is approximated only using its first component, since this is a series of rapidly decreasing terms. \cref{fig:triangle_approx} shows simulation results of the error introduced by approximating the mixture using only its first component. Based on these results the error from this approximation appears very small, when $p(w)$ is in the Mat\'ern-class, which is the case in all experiments in this paper.

\subsection{Periodic ReLU Activation} \label{app:periodic_relu}
Finally, we consider a periodic function that captures the spirit of the ReLU activation function in the form of a repeating rectified linear pattern. This can be modelled as the sum of two triangle waves, the second one shifted by half a period. The resulting periodic function is piece-wise linear and defined as:
\begin{equation}
  \psi(x) = \frac{2}{\pi} \bigg( \bigg((x+\frac{\pi}{2}) - \pi \bigg\lfloor \frac{(x+\frac{\pi}{2})}{\pi} + \frac{1}{2} \bigg\rfloor\bigg) (-1)^{\lfloor \frac{(x+\frac{\pi}{2})}{\pi} + \frac{1}{2} \rfloor} +  \bigg(x - \pi \bigg\lfloor \frac{x}{\pi} + \frac{1}{2} \bigg\rfloor\bigg) (-1)^{\lfloor \frac{x}{\pi} + \frac{1}{2} \rfloor} \bigg)  ,
\end{equation}
assuming a period of $p = 2 \pi$.
In the following derivation we will again use a Fourier series approximation of a slightly differently weighted form of the periodic ReLU function as the activation function, \ie,
\begin{equation}
	\sigma(x) = \lim_{n\to\infty}\sum^{n-1}_{k=0} (-1)^{k} \lambda_k^{-2} \left( \sin(\lambda_k (x + \frac{\pi}{2})) + \sin(\lambda_k x) \right) ,
\end{equation}
and $p(b) = \text{Uniform}(-\pi, \pi)$.
Then, the covariance function is given as:
\begin{multline}
	\kappa(x,x') = \int p(w)\, \int p(b)\, \left[ \lim_{n\to\infty} \sum^{n-1}_{k=0} (-1)^k \lambda_k^{-2} \left( \sin(\lambda_k (wx + \frac{\pi}{2} + b)) + \sin(\lambda_k (wx + b)) \right) \right] \\
	\left[ \lim_{n\to\infty} \sum^{n-1}_{j=0} (-1)^j \lambda_j^{-2} \left( \sin(\lambda_j (wx' + \frac{\pi}{2} + b)) + \sin(\lambda_j (wx' + b)) \right) \right] \dd w \dd b .
\end{multline}

As done previously, we will first solve the inner integral by assuming that $k \neq j$, \ie,
\begin{multline}
	\int p(b)\, \frac{(-1)^{k+j}}{\lambda_k^2\lambda_j^2} \left( \sin(\lambda_k (wx + \frac{\pi}{2} + b)) + \sin(\lambda_k (wx + b)) \right) \\
	\left( \sin(\lambda_j (wx' + \frac{\pi}{2} + b)) + \sin(\lambda_j (wx' + b)) \right) \dd b ,
\end{multline}
where we recognise that,
\begin{equation}
	\sin(\lambda_k (wx + \frac{\pi}{2} + b)) = (-1)^k \cos(\lambda_k (wx + b)) ,
\end{equation}
due to the definition of $\lambda_k$, giving us a series of definite integrals:
\begin{align}
	&\int_{-\pi}^{\pi} (-1)^{k+j} \cos(\lambda_k (x + b)) \cos(\lambda_j (x' + b)) \dd b  \nonumber  \\
	&\qquad= \frac{(-1)^{k+j}}{(\lambda_j - \lambda_k) (\lambda_j + \lambda_k)} \Big(2 \sin(\pi \lambda_j) \cos(\pi \lambda_k) (\lambda_k \sin(\lambda_j x') \sin(\lambda_k x) + \lambda_j \cos(\lambda_j x') \cos(\lambda_k x))  \nonumber \\
	&\qquad\qquad- 2 \cos(\pi \lambda_j) \sin(\pi \lambda_k) (\lambda_j \sin(\lambda_j x') \sin(\lambda_k x) + \lambda_k \cos(\lambda_j x') \cos(\lambda_k x))\Big)  \nonumber \\
	&= 0  ,
\end{align}
which cancels out as $\sin(\pi \lambda_k)$ and $\sin(\pi \lambda_j)$ will be zero for every $j$ and every $k$.
Moreover, we have:
\begin{align}
	&\int_{-\pi}^{\pi} \sin(\lambda_k (x + b)) \sin(\lambda_j (x' + b)) \dd b\nonumber \\
	&\qquad= \frac{1}{(\lambda_j - \lambda_k) (\lambda_j + \lambda_k)} \Big(2 \sin(\pi \lambda_j) \cos(\pi \lambda_k) (\lambda_j \sin(\lambda_j x') \sin(\lambda_k x) + \lambda_k \cos(\lambda_j x') \cos(\lambda_k x)) \nonumber\\
	&\qquad\qquad- 2 \cos(\pi \lambda_j) \sin(\pi \lambda_k) (\lambda_k \sin(\lambda_j x') \sin(\lambda_k x) + \lambda_j \cos(\lambda_j x') \cos(\lambda_k x))\Big)\nonumber \\
	&= 0  ,
\end{align}
which again cancels out for the same reason.
Finally, we have:
\begin{align}
	&\int_{-\pi}^{\pi} (-1)^j \sin(\lambda_k (x + b)) \cos(\lambda_j (x' + b)) \dd b \nonumber\\
	&= \frac{(-1)^j}{(\lambda_j - \lambda_k) (\lambda_j + \lambda_k)} \Big(2 (\cos(\pi \lambda_j) \sin(\pi \lambda_k) (\lambda_j \sin(\lambda_j x') \cos(\lambda_k x) - \lambda_k \cos(\lambda_j x') \sin(\lambda_k x)) \nonumber\\
	&\qquad+ \sin(\pi \lambda_j) \cos(\pi \lambda_k) (\lambda_j \cos(\lambda_j x') \sin(\lambda_k x) - \lambda_k \sin(\lambda_j x') \cos(\lambda_k x)))\Big) \nonumber\\
	&= 0 ,
\end{align}
and the same can be shown for $\int_{-\pi}^{\pi} (-1)^k \sin(\lambda_j (x' + b)) \cos(k (x + b)) \dd b$.

Therefore, our derivations simplify to containing only terms where $k=j$:
\begin{multline}
	\kappa(x,x') = \int p(w)\, \int p(b)\, \Bigg[\lim_{n\to\infty}\sum^{n-1}_{k=0} (-1)^{2k} \lambda_k^{-4} \left( (-1)^k \cos(\lambda_k (wx + b)) + \sin(\lambda_k (wx + b)) \right)  \\
	\left( (-1)^k \cos(\lambda_k (wx' + b)) + \sin(\lambda_k (wx' + b)) \right) \Bigg] \dd w \dd b  .
\end{multline}

Here we can again use the dominated convergence theorem to take the limit outside of the integrals:
\begin{multline}
	\kappa(x,x') = \lim_{n\to\infty}\int p(w)\, \sum^{n-1}_{k=0} (-1)^{2k} \lambda_k^{-4}\int p(b)\, \Bigg[ \left( (-1)^k \cos(\lambda_k (wx + b)) + \sin(\lambda_k (wx + b)) \right)  \\
	\left( (-1)^k \cos(\lambda_k (wx' + b)) + \sin(\lambda_k (wx' + b)) \right) \Bigg] \dd w \dd b .
\end{multline}

We can now start by calculating the inner integral:
\begin{align}
&\int p(b)\, \Big[ \left( (-1)^k \cos(\lambda_k (wx + b)) + \sin(\lambda_k (wx + b)) \right) \nonumber \\
& \qquad \left( (-1)^k \cos(\lambda_k (wx' + b)) + \sin(\lambda_k (wx' + b)) \right) \Big] \dd b \nonumber \\
&=\int p(b)\, \Bigg[ \cos(\lambda_k (wx + b))\cos(\lambda_k (wx' + b)) + \sin(\lambda_k (wx + b))\sin(\lambda_k (wx' + b))  \nonumber \\
&\qquad+ (-1)^k \cos(\lambda_k (wx + b))\sin(\lambda_k (wx' + b)) \nonumber \\
&\qquad+ (-1)^k \cos(\lambda_k (wx' + b))\sin(\lambda_k (wx+ b))\Bigg] \dd b \nonumber \\
&=\int p(b)\, \Bigg[ \cos(\lambda_k (wx + b) - \lambda_k (wx' + b)) + (-1)^k  \sin(\lambda_k (wx + b) + \lambda_k (wx' + b)) \Bigg] \dd b \nonumber \\
&=\int p(b)\,  \cos(\lambda_k w (x - x') \dd b + (-1)^k  \underbrace{\int p(b)\, \sin(\lambda_k w (x + x') + 2\lambda_k b)) \dd b}_{=0} \nonumber \\
&= \cos(\lambda_k w (x - x'))
\end{align}

Inserting this result into the original equation results in:
\begin{equation}
	\kappa(x,x') = \lim_{n\to\infty}\int p(w)\, \sum^{n-1}_{k=0}  \lambda_k^{-4} \cos( w \lambda_k (x - x')) \dd w  ,
\end{equation}

where we recognise that this the exact same equation encountered in \cref{app:triangle_wave} and the terms in the series decrease rapidly towards zero, thus, allowing us to approximate the covariance using only the first term of the sum which gives us (see \cref{app:triangle_wave} for analysis on the approximation error): 
\begin{equation}
	\kappa(x,x') = \int p(w)\, \cos(w(x - x')) \dd w = \int p(w)\, e^{\imag w(x - x')} \dd w  .
\end{equation}

Finally, by letting $r = x-x'$, we find that we again recover the spectral density decomposition of a stationary process given by the Wiener–Khinchin theorem, \ie,
\begin{equation}
\kappa(r) = \int p(w) \,  e^{\imag w r} \dd w  ,
\end{equation}
which concludes the derivation.

\section{Derivations of Correspondence to the Mat\'ern family}\label{app:Matern}
Relating to \cref{sec:kernel-func} in the main paper, we provide the following derivation.
The spectral density of the Mat\'ern family (\cf, \cref{eq:matern}) for the 1D case is given as:
\begin{equation}
S_{\text{Mat.}}(w)= 2\sqrt{\pi}\frac{\Gamma(\nu+\frac{1}{2})}{\Gamma(\nu)}(2\nu)^{\nu}\left( 2\nu+w^2\right)^{-(\nu+\frac{1}{2})}  ,
\end{equation}
where $\nu$ is a smoothness parameter, and we assume unit magnitude and $\ell=1$ for simplicity. Note that we intentionally used $w$ instead of $\omega$ to highlight the correspondence to the prior on the weights.
By assuming the prior $p(w)$ to follow a Student-$t$ distribution \ie,
\begin{equation}
p(w) =\frac{\Gamma(\frac{u+1}{2})}{\sqrt{u\pi}\Gamma(\frac{u}{2})} \left(1+\frac{w^2}{u}\right)^{-\frac{u+1}{2}} ,
\end{equation}
and setting the degree of freedom $u = 2\nu$, we can recover the spectral density of the Mat\'ern family, \ie,
\begin{align}
    p(w) &= \frac{\Gamma(\frac{2\nu+1}{2})}{\sqrt{2\nu\pi}\Gamma(\frac{2\nu}{2})} \left(1+\frac{w^2}{2\nu}\right)^{-\frac{2\nu+1}{2}} &&=\frac{\Gamma(\nu+\frac{1}{2})}{\sqrt{2\nu\pi}\Gamma(\nu)}\left( \frac{1}{2\nu}(2\nu+w^2)\right)^{-(\nu+\frac{1}{2})} \nonumber \\
&=\frac{\Gamma(\nu+\frac{1}{2})}{\sqrt{2\nu\pi}\Gamma(\nu)}(2\nu)^{\nu+\frac{1}{2}}\left( 2\nu+w^2\right)^{-(\nu+\frac{1}{2})} &&=\frac{\Gamma(\nu+\frac{1}{2})}{\sqrt{\pi}\Gamma(\nu)}(2\nu)^{\nu}\left( 2\nu+w^2\right)^{-(\nu+\frac{1}{2})} \nonumber \\
&=\frac{1}{2\pi}2\sqrt{\pi}\frac{\Gamma(\nu+\frac{1}{2})}{\Gamma(\nu)}(2\nu)^{\nu}\left( 2\nu+w^2\right)^{-(\nu+\frac{1}{2})} &&= \frac{1}{2\pi}S_{\text{Mat.}}(w)  .
\end{align}

A summary of the priors on the network weights corresponding to the spectral density of kernels in the Mat\'ern family is given in \cref{tab:priors}.

\begin{table}[t]
\caption{Priors on the weights corresponding to the spectral density of kernels in the Mat\'ern family.}
\label{tab:priors}
\centering
\begin{tabularx}{.5\textwidth}{ll}
	\toprule
	\sc Kernel function & \sc Prior distribution \\
	\midrule
	Exponential ($\nu=\nicefrac{1}{2}$) & $\cauchy(0, 1)$ \\
	Mat\'ern-$\nu$ & $\tdist(2\nu)$ \\
    RBF ($\nu \to \infty$) & $\gaussian(0, 1)$ \\
	\bottomrule
\end{tabularx}
\end{table}%

\section{Additional Insights}\label{app:insights}
Fourier features, the sinusoidal (or other periodic) basis, and the special dual relationship between stationary covariance functions and the associated spectral density,  have been common building blocks in machine learning and signal processing methods for decades. We review connections to Random Fourier features and Fourier methods in GP models.

\subsection{Connection to Random Fourier Features}
Random Fourier features \cite{rahimi2008} are a popular technique for randomized, low-dimensional approximations of kernel functions.
They are motivated by the observation that the spectral density of a RBF covariance function of a Gaussian process prior can be estimated using Monte Carlo integration. Let $\vomega \sim \N(0,1)$ be Gaussian distributed and $\zeta_{\vomega}(\vx) = \exp\left(\imag \, \vomega\T \vx \right)$, then 
\begin{equation}
\E_{\vomega}[ \zeta_{\vomega}(\vx), \zeta_{\vomega}(\vx')^* ] = \int p(\vomega) \, \exp\left(\imag \, \vomega\T \vect{r}\right) \dd \vomega ,
\end{equation}
where $^*$ denotes the complex conjugate, is an estimator of the covariance function.

Assuming that $\vomega$ and $\vx$ are real-valued, let $b$ be a value drawn uniformly from $[-\pi, \pi]$, and by replacing the complex exponentials with cosines $z_{\vomega}(\vx) = \sqrt 2 \cos(\vomega\T\vx + b)$ we obtain:
\begin{align}
    \E_{\vomega}[ z_{\vomega}(\vx) z_{\vomega}(\vx')] &= \E_{\vomega}[ \cos(\vomega\T(\vx - \vx')) ] \nonumber \\
    &= \E_{\vomega}[ \E_{b}[ \sqrt 2 \cos(\vomega\T\vx + b) \,  \sqrt 2 \cos(\vomega\T\vx' + b) \cbar \vomega]]   && \text{(Euler's formula)} \nonumber \\
    &\approx \frac{1}{K} \sum_{j=1}^K \sqrt 2 \cos(\vomega_k\T\vx + b_k) \, \sqrt 2 \cos(\vomega_k\T\vx' + b_k) .
\end{align}
We recognise that the estimate used in random Fourier features is similar to the covariance function of a finite-width single hidden layer network with sinusoidal activation function.

\citet{rahimi2008} additionally proposed a representation using a composition of a sin and a cosine function, which motivated the use of the sine--cosine activation function in this paper. \citet{SutherlandS15} later showed that this representation reduces the variance of the estimates when used to approximate the RBF kernel.

\subsection{Fourier Methods in Gaussian Process Models}
The Fourier duality for stationary covariance functions has been extensively leveraged in Gaussian process models. For gridded inputs, this duality directly allows for leveraging FFT methods to speed up inference and learning. In particular, the sparse spectrum GP (SSGP) method \cite{lazaro2010sparse} uses the spectral representation of the covariance function to draw random samples from the spectrum. These samples are used to represent the GP on a trigonometric basis, \ie, 
\begin{equation}
  \vectb{\phi}(\vx) = 
  \big( 
   \cos(2\pi \, \vs_1^\top \vx) ~ \sin(2\pi \, \vs_1^\top \vx) ~ \ldots 
   \cos(2\pi \, \vs_h^\top \vx) ~ \sin(2\pi \, \vs_h^\top \vx)
  \big) ,
\end{equation}
where the spectral points $\vect{s}_r, r = 1,2,\ldots,h$ ($2h=m$) are sampled from the spectral density of the stationary covariance function (following the normalization convention used in the original paper). The covariance function corresponding to the SSGP can be given in the form (\cf, Mercer's theorem):
\begin{equation} \label{eq:k_SSGP}
  \kappa(\vect{x},\vect{x}') 
  \approx \frac{2}{m} \, \vectb{\phi}(\vx) \, \vectb{\phi}^\top(\vect{x}') 
  = \frac{1}{h} 
    \sum_{r=1}^h \cos\left( 2\pi \, \vs_r^\top (\vect{x}-\vect{x}') \right) .
\end{equation}
This representation of the sparse spectrum method converges to the full GP in the limit of the number of spectral points going to infinity, and is the preferred formulation of the method in one or two dimensions (discussed in \cite{lazaro2010sparse}). We can interpret the SSGP method in \cref{eq:k_SSGP} as a Monte Carlo approximation of the Wiener--Khinchin integral. This interpretation also gives rise to alternative methods for GPs: the methods by \citet{hensman2018variational} and \citet{solin2020hilbert} can be interpreted as a dense/structural (quadrature) approximation to the same integral.

However, for high-dimensional inputs, the SSGP method requires optimization of the frequencies rather than relying on sampling, which is problematic (as discussed in \cite{lazaro2010sparse}), resulting in a tendency to overfit, and loses the interpretation of the original GP prior in the model. Note that these issues have been addressed in subsequent work. As in SSGPs, our method can be seen as a sampling/optimization approach to a rank-reduced approximation of the induced prior covariance structure. However, the connection we derived retains the role of the prior throughout and generalizes the interpretation of the role of the periodic basis.

\begin{figure}[t]
  \centering\scriptsize
  \begin{subfigure}[b]{.48\textwidth}
    \centering\tiny

    \pgfplotsset{axis on top,scale only axis, width=\figurewidth,height=\figureheight,yticklabel={\empty},xticklabel={\empty},ticks=none}
    \setlength{\figurewidth}{.3\textwidth}
    \setlength{\figureheight}{\figurewidth}  
    \begin{minipage}[t]{.05\textwidth}
      \centering
      \tikz\node[rotate=90]{Highly non-stationary};
    \end{minipage} 
    \hfill
    \begin{minipage}[t]{.3\textwidth}
      \centering
      % This file was created by tikzplotlib v0.9.8.
\begin{tikzpicture}

\begin{axis}[
height=\figureheight,
tick align=outside,
tick pos=left,
width=\figurewidth,
x grid style={white!69.0196078431373!black},
xmin=-3, xmax=3,
xtick style={color=black},
y dir=reverse,
y grid style={white!69.0196078431373!black},
ymin=-3, ymax=3,
ytick style={color=black}
]
\addplot graphics [includegraphics cmd=\pgfimage,xmin=-6.10927152317881, xmax=5.8112582781457, ymin=3.99337748344371, ymax=-3.95364238410596] {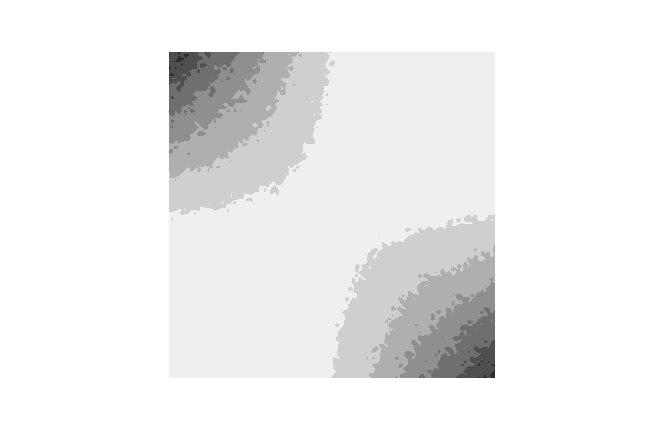};
\end{axis}

\end{tikzpicture}\\
      ArcCos-1 $\sim$ ReLU
    \end{minipage} 
    \hfill 
    \begin{minipage}[t]{.3\textwidth}
      \centering
      % This file was created by tikzplotlib v0.9.8.
\begin{tikzpicture}

\begin{axis}[
height=\figureheight,
tick align=outside,
tick pos=left,
width=\figurewidth,
x grid style={white!69.0196078431373!black},
xmin=-3, xmax=3,
xtick style={color=black},
y dir=reverse,
y grid style={white!69.0196078431373!black},
ymin=-3, ymax=3,
ytick style={color=black}
]
\addplot graphics [includegraphics cmd=\pgfimage,xmin=-6.10927152317881, xmax=5.8112582781457, ymin=3.99337748344371, ymax=-3.95364238410596] {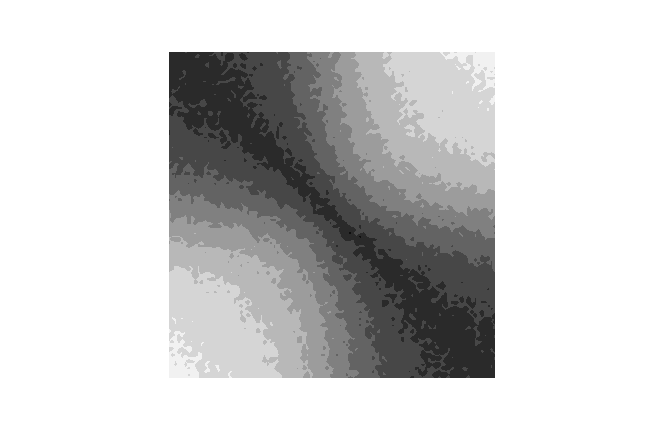};
\end{axis}

\end{tikzpicture}\\
      ArcCos-0 $\sim$ Step
    \end{minipage} 
    \hfill     
    \begin{minipage}[t]{.3\textwidth}
      \centering
      % This file was created by tikzplotlib v0.9.8.
\begin{tikzpicture}

\begin{axis}[
height=\figureheight,
tick align=outside,
tick pos=left,
width=\figurewidth,
x grid style={white!69.0196078431373!black},
xmin=-3, xmax=3,
xtick style={color=black},
y dir=reverse,
y grid style={white!69.0196078431373!black},
ymin=-3, ymax=3,
ytick style={color=black}
]
\addplot graphics [includegraphics cmd=\pgfimage,xmin=-6.10927152317881, xmax=5.8112582781457, ymin=3.99337748344371, ymax=-3.95364238410596] {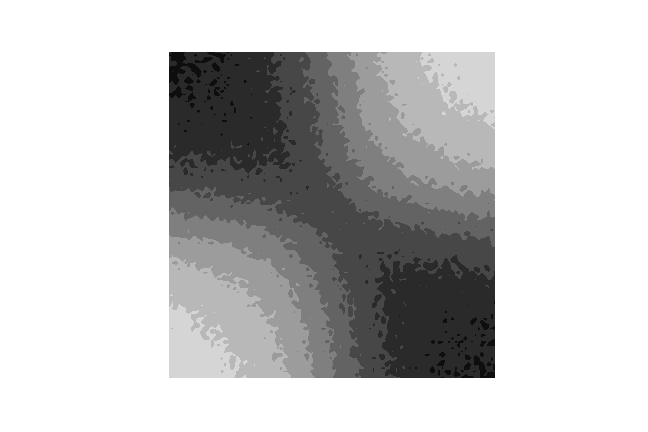};
\end{axis}

\end{tikzpicture}\\
      NN $\sim$ Sigmoid
    \end{minipage}\\[1em]
    \begin{minipage}[t]{.05\textwidth}
      \centering
      \tikz\node[rotate=90]{\hspace*{1em}Locally stationary};
    \end{minipage} 
    \hfill    
    \begin{minipage}[t]{.3\textwidth}
      \centering
      % This file was created by tikzplotlib v0.9.8.
\begin{tikzpicture}

\begin{axis}[
height=\figureheight,
tick align=outside,
tick pos=left,
width=\figurewidth,
x grid style={white!69.0196078431373!black},
xmin=-3, xmax=3,
xtick style={color=black},
y dir=reverse,
y grid style={white!69.0196078431373!black},
ymin=-3, ymax=3,
ytick style={color=black}
]
\addplot graphics [includegraphics cmd=\pgfimage,xmin=-6.10927152317881, xmax=5.8112582781457, ymin=3.99337748344371, ymax=-3.95364238410596] {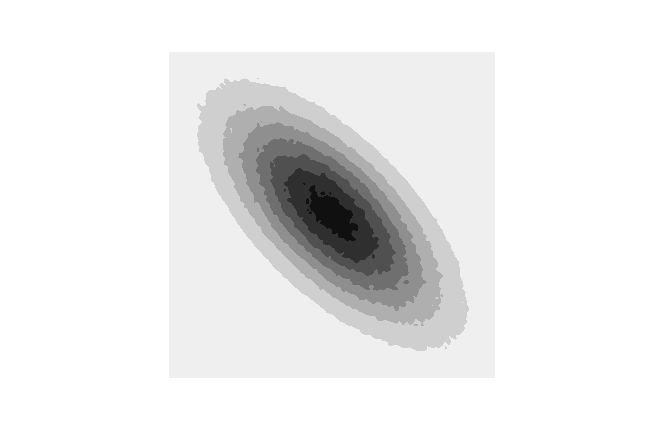};
\end{axis}

\end{tikzpicture}\\
      RBF-NN (local)
    \end{minipage} 
    \hfill 
    \begin{minipage}[t]{.3\textwidth}
      \centering
      % This file was created by tikzplotlib v0.9.8.
\begin{tikzpicture}

\begin{axis}[
height=\figureheight,
tick align=outside,
tick pos=left,
width=\figurewidth,
x grid style={white!69.0196078431373!black},
xmin=-3, xmax=3,
xtick style={color=black},
y dir=reverse,
y grid style={white!69.0196078431373!black},
ymin=-3, ymax=3,
ytick style={color=black}
]
\addplot graphics [includegraphics cmd=\pgfimage,xmin=-6.10927152317881, xmax=5.8112582781457, ymin=3.99337748344371, ymax=-3.95364238410596] {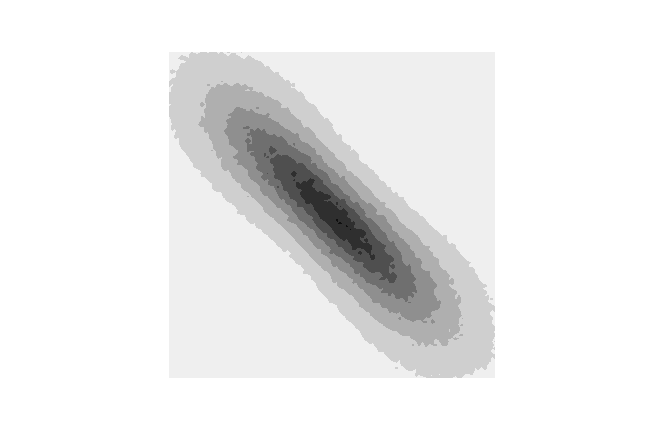};
\end{axis}

\end{tikzpicture}\\
      Mat\'ern-$\nicefrac{3}{2}$-NN (local)
    \end{minipage} 
    \hfill     
    \begin{minipage}[t]{.3\textwidth}
      \centering
      % This file was created by tikzplotlib v0.9.8.
\begin{tikzpicture}

\begin{axis}[
height=\figureheight,
tick align=outside,
tick pos=left,
width=\figurewidth,
x grid style={white!69.0196078431373!black},
xmin=-3, xmax=3,
xtick style={color=black},
y dir=reverse,
y grid style={white!69.0196078431373!black},
ymin=-3, ymax=3,
ytick style={color=black}
]
\addplot graphics [includegraphics cmd=\pgfimage,xmin=-6.10927152317881, xmax=5.8112582781457, ymin=3.99337748344371, ymax=-3.95364238410596] {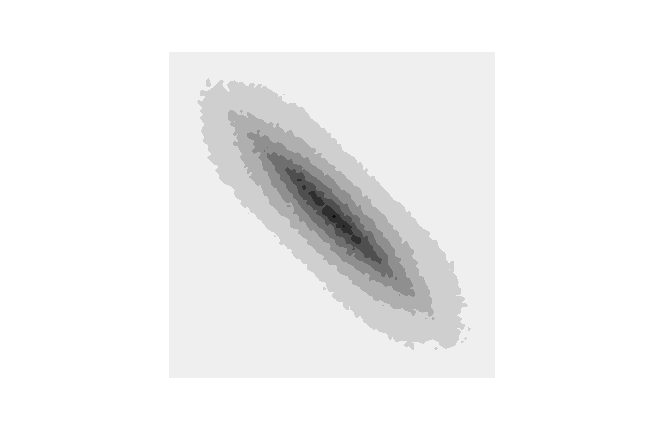};
\end{axis}

\end{tikzpicture}\\
      Exp-NN (local)
    \end{minipage}\\[1em]  
    \begin{minipage}[t]{.05\textwidth}
      \centering
      \tikz\node[rotate=90]{\hspace*{2.5em}Stationary};
    \end{minipage} 
    \hfill      
    \begin{minipage}[t]{.3\textwidth}
      \centering
      % This file was created by tikzplotlib v0.9.8.
\begin{tikzpicture}

\begin{axis}[
height=\figureheight,
tick align=outside,
tick pos=left,
width=\figurewidth,
x grid style={white!69.0196078431373!black},
xmin=-3, xmax=3,
xtick style={color=black},
y dir=reverse,
y grid style={white!69.0196078431373!black},
ymin=-3, ymax=3,
ytick style={color=black}
]
\addplot graphics [includegraphics cmd=\pgfimage,xmin=-6.10927152317881, xmax=5.8112582781457, ymin=3.99337748344371, ymax=-3.95364238410596] {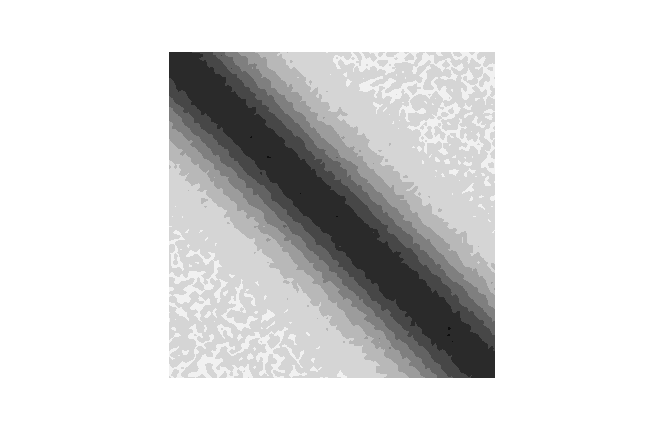};
\end{axis}

\end{tikzpicture}\\
      RBF
    \end{minipage} 
    \hfill 
    \begin{minipage}[t]{.3\textwidth}
      \centering
      % This file was created by tikzplotlib v0.9.8.
\begin{tikzpicture}

\begin{axis}[
height=\figureheight,
tick align=outside,
tick pos=left,
width=\figurewidth,
x grid style={white!69.0196078431373!black},
xmin=-3, xmax=3,
xtick style={color=black},
y dir=reverse,
y grid style={white!69.0196078431373!black},
ymin=-3, ymax=3,
ytick style={color=black}
]
\addplot graphics [includegraphics cmd=\pgfimage,xmin=-6.10927152317881, xmax=5.8112582781457, ymin=3.99337748344371, ymax=-3.95364238410596] {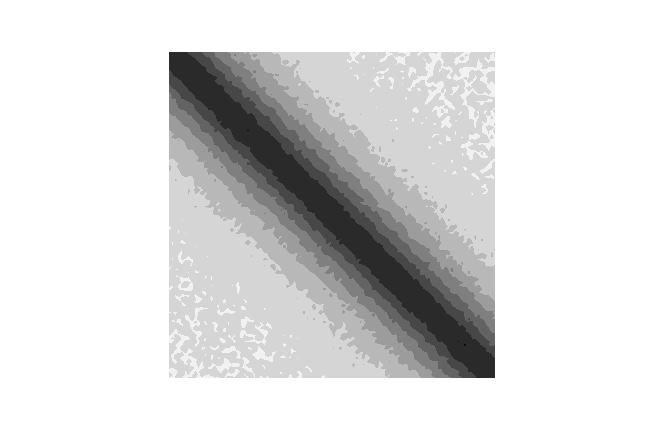};
\end{axis}

\end{tikzpicture}\\
      Mat\'ern-$\nicefrac{3}{2}$
    \end{minipage} 
    \hfill     
    \begin{minipage}[t]{.3\textwidth}
      \centering
      % This file was created by tikzplotlib v0.9.8.
\begin{tikzpicture}

\begin{axis}[
height=\figureheight,
tick align=outside,
tick pos=left,
width=\figurewidth,
x grid style={white!69.0196078431373!black},
xmin=-3, xmax=3,
xtick style={color=black},
y dir=reverse,
y grid style={white!69.0196078431373!black},
ymin=-3, ymax=3,
ytick style={color=black}
]
\addplot graphics [includegraphics cmd=\pgfimage,xmin=-6.10927152317881, xmax=5.8112582781457, ymin=3.99337748344371, ymax=-3.95364238410596] {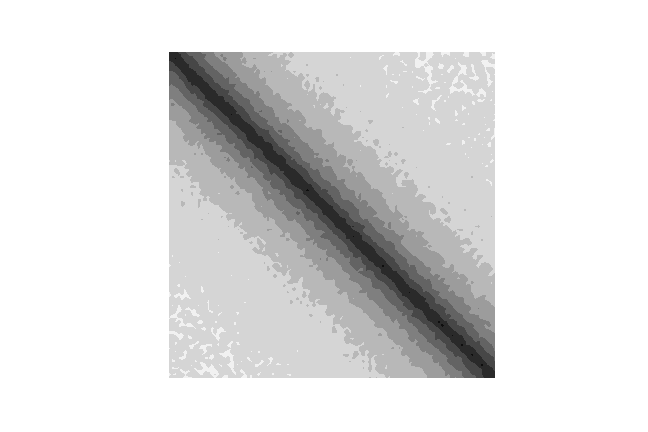};
\end{axis}

\end{tikzpicture}\\
      Exponential
    \end{minipage}
    \caption{Gram matrices}
    \label{fig:gram-finite}
  \end{subfigure}
  \hfill
  \begin{subfigure}[b]{.48\textwidth}
\centering\tiny

    \pgfplotsset{axis on top,scale only axis,width=\figurewidth,height=\figureheight,yticklabel={\empty},xticklabel={\empty},ticks=none}
    \setlength{\figurewidth}{.3\textwidth}
    \setlength{\figureheight}{\figurewidth}  
    \begin{minipage}[t]{.3\textwidth}
      \centering
      \input{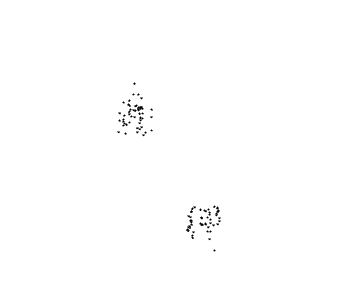}\\
      ArcCos-1 $\sim$ ReLU
    \end{minipage} 
    \hfill 
    \begin{minipage}[t]{.3\textwidth}
      \centering
      \input{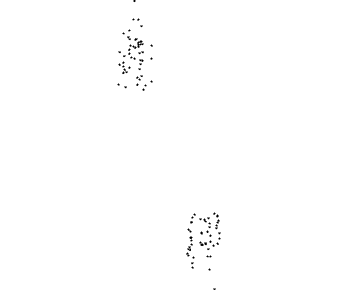}\\
      ArcCos-0 $\sim$ Step
    \end{minipage} 
    \hfill     
    \begin{minipage}[t]{.3\textwidth}
      \centering
      \input{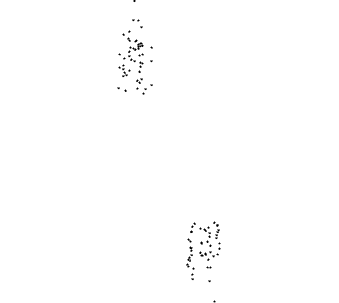}\\
      NN $\sim$ Sigmoid
    \end{minipage}\\[1em]
    \begin{minipage}[t]{.3\textwidth}
      \centering
      \input{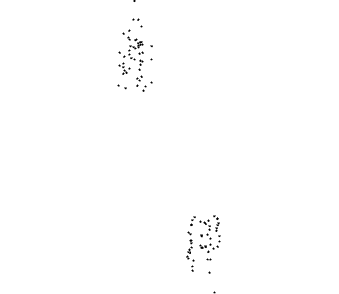}\\
      RBF-NN (local)
    \end{minipage} 
    \hfill 
    \begin{minipage}[t]{.3\textwidth}
      \centering
      \input{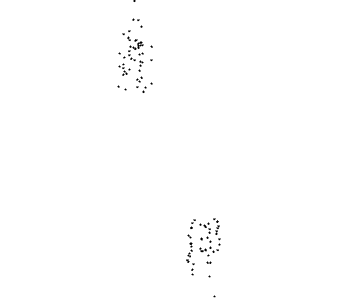}\\
      Mat\'ern-$\nicefrac{3}{2}$-NN (local)
    \end{minipage} 
    \hfill     
    \begin{minipage}[t]{.3\textwidth}
      \centering
      \input{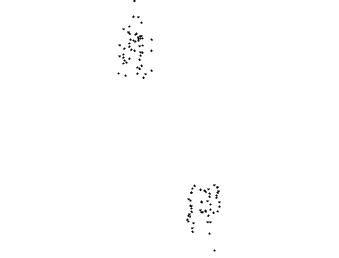}\\
      Exp-NN (local)
    \end{minipage}\\[1em] 
    \begin{minipage}[t]{.3\textwidth}
      \centering
      \input{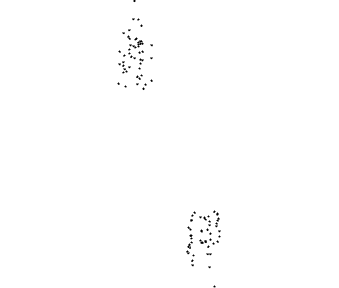}\\
      RBF
    \end{minipage} 
    \hfill 
    \begin{minipage}[t]{.3\textwidth}
      \centering
      \input{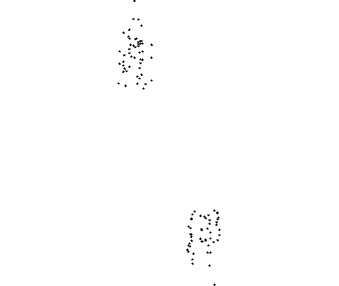}\\
      Mat\'ern-$\nicefrac{3}{2}$
    \end{minipage} 
    \hfill     
    \begin{minipage}[t]{.3\textwidth}
      \centering
      \input{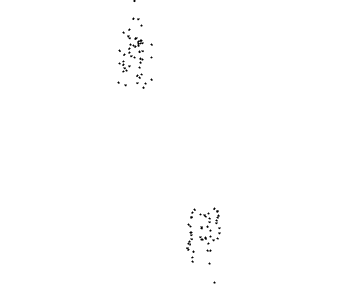}\\
      Exponential
    \end{minipage}
    \caption{Regression results with corresponding prior}
    \label{fig:regression-finite}
  \end{subfigure}
  \caption{{\bf Left:}~Gram matrices (evaluated for $\hat{\kappa}(x,x')$ with 1000 Monte Carlo samples) corresponding to the prior covariance induced by different finite-width NNs (10 hidden units). {\bf Right:}~1D regression results corresponding to the model induced by the prior in the left-hand panels, showing the posterior and predictive 95\% intervals of a BNN with ten hidden units obtained through sampling with dynamic HMC for 5000 iterations.}
  \label{fig:gram-regression-finite}
\end{figure}

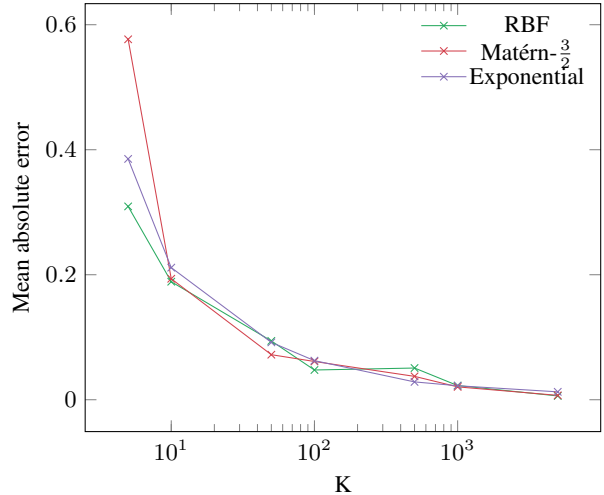
\begin{figure}
\centering\footnotesize
\begin{tikzpicture}
	\begin{semilogxaxis}[
		xlabel=K,
		ylabel={Mean absolute error},
    xtick={10, 100, 1000}
    ]
	\addplot[color=mycolor1,mark=x] coordinates {
		(5,0.309305)
		(10,0.18886)
		(50,0.0940903)
		(100,0.0475922)
		(500,0.0507275)
		(1000,0.0226722)
		(5000,0.00609189)
	};
	\addplot[color=mycolor2,mark=x] coordinates {
		(5,0.576658)
		(10,0.193571)
		(50,0.0720818)
		(100,0.0612155)
		(500,0.0374068)
		(1000,0.0205772)
		(5000,0.00702051)
	};
	\addplot[color=mycolor3,mark=x] coordinates {
		(5,0.385309)
		(10,0.211374)
		(50,0.0917824)
		(100,0.0625705)
		(500,0.0285404)
		(1000,0.0224113)
		(5000,0.0126504)
	};
	\legend{RBF,Mat\'ern-$\frac{3}{2}$,Exponential};
	\end{semilogxaxis}
\end{tikzpicture}
\caption{Simulation estimates of the error between the Gram matrix of the limiting process and the Gram matrix of a finite width model with sinusoidal activation functions under increasing number of hidden units ($K$).}
  \label{fig:gram-comparison}
\end{figure}

\section{Experiment Details and Additional Results}
\label{app:details}

Our main contribution is linking globally stationary GP priors to a corresponding infinite-width NN with one hidden layer by specifying activation functions and priors on weights and biases. Most applications, however, require NNs with a deeper structure. In the experiments, this is achieved by considering the preceding NN structure as a feature extractor. The layer containing the model specification is added as the last hidden fully connected layer after the feature extractor part. This means that the activation function and priors for weights and biases only apply to the last hidden layer of the full NN architecture. It is important to note here that we do not mean the final output layer that produces the class weights or regression output when we refer to the last hidden layer, but the linear layer preceding the output layer. For NN architectures that would not otherwise have a fully connected hidden layer before the output layer, one such layer is added. We refer to this last hidden layer that contains the model implementation as the `model layer'. The weights and biases in the model layer are initialized from the prior distributions defined by the model. \cref{tab:priors} lists prior distributions corresponding to different globally stationary priors. Prior distributions for locally stationary and non-stationary ReLU models are $\gaussian(0, 1)$ following the results from \cite{meronen2020stationary}. Many of these priors require using periodic activation functions instead of the ReLU, which could be expected to affect trainability. In our experiments, we observed that sometimes training neural networks with periodic activation functions can be a bit harder compared to ReLU networks, but usually, slightly adjusting the learning rate is enough to solve these issues. We expect that the issues with trainability were minor as the periodic activation functions are only used in a single layer of the NN, while other parts of the NN are still using ReLU.

Although our theoretical result considers an infinitely wide NN, we observed in practice that a good result can be obtained if the number of hidden units is sufficiently larger than the dimensionality of the preceding feature space. To avoid using an excessively large number of hidden units, we have added a bottleneck layer for NN architectures for which the feature space after the feature-extraction part would have high dimensionality. This way, we could use a model layer with roughly 100 times more hidden units than the number of dimensions in the preceding feature space.

In the derivations for the prior on the weights, we expected $\ell = 1$ for simplicity. It may be necessary for the model to have a different lengthscale depending on the data in practice. We implemented this by multiplying the weights by a lengthscale parameter before multiplying the input by these weights in the forward pass. This way, we do not need to adjust the prior distribution for the weights as the weights themselves will still stay in the space corresponding to $\ell = 1$. The lengthscale parameter is added as a trainable parameter with a prior of $\ell \sim \Gamma(\alpha=2, \beta=0.5)$, which is weakly informative.

The model layer produces hidden features which are mapped to outputs using the linear output layer. The weights for the output layer are initialized from $\gaussian(0, \nicefrac{1}{K})$. The output $f(x)$ dimensionality equals the number of classes in classification tasks and one in regression tasks. Based on this output, the data likelihood needs to be calculated for the loss function. For regression tasks the data likelihood is $\gaussian(y-f(x) \mid 0, s^2)$, where $s$ is the standard deviation of the measurement noise which we include into the model as a trainable parameter with prior $\Gamma(\alpha=0.5, \beta=1)$. For classification, the data likelihood is calculated by applying the softmax function on $f(x)$ to map the outputs to probabilities and then choosing the class probability corresponding to the class $y$. 

To train the model, we construct a loss function that considers both the data likelihood and the prior distributions. Since our goal is to fit an approximate posterior distribution on the model parameters, we use the Bayes formula to obtain a loss function that is directly proportional to the posterior distribution: 
\begin{equation}
	p(w,b,\ell,s \mid y,x) \propto p(y \mid x)\,p(w)\,p(b)\,p(\ell)\,p(s)  .
\end{equation}
For optimization purposes we take a negative logarithm of the product of priors and data likelihood, which also changes this to a minimization problem, resulting in the following loss function: 
\begin{equation}
	\mathcal{L} = -\log p(y \mid x) -\log p(w) -\log p(b) -\log p(\ell) -\log p(s) .
\end{equation}
Here $p(y \mid x)$ is the data likelihood described above (for classification the cross entropy loss directly gives $-\log p(y|x)$ from the outputs $f(x)$), $p(w)$ is the prior on weights, $p(b)$ is the prior on biases, $p(\ell)$ is the prior on lengthscale, and $p(s)$ is the prior on measurement noise variance (for classification $-\log p(s)=0$). For NN architectures that have a feature-extractor part preceding the model layer, we also include standard L2-regularization on the parameters of the feature-extractor network to prevent the model from completely bypassing the defined priors by learning extreme values for feature-extractor network parameters. For globally stationary models $p(b) = \Uni(-\pi,\pi)$, meaning that the bias term is defined on a constrained space. We therefore optimise $\hat{b} \in \mathbb{R}$, which is defined on an unconstrained space, and apply the map/link function $b = 2\pi \, \mathrm{sigmoid}(\hat{b}) - \pi$.

The calculations for obtaining the experiment results were mostly performed using computer resources within the Aalto Science-IT project. These resources included both CPU nodes and GPU nodes (NVIDIA V100 and Tesla P100). Some results were also calculated on local GPU resources (NVIDIA RTX 2080). All of the utilized data sets are publicly available and widely used, and none of them contain any personally identifiable information or offensive content. The illustrative toy BNN examples are implemented using HMC in {Turing.jl}~\cite{Ge2018}, GP regression results use GPflow~\cite{GPflow:2017}, and all other experiments are implemented using {PyTorch}~\cite{PyTorch}.

\subsection{Illustrative Toy Examples}
\label{app:toy}
\begin{figure*}[!t]
  \scriptsize
  \pgfplotsset{hide axis,scale only axis,width=\figurewidth,height=\figureheight}
  \setlength{\figurewidth}{.159\textwidth}
  \setlength{\figureheight}{\figurewidth}
  \begin{tikzpicture}
    \tikzstyle{box} = [minimum width=.99\figurewidth,draw=none,inner sep=1pt,rounded corners=1pt]
    \node[box, rotate=90] at (\figurewidth,.55\figurewidth) {Sin Activation};      
  \end{tikzpicture} 
  \begin{subfigure}[b]{\figurewidth}
    \centering Exponential (mean)
    \includegraphics[width=\linewidth]{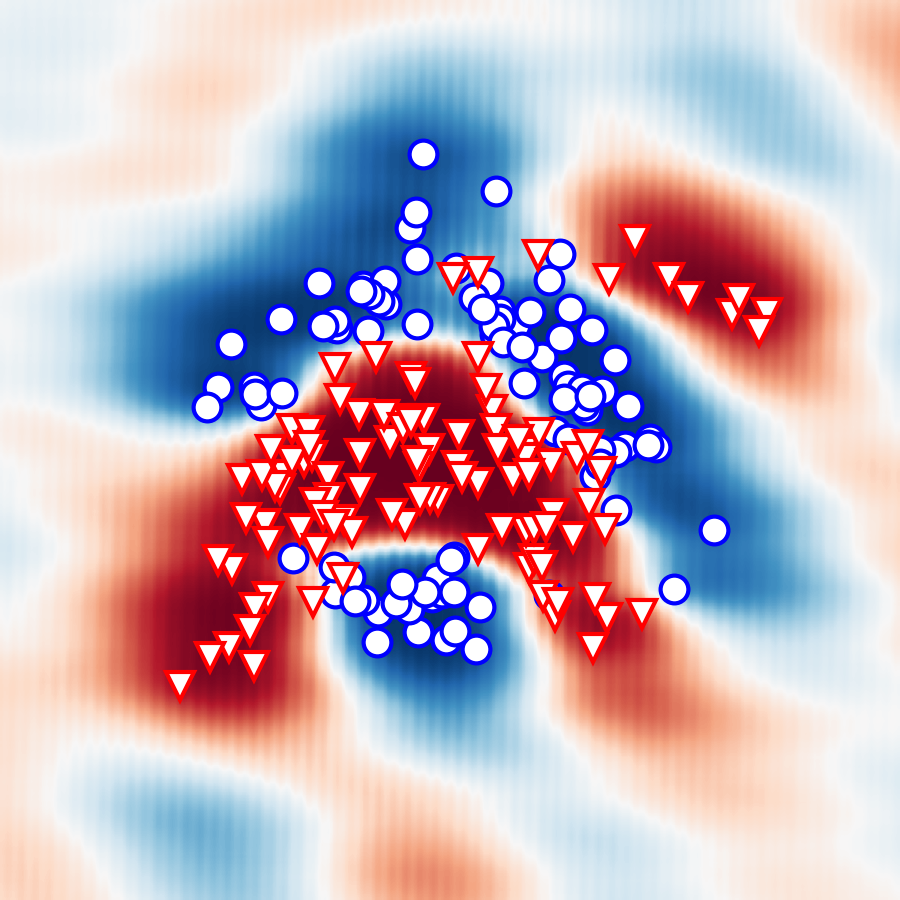}
  \end{subfigure}
  \hfill
    \begin{subfigure}[b]{\figurewidth}
    \centering Exponential (variance)
    \includegraphics[width=\linewidth]{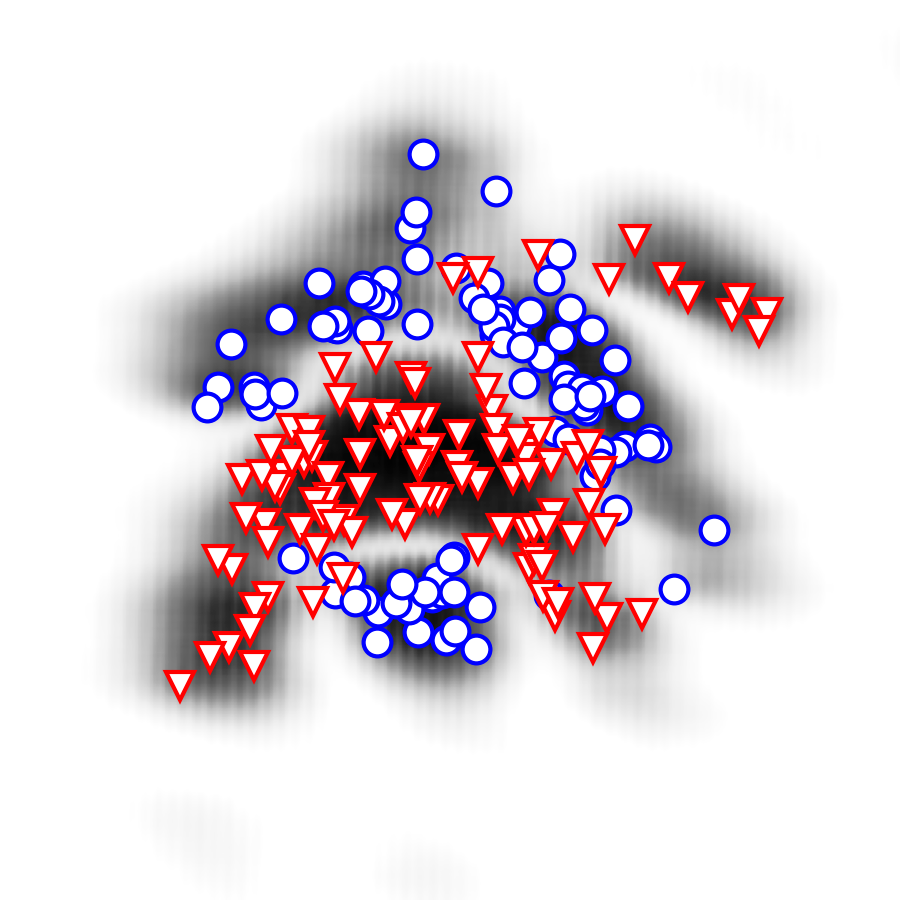}
  \end{subfigure}
  \hfill
    \begin{subfigure}[b]{\figurewidth}
    \centering Mat\'ern-$\frac{3}{2}$ (mean)
    \includegraphics[width=\linewidth]{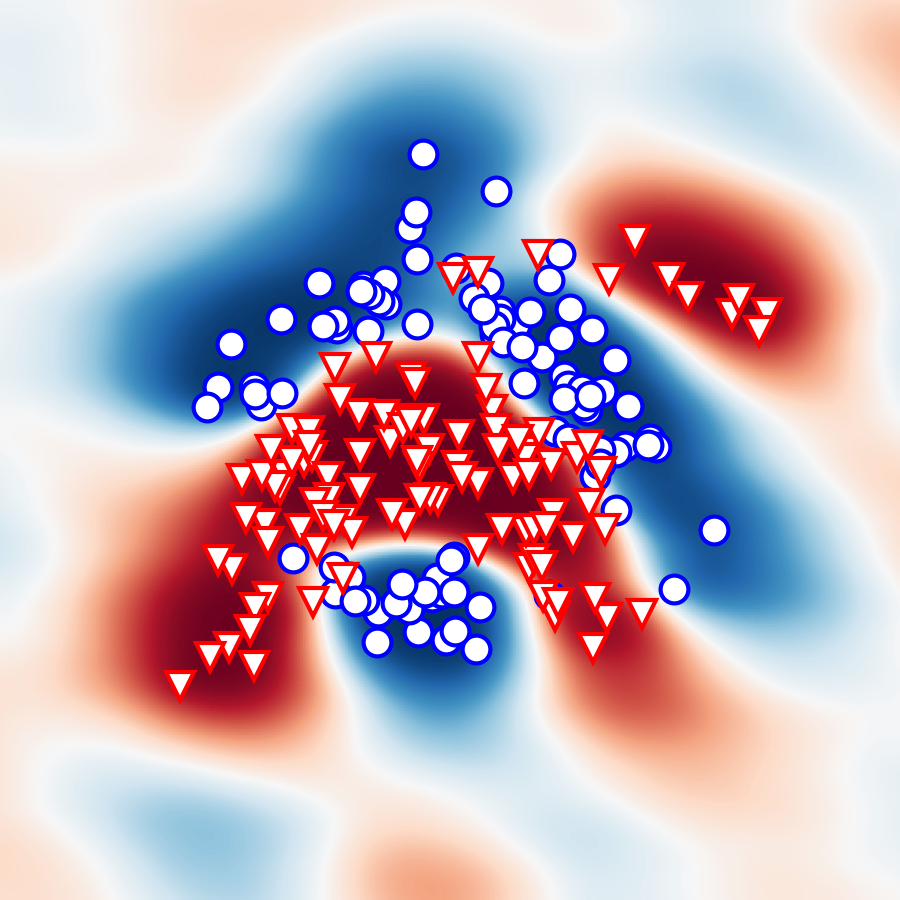}
  \end{subfigure}
  \hfill
  \begin{subfigure}[b]{\figurewidth}
    \centering Mat\'ern-$\frac{3}{2}$ (variance)
    \includegraphics[width=\linewidth]{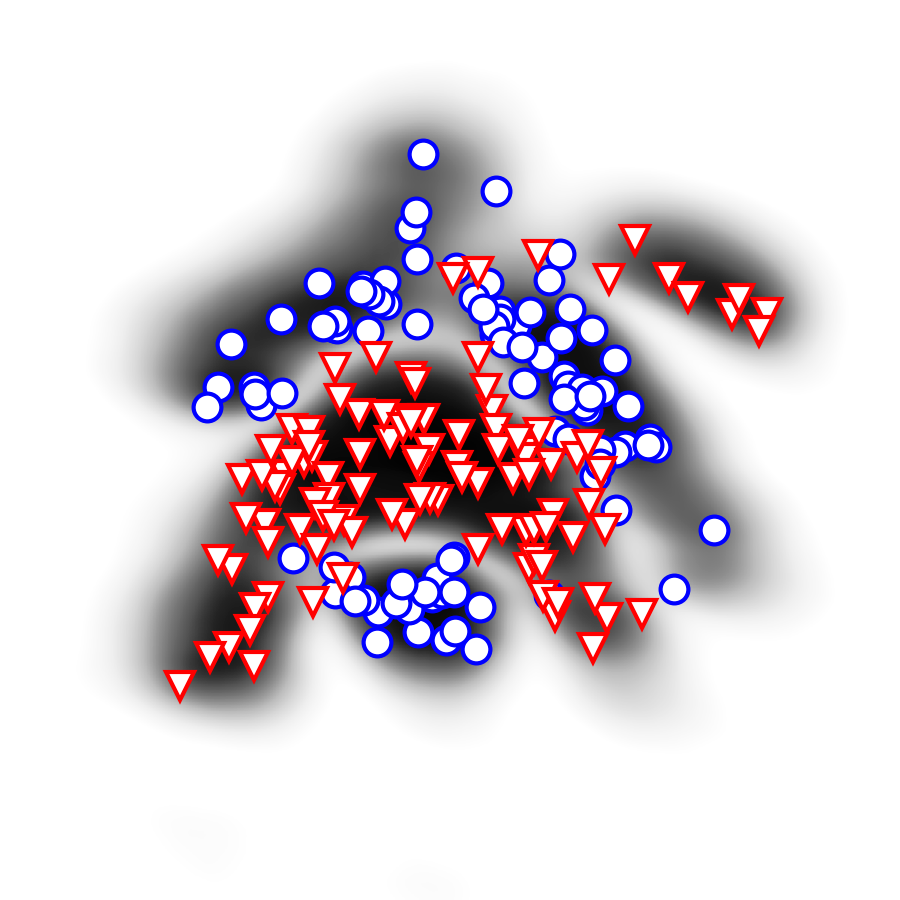}
  \end{subfigure}
  \hfill
  \begin{subfigure}[b]{\figurewidth}
    \centering RBF (mean)
    \includegraphics[width=\linewidth]{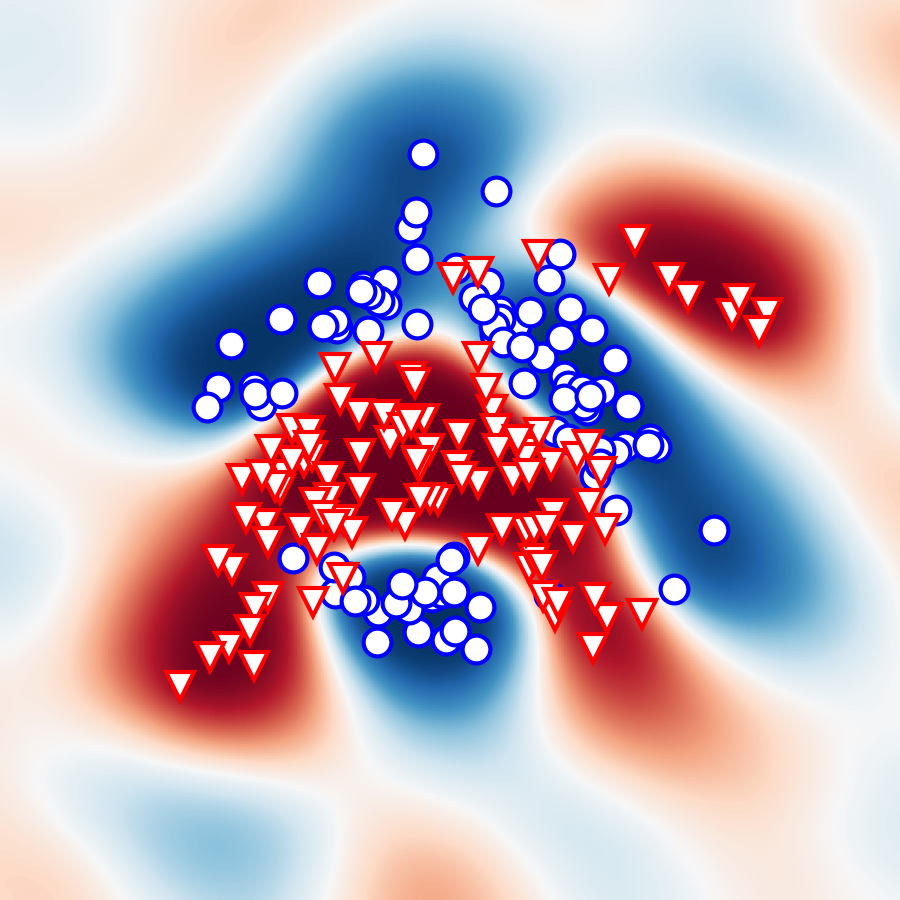}
  \end{subfigure}
  \hfill
  \begin{subfigure}[b]{\figurewidth}
    \centering RBF (variance)
    \includegraphics[width=\linewidth]{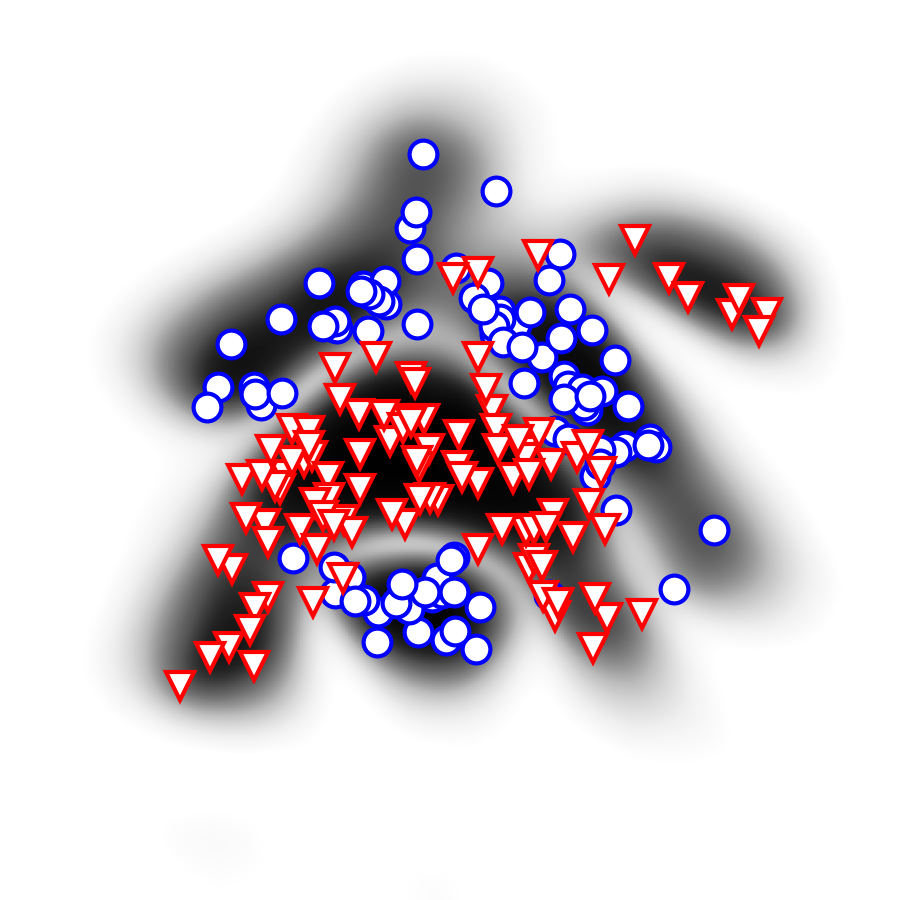}
  \end{subfigure}\\[0.5em]
  \begin{tikzpicture}
    \tikzstyle{box} = [minimum width=.99\figurewidth,inner sep=1pt,rounded corners=1pt]
    \node[box, rotate=90] at (\figurewidth,.55\figurewidth) {Sine Cosine};      
  \end{tikzpicture} 
  \begin{subfigure}[t]{\figurewidth}
     \includegraphics[width=\linewidth]{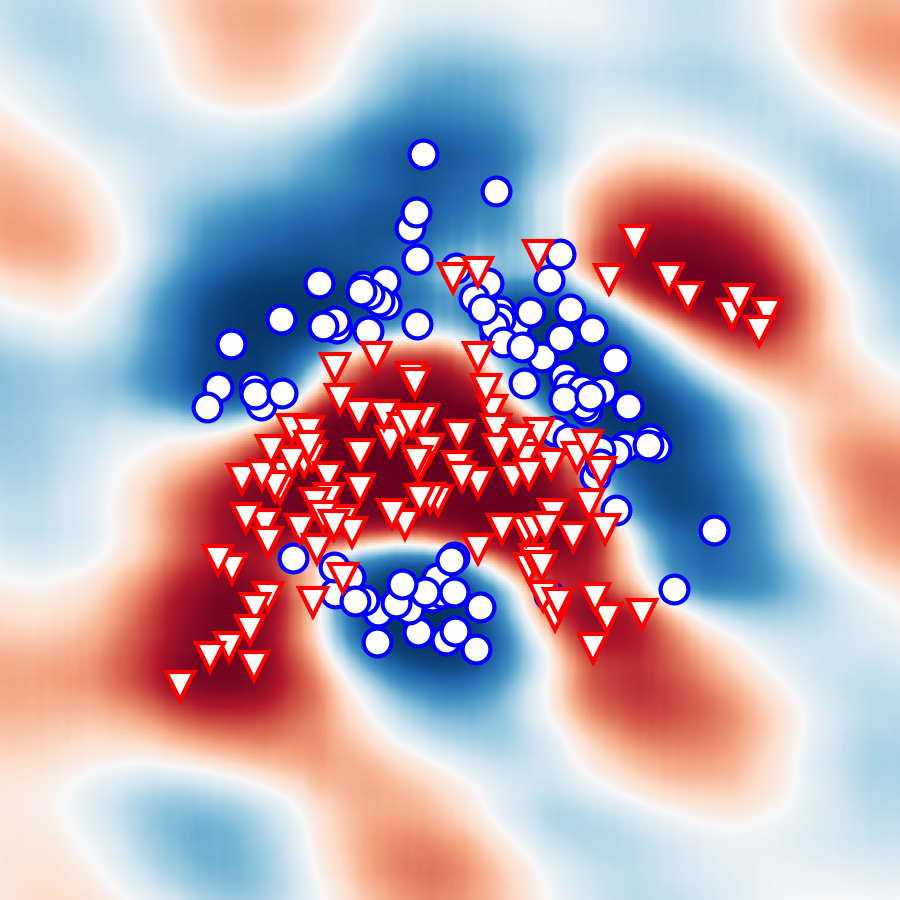}
  \end{subfigure}
  \hfill
    \begin{subfigure}[t]{\figurewidth}
     \includegraphics[width=\linewidth]{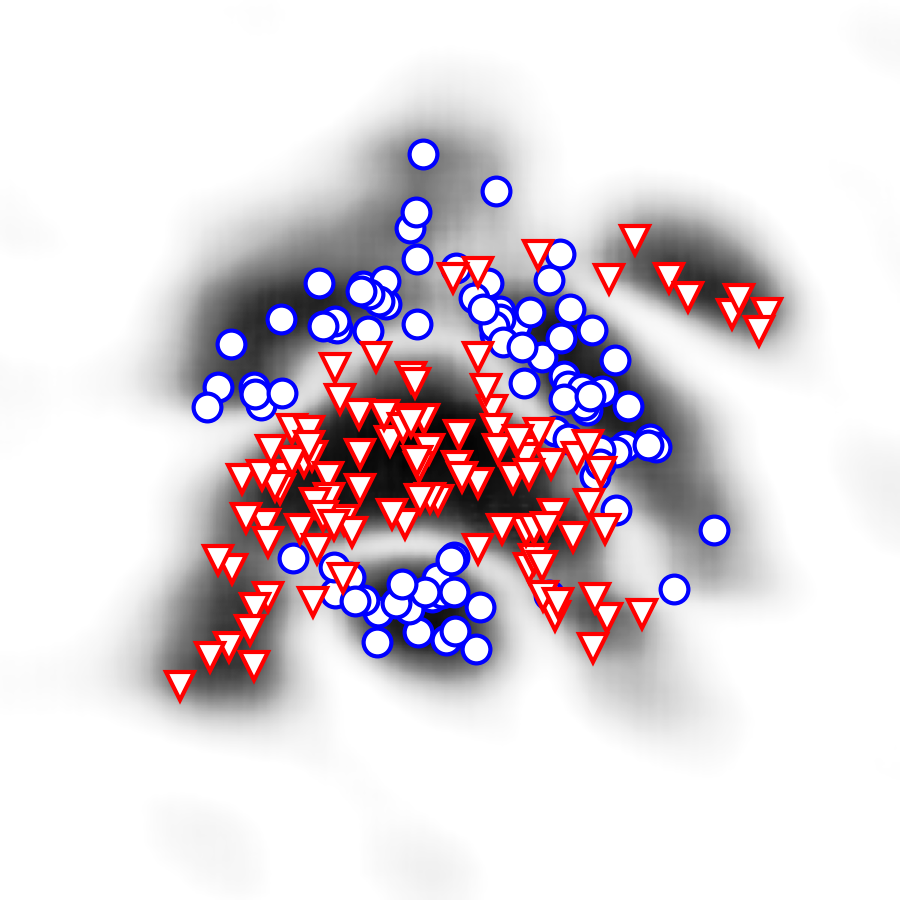}
  \end{subfigure}
  \hfill
    \begin{subfigure}[t]{\figurewidth}
     \includegraphics[width=\linewidth]{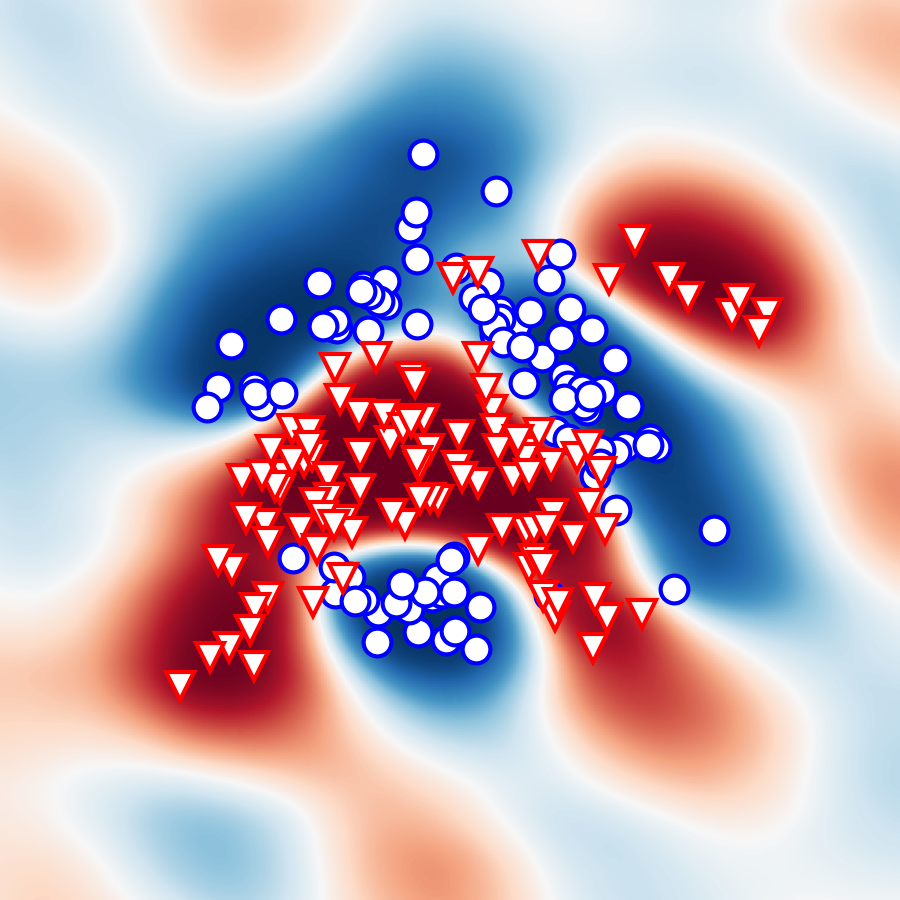}
  \end{subfigure}
      \begin{subfigure}[t]{\figurewidth}
     \includegraphics[width=\linewidth]{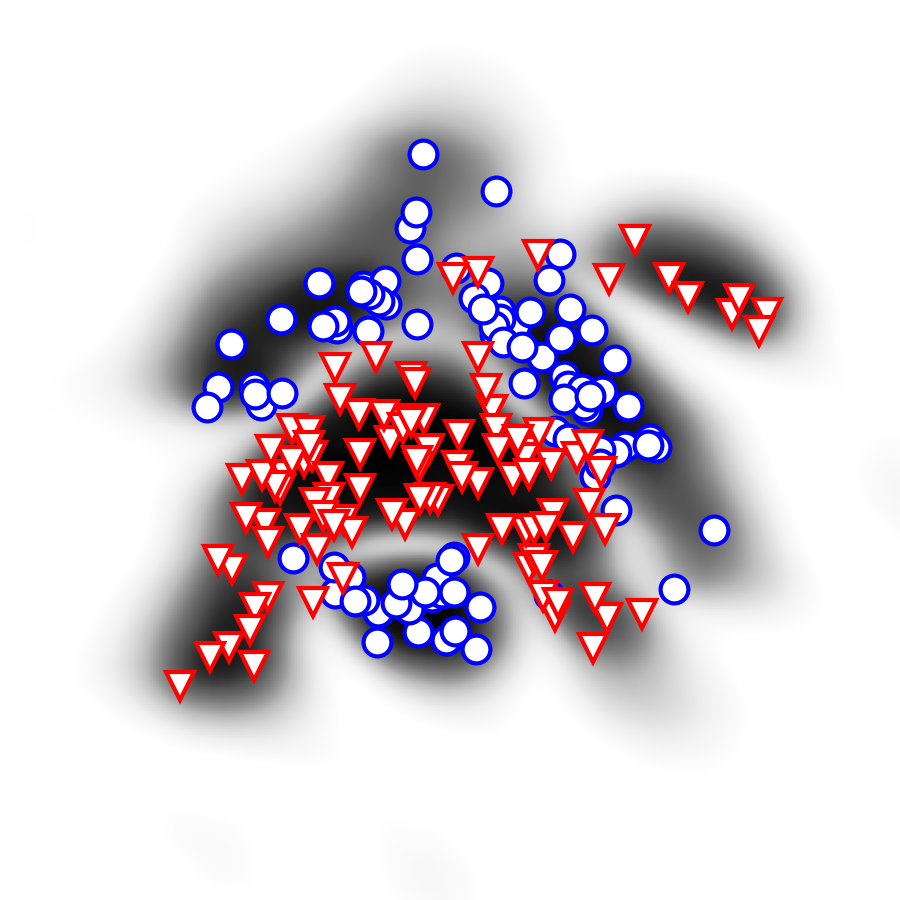}
  \end{subfigure}
    \hfill
  \begin{subfigure}[t]{\figurewidth}
    \includegraphics[width=\linewidth]{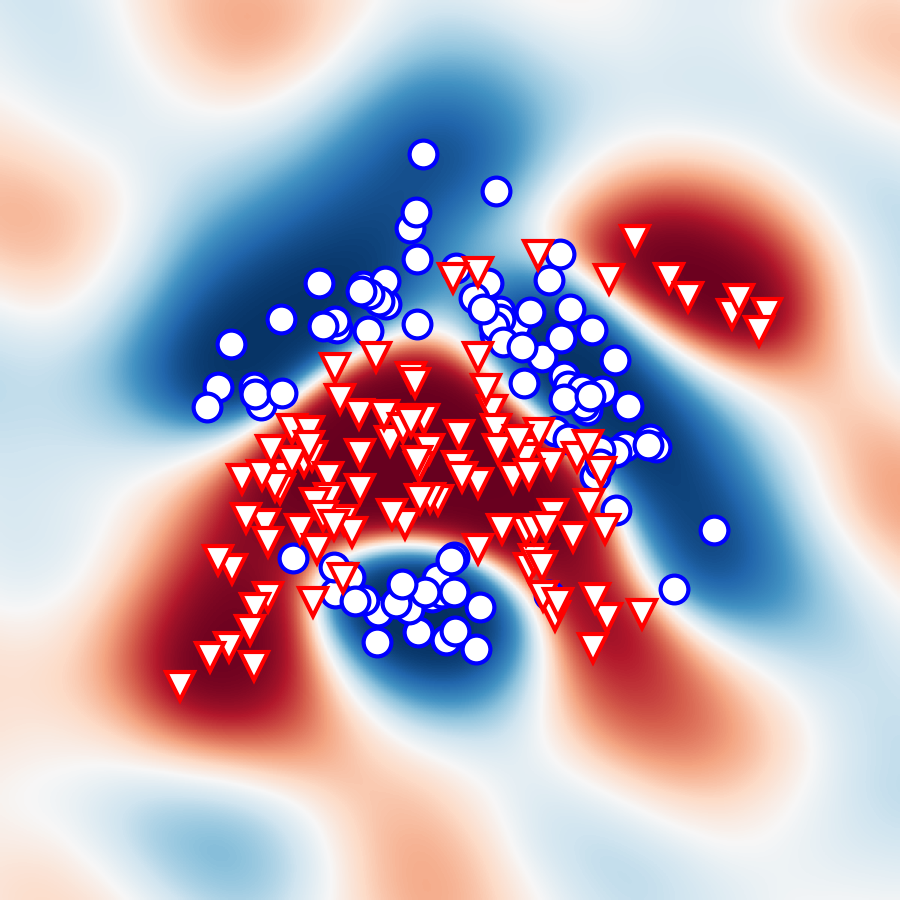}
  \end{subfigure}
   \begin{subfigure}[t]{\figurewidth}
    \includegraphics[width=\linewidth]{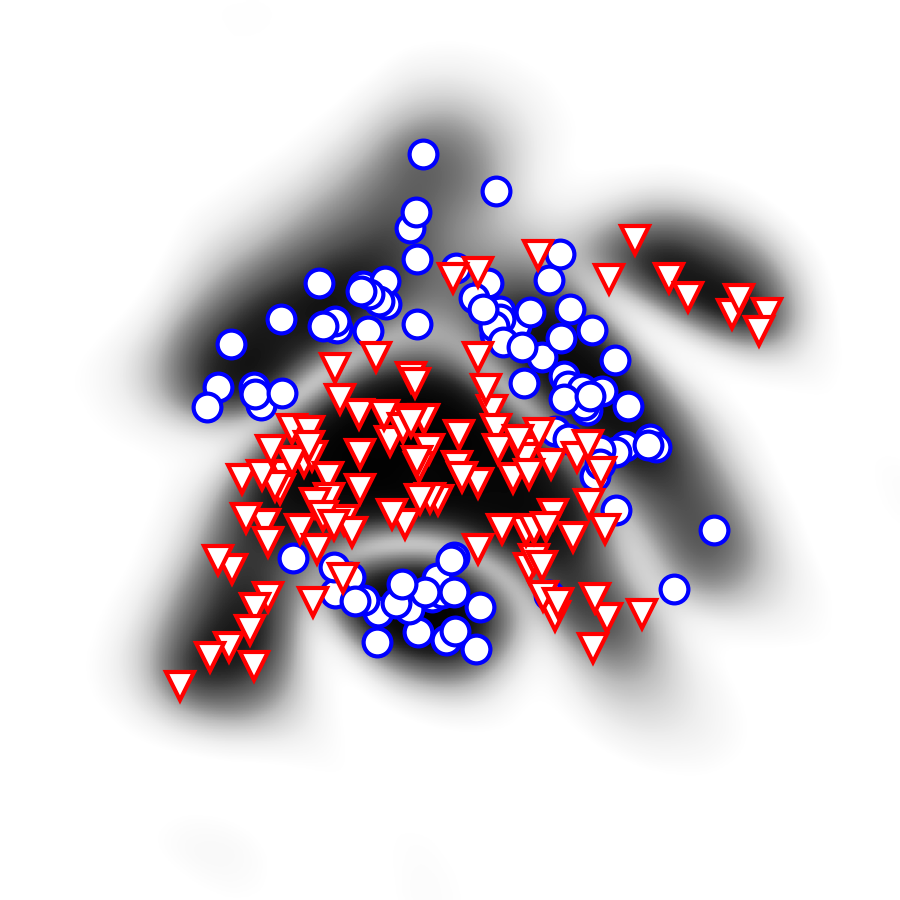}
  \end{subfigure}\\[0.5em]
  \begin{tikzpicture}
    \tikzstyle{box} = [minimum width=.99\figurewidth,inner sep=1pt,rounded corners=1pt]
    \node[box, rotate=90] at (\figurewidth,.55\figurewidth) {Triangle Wave};      
  \end{tikzpicture} 
  \begin{subfigure}[t]{\figurewidth}
     \includegraphics[width=\linewidth]{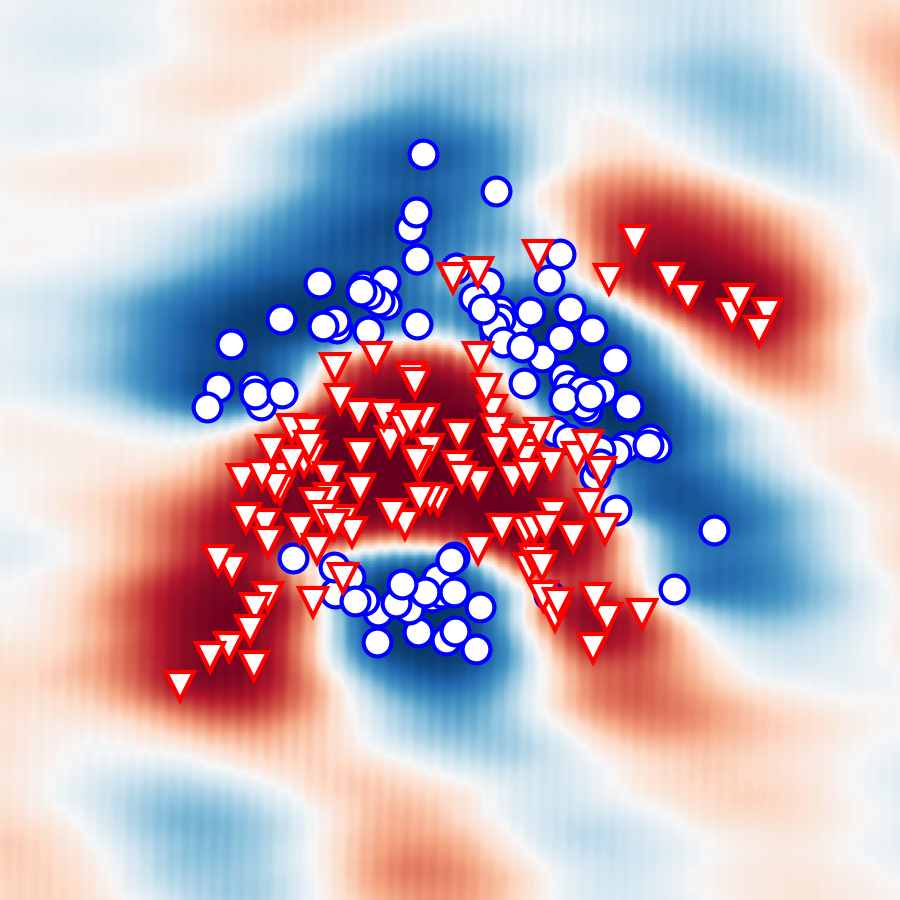}
  \end{subfigure}
  \hfill
    \begin{subfigure}[t]{\figurewidth}
     \includegraphics[width=\linewidth]{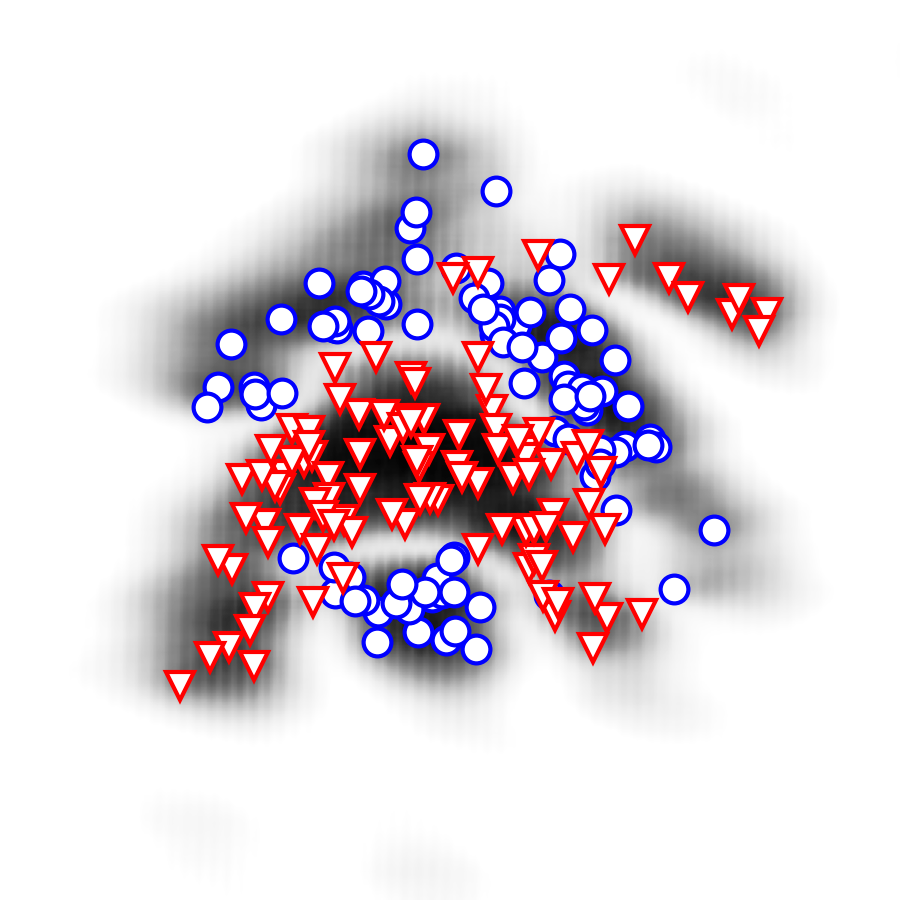}
  \end{subfigure}
  \hfill
    \begin{subfigure}[t]{\figurewidth}
     \includegraphics[width=\linewidth]{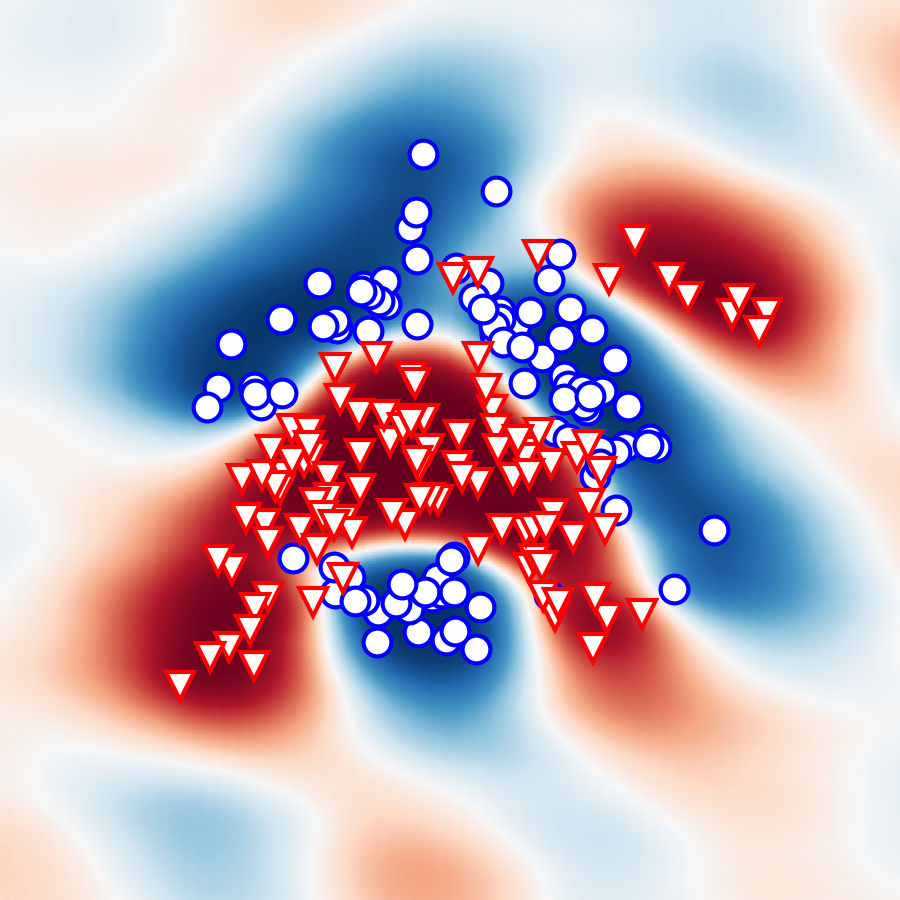}
  \end{subfigure}
      \begin{subfigure}[t]{\figurewidth}
     \includegraphics[width=\linewidth]{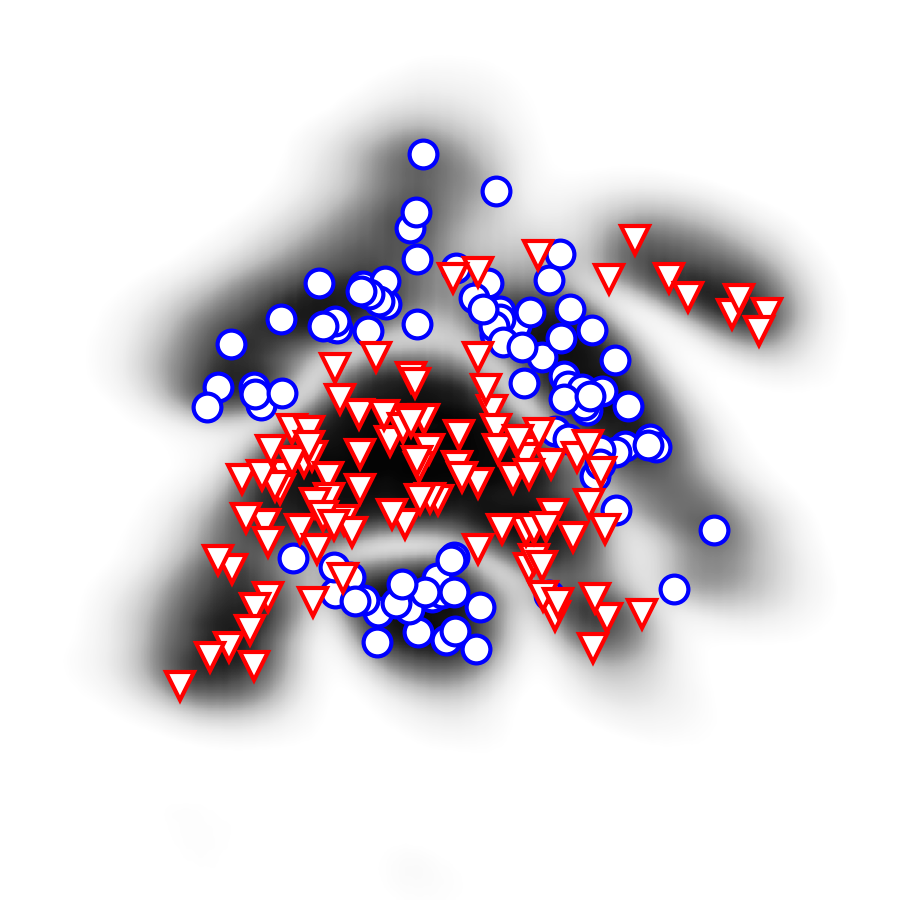}
  \end{subfigure}
  \hfill
  \begin{subfigure}[t]{\figurewidth}
    \includegraphics[width=\linewidth]{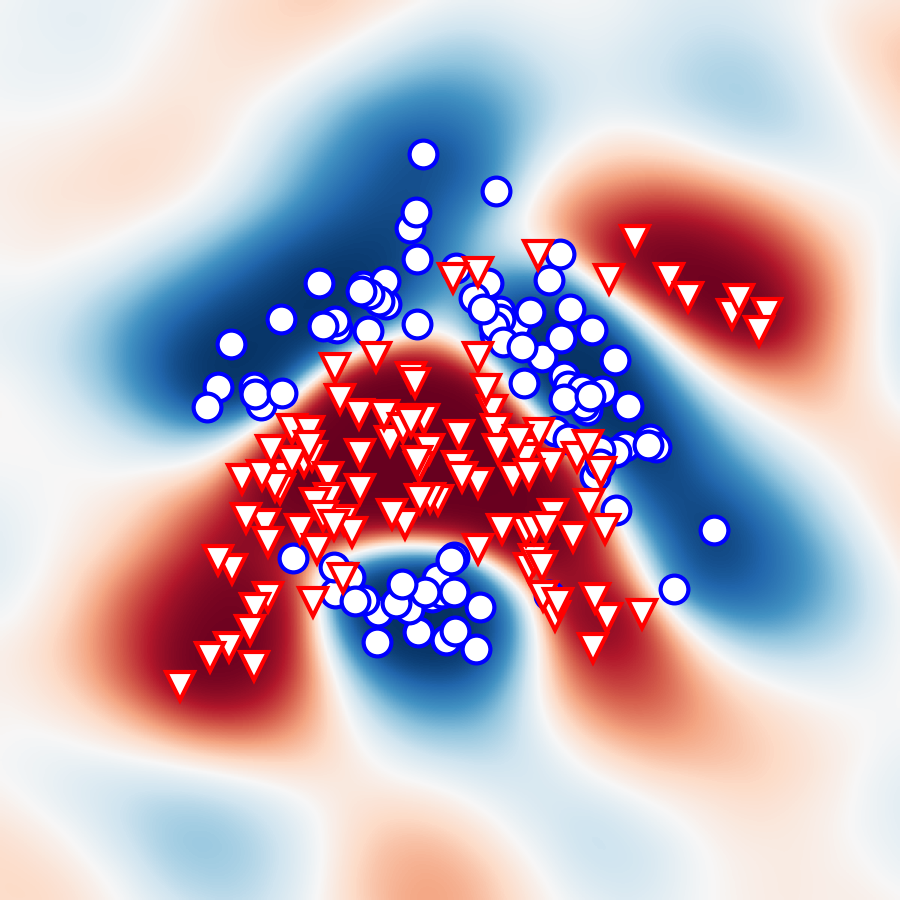}
  \end{subfigure}
   \begin{subfigure}[t]{\figurewidth}
    \includegraphics[width=\linewidth]{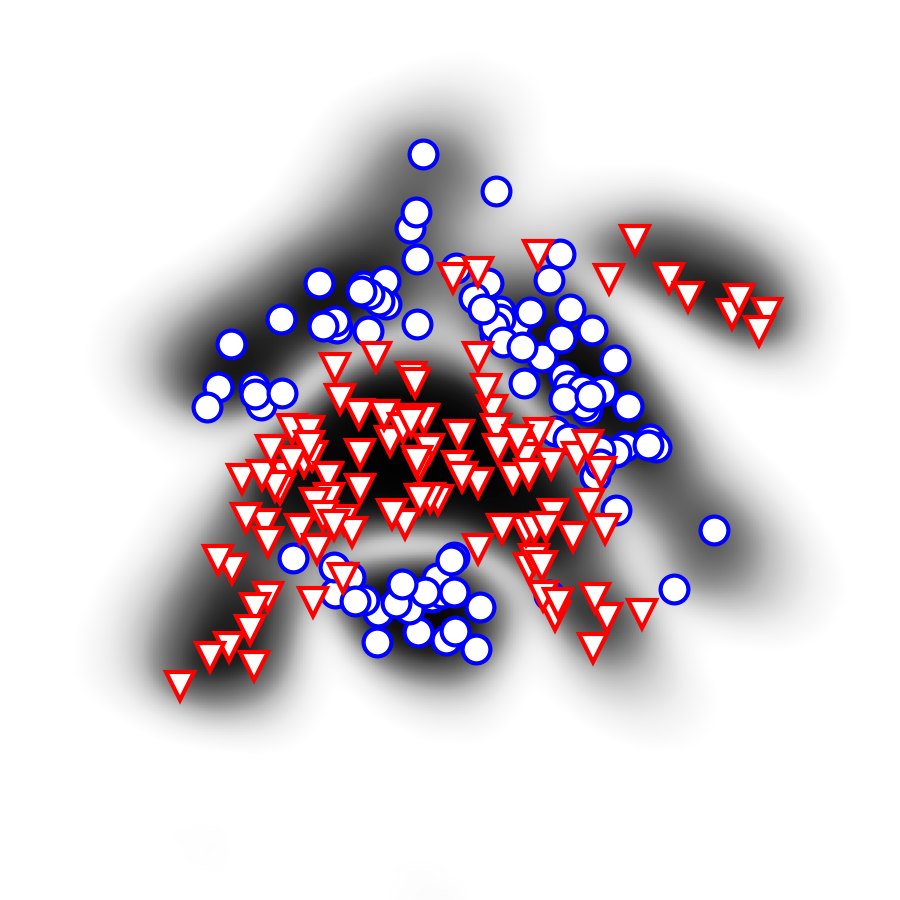}
  \end{subfigure}\\[0.5em]
  \begin{tikzpicture}
    \tikzstyle{box} = [minimum width=.99\figurewidth,inner sep=1pt,rounded corners=1pt]
    \node[box, rotate=90] at (\figurewidth,.55\figurewidth) {Periodic ReLU};      
  \end{tikzpicture} 
  \begin{subfigure}[t]{\figurewidth}
     \includegraphics[width=\linewidth]{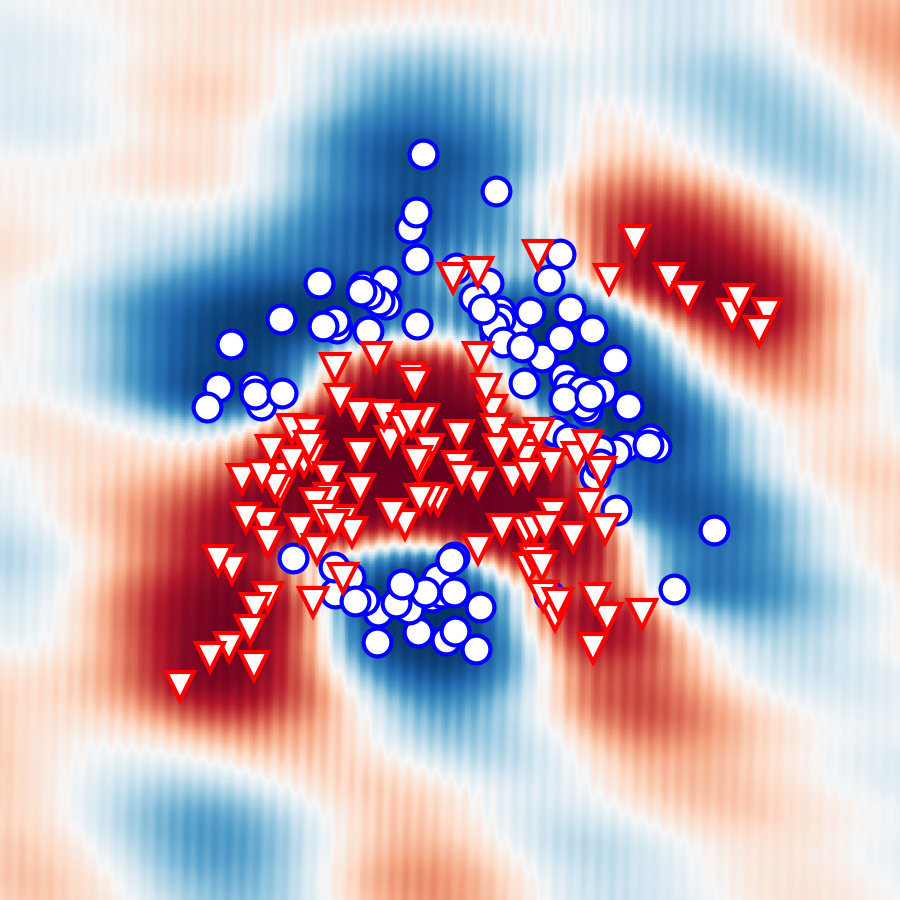}
  \end{subfigure}
  \hfill
    \begin{subfigure}[t]{\figurewidth}
     \includegraphics[width=\linewidth]{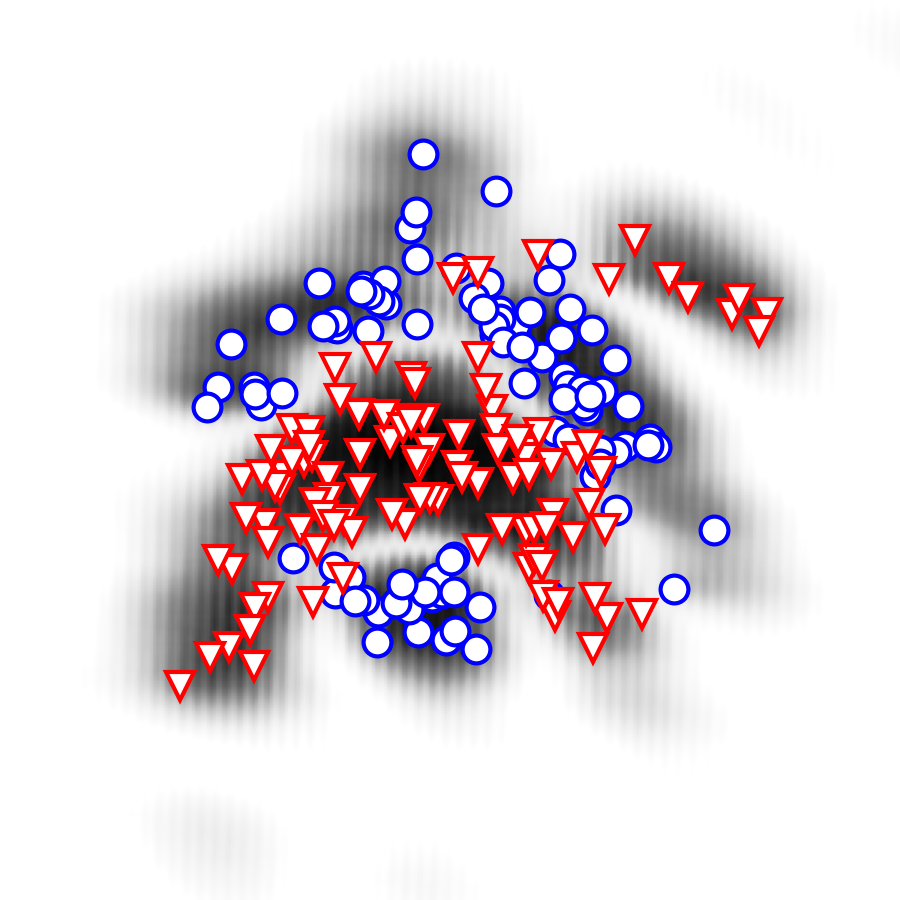}
  \end{subfigure}
  \hfill
    \begin{subfigure}[t]{\figurewidth}
     \includegraphics[width=\linewidth]{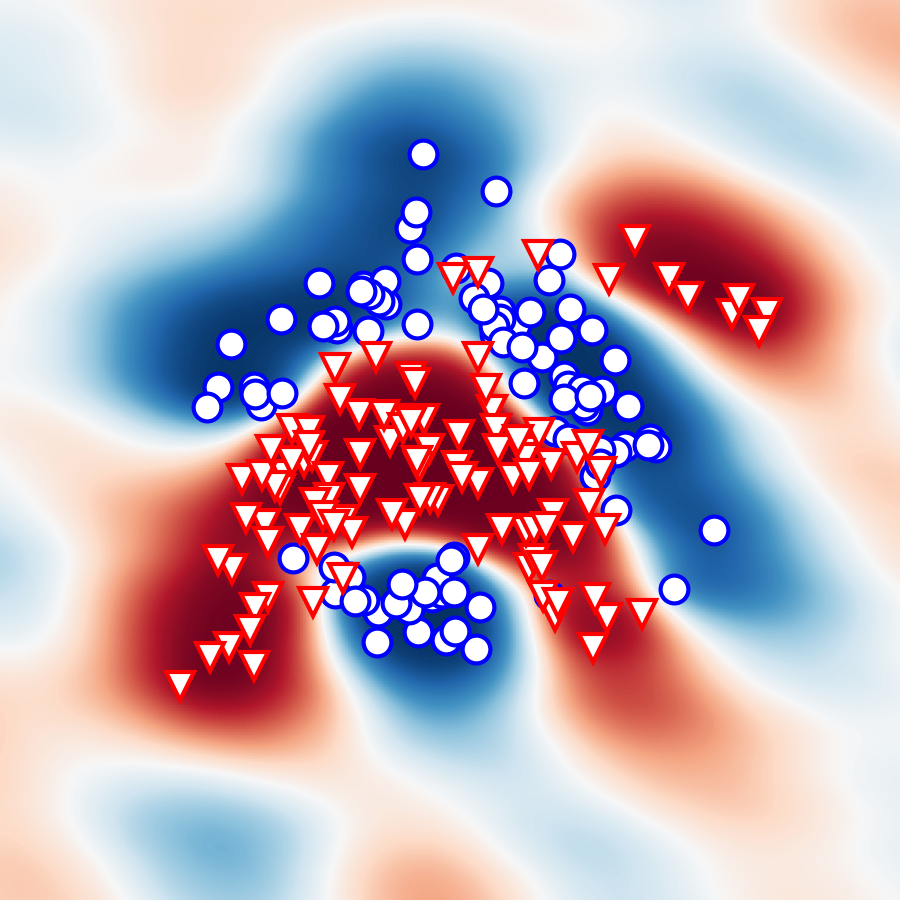}
  \end{subfigure}
      \begin{subfigure}[t]{\figurewidth}
     \includegraphics[width=\linewidth]{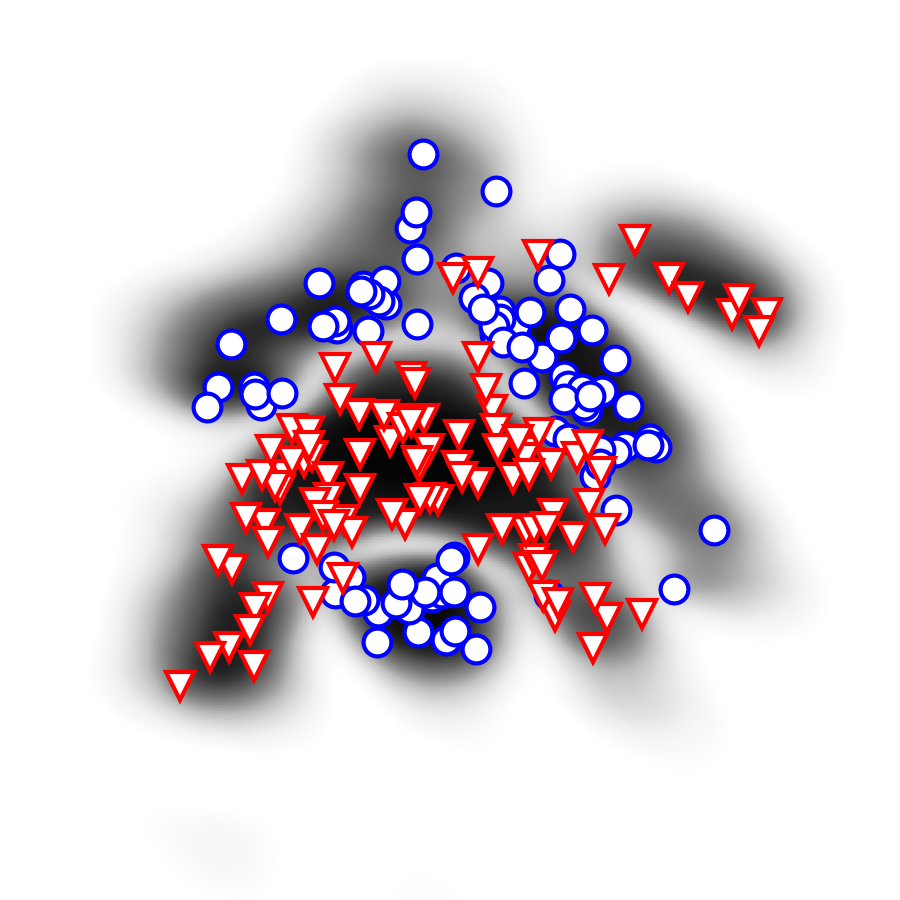}
  \end{subfigure}
    \hfill
  \begin{subfigure}[t]{\figurewidth}
    \includegraphics[width=\linewidth]{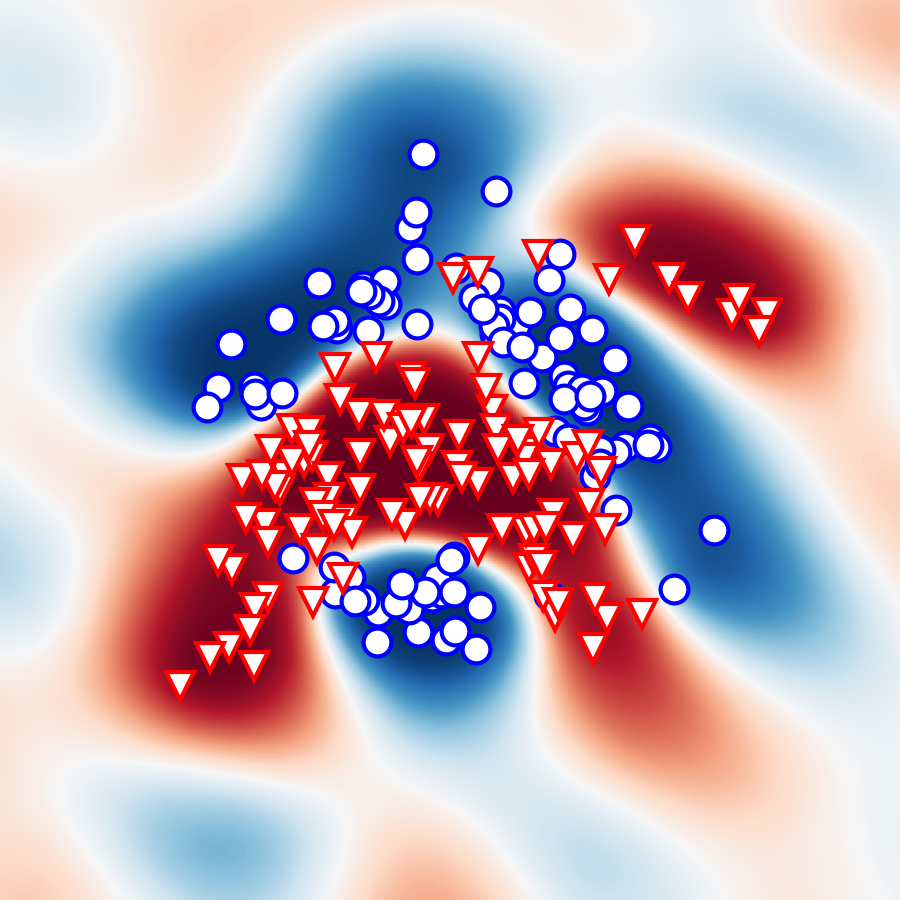}
  \end{subfigure}
   \begin{subfigure}[t]{\figurewidth}
    \includegraphics[width=\linewidth]{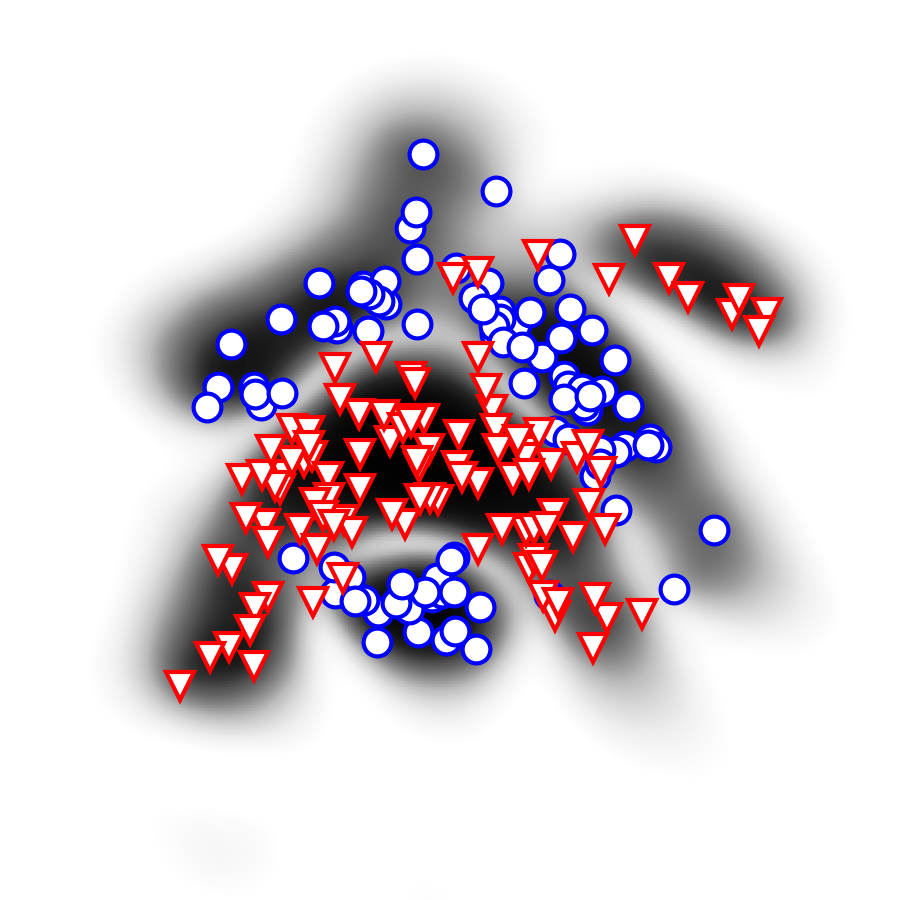}
  \end{subfigure}
  \caption{Posterior predictive densities of globally stationary BNNs with 30 hidden units on the banana classification task.  Table rows show results for different periodic activation functions, and columns show different prior covariance functions. For each model, the resulting predictive mean and variance are plotted side by side. For predictive mean plots, the colour intensity represents confidence in the class prediction. For the variance plots, white colour represents high variance, and black colour represents low variance. We obtain comparable results regardless of the choice of the periodic activation function. Estimated using dynamic HMC run for 10k iterations and 4 chains.}
  \label{fig:comparisons_app_meanvar}
  \vspace*{-1em}
\end{figure*}
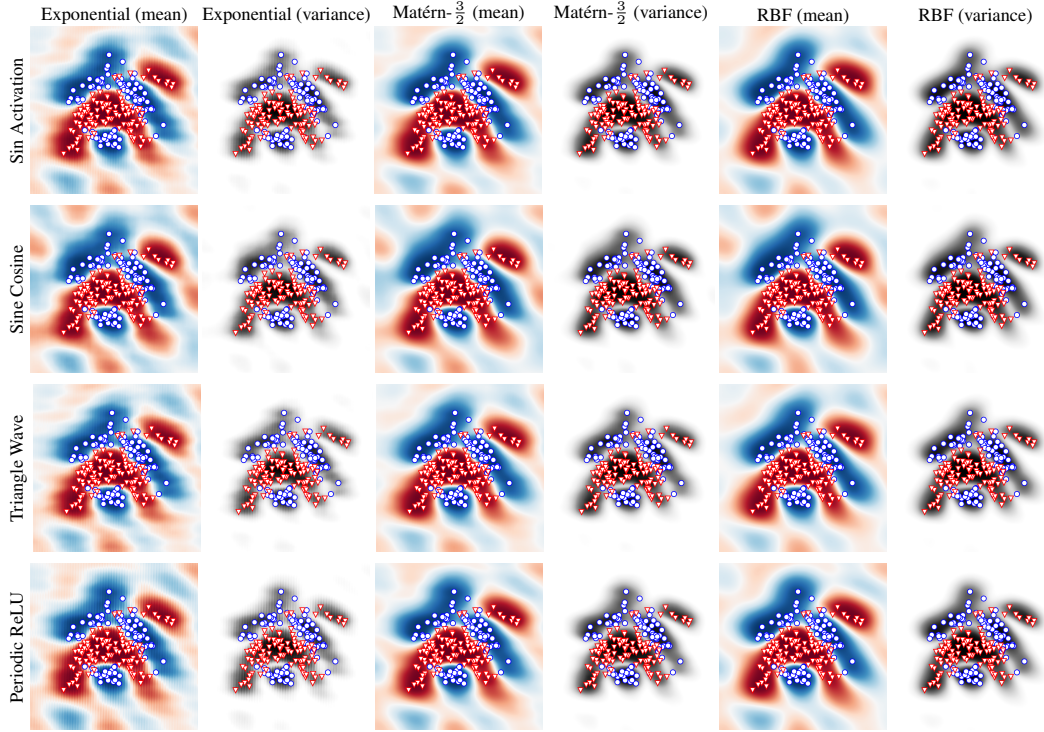

\begin{figure*}[!t]
  \scriptsize
  \pgfplotsset{hide axis,scale only axis,width=\figurewidth,height=\figureheight}
  \setlength{\figurewidth}{.1575\textwidth}
  \setlength{\figureheight}{\figurewidth}
  \begin{tikzpicture}
    \tikzstyle{box} = [minimum width=.99\figurewidth,draw=none,inner sep=1pt,rounded corners=1pt]
    \node[box, rotate=90] at (\figurewidth,.55\figurewidth) {};      
  \end{tikzpicture}
  \begin{subfigure}[b]{.477\textwidth}
  \centering Exponential (mean)
  \end{subfigure}
  \hfill
  \begin{subfigure}[b]{.477\textwidth}
  \centering RBF (mean)
  \end{subfigure} \\
  \begin{tikzpicture}
    \tikzstyle{box} = [minimum width=.99\figurewidth,draw=none,inner sep=1pt,rounded corners=1pt]
    \node[box, rotate=90] at (\figurewidth,.55\figurewidth) {Sin Activation};      
  \end{tikzpicture}
  \begin{subfigure}[b]{\figurewidth}
    \centering $K=10$
    \includegraphics[width=\linewidth]{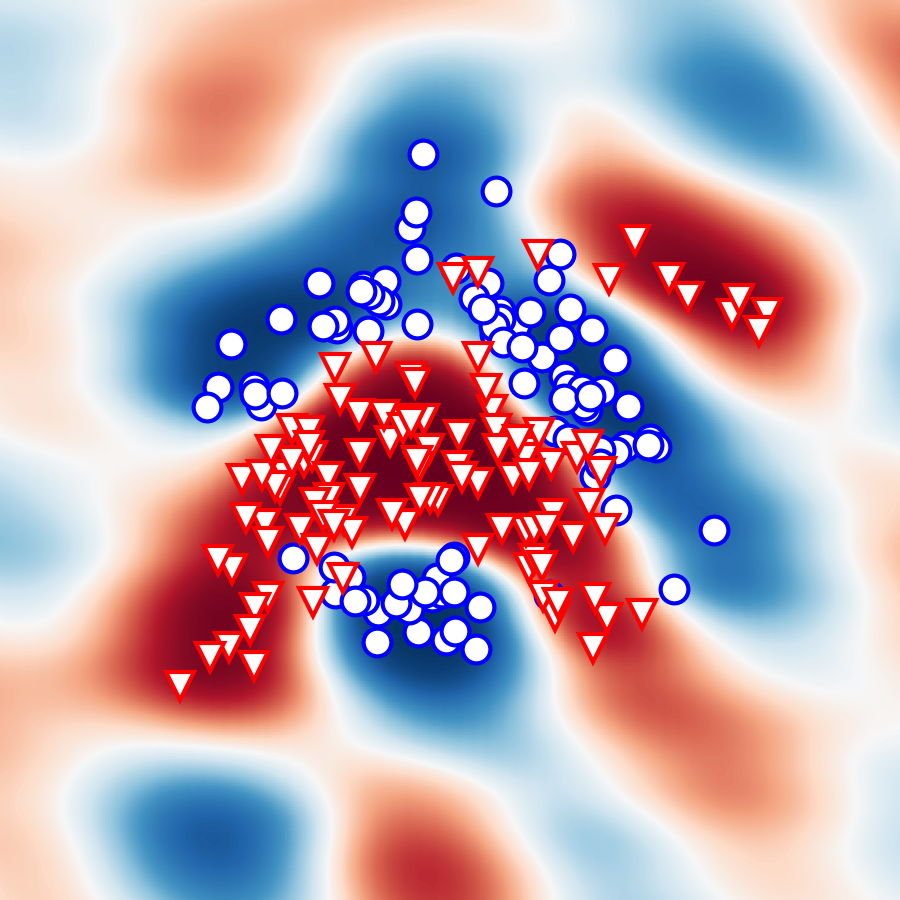}
  \end{subfigure}
    \begin{subfigure}[b]{\figurewidth}
    \centering $K=20$
    \includegraphics[width=\linewidth]{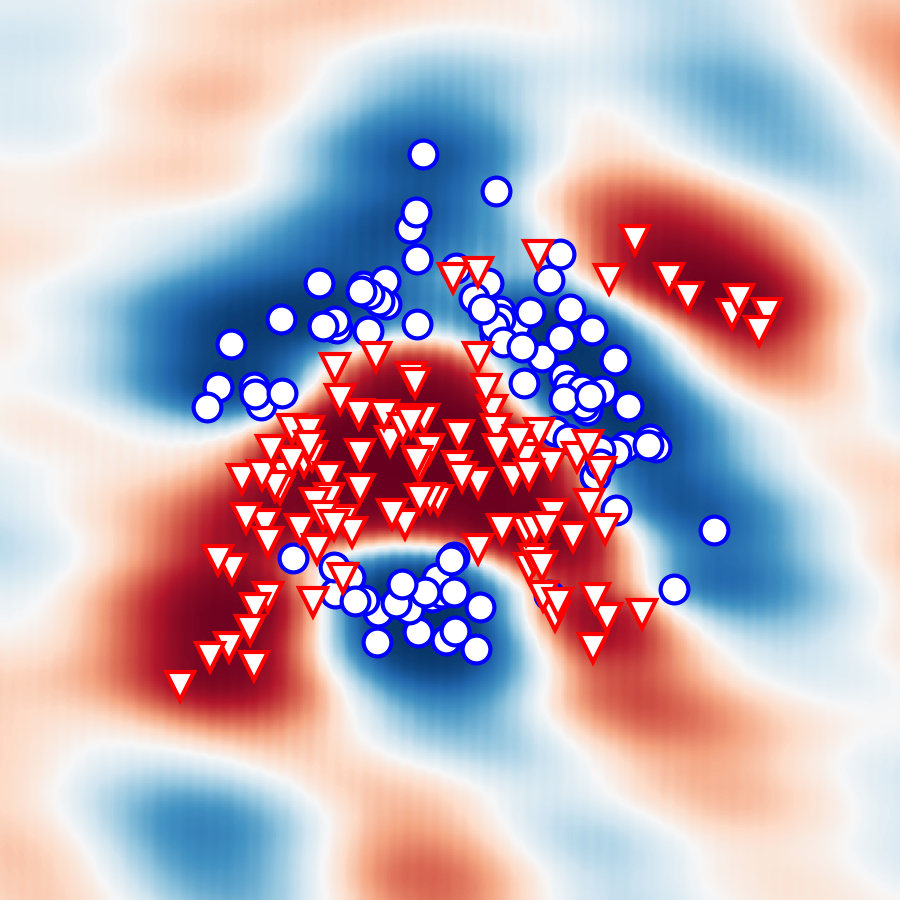}
  \end{subfigure}
    \begin{subfigure}[b]{\figurewidth}
    \centering $K=30$
    \includegraphics[width=\linewidth]{img/banana_Matern_SinActivation_5_30_combined_mean}
  \end{subfigure}
  \hspace{0.5em}
  \begin{subfigure}[b]{\figurewidth}
    \centering $K=10$
    \includegraphics[width=\linewidth]{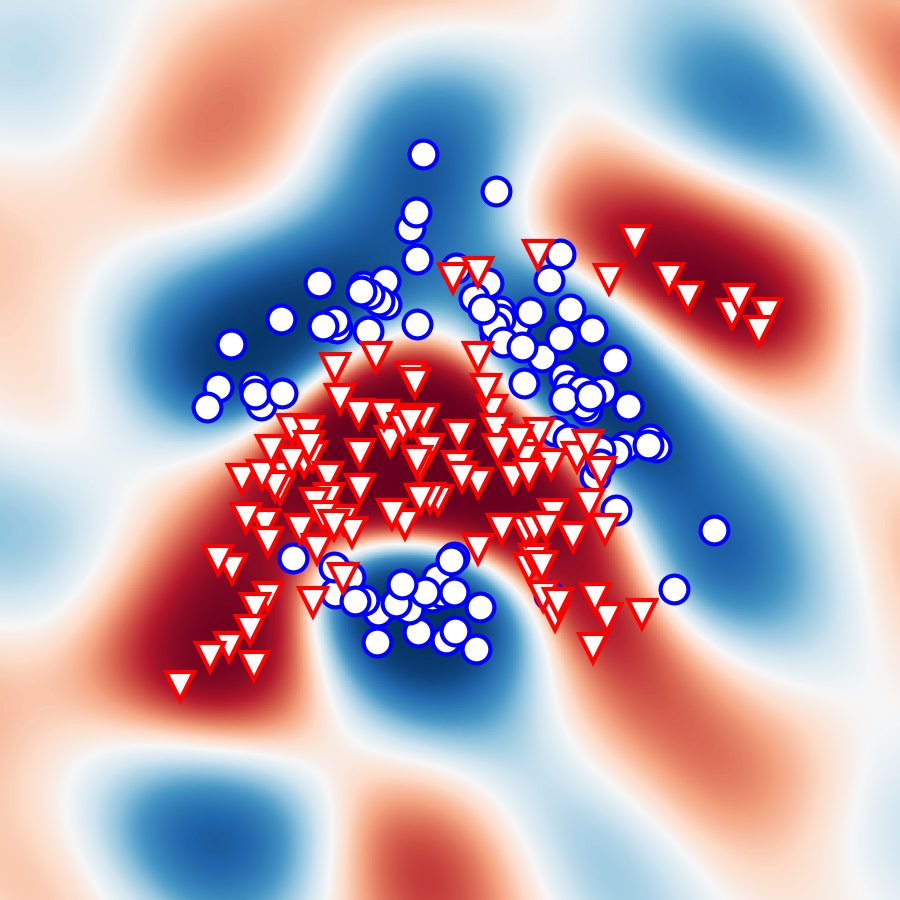}
  \end{subfigure}
  \begin{subfigure}[b]{\figurewidth}
    \centering $K=20$
    \includegraphics[width=\linewidth]{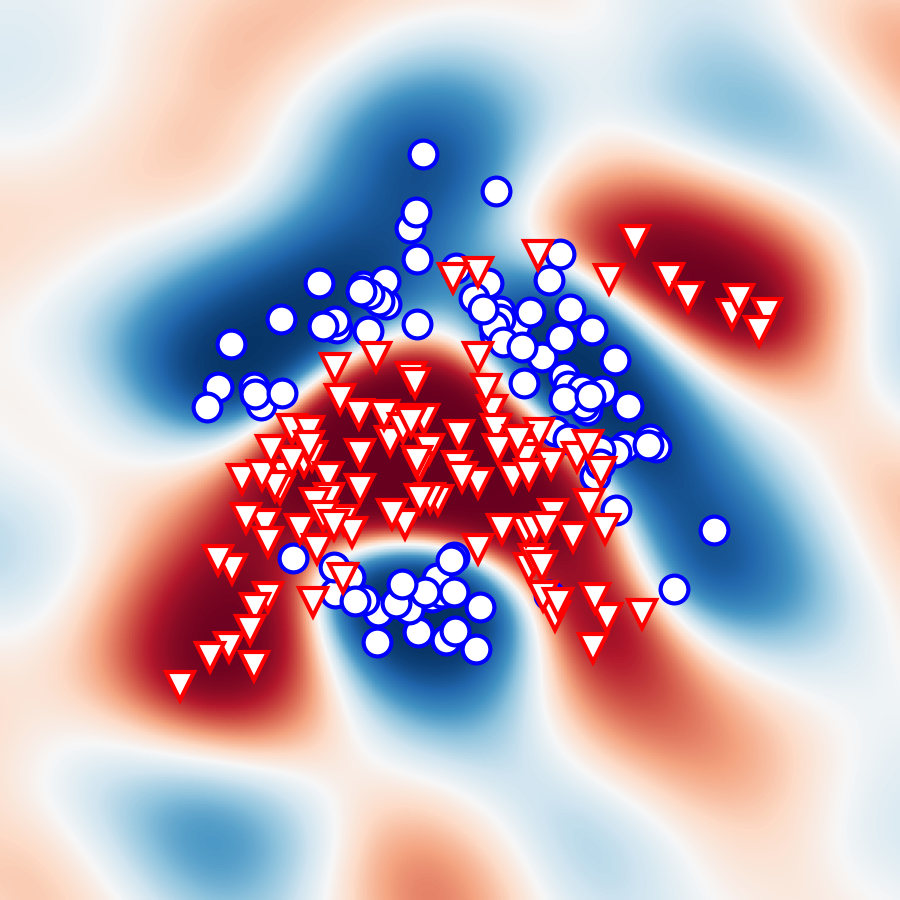}
  \end{subfigure}
  \begin{subfigure}[b]{\figurewidth}
    \centering $K=30$
    \includegraphics[width=\linewidth]{img/banana_RBF_SinActivation_30_combined_mean}
  \end{subfigure}\\[0.5em]
  \begin{tikzpicture}
    \tikzstyle{box} = [minimum width=.99\figurewidth,inner sep=1pt,rounded corners=1pt]
    \node[box, rotate=90] at (\figurewidth,.55\figurewidth) {Sine Cosine};      
  \end{tikzpicture} 
  \begin{subfigure}[t]{\figurewidth}
     \includegraphics[width=\linewidth]{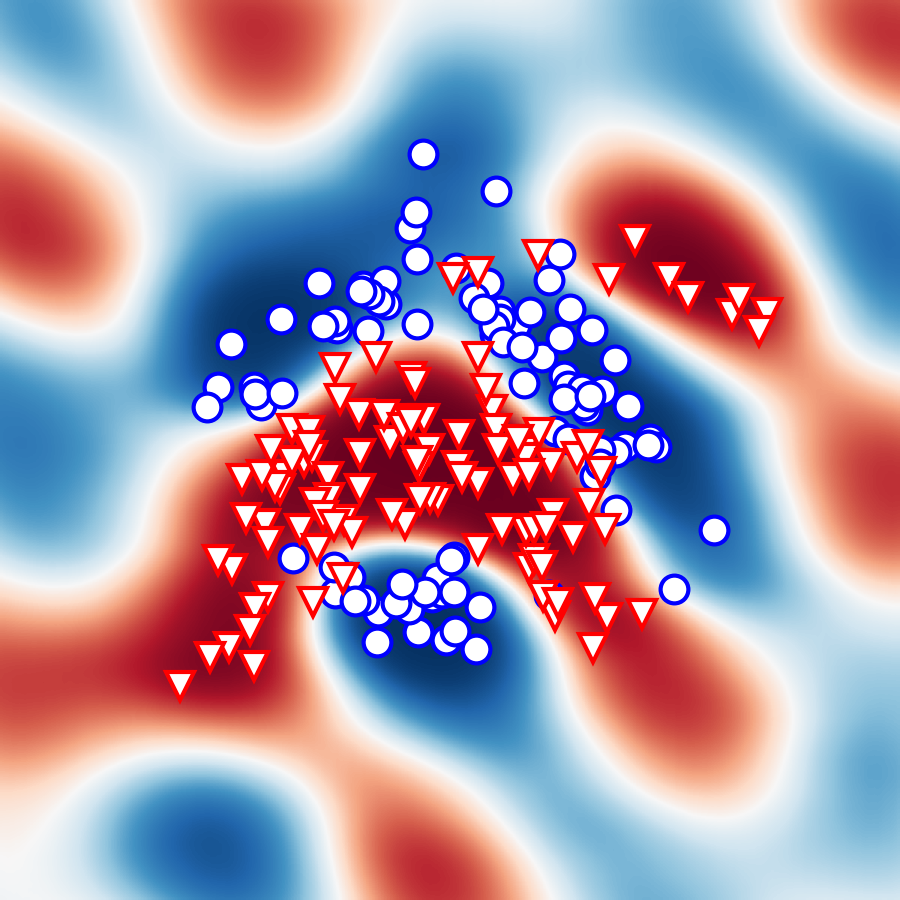}
  \end{subfigure}
    \begin{subfigure}[t]{\figurewidth}
     \includegraphics[width=\linewidth]{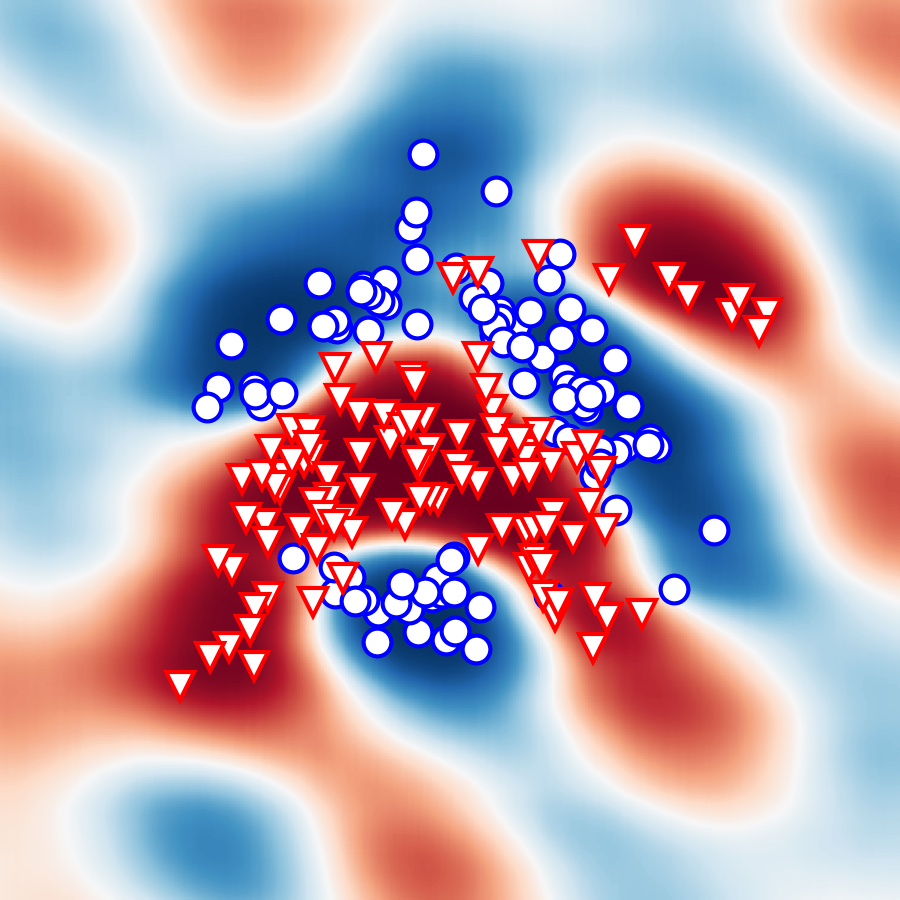}
  \end{subfigure}
    \begin{subfigure}[t]{\figurewidth}
     \includegraphics[width=\linewidth]{img/banana_Matern_SinCosActivation_5_30_combined_mean}
  \end{subfigure}
    \hspace{0.5em}
      \begin{subfigure}[t]{\figurewidth}
     \includegraphics[width=\linewidth]{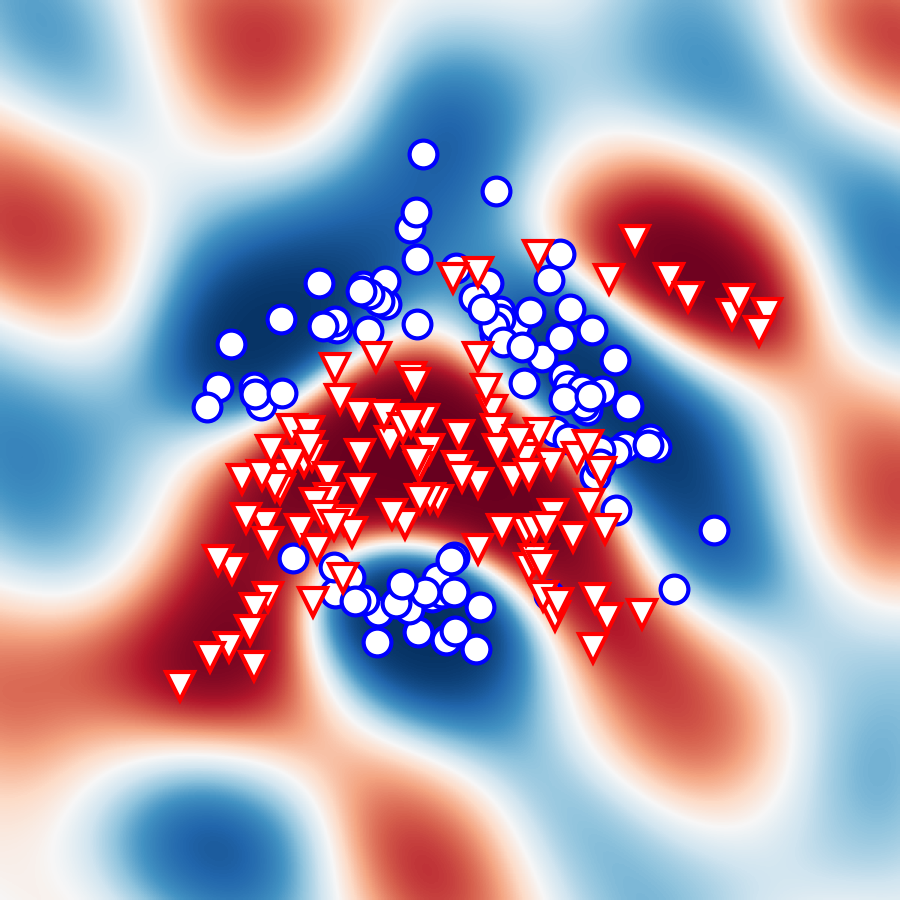}
  \end{subfigure}
  \begin{subfigure}[t]{\figurewidth}
    \includegraphics[width=\linewidth]{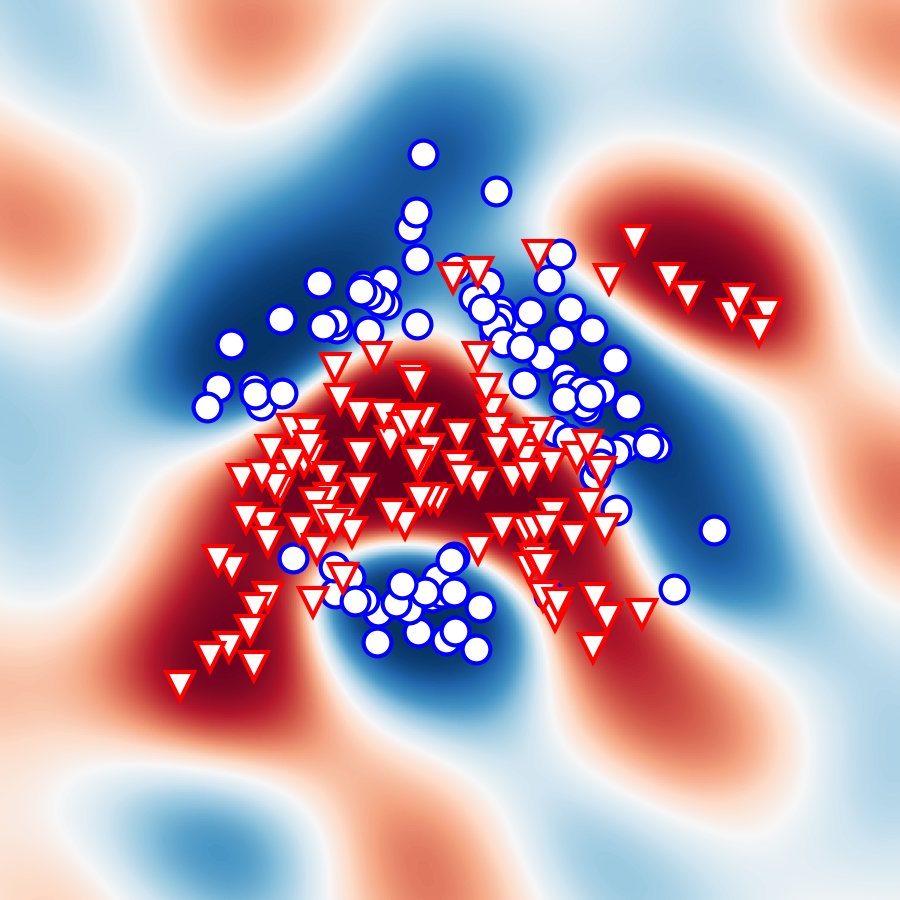}
  \end{subfigure}
   \begin{subfigure}[t]{\figurewidth}
    \includegraphics[width=\linewidth]{img/banana_RBF_SinCosActivation_30_combined_mean}
  \end{subfigure}\\[0.5em]
  \begin{tikzpicture}
    \tikzstyle{box} = [minimum width=.99\figurewidth,inner sep=1pt,rounded corners=1pt]
    \node[box, rotate=90] at (\figurewidth,.55\figurewidth) {Triangle Wave};      
  \end{tikzpicture} 
  \begin{subfigure}[t]{\figurewidth}
     \includegraphics[width=\linewidth]{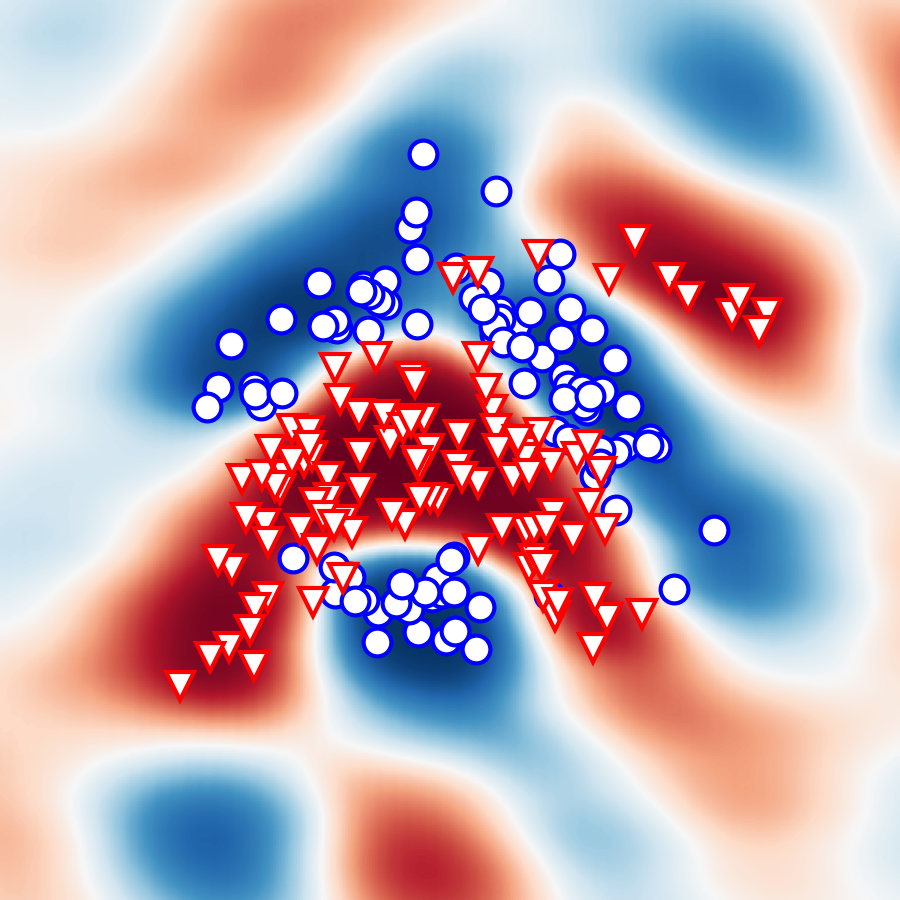}
  \end{subfigure}
    \begin{subfigure}[t]{\figurewidth}
     \includegraphics[width=\linewidth]{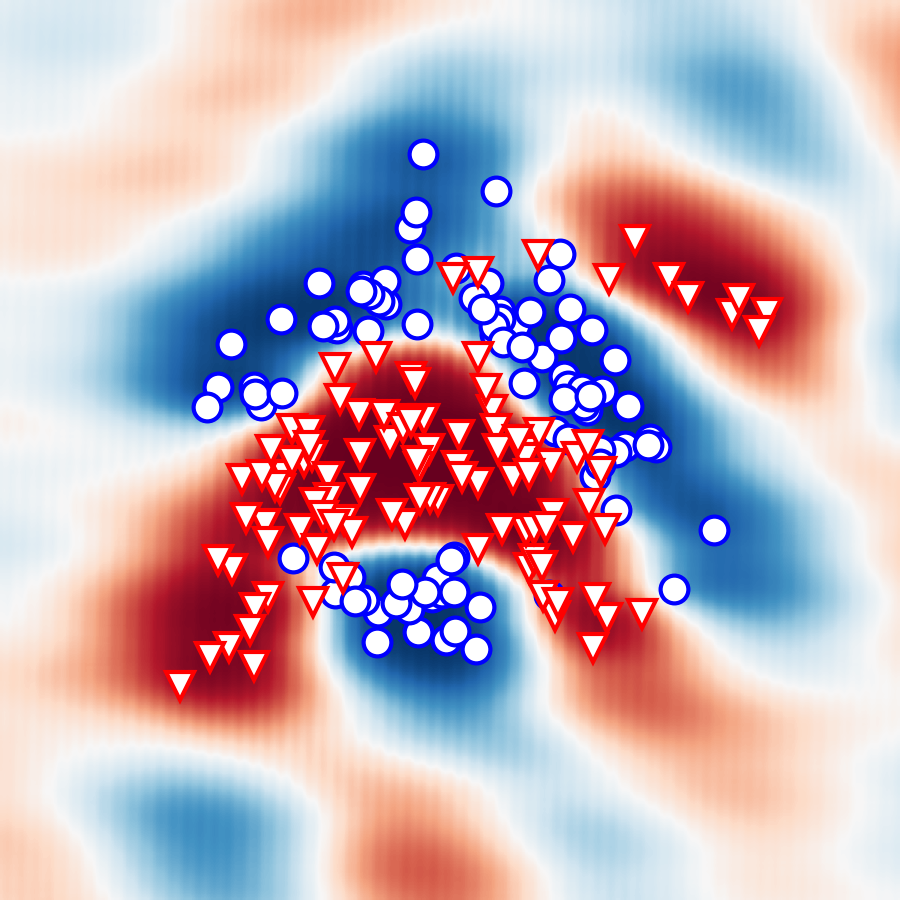}
  \end{subfigure}
    \begin{subfigure}[t]{\figurewidth}
     \includegraphics[width=\linewidth]{img/banana_Matern_TriangleWave_5_30_combined_mean}
  \end{subfigure}
    \hspace{0.5em}
      \begin{subfigure}[t]{\figurewidth}
     \includegraphics[width=\linewidth]{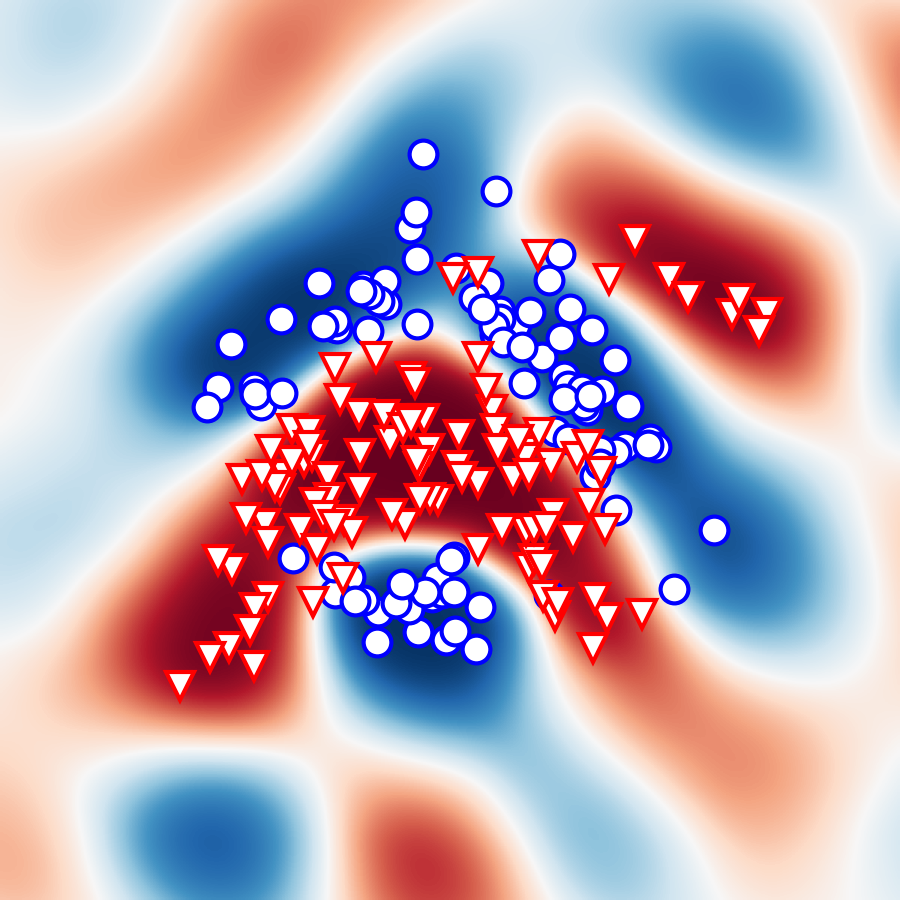}
  \end{subfigure}
  \begin{subfigure}[t]{\figurewidth}
    \includegraphics[width=\linewidth]{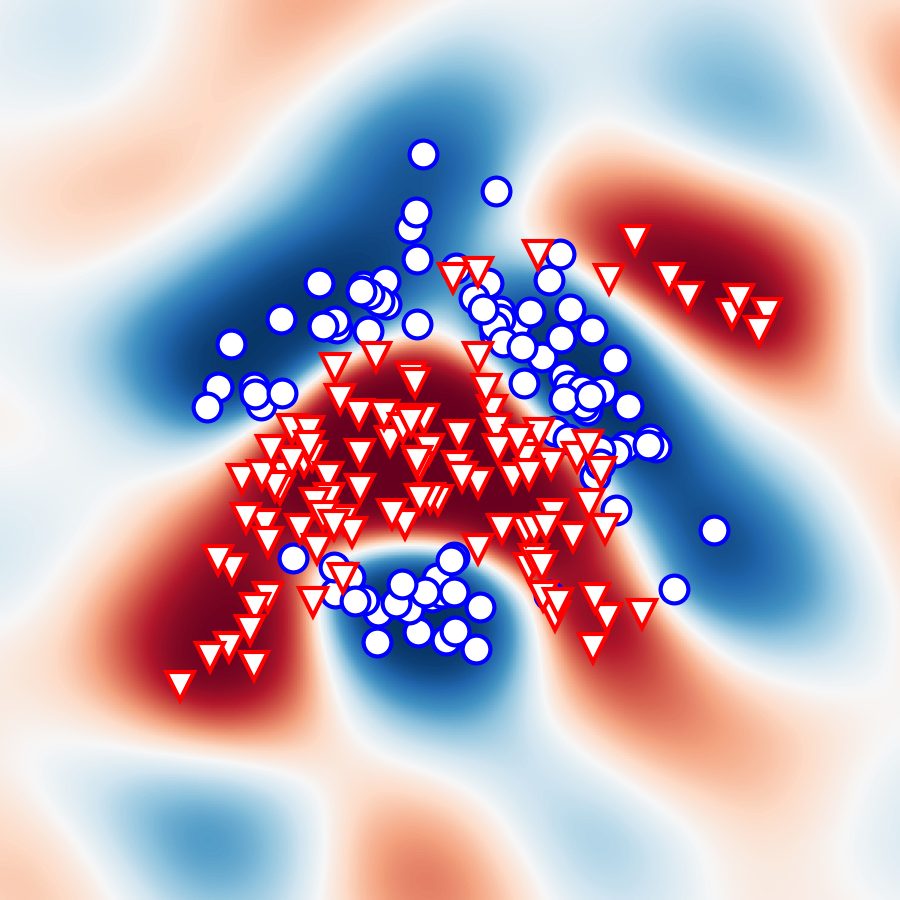}
  \end{subfigure}
   \begin{subfigure}[t]{\figurewidth}
    \includegraphics[width=\linewidth]{img/banana_RBF_TriangleWave_30_combined_mean}
  \end{subfigure}\\[0.5em]
  \begin{tikzpicture}
    \tikzstyle{box} = [minimum width=.99\figurewidth,inner sep=1pt,rounded corners=1pt]
    \node[box, rotate=90] at (\figurewidth,.55\figurewidth) {Periodic ReLU};      
  \end{tikzpicture} 
  \begin{subfigure}[t]{\figurewidth}
     \includegraphics[width=\linewidth]{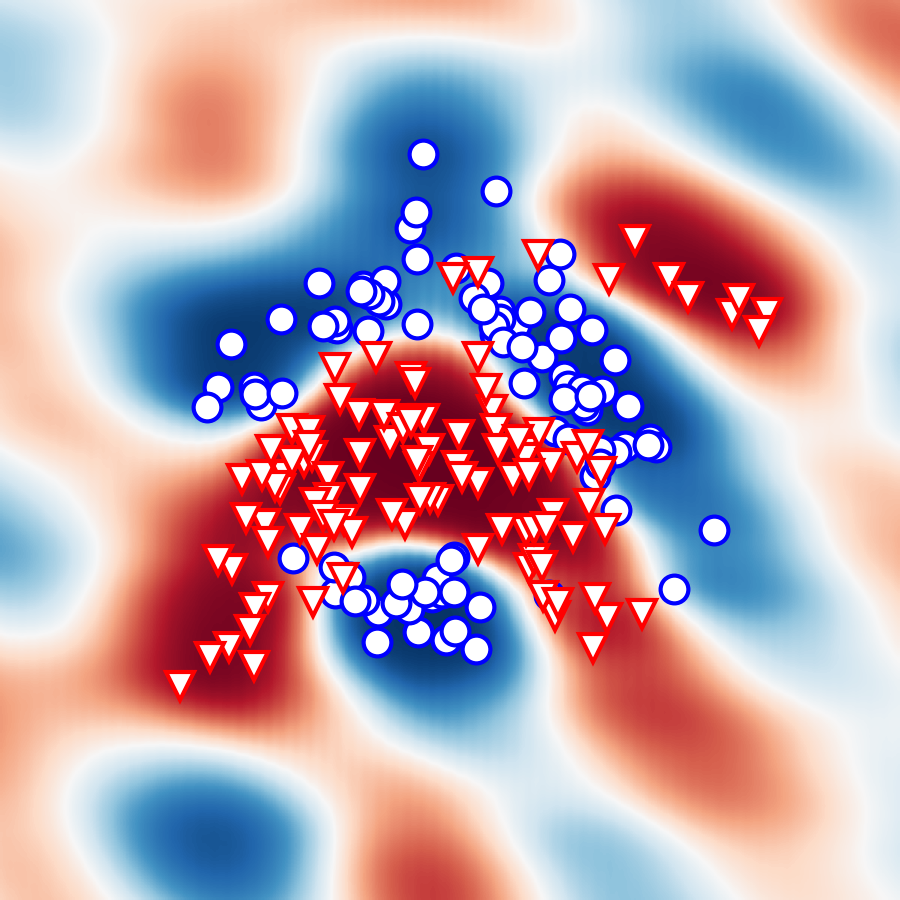}
  \end{subfigure}
    \begin{subfigure}[t]{\figurewidth}
     \includegraphics[width=\linewidth]{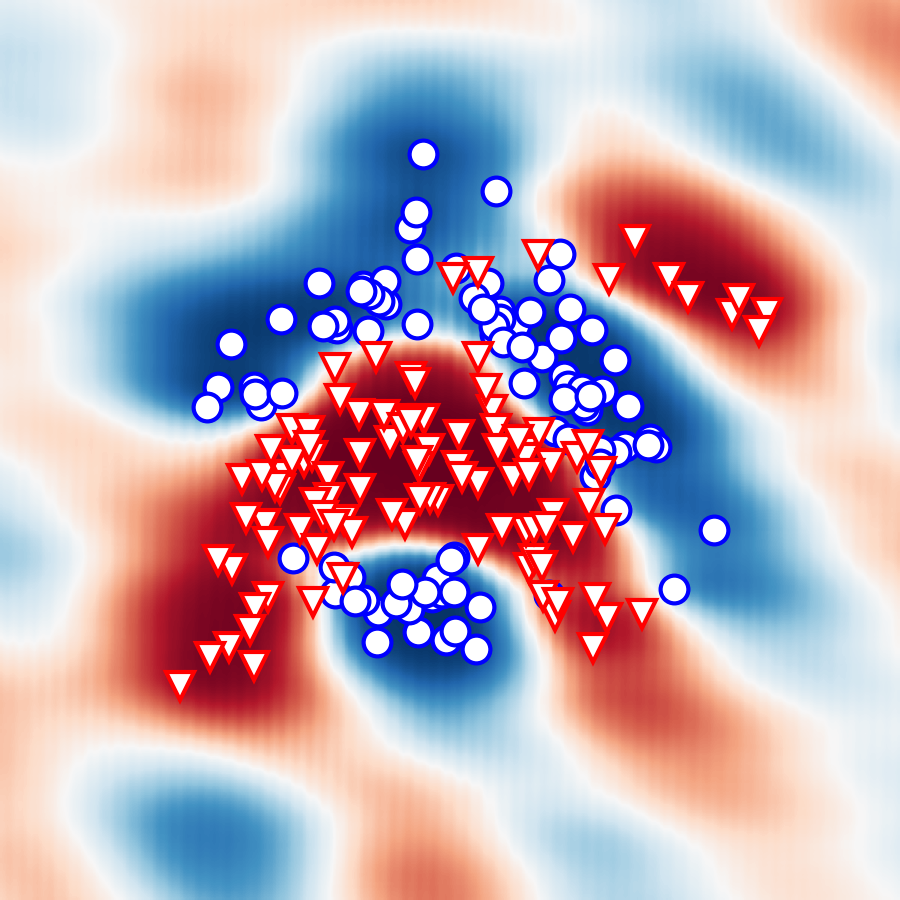}
  \end{subfigure}
    \begin{subfigure}[t]{\figurewidth}
     \includegraphics[width=\linewidth]{img/banana_Matern_PeriodicReLU_5_30_combined_mean}
  \end{subfigure}
    \hspace{0.5em}
      \begin{subfigure}[t]{\figurewidth}
     \includegraphics[width=\linewidth]{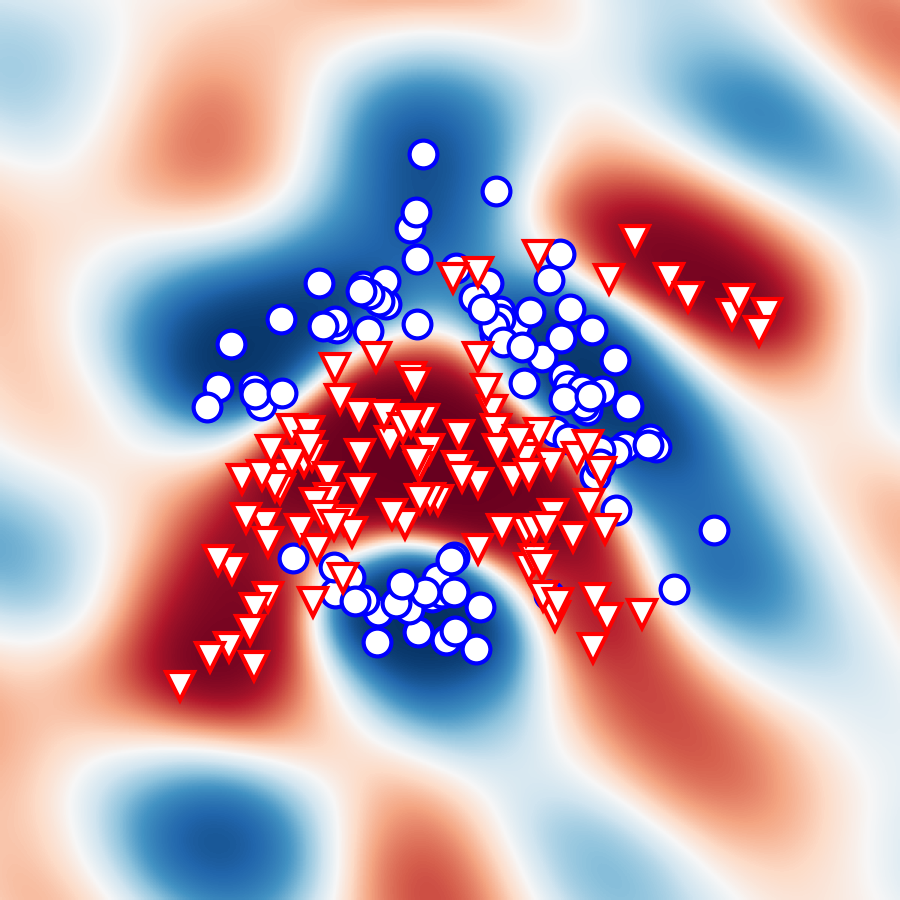}
  \end{subfigure}
  \begin{subfigure}[t]{\figurewidth}
    \includegraphics[width=\linewidth]{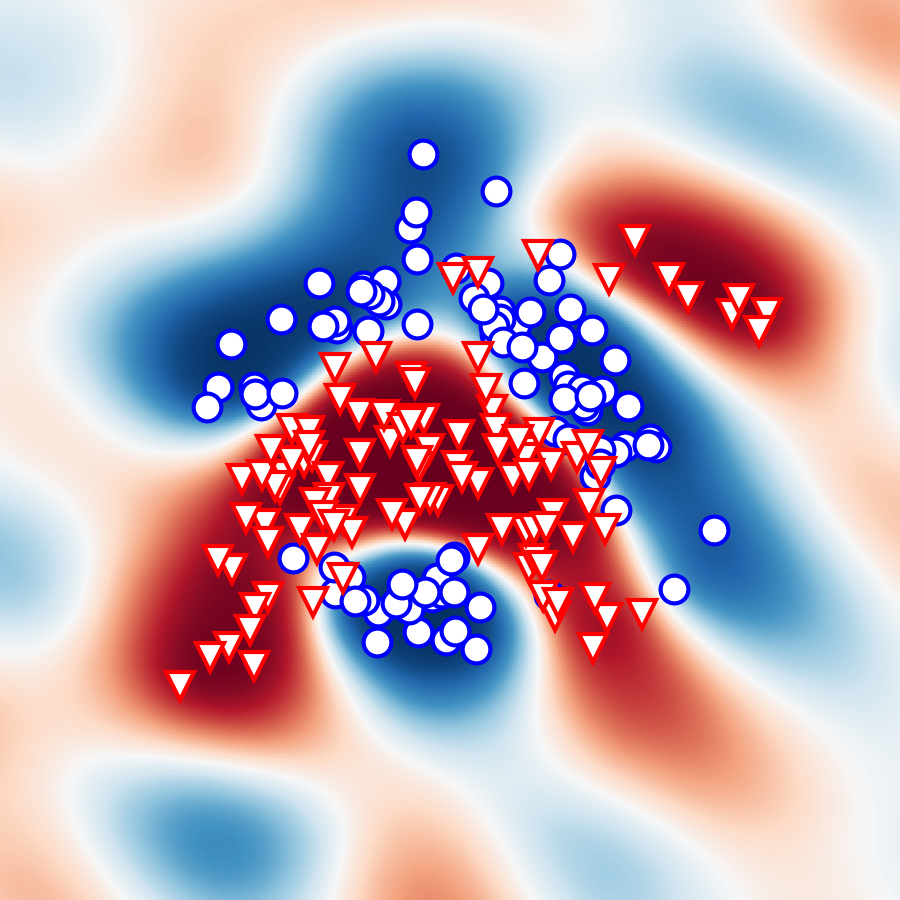}
  \end{subfigure}
   \begin{subfigure}[t]{\figurewidth}
    \includegraphics[width=\linewidth]{img/banana_RBF_PeriodicReLU_30_combined_mean}
  \end{subfigure}
  \caption{Posterior predictive densities of stationary BNNs for varying number of hidden units $K$ on the banana classification task.  Table rows show results for different periodic activation functions, and columns show different number of hidden units, denoted as $K$. Results are shown for the two extreme cases, \ie Exponential and RBF kernel. We obtain comparable results for all periodic activation function. Results are estimated using dynamic HMC run for 10k iterations and 4 chains.}
  \label{fig:comparisons_app}
  \vspace*{-1em}
\end{figure*}

The NN model architecture for the illustrative toy examples contains only the model layer with 30 hidden units and the output layer. The posterior estimates are obtained through dynamic HMC sampling \cite{Ge2018} run for 10k iterations and four independent chains. \cref{fig:teaser} shows predictive densities for non-stationary, locally stationary, and globally stationary activation functions on the banana classification task. The top row illustrates the predictive densities of infinite-width BNNs (GP), and the bottom row shows corresponding results for a finite-width BNN. We observe that models with global stationarity-inducing activation functions revert to the prior outside the data, leading to conservative behaviour (high uncertainty) for out-of-domain samples. Moreover, we see that the finite-width BNNs result in similar behaviour to their infinite-width counterpart, while the locally stationary activation functions in finite-width BNNs exhibit a slower reversion to the mean than their infinite-width corresponding GPs. \cref{fig:comparisons_app_meanvar} shows additional BNN results for the same experiment for globally stationary models using different periodic activation functions. We see that we obtain similar results regardless of the choice of the periodic activation function. As expected, all periodic activation functions result in low variance only for the training data clusters and revert to the prior outside the data. \cref{fig:comparisons_app} shows the effect of varying the number of hidden units in the same experiment.

 Additionally, we include a 1D toy regression study highlighting the differences between different prior assumptions encoded by choice of the activation function. \cref{fig:gram-regression} shows the corresponding prior covariance as well as posterior predictions for the infinite-width (GP) model. In \cref{fig:gram-regression-finite}, we replicate the same study with a finite-width network and recover the same behaviour. For the finite-width results, posterior estimates are obtained through dynamic HMC sampling for 5000 iterations. \cref{fig:gram-comparison} illustrates the error between the Gram matrices of the infinite width and finite width models.
 
\subsection{Benchmark Regression Tasks}
\label{app:uci_reg}
For the UCI \cite{UCI} regression tasks, the NN architecture is a fully connected network with layers $d$-1000-1000-500-25-2000-1. A dropout layer with $p=0.1$ is applied at the 500 nodes wide layer to prevent overfitting. For these experiments, we used a 10-fold cross-validation setup performed for a single repetition per experiment. Each model is trained for 100 epochs using SGD (momentum 0.9). The batch sizes used for each data set are listed in \cref{tbl:benchmarks_reg_full}. The learning rates for the lengthscale parameter $\ell$ and measurement noise standard deviation $s$ were set to 0.01. All learning rates were decreased to 0.72 of the original value during training, using a schedule having square root dependence on the progression through epochs (slower than linear learning rate decay). The lengthscale parameter $\ell$ was initialized with a value of five, and $s$ was initialized with one. For the non-stationary ReLU model, the lengthscale parameter does not have a similar significance as in the global and local stationary models and therefore was initialized with a value of one to prevent vanishing or exploding gradients. Posterior inference was performed using KFAC Laplace \cite{ritter2018a_kfac_laplace}.

Since we observed that the results were sensitive to the SGD learning rate and the variance scale parameter of KFAC Laplace, we performed a grid search over both of these hyperparameters. Therefore, we split the training set such that 80\% is used for training and the remaining 20\% is used as a validation set for the grid search. First, a grid search for the SGD learning rate was performed over the values $[\num{5e-5}, \num{1e-4}, \num{5e-4}, \num{1e-3}]$, choosing the learning rate that achieved the smallest RMSE on the validation set. A single common learning rate was chosen for all ten folds. Subsequently, KFAC Laplace was applied on models trained using the best learning rate, using variance scales in the grid $[0.01, 0.05, 0.1, 0.15, 0.2, 0.25] $. 30 samples were used for model averaging in the grid search. As the objective in the KFAC Laplace variance scale grid search, we used a weighted sum of validation set data negative log-likelihood and an OOD noise validation set negative log-likelihood \cite{pmlr-v119-kristiadi20a}. We used $\lambda=0.2$ as the parameter controlling the weighting between the regular validation set score and the OOD noise set score (one-fifth of the weight on the OOD set). A single common KFAC Laplace variance scale was chosen for all ten folds. After the best learning rate and KFAC Laplace variance scale has been selected, the model is retrained from the start on the full training set of each fold using the best learning rate. KFAC Laplace model with the best variance scale is then fitted on the trained model, and 50 samples from the approximate posterior are used for model averaging to obtain the final results.

The results for the boston (2--3 h), concrete (2--3 h) and airfoil (6--8 h) data sets were calculated using a single CPU per experiment, and the results for the elevators (1--2 h) data set were calculated using a single GPU per experiment. Copyright of the concrete data set: Prof. I-Cheng Yeh \cite{concrete}. 

\cref{tbl:benchmarks_reg_full} shows results on four UCI regression data sets comparing deep neural networks with ReLU, locally stationary RBF \cite{williams1998computation}, and locally stationary Mat\'ern-$\nicefrac{3}{2}$ and Mat\'ern-$\nicefrac{5}{2}$ \cite{meronen2020stationary} against global stationary models. \cref{tbl:benchmarks_reg_full} lists root mean square error (RMSE) and negative log predictive density (NLPD), which captures the predictive uncertainty, while the RMSE only accounts for the mean. The table lists mean and standard deviation values across folds of the 10-fold cross-validation. The values for the best performing models are shown in bold. Especially on small data sets, the standard deviation values are large, which is mostly due to differences between different folds instead of variations in model performance. The table shows that global stationary models provide better estimates of the target distribution in all cases while obtaining comparable RMSEs. Moreover, we observe that the periodic ReLU activation function tends to outperform the sinusoidal activation. Also, the importance of the choice of prior covariance can be seen in \cref{tbl:benchmarks_reg_full}. It appears that the Mat\'ern-$\nicefrac{3}{2}$ covariance is the best choice for the smallest boston data set, while the smoother Mat\'ern-$\nicefrac{5}{2}$ or RBF covariance functions seem to be more suitable for the larger data sets.

\begin{table}[t!]
  \caption{Examples of UCI regression tasks, showing the globally stationary NN model directly gives competitive mean negative log predictive density (NLPD) and root mean square error (RMSE) to locally stationary and non-stationary NN models. KFAC Laplace was used as the inference method.}
  \label{tbl:benchmarks_reg_full}
  \scriptsize
  \tiny

    \setlength{\tabcolsep}{0pt}
  \setlength{\tblw}{0.09\textwidth}  
  \begin{tabularx}{\columnwidth}{l @{\extracolsep{\fill}} C{\tblw}  C{\tblw} C{\tblw} C{\tblw} C{\tblw}  C{\tblw} C{\tblw} C{\tblw}}
  \toprule

%NLPD and RMSE (this table has been automatically generated, do not alter manually)
                          &        \multicolumn{2}{c}{\sc boston}         &       \multicolumn{2}{c}{\sc concrete}        &        \multicolumn{2}{c}{\sc airfoil}        &       \multicolumn{2}{c}{\sc elevators}       \\
($n$, $d$)                &         \multicolumn{2}{c}{(506, 12)}         &         \multicolumn{2}{c}{(1030, 5)}         &         \multicolumn{2}{c}{(1503, 5)}         &        \multicolumn{2}{c}{(16599, 18)}        \\
($c$, $n_\textrm{batch}$) &          \multicolumn{2}{c}{(1, 50)}          &          \multicolumn{2}{c}{(1, 50)}          &          \multicolumn{2}{c}{(1, 50)}          &         \multicolumn{2}{c}{(1, 500)}          \\
\midrule
                          &         NLPD          &         RMSE          &         NLPD          &         RMSE          &         NLPD          &         RMSE          &         NLPD          &         RMSE          \\
\midrule
ReLU                      &    $0.51{\pm}0.32$    &    $0.37{\pm}0.07$    &    $0.78{\pm}0.16$    &    $0.48{\pm}0.04$    &    $0.51{\pm}0.53$    &    $0.41{\pm}0.21$    &    $0.38{\pm}0.03$    &    $0.35{\pm}0.01$    \\
loc RBF                   &    $0.52{\pm}0.30$    &    $0.37{\pm}0.08$    &    $0.78{\pm}0.22$    &    $0.44{\pm}0.05$    &    $0.10{\pm}0.15$    &    $0.26{\pm}0.03$    &    $0.41{\pm}0.04$    &    $0.35{\pm}0.01$    \\
glob RBF (sin)            &    $0.42{\pm}0.34$    &    $0.36{\pm}0.07$    &    $0.74{\pm}0.15$    &    $0.49{\pm}0.05$    &    $0.14{\pm}0.17$    &    $0.29{\pm}0.05$    &    $0.38{\pm}0.03$    &    $0.35{\pm}0.01$    \\
glob RBF (tri)            &    $0.44{\pm}0.38$    &    $0.36{\pm}0.09$    &    $0.75{\pm}0.16$    &    $0.49{\pm}0.05$    &    $0.08{\pm}0.11$    &    $0.27{\pm}0.03$    &    $0.38{\pm}0.03$    &    $0.35{\pm}0.01$    \\
loc Mat-5/2               &    $0.74{\pm}0.42$    &    $0.36{\pm}0.07$    &    $0.87{\pm}0.19$    &    $0.47{\pm}0.04$    &    $0.14{\pm}0.16$    &    $0.27{\pm}0.03$    &    $0.41{\pm}0.04$    &    $0.35{\pm}0.01$    \\
glob Mat-5/2 (sin)        &    $0.41{\pm}0.33$    &    $0.36{\pm}0.08$    &    $0.67{\pm}0.09$    &    $0.47{\pm}0.03$    &    $0.05{\pm}0.12$    &    $0.26{\pm}0.03$    &    $0.37{\pm}0.04$    &    $0.35{\pm}0.01$    \\
glob Mat-5/2 (tri)        &    $0.45{\pm}0.38$    &    $0.36{\pm}0.09$    &    $0.65{\pm}0.09$    &    $0.46{\pm}0.03$    &    $0.05{\pm}0.16$    &    $0.26{\pm}0.03$    &  $\bf 0.37{\pm}0.03$  &  $\bf 0.34{\pm}0.01$  \\
loc Mat-3/2               &    $0.71{\pm}0.38$    &    $0.40{\pm}0.08$    &    $0.84{\pm}0.28$    &  $\bf 0.42{\pm}0.04$  &    $0.11{\pm}0.18$    &    $0.26{\pm}0.03$    &    $0.43{\pm}0.04$    &    $0.35{\pm}0.01$    \\
glob Mat-3/2 (sin)        &    $0.43{\pm}0.27$    &    $0.39{\pm}0.08$    &    $0.73{\pm}0.16$    &    $0.49{\pm}0.05$    &    $0.07{\pm}0.15$    &    $0.27{\pm}0.03$    &    $0.37{\pm}0.03$    &    $0.35{\pm}0.01$    \\
glob Mat-3/2 (tri)        &    $0.46{\pm}0.39$    &    $0.36{\pm}0.09$    &    $0.71{\pm}0.14$    &    $0.48{\pm}0.04$    &    $0.19{\pm}0.43$    &    $0.34{\pm}0.21$    &    $0.37{\pm}0.03$    &    $0.35{\pm}0.01$    \\
glob RBF (sincos)         &    $0.50{\pm}0.36$    &    $0.37{\pm}0.08$    &    $0.73{\pm}0.13$    &    $0.49{\pm}0.04$    &    $0.19{\pm}0.21$    &    $0.30{\pm}0.07$    &    $0.38{\pm}0.02$    &    $0.35{\pm}0.01$    \\
glob Mat-5/2 (sincos)     &    $0.41{\pm}0.23$    &    $0.39{\pm}0.08$    &    $0.72{\pm}0.12$    &    $0.49{\pm}0.04$    &  $\bf 0.04{\pm}0.11$  &    $0.26{\pm}0.03$    &    $0.38{\pm}0.03$    &    $0.35{\pm}0.01$    \\
glob Mat-3/2 (sincos)     &  $\bf 0.35{\pm}0.32$  &  $\bf 0.34{\pm}0.07$  &    $0.68{\pm}0.14$    &    $0.47{\pm}0.04$    &    $0.05{\pm}0.16$    &    $0.27{\pm}0.04$    &    $0.56{\pm}0.54$    &    $0.42{\pm}0.21$    \\
glob RBF (prelu)          &    $0.39{\pm}0.30$    &    $0.36{\pm}0.07$    &    $0.74{\pm}0.14$    &    $0.49{\pm}0.04$    &    $0.05{\pm}0.12$    &  $\bf 0.26{\pm}0.03$  &    $0.74{\pm}0.73$    &    $0.46{\pm}0.21$    \\
glob Mat-5/2 (prelu)      &    $0.50{\pm}0.42$    &    $0.37{\pm}0.08$    &  $\bf 0.64{\pm}0.11$  &    $0.46{\pm}0.04$    &    $0.08{\pm}0.16$    &    $0.27{\pm}0.04$    &    $0.37{\pm}0.03$    &    $0.35{\pm}0.01$    \\
glob Mat-3/2 (prelu)      &    $0.38{\pm}0.22$    &    $0.38{\pm}0.08$    &    $0.72{\pm}0.18$    &    $0.48{\pm}0.05$    &    $0.08{\pm}0.12$    &    $0.27{\pm}0.03$    &    $0.39{\pm}0.03$    &    $0.36{\pm}0.01$    \\

  \bottomrule
  \end{tabularx}
  \vspace*{-1em}
\end{table}

\subsection{Benchmark Classification Tasks}
\label{app:uci_class}

\begin{table}[t]
  \caption{Examples of UCI classification tasks, showing the globally stationary NN model directly gives competitive accuracy and mean negative log predictive density (NLPD) to non-stationary and locally stationary NN models. KFAC Laplace was used as the inference method.}
  \label{tbl:benchmarks_class}
  \scriptsize
  \tiny

    \setlength{\tabcolsep}{0pt}
  \setlength{\tblw}{0.09\textwidth}  
  \begin{tabularx}{\columnwidth}{l @{\extracolsep{\fill}} C{\tblw}  C{\tblw} C{\tblw} C{\tblw} C{\tblw}  C{\tblw} C{\tblw} C{\tblw}}
    \toprule

%NLPD_ACC (this table has been automatically generated, do not alter manually)
                          &       \multicolumn{2}{c}{\sc diabetes}        &         \multicolumn{2}{c}{\sc adult}         &       \multicolumn{2}{c}{\sc connect-4}       &        \multicolumn{2}{c}{\sc covtype}        \\
($n$, $d$)                &         \multicolumn{2}{c}{(768, 8)}          &        \multicolumn{2}{c}{(45222, 14)}        &        \multicolumn{2}{c}{(67556, 42)}        &       \multicolumn{2}{c}{(581912, 54)}        \\
($c$, $n_\textrm{batch}$) &          \multicolumn{2}{c}{(2, 50)}          &         \multicolumn{2}{c}{(2, 500)}          &         \multicolumn{2}{c}{(3, 500)}          &         \multicolumn{2}{c}{(7, 500)}          \\
\midrule
                          &         NLPD          &          ACC          &         NLPD          &          ACC          &         NLPD          &          ACC          &         NLPD          &          ACC          \\
\midrule
ReLU                      &    $0.48{\pm}0.05$    &    $0.76{\pm}0.04$    &  $\bf 0.31{\pm}0.01$  &    $0.85{\pm}0.00$    &    $0.57{\pm}0.01$    &    $0.82{\pm}0.00$    &    $0.18{\pm}0.00$    &    $0.93{\pm}0.00$    \\
loc RBF                   &    $0.48{\pm}0.04$    &    $0.76{\pm}0.04$    &    $0.31{\pm}0.01$    &    $0.85{\pm}0.00$    &    $0.54{\pm}0.01$    &    $0.81{\pm}0.00$    &    $0.19{\pm}0.01$    &    $0.93{\pm}0.00$    \\
glob RBF (sin)            &    $0.48{\pm}0.05$    &    $0.77{\pm}0.04$    &    $0.32{\pm}0.01$    &    $0.85{\pm}0.00$    &    $0.61{\pm}0.02$    &    $0.81{\pm}0.00$    &    $0.18{\pm}0.01$    &    $0.93{\pm}0.00$    \\
glob RBF (tri)            &    $0.48{\pm}0.05$    &    $0.77{\pm}0.04$    &    $0.31{\pm}0.01$    &    $0.85{\pm}0.00$    &    $0.61{\pm}0.01$    &    $0.82{\pm}0.00$    &    $0.18{\pm}0.01$    &    $0.93{\pm}0.00$    \\
loc Mat-5/2               &    $0.52{\pm}0.04$    &  $\bf 0.78{\pm}0.06$  &    $0.32{\pm}0.01$    &    $0.85{\pm}0.00$    &  $\bf 0.50{\pm}0.01$  &    $0.81{\pm}0.00$    &    $0.22{\pm}0.00$    &    $0.92{\pm}0.00$    \\
glob Mat-5/2 (sin)        &    $0.48{\pm}0.04$    &    $0.77{\pm}0.04$    &    $0.31{\pm}0.01$    &    $0.85{\pm}0.00$    &    $0.64{\pm}0.03$    &    $0.82{\pm}0.00$    &  $\bf 0.17{\pm}0.01$  &    $0.93{\pm}0.00$    \\
glob Mat-5/2 (tri)        &    $0.48{\pm}0.04$    &    $0.77{\pm}0.04$    &    $0.31{\pm}0.01$    &    $0.85{\pm}0.00$    &    $0.64{\pm}0.03$    &    $0.82{\pm}0.00$    &    $0.17{\pm}0.00$    &  $\bf 0.93{\pm}0.00$  \\
loc Mat-3/2               &    $0.49{\pm}0.04$    &    $0.76{\pm}0.05$    &    $0.32{\pm}0.01$    &  $\bf 0.85{\pm}0.01$  &    $0.53{\pm}0.01$    &    $0.81{\pm}0.00$    &    $0.19{\pm}0.01$    &    $0.93{\pm}0.00$    \\
glob Mat-3/2 (sin)        &    $0.48{\pm}0.04$    &    $0.77{\pm}0.04$    &    $0.32{\pm}0.01$    &    $0.85{\pm}0.00$    &    $0.66{\pm}0.03$    &    $0.81{\pm}0.00$    &    $0.17{\pm}0.01$    &    $0.93{\pm}0.00$    \\
glob Mat-3/2 (tri)        &    $0.48{\pm}0.04$    &    $0.78{\pm}0.04$    &    $0.32{\pm}0.01$    &    $0.85{\pm}0.00$    &    $0.66{\pm}0.02$    &    $0.81{\pm}0.00$    &    $0.18{\pm}0.01$    &    $0.93{\pm}0.00$    \\
glob RBF (sincos)         &    $0.48{\pm}0.05$    &    $0.77{\pm}0.05$    &    $0.32{\pm}0.01$    &    $0.85{\pm}0.00$    &    $0.61{\pm}0.01$    &    $0.82{\pm}0.00$    &    $0.18{\pm}0.00$    &    $0.93{\pm}0.00$    \\
glob Mat-5/2 (sincos)     &    $0.48{\pm}0.04$    &    $0.76{\pm}0.04$    &    $0.31{\pm}0.01$    &    $0.85{\pm}0.00$    &    $0.64{\pm}0.02$    &  $\bf 0.82{\pm}0.00$  &    $0.17{\pm}0.01$    &    $0.93{\pm}0.00$    \\
glob Mat-3/2 (sincos)     &    $0.48{\pm}0.04$    &    $0.76{\pm}0.04$    &    $0.32{\pm}0.01$    &    $0.85{\pm}0.00$    &    $0.66{\pm}0.03$    &    $0.81{\pm}0.01$    &    $0.18{\pm}0.00$    &    $0.93{\pm}0.00$    \\
glob RBF (prelu)          &    $0.48{\pm}0.05$    &    $0.77{\pm}0.04$    &    $0.31{\pm}0.01$    &    $0.85{\pm}0.00$    &    $0.61{\pm}0.02$    &    $0.82{\pm}0.00$    &    $0.18{\pm}0.00$    &    $0.93{\pm}0.00$    \\
glob Mat-5/2 (prelu)      &  $\bf 0.47{\pm}0.04$  &    $0.77{\pm}0.04$    &    $0.32{\pm}0.01$    &    $0.85{\pm}0.00$    &    $0.64{\pm}0.02$    &    $0.81{\pm}0.00$    &    $0.18{\pm}0.00$    &    $0.93{\pm}0.00$    \\
glob Mat-3/2 (prelu)      &    $0.48{\pm}0.04$    &    $0.77{\pm}0.05$    &    $0.32{\pm}0.01$    &    $0.85{\pm}0.00$    &    $0.65{\pm}0.02$    &    $0.81{\pm}0.00$    &    $0.18{\pm}0.01$    &    $0.93{\pm}0.00$    \\

    \bottomrule
  \end{tabularx}
  \vspace*{-1em}
\end{table}

\begin{table}[t]
  \caption{Examples of UCI classification tasks, showing the globally stationary NN model directly gives competitive area under receiver operating characteristic curve (AUC) to non-stationary and locally stationary NN models. KFAC Laplace was used as the inference method.}
  \label{tbl:benchmarks_class_auc}
  \scriptsize
  \tiny

    \setlength{\tabcolsep}{0pt}
  \setlength{\tblw}{0.09\textwidth}  
  \begin{tabularx}{\columnwidth}{l @{\extracolsep{\fill}} C{\tblw}  C{\tblw} C{\tblw} C{\tblw}}
    \toprule

%AUC (this table has been automatically generated, do not alter manually)
                          &                {\sc diabetes}                 &                  {\sc adult}                  &                {\sc connect-4}                &                 {\sc covtype}                 \\
($n$, $d$)                &                   (768, 8)                    &                  (45222, 14)                  &                  (67556, 42)                  &                 (581912, 54)                  \\
($c$, $n_\textrm{batch}$) &                    (2, 50)                    &                   (2, 500)                    &                   (3, 500)                    &                   (7, 500)                    \\
\midrule
                          &                      AUC                      &                      AUC                      &                      AUC                      &                      AUC                      \\
\midrule
ReLU                      &                $0.84{\pm}0.03$                &              $\bf 0.91{\pm}0.00$              &                $0.90{\pm}0.00$                &                $0.99{\pm}0.00$                \\
loc RBF                   &                $0.84{\pm}0.03$                &                $0.91{\pm}0.00$                &                $0.90{\pm}0.00$                &                $0.99{\pm}0.00$                \\
glob RBF (sin)            &              $\bf 0.84{\pm}0.03$              &                $0.91{\pm}0.00$                &              $\bf 0.90{\pm}0.00$              &              $\bf 0.99{\pm}0.00$              \\
glob RBF (tri)            &                $0.84{\pm}0.03$                &                $0.91{\pm}0.00$                &                $0.90{\pm}0.00$                &                $0.99{\pm}0.00$                \\
loc Mat-5/2               &                $0.84{\pm}0.03$                &                $0.91{\pm}0.00$                &                $0.89{\pm}0.00$                &                $0.99{\pm}0.00$                \\
glob Mat-5/2 (sin)        &                $0.84{\pm}0.03$                &                $0.91{\pm}0.00$                &                $0.90{\pm}0.00$                &                $0.99{\pm}0.00$                \\
glob Mat-5/2 (tri)        &                $0.83{\pm}0.03$                &                $0.91{\pm}0.00$                &                $0.90{\pm}0.00$                &                $0.99{\pm}0.00$                \\
loc Mat-3/2               &                $0.84{\pm}0.03$                &                $0.91{\pm}0.00$                &                $0.90{\pm}0.00$                &                $0.99{\pm}0.00$                \\
glob Mat-3/2 (sin)        &                $0.84{\pm}0.03$                &                $0.91{\pm}0.00$                &                $0.90{\pm}0.00$                &                $0.99{\pm}0.00$                \\
glob Mat-3/2 (tri)        &                $0.84{\pm}0.03$                &                $0.91{\pm}0.00$                &                $0.90{\pm}0.01$                &                $0.99{\pm}0.00$                \\
glob RBF (sincos)         &                $0.83{\pm}0.04$                &                $0.91{\pm}0.00$                &                $0.90{\pm}0.00$                &                $0.99{\pm}0.00$                \\
glob Mat-5/2 (sincos)     &                $0.83{\pm}0.03$                &                $0.91{\pm}0.00$                &                $0.90{\pm}0.00$                &                $0.99{\pm}0.00$                \\
glob Mat-3/2 (sincos)     &                $0.84{\pm}0.03$                &                $0.91{\pm}0.00$                &                $0.90{\pm}0.01$                &                $0.99{\pm}0.00$                \\
glob RBF (prelu)          &                $0.84{\pm}0.03$                &                $0.91{\pm}0.00$                &                $0.90{\pm}0.00$                &                $0.99{\pm}0.00$                \\
glob Mat-5/2 (prelu)      &                $0.84{\pm}0.03$                &                $0.91{\pm}0.00$                &                $0.90{\pm}0.00$                &                $0.99{\pm}0.00$                \\
glob Mat-3/2 (prelu)      &                $0.84{\pm}0.03$                &                $0.91{\pm}0.00$                &                $0.90{\pm}0.01$                &                $0.99{\pm}0.00$                \\

    \bottomrule
  \end{tabularx}
  \vspace*{-1em}
\end{table}

The experimental setup for the UCI \cite{UCI} classification tasks is the same as for the regression tasks, apart from the following details. For the UCI classification tasks, the NN architecture is a fully connected network with layers $d$-1000-1000-500-25-2000-$c$. The batch sizes used for each data set are listed in \cref{tbl:benchmarks_class}. The lengthscale parameter $\ell$ was initialized at value one.

The results for the diabetes (3--4 h) data set were calculated using a single CPU per experiment, and the results for the adult (4--6 h), connect-4 (6--8 h) and covtype (8--12 h) data sets were calculated using a single GPU per experiment. Copyright of the covtype data set: Jock A. Blackard and Colorado State University. 

\cref{tbl:benchmarks_class} and \cref{tbl:benchmarks_class_auc} show results on standard UCI \cite{UCI} classification data sets comparing results for different activation functions. We compare a neural network with a non-stationary model using the ReLU activation function to both local stationary \cite{meronen2020stationary} and global stationary models for different covariance functions. \cref{tbl:benchmarks_class} lists predictive accuracies and negative log predictive densities (NLPD) for each model. Moreover, \cref{tbl:benchmarks_class_auc} lists the area under the receiver operating characteristic curve (AUC) for the different models. The tables lists mean and standard deviation values across folds of the 10-fold cross-validation. The values for the best performing models are shown in bold. The classification results indicate that there are very few differences in the performance between the different models and that global stationary models achieve competitive predictive accuracy, NLPD, and AUC compared to the locally stationary and non-stationary models.

For the UCI classification tasks, we performed additional experiments using SWAG \cite{maddox19_SWAG}, inference instead of KFAC Laplace. The experiment setup here was also 10-fold cross-validation. The NN architecture is a fully connected network with layer widths d-1000-1000-500-50-c for all models. The models were trained for 20 epochs using batch sizes listed in \cref{tbl:benchmarks_swag} with Adam optimizer and a learning rate of $\num{1e-4}$. The learning rate for the lengthscale parameter $\ell$ was separately set to $0.01$ and initialized with one. A schedule was used for the Adam learning rates, decreasing them to one-tenth of the current value at epochs 10 and 15. The SWAG model was collected for 40 epochs (with $M=20$ samples to estimate the covariance matrix) using SGD (momentum 0.9) as the optimizer, updating the posterior estimate once per epoch. The lengthscale parameter is kept fixed during the SWAG model collection, as no SWAG posterior estimate is collected for it. The learning rate for SGD in the SWAG model collection part (the SWAG learning rate) was selected using Bayesian optimization with BoTorch in the range $(\num{1e-4}, 3)$, selecting the value providing the best negative log-likelihood on the validation set. We used one-fifth of the training set of the current fold for validation, and after selecting the best SWAG learning rate we trained the model from the beginning using the full training set for each fold using the best performing SWAG learning rate. We used a fixed value for $\ell$ equal to what the earlier optimization ended at (this is to prevent ending up in a different local optimum where the optimized SWAG learning rate does not provide good results). Each models SWAG learning rate was optimized individually, but a common SWAG learning rate was used for all ten folds of a single experiment. For model averaging, 50 samples from the approximate posterior were sampled. The SWAG results for all UCI classification data sets were calculated on a single GPU per experiment and the rough running times for each data set were diabetes: 2--4 h, adult 6--8 h, connect-4 8--10 h and covtype 15--20 h. For the SWAG results, \cref{tbl:benchmarks_swag} lists predictive accuracies and NLPDs, and \cref{tbl:benchmarks_swag_auc} lists AUC values for the different models. The results show that similar to the KFAC Laplace results, the global stationary models using periodic activation functions achieve competitive predictive accuracy, NLPD, and AUC compared to the locally stationary and non-stationary models. The main difference to results obtained using KFAC Laplace is that SWAG seems to produce more variability in the results between different models. The missing values in \cref{tbl:benchmarks_swag} and \cref{tbl:benchmarks_swag_auc} are due to the optimization diverging in the SWAG model collection phase.

\begin{table}[t]
  \caption{Examples of UCI classification tasks, showing the globally stationary NN model directly gives competitive accuracy and mean negative log predictive density (NLPD) to non-stationary and locally stationary NN models. SWAG was used as the inference method.}
  \label{tbl:benchmarks_swag}
  \scriptsize
  \tiny

    \setlength{\tabcolsep}{0pt}
  \setlength{\tblw}{0.09\textwidth}  
  \begin{tabularx}{\columnwidth}{l @{\extracolsep{\fill}} C{\tblw}  C{\tblw} C{\tblw} C{\tblw} C{\tblw}  C{\tblw} C{\tblw} C{\tblw}}
    \toprule

%NLPD_ACC (this table has been automatically generated, do not alter manually)
                          &       \multicolumn{2}{c}{\sc diabetes}        &         \multicolumn{2}{c}{\sc adult}         &       \multicolumn{2}{c}{\sc connect-4}       &        \multicolumn{2}{c}{\sc covtype}        \\
($n$, $d$)                &         \multicolumn{2}{c}{(768, 8)}          &        \multicolumn{2}{c}{(45222, 14)}        &        \multicolumn{2}{c}{(67556, 42)}        &       \multicolumn{2}{c}{(581912, 54)}        \\
($c$, $n_\textrm{batch}$) &          \multicolumn{2}{c}{(2, 50)}          &         \multicolumn{2}{c}{(2, 500)}          &         \multicolumn{2}{c}{(3, 500)}          &         \multicolumn{2}{c}{(7, 500)}          \\
\midrule
                          &         NLPD          &          ACC          &         NLPD          &          ACC          &         NLPD          &          ACC          &         NLPD          &          ACC          \\
\midrule
ReLU                      &    $0.53{\pm}0.07$    &    $0.75{\pm}0.03$    &    $0.38{\pm}0.15$    &    $0.79{\pm}0.18$    &    $0.54{\pm}0.12$    &    $0.79{\pm}0.04$    &    $0.19{\pm}0.00$    &    $0.92{\pm}0.00$    \\
loc RBF                   &    $0.49{\pm}0.05$    &    $0.76{\pm}0.04$    &    $0.33{\pm}0.01$    &    $0.85{\pm}0.00$    &  $\bf 0.47{\pm}0.01$  &    $0.81{\pm}0.00$    &    $0.19{\pm}0.00$    &    $0.93{\pm}0.00$    \\
glob RBF (sin)            &    $0.51{\pm}0.05$    &  $\bf 0.76{\pm}0.04$  &    $0.35{\pm}0.03$    &    $0.85{\pm}0.00$    &    $0.51{\pm}0.05$    &    $0.81{\pm}0.01$    &    $0.18{\pm}0.00$    &    $0.93{\pm}0.00$    \\
glob RBF (tri)            &    $0.53{\pm}0.07$    &    $0.73{\pm}0.04$    &    $0.33{\pm}0.01$    &    $0.85{\pm}0.00$    &    $0.51{\pm}0.05$    &    $0.81{\pm}0.01$    &    $0.19{\pm}0.01$    &    $0.92{\pm}0.00$    \\
loc Mat-5/2               &    $0.49{\pm}0.05$    &    $0.76{\pm}0.04$    &    $0.32{\pm}0.01$    &  $\bf 0.85{\pm}0.00$  &    $0.47{\pm}0.01$    &  $\bf 0.82{\pm}0.00$  &    $0.25{\pm}0.01$    &    $0.91{\pm}0.00$    \\
glob Mat-5/2 (sin)        &    $0.53{\pm}0.07$    &    $0.74{\pm}0.04$    &    $0.34{\pm}0.02$    &    $0.85{\pm}0.01$    &    $0.49{\pm}0.01$    &    $0.81{\pm}0.00$    &    $0.18{\pm}0.00$    &    $0.93{\pm}0.00$    \\
glob Mat-5/2 (tri)        &    $0.52{\pm}0.05$    &    $0.73{\pm}0.03$    &    $0.34{\pm}0.01$    &    $0.85{\pm}0.00$    &    $0.50{\pm}0.03$    &    $0.81{\pm}0.01$    &    $0.19{\pm}0.00$    &    $0.92{\pm}0.00$    \\
loc Mat-3/2               &  $\bf 0.49{\pm}0.04$  &    $0.75{\pm}0.03$    &  $\bf 0.32{\pm}0.01$  &    $0.85{\pm}0.00$    &    $0.47{\pm}0.01$    &    $0.82{\pm}0.00$    &    $0.23{\pm}0.01$    &    $0.91{\pm}0.00$    \\
glob Mat-3/2 (sin)        &    $0.57{\pm}0.07$    &    $0.73{\pm}0.04$    &    $0.35{\pm}0.02$    &    $0.84{\pm}0.00$    &    $0.50{\pm}0.01$    &    $0.81{\pm}0.00$    &    $0.19{\pm}0.01$    &    $0.93{\pm}0.00$    \\
glob Mat-3/2 (tri)        &    $0.55{\pm}0.07$    &    $0.74{\pm}0.04$    &    $0.34{\pm}0.01$    &    $0.85{\pm}0.01$    &    $0.50{\pm}0.02$    &    $0.80{\pm}0.00$    &    $0.19{\pm}0.00$    &    $0.93{\pm}0.00$    \\
glob RBF (sincos)         &           ---         &          ---          &    $0.37{\pm}0.10$    &    $0.83{\pm}0.07$    &    $0.54{\pm}0.07$    &    $0.80{\pm}0.01$    &    $0.18{\pm}0.01$    &    $0.93{\pm}0.00$    \\
glob Mat-5/2 (sincos)     &           ---         &          ---          &    $0.34{\pm}0.02$    &    $0.85{\pm}0.01$    &    $0.52{\pm}0.03$    &    $0.81{\pm}0.01$    &    $0.18{\pm}0.01$    &  $\bf 0.93{\pm}0.00$  \\
glob Mat-3/2 (sincos)     &    $0.58{\pm}0.05$    &    $0.72{\pm}0.03$    &    $0.39{\pm}0.11$    &    $0.81{\pm}0.08$    &    $0.51{\pm}0.02$    &    $0.81{\pm}0.00$    &  $\bf 0.18{\pm}0.01$  &    $0.93{\pm}0.00$    \\
glob RBF (prelu)          &    $0.52{\pm}0.06$    &    $0.76{\pm}0.04$    &    $0.34{\pm}0.01$    &    $0.85{\pm}0.00$    &    $0.50{\pm}0.01$    &    $0.81{\pm}0.00$    &    $0.19{\pm}0.01$    &    $0.92{\pm}0.00$    \\
glob Mat-5/2 (prelu)      &    $0.53{\pm}0.05$    &    $0.75{\pm}0.03$    &    $0.33{\pm}0.01$    &    $0.85{\pm}0.00$    &    $0.49{\pm}0.01$    &    $0.81{\pm}0.00$    &    $0.19{\pm}0.00$    &    $0.92{\pm}0.00$    \\
glob Mat-3/2 (prelu)      &    $0.58{\pm}0.07$    &    $0.73{\pm}0.03$    &    $0.34{\pm}0.01$    &    $0.85{\pm}0.00$    &    $0.49{\pm}0.01$    &    $0.81{\pm}0.00$    &    $0.19{\pm}0.01$    &    $0.93{\pm}0.00$    \\

    \bottomrule
  \end{tabularx}
  \vspace*{-1em}
\end{table}

\begin{table}[t]
  \caption{Examples of UCI classification tasks, showing the globally stationary NN model directly gives competitive area under receiver operating characteristic curve (AUC) to non-stationary and locally stationary NN models. SWAG was used as the inference method.}
  \label{tbl:benchmarks_swag_auc}
  \scriptsize
  \tiny

    \setlength{\tabcolsep}{0pt}
  \setlength{\tblw}{0.09\textwidth}  
  \begin{tabularx}{\columnwidth}{l @{\extracolsep{\fill}} C{\tblw}  C{\tblw} C{\tblw} C{\tblw}}
    \toprule

%AUC (this table has been automatically generated, do not alter manually)
                          &                {\sc diabetes}                 &                  {\sc adult}                  &                {\sc connect-4}                &                 {\sc covtype}                 \\
($n$, $d$)                &                   (768, 8)                    &                  (45222, 14)                  &                  (67556, 42)                  &                 (581912, 54)                  \\
($c$, $n_\textrm{batch}$) &                    (2, 50)                    &                   (2, 500)                    &                   (3, 500)                    &                   (7, 500)                    \\
\midrule
                          &                      AUC                      &                      AUC                      &                      AUC                      &                      AUC                      \\
\midrule
ReLU                      &                $0.82{\pm}0.04$                &                $0.87{\pm}0.12$                &                $0.86{\pm}0.08$                &                $0.99{\pm}0.00$                \\
loc RBF                   &                $0.83{\pm}0.03$                &                $0.91{\pm}0.00$                &              $\bf 0.90{\pm}0.00$              &                $0.99{\pm}0.00$                \\
glob RBF (sin)            &                $0.82{\pm}0.04$                &                $0.90{\pm}0.01$                &                $0.89{\pm}0.01$                &              $\bf 0.99{\pm}0.00$              \\
glob RBF (tri)            &                $0.81{\pm}0.04$                &                $0.91{\pm}0.00$                &                $0.89{\pm}0.01$                &                $0.99{\pm}0.00$                \\
loc Mat-5/2               &                $0.84{\pm}0.03$                &                $0.91{\pm}0.00$                &                $0.90{\pm}0.00$                &                $0.99{\pm}0.00$                \\
glob Mat-5/2 (sin)        &                $0.80{\pm}0.05$                &                $0.90{\pm}0.01$                &                $0.89{\pm}0.01$                &                $0.99{\pm}0.00$                \\
glob Mat-5/2 (tri)        &                $0.81{\pm}0.04$                &                $0.90{\pm}0.00$                &                $0.89{\pm}0.01$                &                $0.99{\pm}0.00$                \\
loc Mat-3/2               &                $0.83{\pm}0.03$                &              $\bf 0.91{\pm}0.00$              &                $0.90{\pm}0.01$                &                $0.99{\pm}0.00$                \\
glob Mat-3/2 (sin)        &                $0.79{\pm}0.04$                &                $0.90{\pm}0.01$                &                $0.89{\pm}0.00$                &                $0.99{\pm}0.00$                \\
glob Mat-3/2 (tri)        &                $0.81{\pm}0.04$                &                $0.90{\pm}0.01$                &                $0.89{\pm}0.01$                &                $0.99{\pm}0.00$                \\
glob RBF (sincos)         &                      ---                      &                $0.85{\pm}0.16$                &                $0.88{\pm}0.02$                &                $0.99{\pm}0.00$                \\
glob Mat-5/2 (sincos)     &                      ---                      &                $0.90{\pm}0.00$                &                $0.88{\pm}0.01$                &                $0.99{\pm}0.00$                \\
glob Mat-3/2 (sincos)     &                $0.78{\pm}0.03$                &                $0.84{\pm}0.15$                &                $0.88{\pm}0.01$                &                $0.99{\pm}0.00$                \\
glob RBF (prelu)          &                $0.81{\pm}0.04$                &                $0.91{\pm}0.01$                &                $0.89{\pm}0.00$                &                $0.99{\pm}0.00$                \\
glob Mat-5/2 (prelu)      &                $0.81{\pm}0.03$                &                $0.90{\pm}0.01$                &                $0.89{\pm}0.01$                &                $0.99{\pm}0.00$                \\
glob Mat-3/2 (prelu)      &                $0.79{\pm}0.03$                &                $0.91{\pm}0.01$                &                $0.89{\pm}0.01$                &                $0.99{\pm}0.00$                \\

    \bottomrule
  \end{tabularx}
  \vspace*{-1em}
\end{table}

\subsection{Detection of Distribution Shift with Rotated MNIST}
\label{app:mnist}
For the MNIST (\cite{MNIST}, available under CC~BY-SA~3.0) digit classification experiment, the feature extractor part of the NN architecture has two convolutional layers (32 and 64 channels, both using a $3\times3$ kernel) followed by a fully connected layer taking the dimensionality down to 25, and the following model layer has 2000 hidden units. The models were trained on the MNIST training set for 50 epochs using a batch size of 64 with an SGD optimizer (learning rate $\num{1e-3}$, momentum $0.9$), using only unrotated images. A schedule was used for the SGD learning rates, decreasing them to 0.9 of the current value at epochs 25 and 37. The learning rate for the lengthscale parameter $\ell$ was set to $\num{1e-4}$, and was initialized with $0.2$. The posterior inference was performed using KFAC Laplace \cite{ritter2018a_kfac_laplace} with a fixed variance scale of one. For model averaging, we used 30 posterior samples. We tested the trained model on the standard unrotated MNIST test set and rotated versions of the same test set for rotation angles every $10^\circ$ up to $360^\circ$. Running this experiment for one model on one GPU took roughly 2 hours due to testing the model on multiple test sets.

\cref{fig:mnist} shows the results on the rotated MNIST experiment for different models. We evaluate the predictive accuracy, mean confidence of the predicted class, and NLPD on the rotated test sets. The results indicate that all models obtain similar accuracy results, while only local and global stationary models do not result in over-confident uncertainty estimates. For an ideally calibrated model, the mean confidence would decrease as low as the accuracy curve when the digits are rotated, which would keep the NLPD values as low as possible. We see that even for the local and global stationary models, the mean confidence has a minimum of around $0.55$ while the accuracy decreases below $0.2$. We also observe that the NLPD curves rise to high values (over 3) for the better performing models. The accuracy of all models increases near the $180^\circ$ rotation. This is most likely due to numbers 0, 1 and 8 appearing similar with $0^\circ$ and $180^\circ$ rotations. Interestingly, the NLPD values for the local and global stationary models hardly decrease for the rotation angle of $180^\circ$ although the accuracy for this angle increases compared to adjacent angles. This could be due to number 6 looking like number 9 at $180^\circ$ rotation, and vice versa, causing the model to make overconfident incorrect predictions. Although, this kind of overconfident misclassification cannot be prevented even with a correctly calibrated model, as samples of one class genuinely appear to belong to another class.

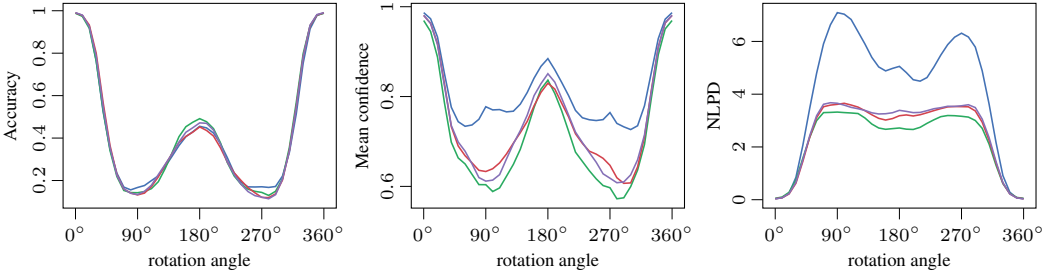
\begin{figure}[b]
  \centering\scriptsize
  \setlength{\figurewidth}{.26\textwidth}
  \setlength{\figureheight}{.75\figurewidth}
  \pgfplotsset{scale only axis,y tick label style={rotate=90},
    xtick={0,90,180,270,360},
    xticklabels={$0^\circ$,$90^\circ$,$180^\circ$,$270^\circ$,$360^\circ$}}
  \begin{subfigure}[t]{.33\textwidth}
    \centering
    \pgfplotsset{ylabel={Accuracy}}
    % This file was created by tikzplotlib v0.9.8.
\begin{tikzpicture}

\definecolor{color0}{rgb}{0,0.75,0.75}

\begin{axis}[
height=\figureheight,
legend cell align={left},
legend style={
  fill opacity=0.8,
  draw opacity=1,
  text opacity=1,
  at={(0.5,0.91)},
  anchor=north,
  draw=white!80!black
},
tick align=outside,
tick pos=left,
width=\figurewidth,
x grid style={white!69.0196078431373!black},
xlabel={rotation angle},
xmin=-18, xmax=378,
xtick style={color=black},
y grid style={white!69.0196078431373!black},
ymin=0.07005, ymax=1.03475,
ytick style={color=black}
]
\addplot [semithick, mycolor0]
table {%
0 0.9894
10 0.9773
20 0.9259
30 0.7646
40 0.525
50 0.3483
60 0.2345
70 0.1691
80 0.1559
90 0.1672
100 0.1764
110 0.1995
120 0.222
130 0.2653
140 0.3144
150 0.3647
160 0.4067
170 0.4298
180 0.4554
190 0.4479
200 0.4238
210 0.3741
220 0.3081
230 0.2356
240 0.1927
250 0.1701
260 0.1695
270 0.1705
280 0.1675
290 0.1719
300 0.2209
310 0.3326
320 0.5143
330 0.7531
340 0.9146
350 0.9779
360 0.9894
};
%\addlegendentry{ReLU}
\addplot [semithick, mycolor1]
table {%
0 0.9881
10 0.9734
20 0.9169
30 0.7625
40 0.5314
50 0.3419
60 0.2187
70 0.1532
80 0.1432
90 0.1425
100 0.1478
110 0.1617
120 0.1942
130 0.2565
140 0.3307
150 0.4065
160 0.4536
170 0.4746
180 0.4915
190 0.4769
200 0.4446
210 0.3785
220 0.2849
230 0.2214
240 0.1823
250 0.1593
260 0.1491
270 0.1439
280 0.1297
290 0.1493
300 0.2083
310 0.3561
320 0.5683
330 0.7981
340 0.929
350 0.9792
360 0.9881
};
%\addlegendentry{local RBF}
\addplot [semithick, mycolor2]
table {%
0 0.9898
10 0.9782
20 0.9326
30 0.7998
40 0.5685
50 0.3659
60 0.228
70 0.1601
80 0.141
90 0.1322
100 0.1407
110 0.1714
120 0.2169
130 0.2739
140 0.3264
150 0.379
160 0.4077
170 0.4278
180 0.4526
190 0.4358
200 0.4096
210 0.3549
220 0.2883
230 0.2363
240 0.2
250 0.1664
260 0.1422
270 0.1205
280 0.1194
290 0.1371
300 0.202
310 0.3418
320 0.5482
330 0.7833
340 0.9293
350 0.9799
360 0.9898
};
%\addlegendentry{global RBF (sin)}
\addplot [semithick, mycolor3]
table {%
0 0.9909
10 0.9793
20 0.9251
30 0.7854
40 0.5497
50 0.3551
60 0.2251
70 0.1606
80 0.1398
90 0.1337
100 0.1493
110 0.1779
120 0.2277
130 0.2761
140 0.3255
150 0.3815
160 0.4274
170 0.4487
180 0.4716
190 0.4695
200 0.4377
210 0.372
220 0.2889
230 0.2281
240 0.1821
250 0.148
260 0.1287
270 0.1214
280 0.1139
290 0.1333
300 0.2067
310 0.3389
320 0.5443
330 0.7821
340 0.9323
350 0.982
360 0.9909
};
%\addlegendentry{global RBF (sincos)}
\end{axis}

\end{tikzpicture}
  \end{subfigure}
  \hfill
  \begin{subfigure}[t]{.33\textwidth}
    \centering
    \pgfplotsset{ylabel={Mean confidence}}
    % This file was created by tikzplotlib v0.9.8.
\begin{tikzpicture}

\definecolor{color0}{rgb}{0,0.75,0.75}

\begin{axis}[
height=\figureheight,
legend cell align={left},
legend style={
  fill opacity=0.8,
  draw opacity=1,
  text opacity=1,
  at={(0.5,0.91)},
  anchor=north,
  draw=white!80!black
},
tick align=outside,
tick pos=left,
width=\figurewidth,
x grid style={white!69.0196078431373!black},
xlabel={rotation angle},
xmin=-18, xmax=378,
xtick style={color=black},
y grid style={white!69.0196078431373!black},
ymin=0.551662600067779, ymax=1.00752924106814,
ytick style={color=black}
]
\addplot [semithick, mycolor0]
table {%
0 0.986808030113578
10 0.973201271280646
20 0.932509878438711
30 0.851639797693491
40 0.776243603672087
50 0.74401922993809
60 0.733813543771207
70 0.736705477371812
80 0.749341819496453
90 0.777350408600271
100 0.770257599553466
110 0.771011524873972
120 0.765857543881238
130 0.767413622687757
140 0.783077833661437
150 0.808896014085412
160 0.84112792122364
170 0.867321238091588
180 0.884482622718811
190 0.859663247078657
200 0.831506223995984
210 0.802131471525133
220 0.769092435590923
230 0.752795311631262
240 0.748354658788443
250 0.746590836462378
260 0.748664592123032
270 0.764338696241379
280 0.740216035939753
290 0.732605578437448
300 0.72651723266989
310 0.734959142491221
320 0.779379701694846
330 0.854100380784273
340 0.929003563915193
350 0.972177600827813
360 0.986808030113578
};
%\addlegendentry{ReLU}
\addplot [semithick, mycolor1]
table {%
0 0.969271202667057
10 0.94479585352987
20 0.887832222336531
30 0.786721962553263
40 0.698337694203854
50 0.663311849060655
60 0.649955681405961
70 0.62685513945967
80 0.603927714048326
90 0.603779025371373
100 0.588477497296035
110 0.596622700802982
120 0.622647976149619
130 0.649124596320093
140 0.682527582891285
150 0.727462154877186
160 0.775814565764368
170 0.816438531251252
180 0.836501944155991
190 0.806853370702267
200 0.768234280474484
210 0.722828339667618
220 0.676210015454888
230 0.655616202194989
240 0.635014576251805
250 0.616848098105192
260 0.601829960769415
270 0.596970109887421
280 0.572383811022341
290 0.574584718035162
300 0.599470624107122
310 0.636265324497223
320 0.693235947997868
330 0.797003128673136
340 0.893340540593863
350 0.950513058947027
360 0.969271202667057
};
%\addlegendentry{local RBF}
\addplot [semithick, mycolor2]
table {%
0 0.98059965454638
10 0.963131069150567
20 0.91794869863838
30 0.829139077426493
40 0.733078937846422
50 0.689938621881604
60 0.668280015636981
70 0.652330167470872
80 0.635755693398416
90 0.633123172439635
100 0.639254808442295
110 0.652278632135689
120 0.672192109708488
130 0.696382996308804
140 0.719504019732773
150 0.744577683630586
160 0.779507695473731
170 0.809466393846273
180 0.829744100518525
190 0.816754542067647
200 0.785106913022697
210 0.754595128405094
220 0.717128386877477
230 0.695747672151029
240 0.678384309716523
250 0.671876093122363
260 0.662469867229462
270 0.646853152945638
280 0.619207475244999
290 0.606795843498409
300 0.607708546575904
310 0.64336448174566
320 0.709670316757262
330 0.81470299269259
340 0.910322864064574
350 0.963715863214433
360 0.98059965454638
};
%\addlegendentry{global RBF (sin)}
\addplot [semithick, mycolor3]
table {%
0 0.980881535884738
10 0.962760939055681
20 0.915769310812652
30 0.824296689674258
40 0.735506895385683
50 0.698830663825572
60 0.678274271513522
70 0.650167554502189
80 0.619592348769307
90 0.611718726289272
100 0.613935938346386
110 0.62610617736727
120 0.661149793668091
130 0.694904668238759
140 0.724088194049895
150 0.758338385100663
160 0.802145968744159
170 0.828787429577112
180 0.850824243809283
190 0.83213881816566
200 0.796226164737344
210 0.75874702257961
220 0.72089146836549
230 0.697678053319454
240 0.675384826521575
250 0.653809345284104
260 0.627942215409875
270 0.618196920141578
280 0.608138467538357
290 0.609860615289211
300 0.625630161769688
310 0.659171994511783
320 0.714546117295325
330 0.811244914884865
340 0.90880887786448
350 0.964706841744483
360 0.980881535884738
};
%\addlegendentry{global RBF (sincos)}
\end{axis}

\end{tikzpicture}  
  \end{subfigure}  
  \hfill
  \begin{subfigure}[t]{.33\textwidth}
    \centering
    \pgfplotsset{ylabel={NLPD}}
    % This file was created by tikzplotlib v0.9.8.
\begin{tikzpicture}

\definecolor{color0}{rgb}{0,0.75,0.75}

\begin{axis}[
height=\figureheight,
legend cell align={left},
legend style={
  fill opacity=0.8,
  draw opacity=1,
  text opacity=1,
  at={(0.5,0.09)},
  anchor=south,
  draw=white!80!black
},
tick align=outside,
tick pos=left,
width=\figurewidth,
x grid style={white!69.0196078431373!black},
xlabel={rotation angle},
xmin=-18, xmax=378,
xtick style={color=black},
y grid style={white!69.0196078431373!black},
ymin=-0.317728671341426, ymax=7.4435684661468,
ytick style={color=black}
]
\addplot [semithick, mycolor0]
table {%
0 0.0359461695163048
10 0.0739641479233599
20 0.244089897698077
30 0.846575057450566
40 2.05660972574972
50 3.46678850178669
60 4.79529505938238
70 5.90900807134627
80 6.63036753851327
90 7.09078223262461
100 7.0325210662337
110 6.83210460245876
120 6.36908744202303
130 5.90398732600434
140 5.39576712686765
150 5.03242608027488
160 4.88294628962424
170 4.97704442216085
180 5.05434527419239
190 4.81979971712058
200 4.54847090442721
210 4.49410265556114
220 4.63047724685272
230 5.03100186575715
240 5.48282519753944
250 5.87872691894968
260 6.16080067021189
270 6.31151231292246
280 6.1682255483034
290 5.78596673001035
300 4.88578976580472
310 3.62323810380715
320 2.172974192335
330 0.914489967468447
340 0.284347278474543
350 0.0717311038145464
360 0.0359461695163048
};
%\addlegendentry{ReLU}
\addplot [semithick, mycolor1]
table {%
0 0.0523426355045362
10 0.102377042482143
20 0.278651293012255
30 0.74245567833089
40 1.57098411656028
50 2.36251782345096
60 2.99806892965618
70 3.30131022625362
80 3.30959222571746
90 3.3214906309854
100 3.302563994485
110 3.28748432351677
120 3.26675464538109
130 3.09865368713433
140 2.86822125424785
150 2.71834301126055
160 2.66895863858634
170 2.69410331028401
180 2.71841418363494
190 2.67406171635607
200 2.6623055917601
210 2.74343033752277
220 2.89503112171771
230 3.02902539434979
240 3.14782558619554
250 3.19141543196862
260 3.18414513784787
270 3.16231218879455
280 3.13153041763346
290 3.00275104433647
300 2.7057631912766
310 2.10649215795543
320 1.3074093732458
330 0.609273756507569
340 0.238047243801727
350 0.0878432127295169
360 0.0523426355045362
};
%\addlegendentry{local RBF}
\addplot [semithick, mycolor2]
table {%
0 0.0365139102804614
10 0.0753361357810631
20 0.216904764425872
30 0.624004913079834
40 1.44953525683671
50 2.36971958713612
60 3.11222220713496
70 3.52436395883421
80 3.58622510336864
90 3.61478999460227
100 3.65528102419632
110 3.58932990052024
120 3.51076894868005
130 3.37420716830077
140 3.21936218667535
150 3.08367989087917
160 3.02051638233961
170 3.07543926807911
180 3.18340861320808
190 3.21383642022675
200 3.1743880944972
210 3.22052083405342
220 3.29624519607517
230 3.37709612712322
240 3.45262515053011
250 3.51789262785627
260 3.53145003190852
270 3.52772280430106
280 3.52128735182045
290 3.36458943338114
300 2.93804639145256
310 2.24969264132884
320 1.39245854703205
330 0.630033174287377
340 0.221883736031563
350 0.0728310438082938
360 0.0365139102804614
};
%\addlegendentry{global RBF (sin)}
\addplot [semithick, mycolor3]
table {%
0 0.0350575621807663
10 0.073790503918998
20 0.229562407083471
30 0.65758774713311
40 1.510193794352
50 2.42565445868591
60 3.18582333405017
70 3.61766056721543
80 3.67662645798662
90 3.65582927710949
100 3.58484120148702
110 3.50071203275774
120 3.44858193865155
130 3.38537634073144
140 3.2945672368871
150 3.25016096712195
160 3.259025722969
170 3.29654001087817
180 3.38774304485098
190 3.34804172690663
200 3.30051869952328
210 3.30926568037895
220 3.37478524659808
230 3.44864260388843
240 3.4881963329413
250 3.5421869334306
260 3.53756573056278
270 3.56023088971032
280 3.5999676065266
290 3.49232830277575
300 3.05604351950352
310 2.35271349636051
320 1.44833372931817
330 0.648617536184316
340 0.219365097637988
350 0.0679391465332454
360 0.0350575621807663
};
%\addlegendentry{global RBF (sincos)}
\end{axis}

\end{tikzpicture}
  \end{subfigure}\\[-1.2em]
  \caption{Rotated MNIST maximum a posteriori (MAP) results: The models have been trained on unrotated digits. The test-set digits are rotated at test time to show the sensitivity of the trained model to perturbations. Model predictions are based on a MAP estimate of model parameters. All models perform equally in terms of accuracy, while ReLU (\ref{plt:relu}) shows overconfidence in terms of mean confidence and NLPD. The stationary RBF models (\ref{plt:locrbf}~local, \ref{plt:sin}~sin, \ref{plt:sincos}~sin--cos) capture uncertainty.}
  \label{fig:mnist_map}
\end{figure}

\cref{fig:mnist_map} shows additional results on the rotated MNIST experiment for different models, using only a maximum a posteriori (MAP) estimate for the model parameters. The models used for the MAP results are the same trained models that were used for the results in \cref{fig:mnist}, but for the MAP results the KFAC Laplace inference step is skipped. The MAP results are almost identical to the KFAC Laplace results, except that 90$^{\circ}$ and 270$^{\circ}$ rotations for mean confidence have slightly higher values for the MAP results. This suggests that the KFAC Laplace inference might not be very successful in improving data set shift detection properties in this experiment.

\subsection{Out-of-distribution Detection Using CIFAR-10, CIFAR-100, and SVHN}
\label{app:cifar}
For this experiment, the feature extractor part is a GoogLeNet \cite{szegedy2015going} followed by a 512 node wide model layer. Pre-trained weights are used for the feature extractor part of the NN, and kept unchanged during the model training (pre-trained model from \url{https://github.com/huyvnphan/PyTorch_CIFAR10}). The models were trained on CIFAR-10 \cite{CIFAR} for 20 epochs using a batch size of 128 with Adam optimiser and a learning rate of $\num{1e-4}$. The learning rate for the lengthscale parameter $\ell$ was separately set to $\num{1e-5}$, and the lengthscale parameter was initialised with one. The posterior inference was performed using SWAG \cite{maddox19_SWAG}. The SWAG model was collected for 40 epochs ($M=20$) using SGD (momentum 0.9) as the optimiser, updating the posterior estimate once per epoch. The lengthscale parameter is kept fixed during the SWAG model collection. The learning rate for SGD in the SWAG model collection part (the SWAG learning rate) was selected using Bayesian optimisation with BoTorch in the range $(\num{1e-4}, 3)$ based on the negative log-likelihood on the validation set. As validation data, we used the CIFAR-10 test set. Using the CIFAR-10 test set as the validation set for selecting hyperparameters is valid here; as for this experiment, the focus is not on measuring the performance on the test set but evaluating OOD detection performance on the CIFAR-100 and SVHN test sets. After selecting the best SWAG learning rate, we trained the model from the beginning using the best performing SWAG learning rate and a fixed value for $\ell$ equal to the earlier optimisation. For each model, the SWAG learning rate was optimised individually. For model averaging, 30 samples from the approximate posterior of the parameters were sampled to calculate the predictions on CIFAR-10, CIFAR-100 \cite{CIFAR}, and SVHN \cite{SVHN}. Running this experiment for one model on one GPU took roughly one day due to the Bayesian optimization process.

\cref{fig:histograms_app} compares model performance on out-of-distribution detection for image classification for non-stationary, local stationary and global stationary models. Both CIFAR-100 (more similar) and SVHN (more dissimilar to CIFAR-10) images are OOD data, and the models should show high uncertainties (high predictive entropy, high predictive variance) for the respective test images. The histograms of predictive entropies for different test sets show that most models can separate between in-distribution and OOD data based on this metric. However, the predictive marginal variance histograms show that the global stationary models can better detect the OOD samples compared to ReLU and local stationary models. Interestingly, the ReLU model shows higher variance on CIFAR-100 images compared to SVHN images, although SVHN images are more different from the training set images. For global stationary models, both entropy and variance histograms show that the models clearly consider SVHN more OOD than CIFAR-100, which is intuitive as CIFAR-100 resembles CIFAR-10 more than SVHN. \cref{fig:histograms_app} also shows sample images for most/least similar to the training data distribution that the model has learned. Looking at these images for different models, we can see that for CIFAR-10, images with dark background result in high uncertainty for all models. For the CIFAR-100 sample images, the global stationary Mat\'ern-$\nicefrac{3}{2}$ model has classified pictures of animals and humans with the highest confidence, which seems intuitive as these could be considered resembling some of the CIFAR-10 classes (for example, dogs or cats). Moreover, the images with the highest uncertainty seem visually very different from CIFAR-10 images. For the SVHN sample images, all models seem to be most confident about clear and sharp images and blurry images result in high uncertainty, which is reasonable as CIFAR-10 images usually have clear shapes. However, this is again most apparent for the global stationary models, suggesting the model has learned meaningful representations of the input space.

Using the same results that are visualized in the histograms in \cref{fig:histograms_app}, we calculated area under receiver operating characteristic curve (AUC) and area under precision-recall curve (AUPR) values for OOD detection in the CIFAR-10 experiment to provide additional quantitative results, treating either CIFAR-100 or SVHN as the OOD data set. We calculated the AUC and AUPR measures using the marginal variance as the metric to determine whether a sample is OOD or not. We consider this a better metric for OOD detection compared to predictive entropy, as in-distribution samples that are hard to classify are expected to have high predictive entropy, not necessarily allowing the detection of OOD samples based on this metric. \cref{tbl:cifar_auroc} lists the calculated AUC and AUPR numbers. The results for SVHN as the OOD data set indicate a clear difference between the non-stationary (ReLU) and the stationary (local and global) models. The globally stationary RBF model achieves the best AUC score. Treating CIFAR-100 as an OOD set is not as straightforward considering numerical AUC and AUPR comparisons. CIFAR-100 images are visually very similar to CIFAR-100 images but representing different classes, and hence can be considered not strictly OOD. For example, it is reasonable to expect that even a correctly operating model may consider some CIFAR-100 images more in-distribution than the most difficult or visually peculiar CIFAR-10 test images. For this reason, it is reasonable to compare AUPR and AUC numbers of each model for the two OOD data sets, CIFAR-100 and SVHN, and observe which data set is considered more OOD. We expect SVHN to be considered more OOD, which is true except for the ReLU model based on the AUC metric.

\begin{table}[t]
  \caption{Table of numerical results on the image classification OOD task. The results used to calculate the numbers in this table are the same that were used to create histograms in \cref{fig:histograms_app}. The table lists the area under receiver operating characteristic curve (AUC) and the area under precision-recall curve (AUPR) for each of the models. Numbers are calculated both for considering CIFAR-100 or SVHN as the OOD set, while CIFAR-10 is the in-distribution data. The table also visualizes whether each model considers CIFAR-100 or SVHN more OOD based on which data set is detected as OOD more effectively.}
  \label{tbl:cifar_auroc}
  \centering
  \scriptsize
  \tiny

    \setlength{\tabcolsep}{0pt}
  \setlength{\tblw}{0.09\textwidth}
  \begin{tabularx}{0.5\columnwidth}{l @{\extracolsep{\fill}} C{\tblw}  C{0.2\tblw} C{\tblw} C{\tblw} C{0.2\tblw} C{\tblw}}
   \toprule
                          &       \multicolumn{3}{c}{AUC}        &         \multicolumn{3}{c}{AUPR}        \\
\midrule
 OOD data set                        &         CIFAR-100          &   &       SVHN         &         CIFAR-100          &  &       SVHN        \\
\midrule
ReLU                      &    $0.974$ &  >&  $0.961$    &    $0.963$   &   <  &  $0.970$    \\
loc RBF                   &  $\bf 0.976$ &< &   $0.987$ &    $\bf 0.976$     &  < &    $\bf 0.995$    \\
glob RBF (sin)            &    $0.942$  &  <&  $\bf 0.988$    &  $0.940$  & < &    $\bf 0.995$     \\
loc Mat-3/2              &    $0.973$ &<  & $0.983$      &    $0.972$    &< &   $0.993$     \\
glob Mat-3/2 (sin)         &    $0.965$  &< &  $0.981$  &    $0.965$   & < &  $0.993$      \\
    \bottomrule
  \end{tabularx}
  \vspace*{-1em}
\end{table}

\newsavebox{\sinfunc}
\savebox{\sinfunc}{%
\begin{tikzpicture}[font=\small,>=stealth,yscale=0.05,xscale=0.05]
  \draw[black!75] (3,0) sin (4,1) cos (5,0) sin (6,-1) cos (7,0)
          sin (8,1) cos (9,0) sin (10,-1) cos (11,0);
\end{tikzpicture}%
}

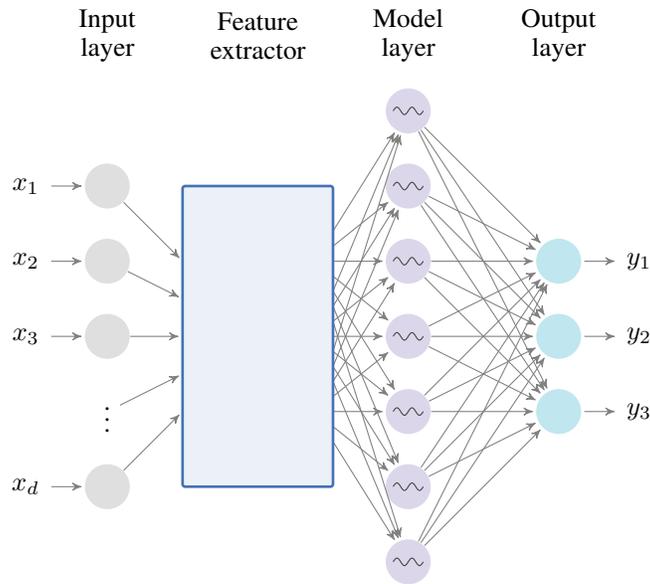
\begin{figure}[t]
\vspace*{1.0cm}
\centering
\begin{tikzpicture}[shorten >=1pt,->,draw=black!50]

% Define block/line styles
\tikzstyle{block}=[shape=rectangle, draw=mycolor0, line width=1, fill=mycolor0!10, rounded corners=1pt]
\tikzstyle{every pin edge}=[<-,shorten <=1pt]
\tikzstyle{neuron}=[circle,fill=black!25,minimum size=17pt,inner sep=0pt]
\tikzstyle{annot} = [text width=4em, text centered]

\foreach \name / \y in {1,...,3}
  \node[neuron, pin=left:{$x_{\y}$}, fill=mylcolor6] (I-\name) at (0,-\y) {};
  
\node (I-4) at (0,-4) {\vdots};
\node[neuron, pin=left:{$x_{d}$}, fill=mylcolor6] (I-5) at (0,-5) {};

\foreach \name / \y in {2,...,4}
  \node[] (H-\name) at (2.75,-\y) {};

\foreach \name / \y in {0,...,6}
  \node[neuron, fill=mylcolor3] (P-\name) at (4,-\y) {\usebox{\sinfunc}};

\foreach \name / \y in {1,...,3}
  \node[neuron, fill=mylcolor5, pin={[pin edge={->}]right:{$y_{\y}$}}] (O-\name) at (6,-\y-1) {};

\foreach \source in {2,...,4}
  \foreach \dest in {0,...,6}
    \path (H-\source) edge (P-\dest);
  
\foreach \source in {0,...,6}
  \foreach \dest in {1,...,3}
    \path (P-\source) edge (O-\dest);
    
\node[block, minimum size=2cm, minimum height = 4cm](feature-NN) at (2,-3) {};

\foreach \source in {1,...,5} {
  \path (I-\source) edge (feature-NN);
}

\node[annot, above of=feature-NN, node distance=4cm] (hl) {Feature extractor};
\node[annot, left of=hl, node distance=2cm] (hl2) {Input layer};
\node[annot, right of=hl, node distance=2cm] (hl3) {Model layer};
\node[annot, right of=hl3, node distance=2cm] (hl4) {Output layer};

\end{tikzpicture}
  \caption{An illustrative figure describing the model architecture. The model first passes an input $\textbf{x} \in \mathbb{R}^d$ into a feature extractor. The feature extractor part is a task dependent neural network architecture, which can be for example a fully connected structure or some convolutional layers. The feature extractor results in some $L$-dimensional representation (in the illustration $L=3$), which is followed by a fully connected hidden layer (referred to as model layer in the text) resulting in a $K$ dimensional representation ($K=7$ in the figure). The model specific activation function is applied on this $K$-dimensional representation (sinusoidal activation in the figure). The output $\textbf{y} \in \mathbb{R}^c$ is produced by a fully connected output layer. Here $c$ is the number of classes in case of a classification task ($c=3$ in the figure).}
  \label{fig:architecture}
\end{figure}

\begin{figure}[t]

  \newcommand{\plotimages}[5]{%
    \foreach \x [count=\i] in {0,1,2,3,4,5,6,7,8,9} {

        \node[draw=white,fill=black!20,minimum size=\figurewidth,inner sep=0pt]
          (\i) at ({\figurewidth*mod(\i-1,5)},{\figureheight*int((\i-1)/5)})
          {\includegraphics[width=\figurewidth]{./fig/cifarimgs_#3/#2/#1_image_low\x.png}};
        \node[draw=white,fill=black!20,minimum size=\figurewidth,inner sep=0pt]
          (\i) at ({\figurewidth*mod(\i-1,5)+5.25*\figurewidth},{\figureheight*int((\i-1)/5)})
          {\includegraphics[width=\figurewidth]{./fig/cifarimgs_#3/#2/#1_image_high\x.png}};  
    }%
    \node[draw=#4,thick,fill=#4!10,rounded corners=1pt,inner sep=2pt] at ({4.625*\figurewidth},{-1.15*\figurewidth}) {\bf\color{#4} #5};  
    \node at ({1.1*\figurewidth},{-1.15*\figurewidth}) {\tiny $\leftarrow$~Most similar};
    \node at ({8.15*\figurewidth},{-1.15*\figurewidth}) {\tiny Least similar~$\rightarrow$};                
  }

  \centering\scriptsize

  \textbf{Results with ReLU model}\\
  \vspace*{1em}
  \pgfplotsset{every x tick scale label/.style={at={(rel axis cs:0.9,-0.45)},anchor=south west,inner sep=1pt},scaled ticks=false}
  \begin{subfigure}{.48\textwidth}
    \centering
    \setlength{\figurewidth}{\linewidth}
    \setlength{\figureheight}{.5\textwidth}
    \pgfplotsset{xlabel={Predictive entropy}, ytick=\empty, axis x line*=bottom, axis y line=none}

    \pgfplotsset{legend image code/.code={\draw[#1,fill=#1,fill opacity=0.2] (0cm,-0.1cm) rectangle (0.6cm,0.1cm);}, legend reversed=true, legend style={inner xsep=1pt, inner ysep=1pt, row sep=0pt, rounded corners=1pt, draw opacity=0.8, draw=white!80!black},}
     \input{./fig/cifarimgs_swag/relu_entropy.tex}
  \end{subfigure}
  \hfill
  \begin{subfigure}{.48\textwidth}
    \centering
    \setlength{\figurewidth}{\linewidth}
    \setlength{\figureheight}{.5\textwidth}
    \pgfplotsset{xlabel={Predictive entropy}, ytick=\empty, axis x line*=bottom, axis y line=none}
    \pgfplotsset{legend image code/.code={\draw[#1,fill=#1,fill opacity=0.2] (0cm,-0.1cm) rectangle (0.6cm,0.1cm);}, legend reversed=true, legend style={inner xsep=1pt, inner ysep=1pt, row sep=0pt, rounded corners=1pt, draw opacity=0.8, draw=white!80!black},}
     \pgfplotsset{xlabel={Predictive marginal variance}}
    \input{./fig/cifarimgs_swag/relu_variance.tex}
  \end{subfigure}\\

  \setlength{\figurewidth}{0.031\textwidth}
  \setlength{\figureheight}{\figurewidth}  
  \begin{subfigure}{.32\textwidth}
    \centering
    \begin{tikzpicture}[outer sep=0]
      \plotimages{cifar10}{relu}{swag}{mycolor1}{CIFAR-10}
    \end{tikzpicture}
  \end{subfigure}
  \hfill
  \begin{subfigure}{.32\textwidth}
    \centering
    \begin{tikzpicture}[outer sep=0]
      \plotimages{cifar100}{relu}{swag}{mycolor0}{CIFAR-100}
    \end{tikzpicture}
  \end{subfigure}
  \hfill
  \begin{subfigure}{.32\textwidth}
    \centering
    \begin{tikzpicture}[outer sep=0]
      \plotimages{svhn}{relu}{swag}{mycolor2}{SVHN}
    \end{tikzpicture}
  \end{subfigure}
  
  \vspace*{3em}
  \par\noindent\rule{\textwidth}{0.4pt}
  \medskip

  \textbf{Results with locally stationary RBF model} \\
  \vspace*{1em}
  \pgfplotsset{every x tick scale label/.style={at={(rel axis cs:0.9,-0.45)},anchor=south west,inner sep=1pt}}
  \begin{subfigure}{.48\textwidth}
    \centering
    \setlength{\figurewidth}{\linewidth}
    \setlength{\figureheight}{.5\textwidth}
    \pgfplotsset{xlabel={Predictive entropy}, ytick=\empty, axis x line*=bottom, axis y line=none}

    \pgfplotsset{legend image code/.code={\draw[#1,fill=#1,fill opacity=0.2] (0cm,-0.1cm) rectangle (0.6cm,0.1cm);}, legend reversed=true, legend style={inner xsep=1pt, inner ysep=1pt, row sep=0pt, rounded corners=1pt, draw opacity=0.8, draw=white!80!black},}
     \input{./fig/cifarimgs_swag/RBF_local_entropy.tex}
  \end{subfigure}
  \hfill
  \begin{subfigure}{.48\textwidth}
    \centering
    \setlength{\figurewidth}{\linewidth}
    \setlength{\figureheight}{.5\textwidth}
    \pgfplotsset{xlabel={Predictive entropy}, ytick=\empty, axis x line*=bottom, axis y line=none}
    \pgfplotsset{legend image code/.code={\draw[#1,fill=#1,fill opacity=0.2] (0cm,-0.1cm) rectangle (0.6cm,0.1cm);}, legend reversed=true, legend style={inner xsep=1pt, inner ysep=1pt, row sep=0pt, rounded corners=1pt, draw opacity=0.8, draw=white!80!black},}
     \pgfplotsset{xlabel={Predictive marginal variance}}
    \input{./fig/cifarimgs_swag/RBF_local_variance.tex}
  \end{subfigure}\\

  \setlength{\figurewidth}{0.031\textwidth}
  \setlength{\figureheight}{\figurewidth}  
  \begin{subfigure}{.32\textwidth}
    \centering
    \begin{tikzpicture}[outer sep=0]
      \plotimages{cifar10}{RBF_local}{swag}{mycolor1}{CIFAR-10}
    \end{tikzpicture}
  \end{subfigure}
  \hfill
  \begin{subfigure}{.32\textwidth}
    \centering
    \begin{tikzpicture}[outer sep=0]
      \plotimages{cifar100}{RBF_local}{swag}{mycolor0}{CIFAR-100}
    \end{tikzpicture}
  \end{subfigure}
  \hfill
  \begin{subfigure}{.32\textwidth}
    \centering
    \begin{tikzpicture}[outer sep=0]as inference method
      \plotimages{svhn}{RBF_local}{swag}{mycolor2}{SVHN}
    \end{tikzpicture}
  \end{subfigure}

  \vspace*{3em}
  \par\noindent\rule{\textwidth}{0.4pt}
  \medskip

  \textbf{Results with globally stationary RBF model (sinusoidal)} \\
  \vspace*{1em}
  \pgfplotsset{every x tick scale label/.style={at={(rel axis cs:0.9,-0.45)},anchor=south west,inner sep=1pt}}
  \begin{subfigure}{.48\textwidth}
    \centering
    \setlength{\figurewidth}{\linewidth}
    \setlength{\figureheight}{.5\textwidth}
    \pgfplotsset{xlabel={Predictive entropy}, ytick=\empty, axis x line*=bottom, axis y line=none}

    \pgfplotsset{legend image code/.code={\draw[#1,fill=#1,fill opacity=0.2] (0cm,-0.1cm) rectangle (0.6cm,0.1cm);}, legend reversed=true, legend style={inner xsep=1pt, inner ysep=1pt, row sep=0pt, rounded corners=1pt, draw opacity=0.8, draw=white!80!black},}
     \input{./fig/cifarimgs_swag/RBF_sin_entropy.tex}
  \end{subfigure}
  \hfill
  \begin{subfigure}{.48\textwidth}
    \centering
    \setlength{\figurewidth}{\linewidth}
    \setlength{\figureheight}{.5\textwidth}
    \pgfplotsset{xlabel={Predictive entropy}, ytick=\empty, axis x line*=bottom, axis y line=none}
    \pgfplotsset{legend image code/.code={\draw[#1,fill=#1,fill opacity=0.2] (0cm,-0.1cm) rectangle (0.6cm,0.1cm);}, legend reversed=true, legend style={inner xsep=1pt, inner ysep=1pt, row sep=0pt, rounded corners=1pt, draw opacity=0.8, draw=white!80!black},}
     \pgfplotsset{xlabel={Predictive marginal variance}}
    \input{./fig/cifarimgs_swag/RBF_sin_variance.tex}
  \end{subfigure}\\

  \setlength{\figurewidth}{0.031\textwidth}
  \setlength{\figureheight}{\figurewidth}  
  \begin{subfigure}{.32\textwidth}
    \centering
    \begin{tikzpicture}[outer sep=0]
      \plotimages{cifar10}{RBF_sin}{swag}{mycolor1}{CIFAR-10}
    \end{tikzpicture}
  \end{subfigure}
  \hfill
  \begin{subfigure}{.32\textwidth}
    \centering
    \begin{tikzpicture}[outer sep=0]
      \plotimages{cifar100}{RBF_sin}{swag}{mycolor0}{CIFAR-100}
    \end{tikzpicture}
  \end{subfigure}
  \hfill
  \begin{subfigure}{.32\textwidth}
    \centering
    \begin{tikzpicture}[outer sep=0]as inference method
      \plotimages{svhn}{RBF_sin}{swag}{mycolor2}{SVHN}
    \end{tikzpicture}
  \end{subfigure}

  \vspace*{3em}
  \par\noindent\rule{\textwidth}{0.4pt}
  \medskip

  \textbf{Results with locally stationary Mat\'ern-$\frac{3}{2}$ model}\\
  \vspace*{1em}
  \pgfplotsset{every x tick scale label/.style={at={(rel axis cs:0.9,-0.45)},anchor=south west,inner sep=1pt}}
  \begin{subfigure}{.48\textwidth}
    \centering
    \setlength{\figurewidth}{\linewidth}
    \setlength{\figureheight}{.5\textwidth}
    \pgfplotsset{xlabel={Predictive entropy}, ytick=\empty, axis x line*=bottom, axis y line=none}

    \pgfplotsset{legend image code/.code={\draw[#1,fill=#1,fill opacity=0.2] (0cm,-0.1cm) rectangle (0.6cm,0.1cm);}, legend reversed=true, legend style={inner xsep=1pt, inner ysep=1pt, row sep=0pt, rounded corners=1pt, draw opacity=0.8, draw=white!80!black},}
     \input{./fig/cifarimgs_swag/matern32_local_entropy.tex}
  \end{subfigure}
  \hfill
  \begin{subfigure}{.48\textwidth}
    \centering
    \setlength{\figurewidth}{\linewidth}
    \setlength{\figureheight}{.5\textwidth}
    \pgfplotsset{xlabel={Predictive entropy}, ytick=\empty, axis x line*=bottom, axis y line=none}
    \pgfplotsset{legend image code/.code={\draw[#1,fill=#1,fill opacity=0.2] (0cm,-0.1cm) rectangle (0.6cm,0.1cm);}, legend reversed=true, legend style={inner xsep=1pt, inner ysep=1pt, row sep=0pt, rounded corners=1pt, draw opacity=0.8, draw=white!80!black},}
     \pgfplotsset{xlabel={Predictive marginal variance}}
    \input{./fig/cifarimgs_swag/matern32_local_variance.tex}
  \end{subfigure}\\

  \setlength{\figurewidth}{0.031\textwidth}
  \setlength{\figureheight}{\figurewidth}  
  \begin{subfigure}{.32\textwidth}
    \centering
    \begin{tikzpicture}[outer sep=0]
      \plotimages{cifar10}{matern32_local}{swag}{mycolor1}{CIFAR-10}
    \end{tikzpicture}
  \end{subfigure}
  \hfill
  \begin{subfigure}{.32\textwidth}
    \centering
    \begin{tikzpicture}[outer sep=0]
      \plotimages{cifar100}{matern32_local}{swag}{mycolor0}{CIFAR-100}
    \end{tikzpicture}
  \end{subfigure}
  \hfill
  \begin{subfigure}{.32\textwidth}
    \centering
    \begin{tikzpicture}[outer sep=0]
      \plotimages{svhn}{matern32_local}{swag}{mycolor2}{SVHN}
    \end{tikzpicture}
  \end{subfigure}

  \vspace*{3em}
  \par\noindent\rule{\textwidth}{0.4pt}
  
  \end{figure} 
  
\setlength\@fptop{0pt}
\begin{figure}[t!]
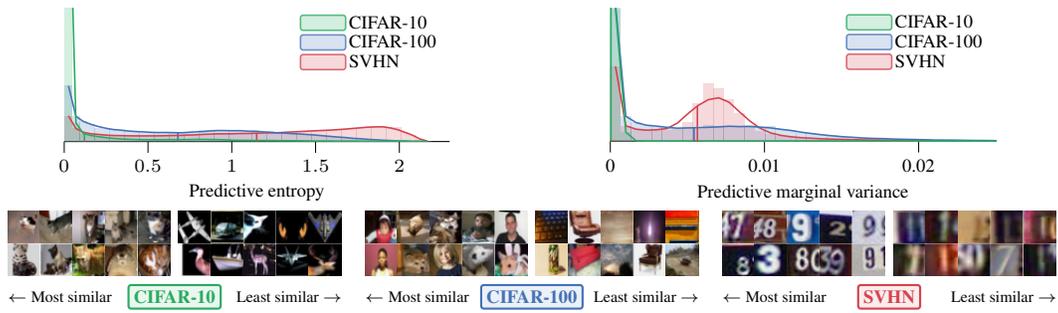
\ContinuedFloat

  \newcommand{\plotimages}[5]{%
    \foreach \x [count=\i] in {0,1,2,3,4,5,6,7,8,9} {

        \node[draw=white,fill=black!20,minimum size=\figurewidth,inner sep=0pt]
          (\i) at ({\figurewidth*mod(\i-1,5)},{\figureheight*int((\i-1)/5)})
          {\includegraphics[width=\figurewidth]{./fig/cifarimgs_#3/#2/#1_image_low\x.png}};
        \node[draw=white,fill=black!20,minimum size=\figurewidth,inner sep=0pt]
          (\i) at ({\figurewidth*mod(\i-1,5)+5.25*\figurewidth},{\figureheight*int((\i-1)/5)})
          {\includegraphics[width=\figurewidth]{./fig/cifarimgs_#3/#2/#1_image_high\x.png}};  
    }%
    \node[draw=#4,thick,fill=#4!10,rounded corners=1pt,inner sep=2pt] at ({4.625*\figurewidth},{-1.15*\figurewidth}) {\bf\color{#4} #5};  
    \node at ({1.1*\figurewidth},{-1.15*\figurewidth}) {\tiny $\leftarrow$~Most similar};
    \node at ({8.15*\figurewidth},{-1.15*\figurewidth}) {\tiny Least similar~$\rightarrow$};                
  }

  \centering\scriptsize

  \textbf{Results with globally stationary Mat\'ern-$\frac{3}{2}$ model (sinusoidal)} \\
  \vspace*{1em}
  \pgfplotsset{every x tick scale label/.style={at={(rel axis cs:0.9,-0.45)},anchor=south west,inner sep=1pt},scaled ticks=false}
  \begin{subfigure}{.48\textwidth}
    \centering
    \setlength{\figurewidth}{\linewidth}
    \setlength{\figureheight}{.5\textwidth}
    \pgfplotsset{xlabel={Predictive entropy}, ytick=\empty, axis x line*=bottom, axis y line=none}

    \pgfplotsset{legend image code/.code={\draw[#1,fill=#1,fill opacity=0.2] (0cm,-0.1cm) rectangle (0.6cm,0.1cm);}, legend reversed=true, legend style={inner xsep=1pt, inner ysep=1pt, row sep=0pt, rounded corners=1pt, draw opacity=0.8, draw=white!80!black},}
     \input{./fig/cifarimgs_swag/matern32_sin_entropy.tex}
  \end{subfigure}
  \hfill
  \begin{subfigure}{.48\textwidth}
    \centering
    \setlength{\figurewidth}{\linewidth}
    \setlength{\figureheight}{.5\textwidth}
    \pgfplotsset{xlabel={Predictive entropy}, ytick=\empty, axis x line*=bottom, axis y line=none}
    \pgfplotsset{legend image code/.code={\draw[#1,fill=#1,fill opacity=0.2] (0cm,-0.1cm) rectangle (0.6cm,0.1cm);}, legend reversed=true, legend style={inner xsep=1pt, inner ysep=1pt, row sep=0pt, rounded corners=1pt, draw opacity=0.8, draw=white!80!black},}
     \pgfplotsset{xlabel={Predictive marginal variance}}
    \input{./fig/cifarimgs_swag/matern32_sin_variance.tex}
  \end{subfigure}\\

  \setlength{\figurewidth}{0.031\textwidth}
  \setlength{\figureheight}{\figurewidth}  
  \begin{subfigure}{.32\textwidth}
    \centering
    \begin{tikzpicture}[outer sep=0]
      \plotimages{cifar10}{matern32_sin}{swag}{mycolor1}{CIFAR-10}
    \end{tikzpicture}
  \end{subfigure}
  \hfill
  \begin{subfigure}{.32\textwidth}
    \centering
    \begin{tikzpicture}[outer sep=0]
      \plotimages{cifar100}{matern32_sin}{swag}{mycolor0}{CIFAR-100}
    \end{tikzpicture}
  \end{subfigure}
  \hfill
  \begin{subfigure}{.32\textwidth}
    \centering
    \begin{tikzpicture}[outer sep=0]as inference method
      \plotimages{svhn}{matern32_sin}{swag}{mycolor2}{SVHN}
    \end{tikzpicture}
  \end{subfigure}
  
  \caption{OOD detection experiment results for models trained on CIFAR-10 and tested on CIFAR-10, CIFAR-100, and SVHN. Predictive entropy histograms of test image results are on the left, and predictive marginal variance histograms are on the right. On the bottom are sample images from each test set: left-side images with lowest entropy/highest confidence, and right-side images with highest entropy/lowest confidence.}
  \label{fig:histograms_app}
\end{figure}

\end{document}